\documentclass{article}
\usepackage[preprint]{neurips_2024}
\usepackage{natbib}

\usepackage[utf8]{inputenc} 
\usepackage[T1]{fontenc}    
\usepackage{hyperref}       
\usepackage{url}            
\usepackage{booktabs}       
\usepackage{amsfonts}       
\usepackage{nicefrac}       
\usepackage{microtype}      
\usepackage{xcolor}         

\usepackage{float} 
\usepackage{caption}
\usepackage{subcaption}
\usepackage{wrapfig}
\usepackage{setspace}
\usepackage{amsmath}
\usepackage{amssymb}
\usepackage{mathtools}
\usepackage{amsthm}
\usepackage{comment}
\usepackage[export]{adjustbox}
\usepackage{multirow}

\usepackage{algorithm}
\usepackage{algorithmic}

\theoremstyle{plain}
\newtheorem{theorem}{Theorem}[section]

\theoremstyle{definition}

\theoremstyle{remark}

\usepackage[normalem]{ulem}

\usepackage{slashbox}

\author{%
  Michal Nauman \\
  Ideas NCBR \& University of Warsaw\\
  \texttt{nauman.mic@gmail.com} \\
  \And
  Marek Cygan \\
  Nomagic \& University of Warsaw \\
  \texttt{ma.cygan@uw.edu.com} \\
}

\title{On the Theory of Risk-Aware Agents: Bridging Actor-Critic and Economics}

\begin{document}

\maketitle

\begin{abstract}
    Risk-aware Reinforcement Learning (RL) algorithms like SAC and TD3 were shown empirically to outperform their risk-neutral counterparts in a variety of continuous-action tasks. However, the theoretical basis for the pessimistic objectives these algorithms employ remains unestablished, raising questions about the specific class of policies they are implementing. In this work, we apply the expected utility hypothesis, a fundamental concept in economics, to illustrate that both risk-neutral and risk-aware RL goals can be interpreted through expected utility maximization using an exponential utility function. This approach reveals that risk-aware policies effectively maximize value certainty equivalent, aligning them with conventional decision theory principles. Furthermore, we propose Dual Actor-Critic (DAC). DAC is a risk-aware, model-free algorithm that features two distinct actor networks: a pessimistic actor for temporal-difference learning and an optimistic actor for exploration. Our evaluations of DAC across various locomotion and manipulation tasks demonstrate improvements in sample efficiency and final performance. Remarkably, DAC, while requiring significantly less computational resources, matches the performance of leading model-based methods in the complex dog and humanoid domains.
\end{abstract}

\section{Introduction}

Deep Reinforcement Learning (RL) is still in its infancy, with a variety of tasks unsolved \citep{sutton2018reinforcement, hafner2023mastering} or solved within an unsatisfactory amount of environment interactions \citep{zawalski2022fast, schwarzer2023bigger}. Whereas increasing the Replay Ratio (RR) (ie. the number of parameter updates per environment interactions step) is a promising general approach for increasing sample efficiency and final performance of RL agents \citep{janner2019trust, chen2020randomized, nikishin2022primacy}, it is characterized by quickly diminishing gains \citep{d2022sample} combined with linearly increasing computational cost \citep{rumelhart1986learning, kingma2014adam}. Moreover, the limitations of robot hardware and data acquisition frequency constrain the maximum achievable replay ratio \citep{smith2022walk}. As such, it is worthwhile to pursue orthogonal techniques such as enhancing the properties of the underlying model-free agents. One continuously researched theme deals with finding risk attitudes that efficiently handle the \textit{exploration-exploitation} dilemma \citep{ciosek2019better, moskovitz2021tactical}. 

\begin{figure}[ht!]
\begin{center}
\begin{minipage}[h]{1.0\linewidth}
    \begin{subfigure}{1.0\linewidth}
    \hfill
    \includegraphics[width=0.49\linewidth]{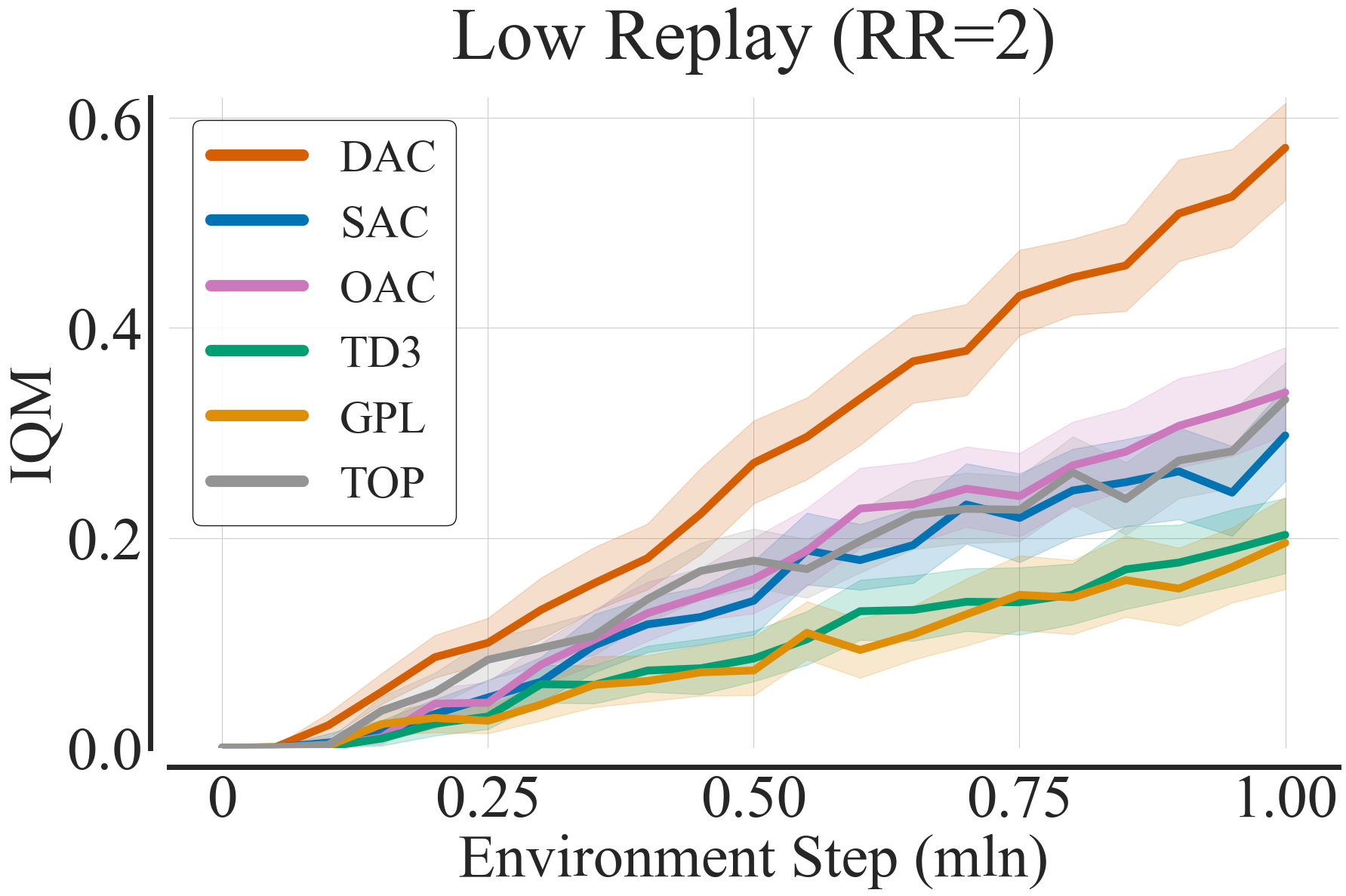}
    \hfill
    \includegraphics[width=0.49\linewidth]{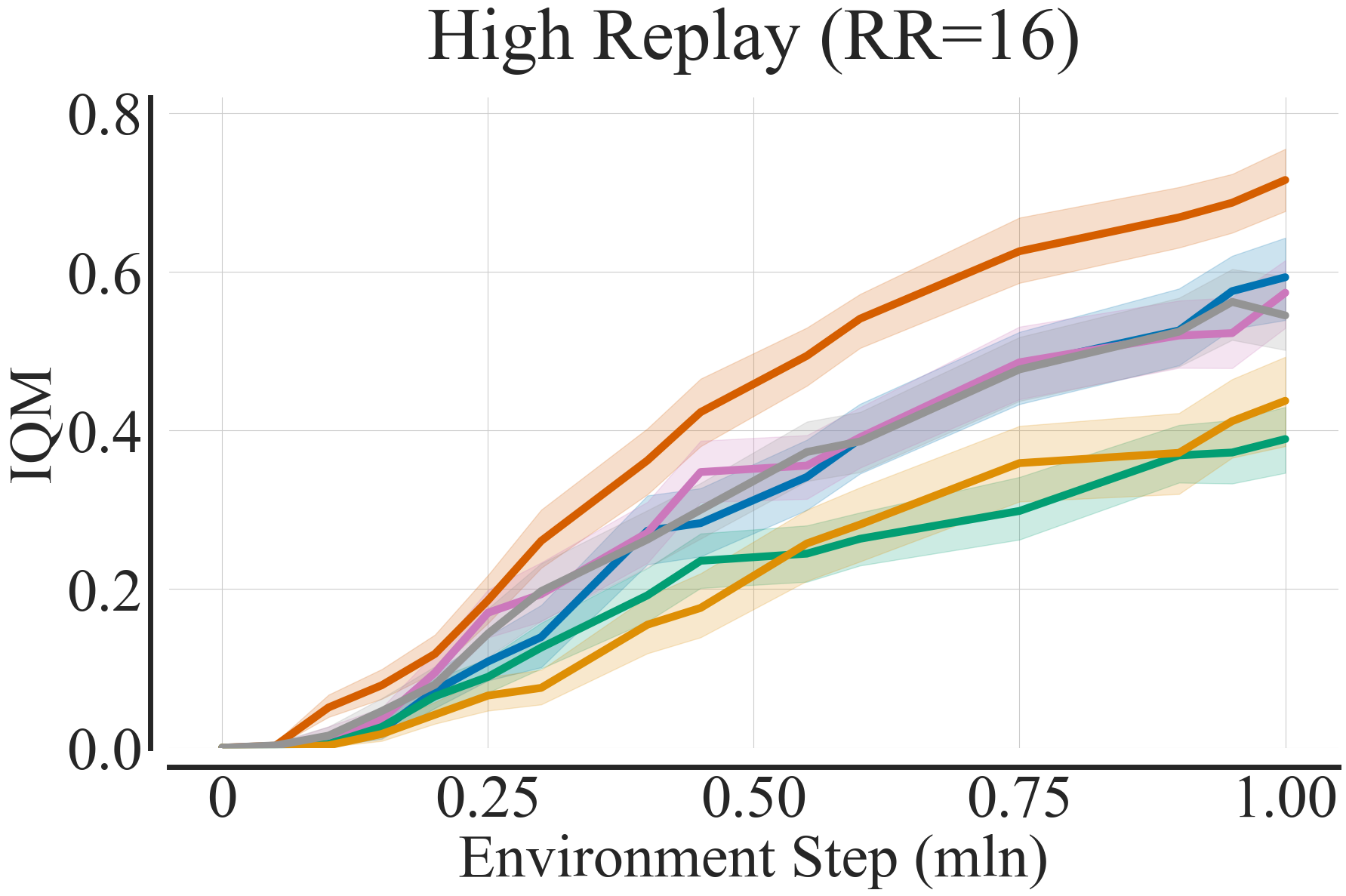}
    \hfill
    \end{subfigure}
\end{minipage}
\vspace{-0.1in} 
\caption{We test the proposed approach (DAC) against various risk-aware and risk-neutral actor-critic baselines in $30$ tasks listed in Table \ref{tab:all_tasks}. Due to a controlled experimental setup described in Section \ref{section:experiments}, the performance differences between algorithms stem solely from their risk-management. $Y$-axis reports IQM with $95$\% CI calculated using 10 seeds, with $1.0$ representing the maximal score.}
\label{fig:sample_efficiency}
\end{center}
\vspace{-0.1in} 
\end{figure}

The attitudes towards risk of algorithmic agents have been researched in multiple contexts. For instance, a risk-seeking approach of optimism in the face of uncertainty has been identified as an effective exploration strategy, minimizing regret during the learning process \citep{wang2020optimism, neu2020unifying}. Conversely, risk-averse, pessimistic Q-learning strategies have proven beneficial in counteracting value overestimation caused by temporal difference errors \cite{hasselt2010double, fujimoto2018addressing}. However, there is a disconnect between these risk-aware strategies and the foundational theories of RL, particularly in how they relate to the goal of value maximization. As a result, despite the empirical success of risk-aware agents like TD3 \citep{fujimoto2018addressing} or SAC \citep{haarnoja2018soft}, the specific class of policies they implement remains ambiguous.

In this paper, we study the reinforcement learning objective aligned with the principles of decision theory. We show that, in contrast to pure value maximization, the decision theoretic perspective allows for derivation of both risk-neutral (e.g., DDPG \citep{silver2014deterministic}) and risk-aware approaches (e.g., TD3 \citep{fujimoto2018addressing}). As such, we demonstrate that the pessimistic updates in state-of-the-art algorithms such as SAC \citep{haarnoja2018soft}, REDQ \citep{chen2020randomized}, or TOP \cite{moskovitz2021tactical} can be derived from expected utility maximization under an exponential utility function. Furthermore, we introduce Dual Actor-Critic (DAC), a risk-aware algorithm with two actors: optimistic and pessimistic. In DAC, each actor is trained independently using gradient backpropagation of a distinct objective: the optimistic actor aims to maximize an upper value bound for exploration, while the pessimistic actor focuses on a lower bound for temporal-difference (TD) targets \citep{fujimoto2018addressing, haarnoja2018soft}. DAC also features an automatic adjustment mechanism to balance the divergence between these actors, allowing for adaptability to various tasks without hyperparameter tuning. We evaluate DAC on a diverse set of locomotion and manipulation tasks and find that, despite its simplicity, DAC achieves significant improvements in sample efficiency and performance. Below, we outline our contributions:

$\mathbf{1}$. We consider a generalized utility-based actor-critic objective, capable of formalizing both risk-neutral and risk-aware actor-critic algorithms. We demonstrate that the policies enacted by Clipped Double Q-Learning and generalizations thereof approximately align with the certainty equivalent of value under an exponential utility, and as such are optimal in the decision theoretic context. 

$\mathbf{2}$. We introduce Dual Actor-Critic (DAC), a risk-aware actor-critic framework featuring a dual actor network configuration. In DAC, each actor is trained through gradient backpropagation stemming from a specific objective that reflects different degrees of risk appetite. We establish the optimistic policy loss function and implement a system for online gradient-based adjustment of optimism hyperparameters. This feature enables DAC to effectively adapt to varying degrees of uncertainties, as well as different reward scales, without the need for manual hyperparameter tuning.
    
$\mathbf{3}$. We show that DAC outperforms benchmark algorithms in terms of both sample efficiency and final performance. Notably, DAC solves the dog domain, reaching the performance of significantly more complex model-based methods. To facilitate further research, we perform ablations on various design and hyperparameter choices (over $5000$ training runs). We release the implementation of DAC under the following URL

\section{Background}
\label{sec:preliminaries}

\paragraph{Reinforcement Learning}

We consider an infinite-horizon Markov Decision Process (MDP) \citep{puterman2014markov} which is described with a tuple $(S, A, r, p, \gamma)$, where states $S$ and actions $A$ are continuous, $r_{s,a}$ is the transition reward, $p$ is a deterministic transition mapping, with $p_0$ being the starting state distribution, and $\gamma \in (0,1]$ is a discount factor. A policy $\pi(a|s)$ is a state-conditioned action distribution with entropy denoted as $\mathcal{H}(\pi(s))$. Soft Value \citep{haarnoja2018soft} is the sum of expected discounted return and state entropies from following the policy at a given state $V^{\pi} (s) = \mathrm{E}_{a\sim\pi} \left[r_{s,a} + \alpha \mathcal{H}(\pi(s)) + \gamma V^{\pi}(s') \right]$, with $\alpha$ denoting the entropy temperature parameter. Q-value is the expected discounted return from performing an action and following the policy thereafter $Q^{\pi}(s, a) = \left[ r_{s,a} + \gamma V^{\pi}(s') \right]$. A policy is said to be optimal if it maximizes the expected value of the possible starting states $s_0$, such that $\pi^{*} = \arg\max_{\pi\in\Pi} \mathrm{E}_{s_0 \sim p_0}~V^{\pi}(s_0)$, with $\pi^{*}$ denoting the optimal policy and $\Pi$ denoting the considered set of policies (eg. Gaussian). Off-policy actor-critic algorithms perform gradient-based learning of both Q-values (ie. critic or value model) and the policy (ie. actor). The critic parameters $\theta$ are updated by minimizing SARSA temporal-difference loss $\mathcal{L}_{\theta}$ on transitions $T=(s,a,r,s')$ which are sampled from past experiences \citep{fujimoto2018addressing, haarnoja2018soft} according to $\mathcal{L}_{\theta} = \mathrm{E}_{T \sim \mathcal{D}} ~ ( Q_{\theta} (s,a) - r_{s,a} - \gamma V^{\pi_{\phi}}(s'))^{2}$. Here, $Q_{\theta} (s,a)$ is the critic output, $V^{\pi}(s')$ is the bootstrap value derived from a target network, and $\mathcal{D}$ is the experience buffer \citep{mnih2015human}. The policy parameters $\phi$ are updated to maximize values approximated by the critic \citep{ciosek2020expected}: $\mathcal{L}_{\phi} = \mathrm{E}_{s \sim \mathcal{D}}~ V^{\pi_{\phi}}(s)$. In this equation, $\pi_{\phi}$ represents the actor, and $V^{\pi_{\phi}}(s)$ is the state value under $\pi_{\phi}$. 

\paragraph{Risk-Awareness in Actor-Critic}

In standard actor-critic the critic represents the expected future discounted return and the actor is optimized with respect to such critic \citep{lillicrap2015continuous}. In contrast, popular algorithms like SAC or TD3 learn a critic that represents the lower bound of Q-values \citep{fujimoto2018addressing, haarnoja2018soft2}. Employing the lower-bound has proven effective in mitigating value overestimation in Temporal Difference (TD) learning \citep{van2016deep, fujimoto2018addressing}. A popular approach is Clipped Double Q-Learning (CDQL), where the lower-bound is calculated by taking the minimum value of an ensemble of critics, most often two \citep{fujimoto2018addressing, haarnoja2018soft2, ciosek2019better, hansen2022temporal}: $V^{\pi_{\phi}}(s) \approx \min ( Q_{\theta}^{1}(s, a), Q_{\theta}^{2}(s, a)) - \log \pi_{\phi} (a|s)$, with $a \sim \pi_{\phi} (s)$. Here, $Q_{\theta}^{1}(s,a)$ and $Q_{\theta}^{2}(s,a)$ denote the first and the second critic in the ensemble, and $\log \pi_{\phi}$ is the state-action entropy. Furthermore, it was shown that using the minimum is equivalent to ensemble statistics \citep{ciosek2019better}: 

\begin{equation}
\label{eq:CDQL}
\begin{split}
    \min \bigl( Q_{\theta}^{1}(s, a), Q_{\theta}^{2}(s, a) \bigr) = \underbrace{\frac{1}{2}\bigl( Q_{\theta}^{1}(s,a) + Q_{\theta}^{2}(s,a)\bigr)}_\text{Mean} - \underbrace{\frac{1}{2}~|Q_{\theta}^{1}(s,a) - Q_{\theta}^{2}(s,a)|}_\text{Standard Deviation}.
\end{split}
\end{equation}

This led to generalizations of the lower-bound as to include varying levels of pessimism \citep{ciosek2019better, moskovitz2021tactical}:

\begin{equation}
\label{eq:lowerbounds}
\begin{split}
    Q^{\beta}(s, a) = Q^{\mu}(s, a) + \beta Q^{\sigma}(s, a), \quad \text{and} \quad V^{\beta}(s) = V^{\mu}(s) + \beta V^{\sigma}(s)
\end{split}
\end{equation}

Above, $Q^{\beta}(s, a)$ and $V^{\beta}(s)$ are the pessimistic Q-value and values respectively, $\beta$ represents the degree of pessimism, and $(\mu, \sigma)$ denote the mean and standard deviation of the critic model ensemble. Furthermore, the pessimistic values and Q-values are related by $V^{\beta}(s) = \mathrm{E}_{a \sim \pi} ( Q^{\beta}(s, a) - \log\pi_{\theta}(a|s))$. In this setup, the risk level of the lower bound is modulated by $\beta$. As shown in Equation \ref{eq:CDQL}, in case of $\beta = -1$ it is true that $Q^{\beta} =  \min ( Q_{\theta}^{1}(s, a), Q_{\theta}^{2}(s, a))$. In setup of pessimistic learning the value targets are affected by the uncertainty measured via the value model ensemble disagreement. A non-zero $\beta$ indicates a departure from risk-neutrality: positive $\beta$ yields a pessimistic lower bound, while negative $\beta$ results in an optimistic upper bound. Thus, for given $\phi$ the difference between the risk-aware value (resulting from bootstrapping with $Q^{\beta}$), and risk-neutral values (resulting from bootstrapping with $Q^{\mu}$) is proportional to the expected values of the discounted sum of critic disagreements \citep{fujimoto2018addressing}. As such, it is clear that iterating the pessimistic objective does not result in the optimal policy as long as the critic disagreement is not equal to zero for all state-actions \citep{kumar2020discor}. Despite this, numerous effective off-policy algorithms adopt a risk-aware objective, leaning towards either pessimism or optimism \citep{fujimoto2018addressing, ciosek2019better, cetin2023learning}. This understanding of risk-awareness is slightly different to the risk-awareness considered in the distributional RL \citep{bellemare2017distributional, dabney2018distributional, dabney2018implicit} or the CVaR series \citep{la2013actor, hiraoka2019learning, lim2022distributional}. There, the distribution of returns is explicitly modelled via a distribution and the risk-awareness is calculated with respect to the aleatoric uncertainty \citep{dabney2018implicit, lim2022distributional}. In contrast, CDQL and associated techniques use regular critics and consider the total estimation uncertainty associated with using point-estimate to model Q-values \citep{fujimoto2018addressing, hiraoka2021dropout}.

\paragraph{Expected Utility Theorem}
  
The Von Neumann–Morgenstern Theorem \citep{von1947theory} posits that an agent whose preferences adhere to four axioms (completeness, transitivity, continuity, and independence), has a utility function $~ \mathcal{U}(x)$ that enables the comparison of the preferences. The expected utility hypothesis \citep{von1947theory, kahneman1979prospect} states that agents choose between risky options by comparing their expected utility. For example, given random variables $X_i$ representing different propositions the agent ought to choose from, it follows that $X_1 \succeq X_2 \iff \mathrm{E}~ U(X_1) \geq \mathrm{E} ~U(X_2)$. In such setting, the goal of the agent is optimize the utility rather than its input \citep{stiglitz1997microeconomics}: $x^{*} = \arg\max_{x} \mathrm{E}_{x_i \sim X} ~ \mathcal{U}(x_i)$. Here, risk stems from the potential decrease in utility due to uncertainty in its input space, making risk preference an attribute of the utility function under consideration. In particular, due to potential non-linearities in the utility there can be a discrepancy between $\arg\max\mathrm{E} ~ U(x_i)$ and $\arg\max\mathrm{E} ~ x_i$. Certainty equivalent of $X$, denoted as $X_{c}$, measures the impact of uncertainty on utility-optimal choices: $\mathcal{U}(X_{c}) = \mathrm{E}~\mathcal{U}(X)$. Certainty equivalent presents a deterministic amount offering the same utility as a random event and varies based on risk preferences. Risk-averse utilities yield a certainty equivalent lower than the expected value, while risk-seeking leads to a higher certainty equivalent. The exponential utility, $\mathcal{U}(x, \beta) = e^{\beta x}$, is a simple model for risk-awareness, with $\beta$ determining the risk preference. We show the basic risk-averse and risk-loving utilities in Figure \ref{fig:risks}.

\section{Theory of Risk-Aware Actor-Critic}
\label{section:risk_aware}

In RL, risks can be associated with any uncertainty present in the optimization of the learning objectives. In this manuscript we focus on risks associated with the uncertainty of value approximation. In particular, we focus on off-policy agents that perform updates according to the risk-aware pessimistic Q-values $Q^{\beta}$ presented in Equation \ref{eq:lowerbounds}, such as SAC \citep{haarnoja2018soft}, TD3 \citep{fujimoto2018addressing}, OAC \citep{ciosek2019better}, TOP \citep{moskovitz2021tactical} or GPL \citep{cetin2023learning}. Consequently, the policy derived from this approach deviates from the optimal policy stemming from pure value maximization, as indicated by the regret defined in the Background section. Notably, many state-of-the-art algorithms adopt non-risk-neutral strategies \citep{moskovitz2021tactical, hiraoka2021dropout, d2022sample}. However, the risk-aware correction, as outlined in Section \ref{sec:preliminaries}, is not fully explicated by existing RL theory and does not emerge from pure value maximization problems. To this end, the efficacy of risk-aware agents is often attributed to two factors: risk-averse pessimism is justified by well-documented value overestimation in temporal learning \citep{hasselt2010double, fujimoto2018addressing, kumar2020discor}, while risk-loving optimism is supported by lower regret guarantees for exploration \cite{chen2017ucb, ciosek2019better, schrittwieser2020mastering}. In this section, we posit that both risk-aware and risk-neutral algorithms can be formalized as optimizing an expected utility objective. We consider a cycle of policy evaluation and improvement steps and analyze a single step thereof. In every step of iteration, the values are assumed to be samples from the distribution $\mathcal{V}(s)$, which is assumed to have finite moments and expected value denoted by $V^{\mu}$, such that $V^{\mu}(s) = \mathrm{E}_{i \sim \mathcal{V}} ~ V_{i}(s)$. Assuming that the expected value of $V_{i}(s)$ is an unbiased estimator of the on-policy value, the standard approach requires the policy to optimize for $V^{\mu}(s)$, such that $\pi^{*} = \arg\max_{\pi \in \Pi} \mathrm{E}_{s \sim \mathcal{D}} V^{\mu}(s)$. However, this is not the approach applied by the risk-aware algorithms, such as SAC or TD3. Given an invertible, increasing utility function $~\mathcal{U}$, we define the \textit{certainty equivalent value}, denoted as $V^{c}(s)$:

\begin{equation}
    V^{c}(s) = \mathcal{I} ~ \underset{i \sim \mathcal{V}}{\mathrm{E}} ~ \mathcal{U} ~ V_{i}(s) = V^{\mu}(s) + \Upsilon(s).
\end{equation}

Above, the inverted utility function is denoted by $\mathcal{I} = \mathcal{U}^{-1}$ and $\Upsilon(s)$ denotes the risk premium. In this context, the certainty equivalent value represents the deterministic amount that amortizes the uncertainties associated with the value approximation stochasticity. As such, if the utility function implies risk-averse behaviour, then $\Upsilon(s) < 0$ resulting in certainty equivalent value that is smaller than $V^{\mu}(s)$. Building on the expected utility objective, we define the \textit{certainty equivalence policy} that seeks to amortize the uncertainty associated with approximation of values:

\begin{equation}
\begin{split}
    \pi^{c} & = \underset{\pi \in \Pi}{\arg\max} ~ \underset{s \sim p}{\mathrm{E}} ~ \underset{i \sim \mathcal{V}}{\mathrm{E}} ~ \mathcal{U} ~ V_{i}(s) =  \underset{\pi \in \Pi}{\arg\max} \underset{s \sim p}{\mathrm{E}} V^{c}(s).
\end{split}
\end{equation}
    
Above, $\pi^{c}$ represents the policy that optimizes for the expected utility of values, which we term the \textit{certainty equivalent policy}. As follows, the certainty equivalent policy is greedy with respect to the certainty equivalent value and aligns with the expected utility hypothesis, wherein the agent seeks to maximize the expected utility of returns. This outcome contrasts with achieving the maximal value possible, as is typically sought in standard RL objectives. The certainty equivalent policy $\pi^{c}$ optimizes for values that are not generally equal to the risk-neutral values. Specifically, the $\pi^{c}$ and $\pi^{*}$ are equivalent only iff $\Upsilon(s) = 0$, thus making $V^{c}(s) = V^{\mu}(s)$. This highlights the fact that the utility-based objective is a generalization of the traditional RL objective, with traditional RL objective being understood as utility maximization under a linear utility function. Hence, risk-neutral algorithms like DDPG can be interpreted as linear utility agents optimizing the certainty equivalent. Te explore scenarios where the utility is non-linear and diverges from the traditional objective, one has to evaluate the risk premium. Given that $~\mathcal{U}$ is invertible, infinitely differentiable and its Taylor expansion is convergent, then the risk premium can be evaluated via Taylor series:

\begin{equation}
\begin{split}
    \Upsilon(s) & \approx \mathcal{I} \underset{i \sim \mathcal{V}}{\mathrm{E}} \sum_{n=1}^{\infty}\frac{\mathcal{U}^{n}\bigl(V^{\mu}(s)\bigr)}{n!}\bigl(V_{i}(s) - V^{\mu}(s)\bigr)^{n}.
\end{split}
\end{equation}

Above, we denote $~\mathcal{U}^{n}(V^{\mu}(s))$ as the $n$th derivative of $~\mathcal{U}$ calculated at $V^{\mu}(s)$. As such, the discrepancy between the certainty equivalent value and the risk-neutral value depends on moments of $\mathcal{V}(s)$. Now, we show that the risk-aware value $V^{\beta}(s)$ stems from an exponential utility.

\begin{theorem}
\label{theorem:1}
Denote exponential utility $~\mathcal{U}(V_{i}, \beta) = e^{2\beta V_{i}(s)}$, then the certainty equivalent value $V^{c}(s)$ is approximately equal to the pessimistic value $V^{\beta}(s)$:

\begin{equation}
\begin{split}
    V^{c}(s) \approx \underbrace{\underset{i \sim \mathcal{V}}{\mathrm{E}}  V_{i}(s)}_\text{Ensemble Mean} + \beta \underbrace{\underset{i \sim \mathcal{V}}{\mathrm{E}} \bigl( V_{i}(s) - V^{\mu}(s) \bigr)^{2}}_\text{Ensemble Variance}.
\end{split}
\end{equation}
\end{theorem}

As detailed in Appendix \ref{appendix:derivations}, we achieve this result by using convergent Taylor approximation, with the key step of approximating $e^{x}$ with second-order series, which since $x = V_{i}(s) - V^{\mu}(s)$ is centered at $0$. In consequence of the above Theorem, risk-aware agents like TD3 or SAC can be understood as optimizing for the utility objective, pursuing certainty equivalent rather than the expected values. 

\section{Dual Actor-Critic}
\label{section:dac}

In this section, we propose the Dual Actor-Critic (DAC). DAC is a risk-aware, off-policy algorithm that addresses the exploration and temporal-difference (TD) learning dichotomy in actor-critic algorithms. Usually, actor-critic algorithms use a single actor for both exploration (sampling actions for new transitions) and TD learning (calculating TD targets), which requires balancing between optimism for exploration \citep{wang2020optimism} and pessimism for avoiding value overestimation \citep{hasselt2010double}. DAC resolves this by employing optimistic and pessimistic actors, with each actor being updated to optimize the certainty equivalent value stemming from utilities with unique risk preferences. Following Soft Actor-Critic, DAC pursues the maximum entropy objective in a policy iteration performed on dataset of previous experiences \cite{haarnoja2018soft}. 

Whereas Theorem \ref{theorem:1} validates the use of risk-aware policies in terms of the optimality of the pursued solutions, we further develop DAC basing on the principles of SARSA, off-policy learning and Optimism in the Face of Uncertainty. Firstly, following SAC, we update the critic network according to a pessimistic value target sampled from the pessimistic actor. This design choice guarantees that the critic learns the pessimistic values under the pessimistic policy \cite{van2009theoretical}, while tackling critic value overestimation \cite{fujimoto2018addressing}. Secondly, by performing off-policy value updates we allow for exploration via a different policy than the one used for value updates. In particular, we consider a policy that is optimistic and learns to perform actions that yield critic disagreement, thus tackling the issue of pessimistic underexploration \cite{ciosek2019better}. In DAC, the optimistic actor is trained to maximize the upper Q-value bound and is solely used for exploration, while the pessimistic actor, guided by the lower Q-value bound, is used for TD learning and evaluation. This separation allows DAC to explore efficiently without the risk of value overestimation. DAC also addresses the shortcomings of Optimistic Actor-Critic (OAC) \citep{ciosek2019better}. By relaxing the first-order approximation and explicitly modeling the optimistic policy via a neural network DAC can approximate the maximum of arbitrary upper bound \citep{hornik1989multilayer}. DAC adjusts the optimism levels associated with the upper bound such that the two policies reach a predefined divergence target, thus alleviating the tandem problem induced by off-policy learning using two actors. DAC is structured around two components: a critic ensemble of $k$ models (following the standard SAC/TD3 implementation we use $k=2$ models) and the dual actor comprising pessimistic and optimistic actors denoted $\pi_{\phi}^{p}$ and $\pi_{\eta}^{o}$ respectively. As shown in Theorem \ref{theorem:1}, each actor can be interpreted as optimizing exponential utility function with different levels of risk appetite, denoted as $\beta^{p}$ and $\beta^{o}$ for the pessimistic and optimistic actors.

\paragraph{Optimistic Actor} The optimistic actor $\pi_{\eta}^{o}$, pursues an objective function that maximizes a utility expression that accounts for the divergence between the two actors:

\begin{equation}
\label{eq:upd_optiactor}
\begin{split}
    \mathcal{L}_{\eta} & = - \underset{s \sim \mathcal{D}}{\mathrm{E}} ~ \underset{i \sim \mathcal{V}}{\mathrm{E}} e^{~2\beta^{o}(V_{i}(s)-\tau KL(\pi_{\phi}^{p}(s)|\Bar{\pi}_{\eta}^{o}(s)))} \\
    & \approx -\underset{s \sim \mathcal{D}}{\mathrm{E}} \bigl( Q^{\mu}_{\theta}(s, a) + \beta^{o} Q^{\sigma}_{\theta}(s, a) - \tau KL\bigl(\pi_{\phi}^{p}(s)|\Bar{\pi}_{\eta}^{o}(s)\bigr) \bigr) \quad \text{with} \quad a \sim \pi^{o}_{\eta}(s).
\end{split}
\end{equation}

Above, $KL$ represents the empirical Kullback-Leibler divergence between the actors, $\beta^{o}$ is the optimism parameter, $\tau$ is the penalty weight associated with KL divergence, and $\Bar{\pi}_{\eta}^{o}(s)$ is the transformed pessimistic policy. Incorporating two actors that are used for TD and exploration respectively, it is natural to assign distinct entropy levels to each policy. Our approach involves computing the KL divergence between the pessimistic policy $\pi_{\phi}^{p}$ and a modified optimistic policy, denoted as $\Bar{\pi}_{\eta}^{o}$. This modified policy is characterized by a standard deviation that is $m$-times smaller than that of the optimistic policy $\pi_{\eta}^{o}$, with $m\in(0,\infty)$. We refer to $m$ as exploration variance multiplier. We present ablations on different values of $m$ in Figure \ref{fig:hyperparams}, as well as tests regarding the importance of KL divergence in Table \ref{tab:design_ablation}. We detail methodology for KL divergence calculation as well as technical details on implementing the optimistic actor via a neural network in Appendix \ref{appendix:dac}. 

\begin{figure}[t!]
\begin{center}
\begin{minipage}[h]{0.245\linewidth}
    \begin{subfigure}{1.0\linewidth}
    \hfill
    \subcaption{$Q^{\sigma} / Q^{\mu}$}
    \label{fig:adjust1}
    \includegraphics[width=1.0\linewidth]{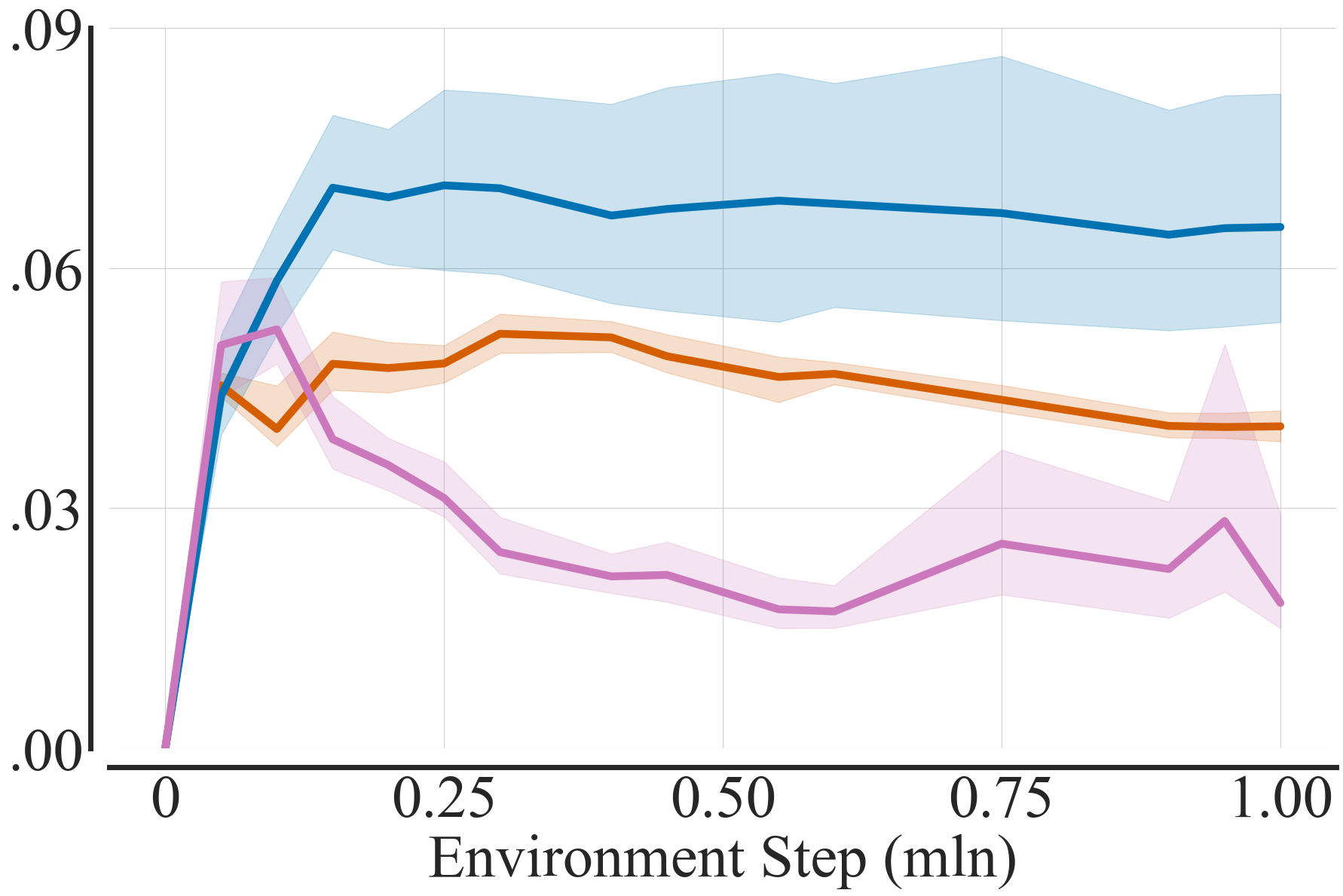}
    \hfill
    \end{subfigure}
\end{minipage}
\begin{minipage}[h]{0.245\linewidth}
    \begin{subfigure}{1.0\linewidth}
    \hfill
    \subcaption{Optimism ($\beta^{o}$)}
    \label{fig:adjust2}
    \includegraphics[width=1.0\linewidth]{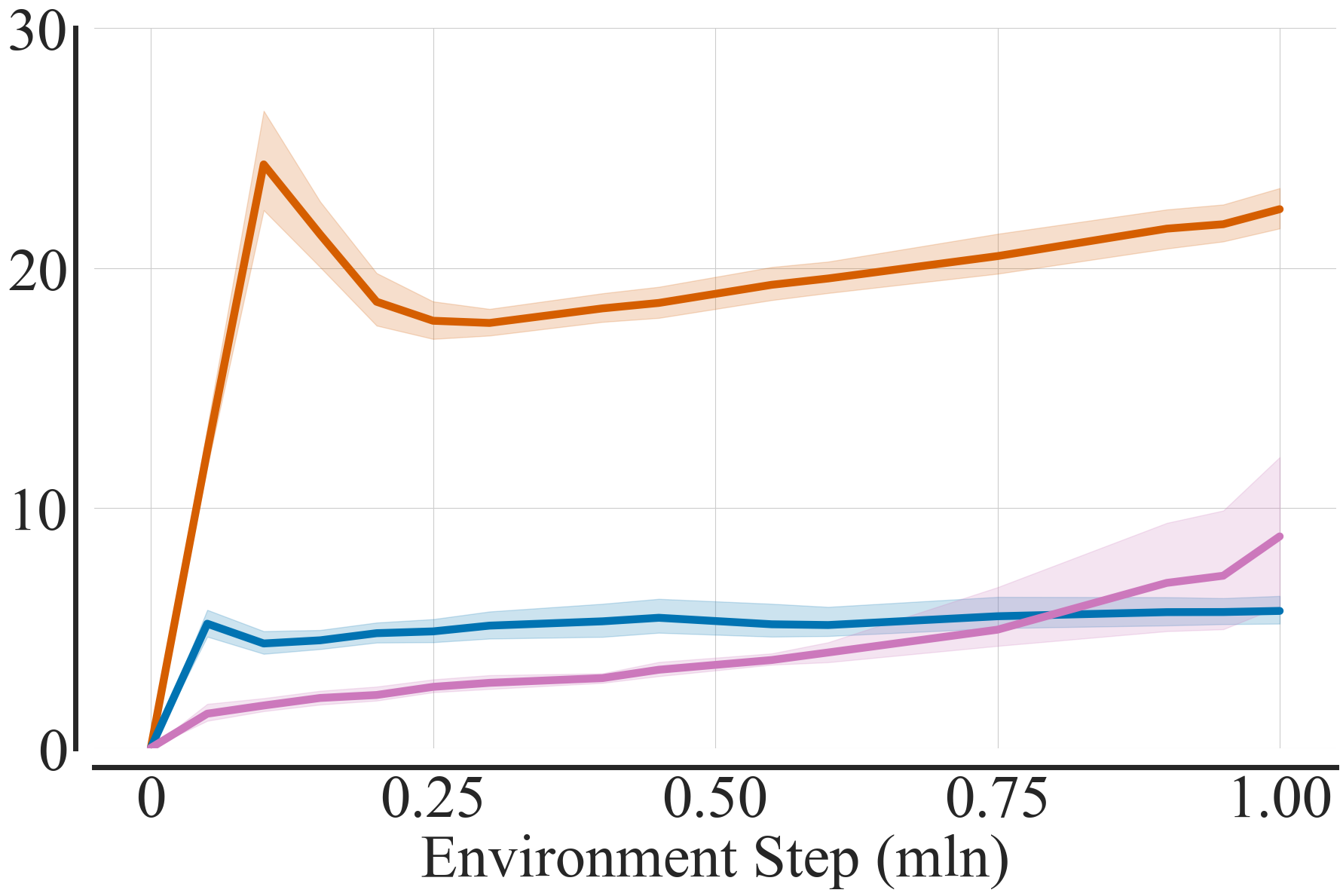}
    \hfill
    \end{subfigure}
\end{minipage}
\begin{minipage}[h]{0.245\linewidth}
    \begin{subfigure}{1.0\linewidth}
    \hfill
    \subcaption{KL Weight ($\tau$)}
    \label{fig:adjust3}
    \includegraphics[width=1.0\linewidth]{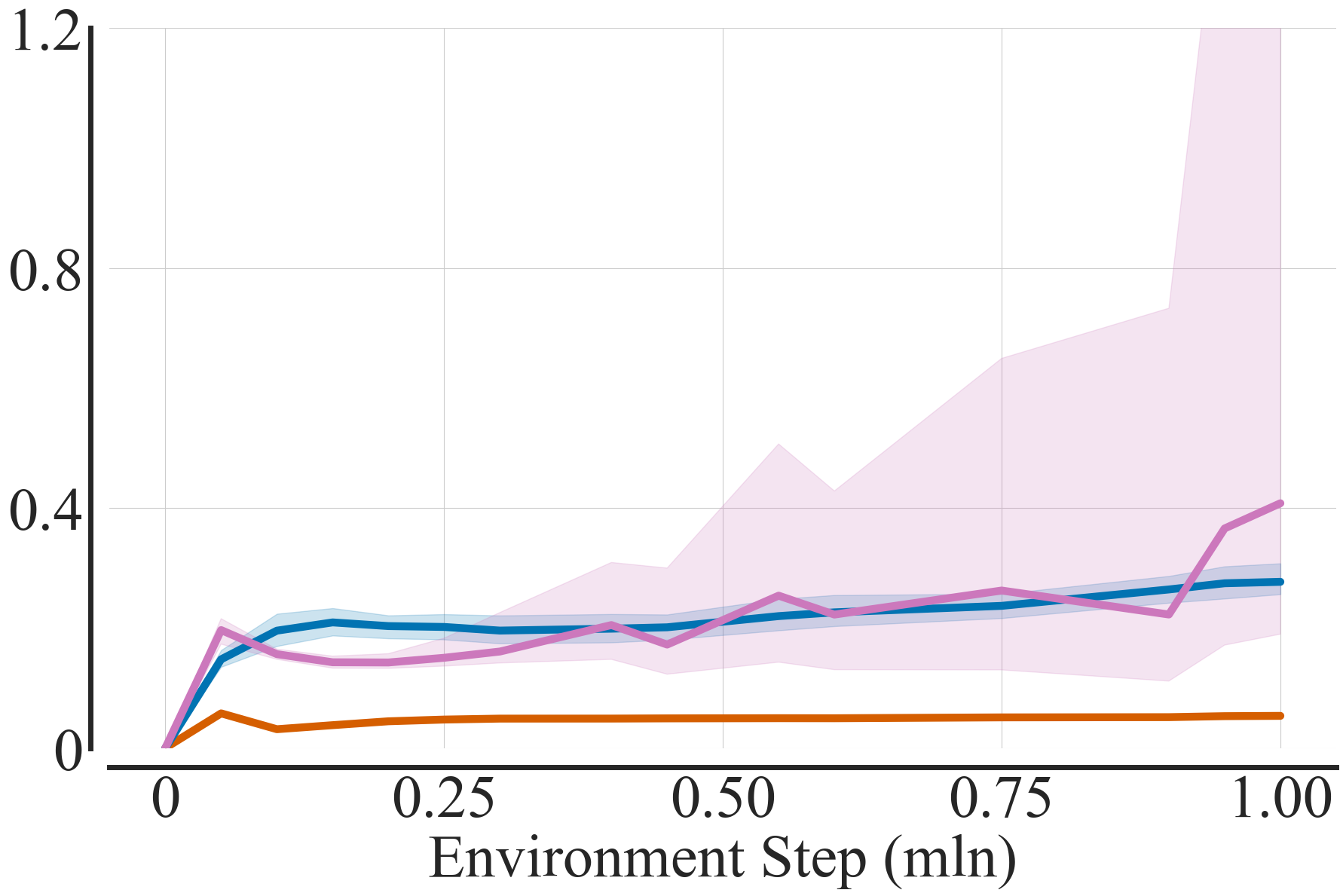}
    \hfill
    \end{subfigure}
\end{minipage}
\begin{minipage}[h]{0.245\linewidth}
    \begin{subfigure}{1.0\linewidth}
    \hfill
    \subcaption{Empirical KL}
    \label{fig:adjust4}
    \includegraphics[width=1.0\linewidth]{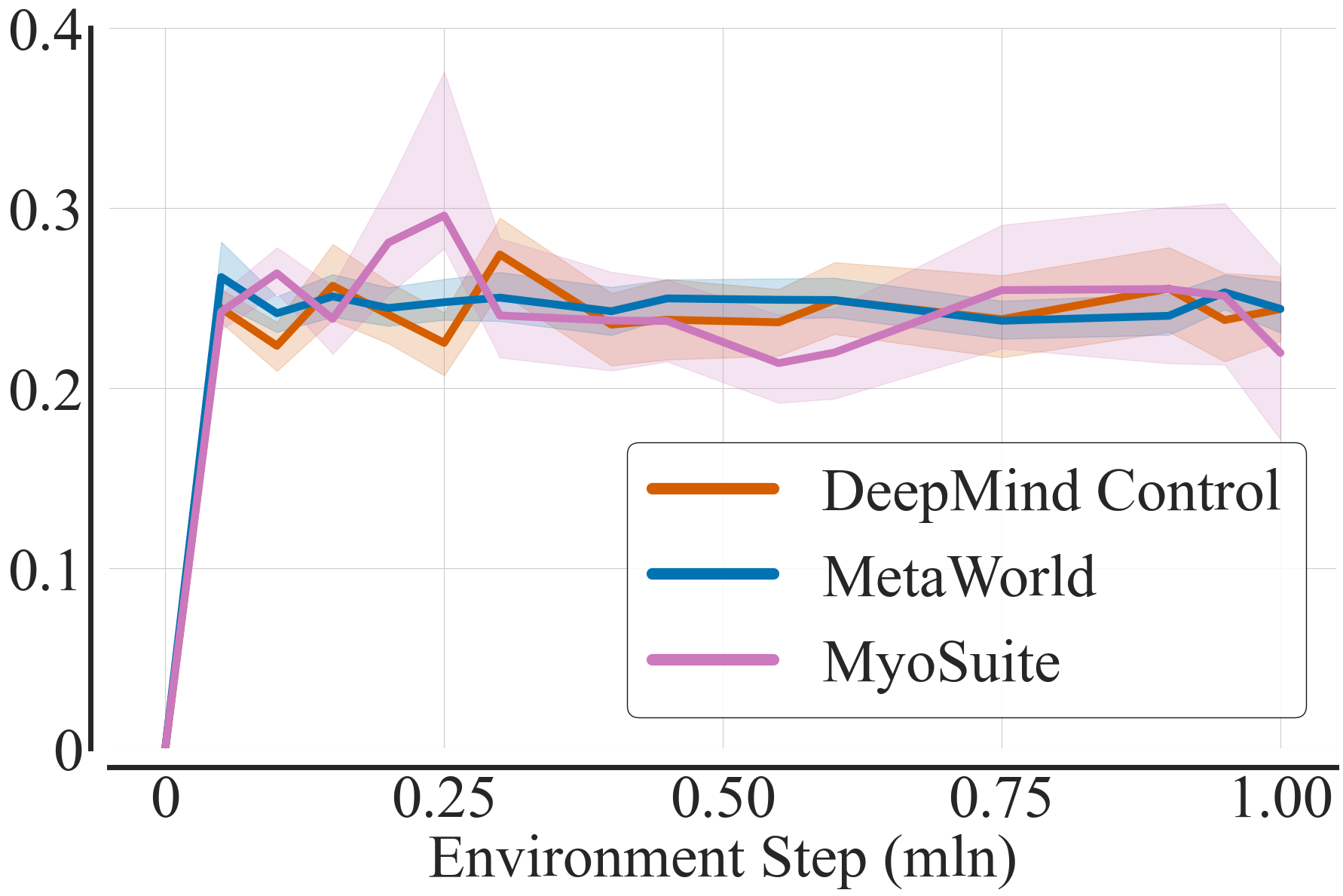}
    \hfill
    \end{subfigure}
\end{minipage}
\vspace{-0.2in} 
\caption{We report relative size of critic disagreement to critic means (\ref{fig:adjust1}), optimism (\ref{fig:adjust2}), KL penalty weight (\ref{fig:adjust3}), and the empirical KL between two actors (\ref{fig:adjust4}) during $1$mln environment steps training on tasks listed in Table \ref{tab:all_tasks}. DAC adjustment mechanism accommodates different scales of Q-values and allows for maintaining a predefined KL divergence between the two policies, despite significant differences between each task. $30$ tasks, $10$ seeds per task.}
\label{fig:adjustments}
\end{center}
\vspace{-0.1in} 
\end{figure}

\paragraph{Pessimistic Actor and Critic} The pessimistic actor is updated to maximize the utility:

\begin{equation}
\label{eq:upd_pessactor}
\begin{split}
    \mathcal{L}_{\phi} & = - \underset{s \sim \mathcal{D}}{\mathrm{E}} ~ \underset{i \sim \mathcal{V}}{\mathrm{E}}  e^{~2\beta^{p} V_{i}(s)} \approx \underset{s \sim \mathcal{D}}{\mathrm{E}} \bigl(\alpha \log\pi^{p}_{\phi}(a|s) - Q^{\mu}_{\theta}(s, a) - \beta^{p} Q^{\sigma}_{\theta}(s, a)\bigr) ~\text{with} ~ a \sim \pi^{p}_{\phi}(s).
\end{split}
\end{equation}

Above, $\mathcal{L}_{\phi}$ is the pessimistic actor loss, $\mathcal{D}$ is the experience buffer, $\alpha$ is the entropy temperature, $\beta^{p}$ is the pessimism parameter, $Q^{\mu}_{\theta}$ and $Q^{\sigma}_{\theta}$ denote the critic ensemble mean and standard deviation respectively. As shown in Theorem \ref{theorem:1}, optimizing for the expected utility objective under an exponential utility function leads to a risk-aware value. As such, the pessimistic actor convergences to the optimal policy under CDQL assumptions \citep{fujimoto2018addressing}. The critic parameters are updated via SARSA under the pessimistic policy:

\begin{equation}
\label{eq:upd_critic}
    \mathcal{L}_{\theta} = \underset{T \sim \mathcal{D}}{\mathrm{E}} \bigl(Q_{\theta} (s,a) - (r + \gamma Q^{\beta^{p}}_{\Bar{\theta}}(s',a') - \alpha \log\pi^{p}_{\phi}(a'|s') \bigr)^{2} ~~~~ \text{with} ~~~~ a' \sim \pi^{p}_{\phi}(s').
\end{equation}

Here, $\mathcal{L}_{\theta}$ is the critic loss function which is optimized off-policy on the dataset of previous experiences $\mathcal{D}$ with $T=(s,a,r,s')$, and $\Bar{\theta}$ denotes the target network parameters which are updated via standard Polyak averaging \citep{fujimoto2018addressing, haarnoja2018soft}. This setup aligns the objectives of the critic and the pessimistic actor in DAC with the traditional pessimistic approaches like SAC. 

\paragraph{Optimism and KL penalty adjustment} The goal of KL regularization used by the optimistic actor is to achieve divergence small enough to ensure good coverage of the pessimistic policy in exploration data, yet large enough to maintain the optimistic policy's exploration effectiveness. In practice, the level of KL is influenced by three factors: the KL penalty weight $\tau$, the difference in risk appetites between the actors $(\beta^{o} - \beta^{p})$, and the critic ensemble disagreement $Q_{\theta}^{\sigma}(s,a)$, which varies based on reward scale and environment uncertainty. To achieve a balanced divergence between the two actors, we allow gradient-based adjustments in $\tau$ and $(\beta^{o} - \beta^{p})$, while assuming a fixed $\beta^{p}$:

\begin{equation}
\label{eq:optimism}
    \mathcal{L}_{\beta^{o}} = \underset{s \sim \mathcal{D}}{\mathrm{E}} (\beta^{o} - \beta^{p}) D(s) ~~ \text{for} ~~ \beta^{o}\in(\beta^{p},\infty), ~~~ \text{and} ~~~ \mathcal{L}_{\tau} = \underset{s \sim \mathcal{D}}{\mathrm{E}} -\tau D(s) ~~ \text{for} ~~ \tau\in(0,\infty).
\end{equation}

Above, $\mathcal{L}_{\beta^{o}}$ and $\mathcal{L}_{\tau}$ represent the loss functions of the optimism and the KL penalty weight respectively. Where $D(s)$ represents the discrepancy between the recorded and target KL divergence:

\begin{equation}
    D(s) = \frac{1}{|A|} KL\bigl(\pi_{\phi}^{p}(s)|\pi_{\eta}^{o}(s)\bigr) - \mathcal{KL}^{*}.
\end{equation}

The mechanism adjusts $\beta^{o}$ and $\tau$ based on the distance between the empirical and target KL divergences. When the empirical divergence exceeds the target, $\beta^{o}$ decreases to a limit of $\beta^{p}$, and $\tau$ increases. Conversely, a smaller empirical divergence than the target leads to an increase in $\beta^{o}$ and a decrease in $\tau$. This dual adjustment allows DAC to regulate the divergence between two policy actors even when $\beta^{o}$ reaches its lower bound. The optimization objectives for $\tau$ and $\beta^{o}$ are thus formulated to adapt to varying conditions, ensuring efficient exploration and exploitation balance in different environments. We test the effectiveness of these adjustment mechanisms in Table \ref{tab:design_ablation}, and depict the different levels of optimism achieved in various tasks in Figure \ref{fig:adjustments}. We expand our discussion of DAC in Appendix \ref{appendix:dac}, as well as present pseudocode in Figure \ref{fig:pseudocode}. 

\section{Experiments}
\label{section:experiments}

Our experimental framework is based on JaxRL \citep{jaxrl}. We assess the performance of DAC across a diverse set of over 30 locomotion and manipulation tasks listed in Table \ref{tab:all_tasks}, sourced from the DeepMind Control (DMC) \citep{tassa2018deepmind}, MetaWorld (MW) \citep{yu2020meta} and MyoSuite (MYO) benchmarks. In DMC we report the returns, whereas in MW and MYO we report the success rates. We calculate robust evaluation statistics using RLiable \citep{agarwal2021deep}. 

\begin{figure}[ht!]
\begin{center}
\begin{minipage}[h]{1.0\linewidth}
    \begin{subfigure}{1.0\linewidth}
    \hfill
    \includegraphics[width=1.0\linewidth]{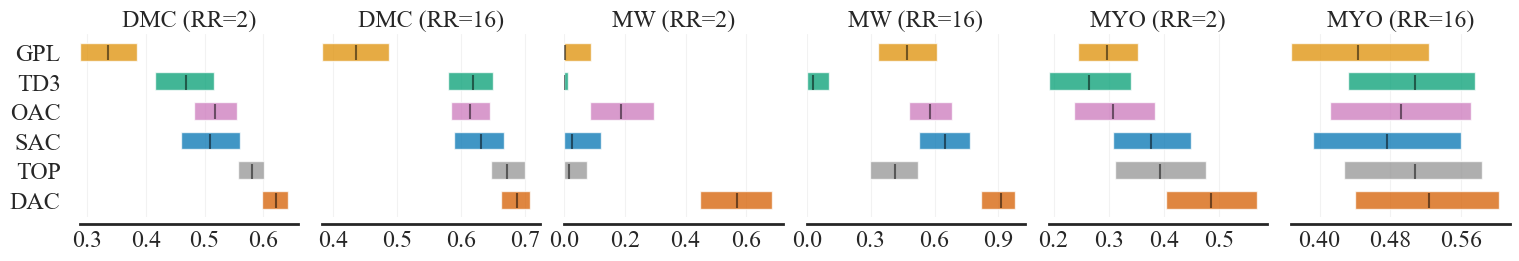}
    \hfill
    \end{subfigure}
\end{minipage}
\vspace{-0.2in} 
\caption{We report final IQM in $30$ tasks (Table \ref{tab:all_tasks}) in low ($RR=2$) and high replay settings ($RR=16$). $1.0$ denotes the maximal possible score, $95$\% CI calculated with $10$ random seeds.}
\label{fig:modelfree}
\end{center}
\vspace{-0.1in} 
\end{figure}

\paragraph{Model-free benchmark}

We test DAC against an array of both risk-neutral and risk-aware baselines, including TD3 \citep{fujimoto2018addressing}, SAC \citep{haarnoja2018soft}, OAC \citep{ciosek2019better}, TOP \citep{moskovitz2021tactical}, and GPL \citep{cetin2023learning}. We align our experiments with the state-of-the-art SAC implementation, standardizing the common hyperparameters across all algorithms \citep{d2022sample}. We employ uniform network architectures and ensemble of two critics, as advocated in previous work \citep{fujimoto2018addressing, haarnoja2018soft, ciosek2019better, moskovitz2021tactical, cetin2023learning}. Importantly, by using uniform network architectures and hyperparameters, we ensure that the performance differences between algorithms stem solely from the different risk-preferences in tackling the exploration-exploitation dilemma. Each of $30$ tasks is run for $1$mln environment steps. We investigate two replay regimes: a compute-efficient regime, involving $2$ gradient updates per environment step, and a sample-efficient regime, using $16$ gradient updates per step with full-parameter resets every $160k$ environment steps \citet{d2022sample}. As evidenced by Figure \ref{fig:sample_efficiency}, DAC particularly excels in the earlier phases of the training, achieving final performance of the best baseline in only $60$\% of environment steps. Furthermore, as shown in Figure \ref{fig:modelfree}, DAC approach to balancing risk-averse exploitation with risk-loving exploration yields significant performance benefits, particularly visible in the low replay ratio setting. 

\paragraph{Dog and Humanoid tasks}

\begin{figure}[t!]
\begin{center}
\begin{minipage}[h]{1.0\linewidth}
    \begin{subfigure}{1.0\linewidth}
    \includegraphics[width=0.325\linewidth]{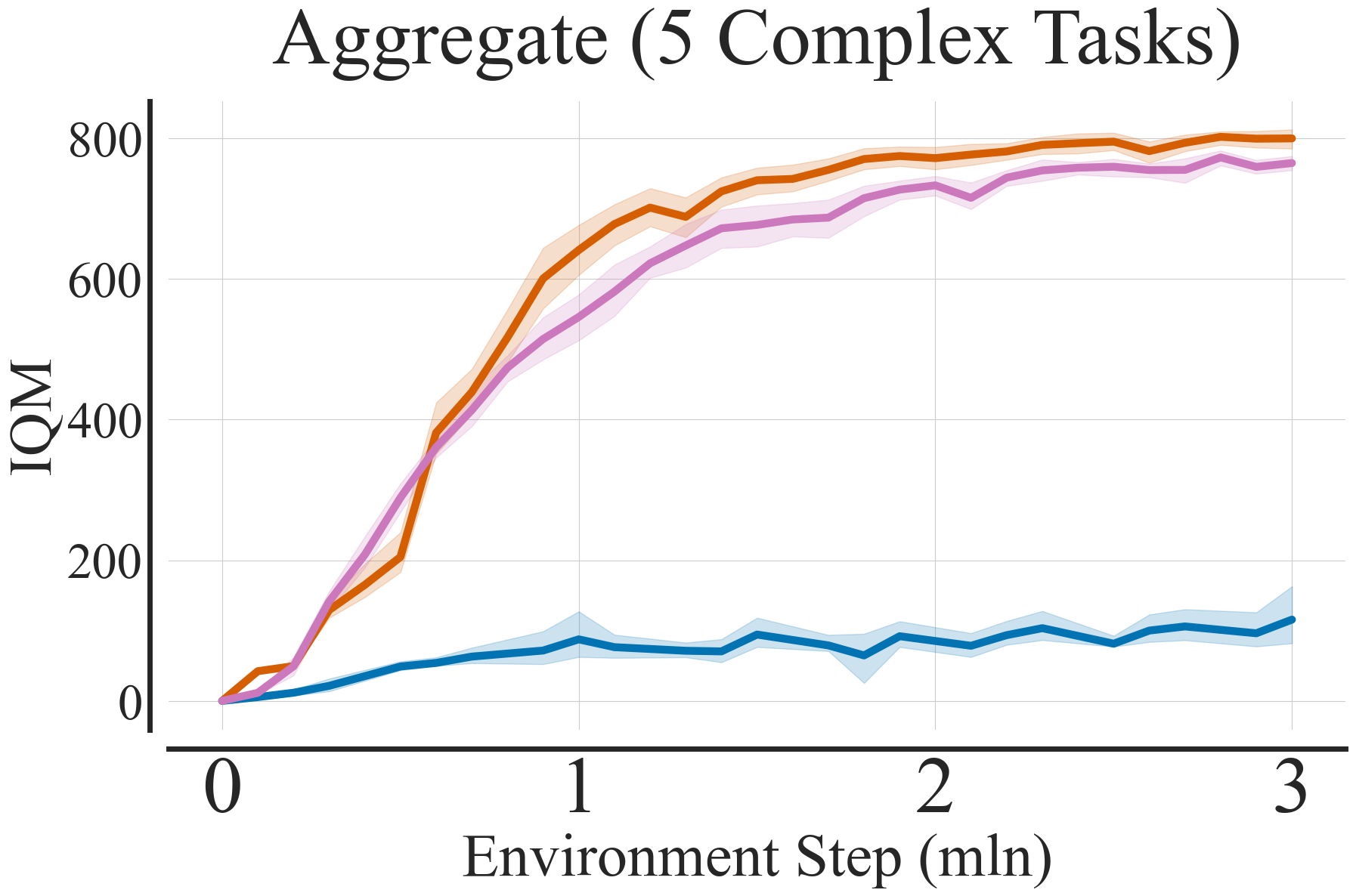}
    \hfill
    \includegraphics[width=0.325\linewidth]{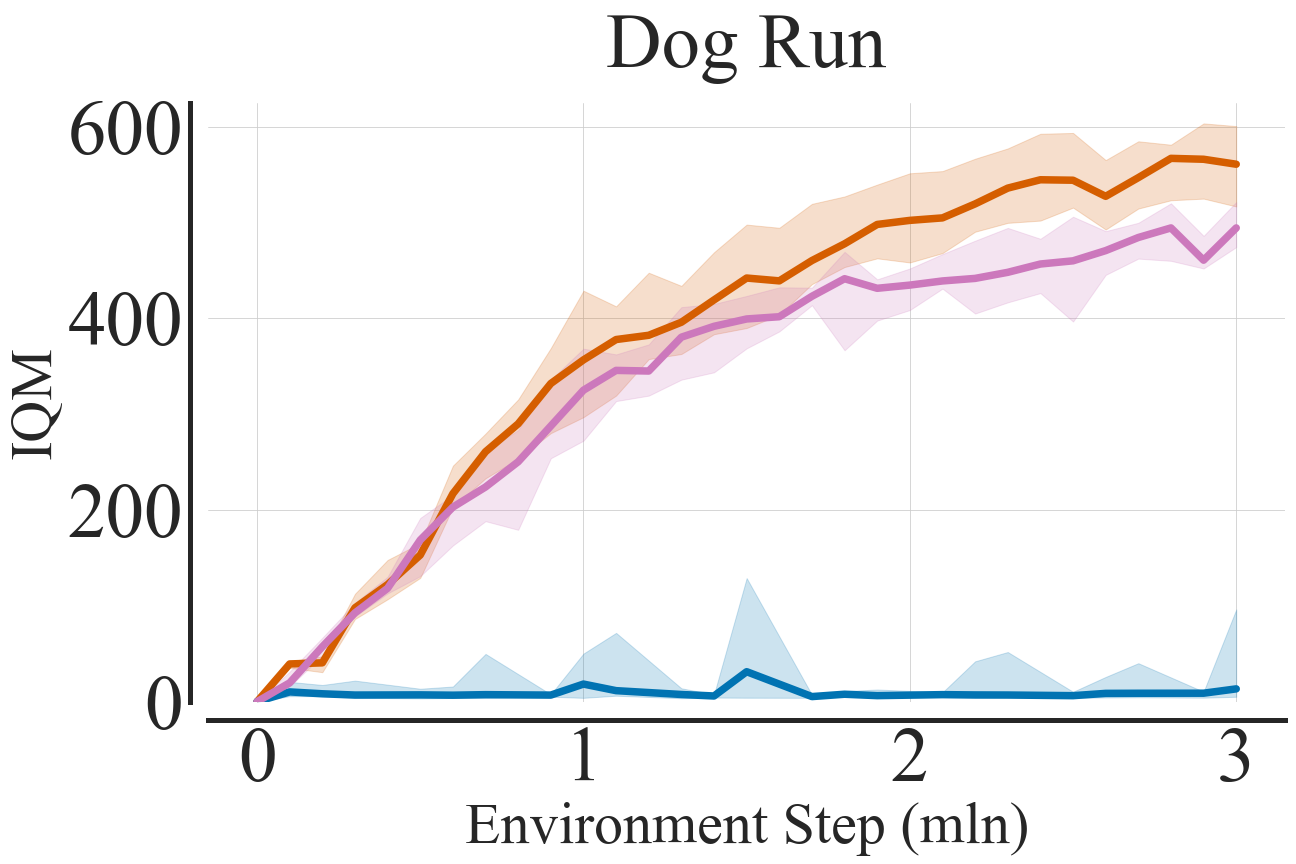}
    \hfill
    \includegraphics[width=0.325\linewidth]{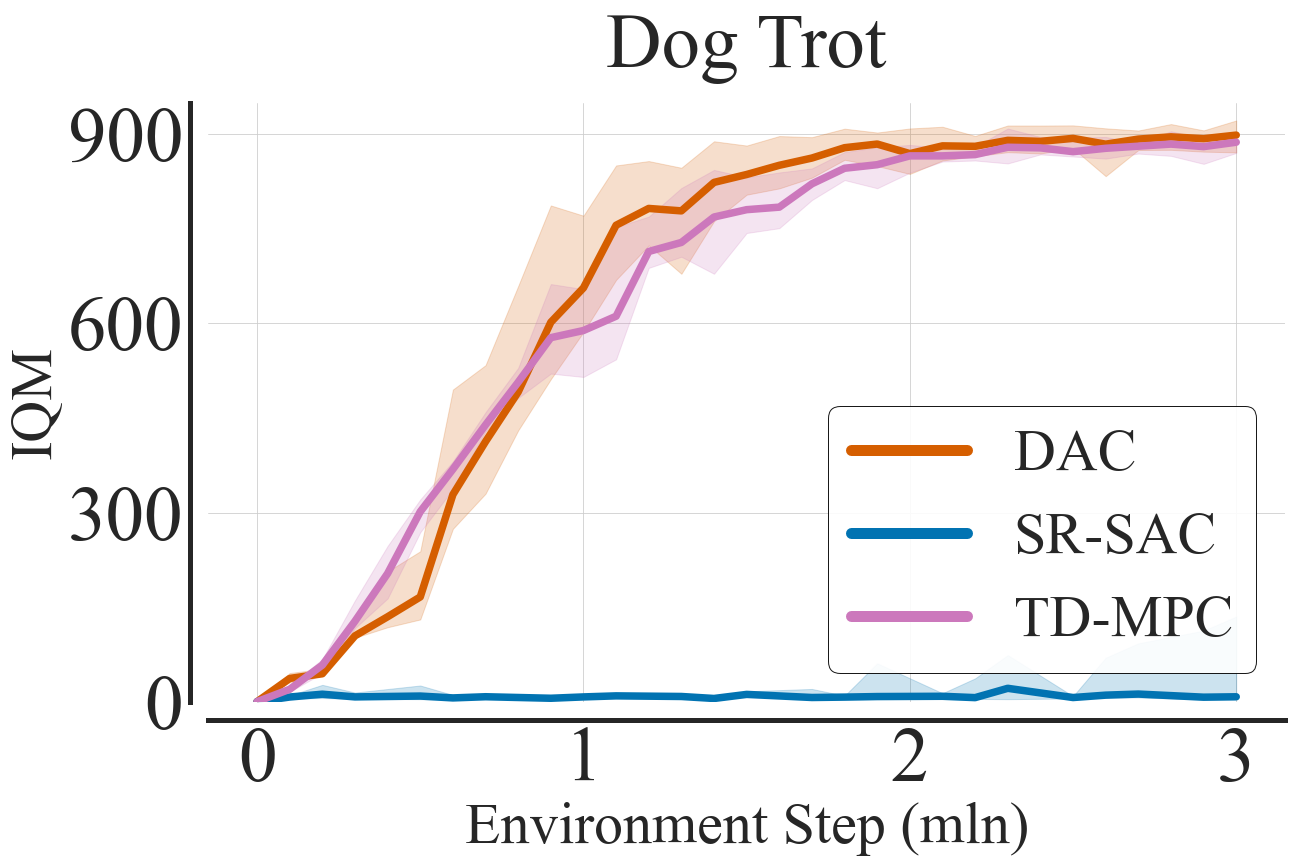}
    \end{subfigure}
\end{minipage}
\caption{We compare model-free DAC, model-free SR-SAC to model-based TD-MPC. $Y$-axis reports IQM, and $X$-axis denotes environment steps. $3$mln environment steps, $5$ seeds.}
\label{fig:hardtasks}
\end{center}
\vspace{-0.1in} 
\end{figure}

Furthermore, we assess the performance of DAC when training for longer than $1$mln environment steps. To this end, we choose $4$ challenging locomotion tasks, two from the Dog domain (run, trot) and two from the Humanoid domain (run, walk), where we train for $3$mln environment steps. The dog domain is particularly notable for its complexity, with only few methods showing non-random performance \citep{hansen2022temporal, xu2023drm}. Here, we compare DAC against compute-heavy, sample-efficient baselines: SR-SAC with $32$ gradient steps per environment step \citep{d2022sample}; and TD-MPC, a model-based approach that leverages a learned world model to perform trajectory optimization \citep{hansen2022temporal}. In particular, TD-MPC has been documented to achieve effective policies in both dog and humanoid domains \citep{hansen2022temporal}, however this performance comes at a significant compute cost, with TD-MPC using around 3 times more learnable parameters than DAC. To partially offset this disparity, we modified DAC critic to mimic the critic network used in TD-MPC, by adding a single hidden layer and layer normalization \citep{ball2023efficient}. As presented in Figure \ref{fig:hardtasks}, despite being a model-free approach, DAC achieves competitive performance in the dog domain, marginally surpassing the model-based TD-MPC. 

\begin{wraptable}[15]{r}{0.5\textwidth}
\vspace{-0.5cm}
\caption{We evaluate the performance impact of DAC components. $15$ tasks, $500k$ steps, $10$ seeds.}
{\renewcommand{\arraystretch}{1.12}
\begin{tabular}{l|cc}
\toprule 
               & $RR=2$ & $RR=16$  \\\midrule
\textsc{Only} $\pi_{\eta}^{o}$ & 0.08 & 0.17 \\
\textsc{No KL Reg} & 0.42 & 0.77 \\
\textsc{Deterministic} $\pi_{\eta}^{o}$ & 0.82 & 0.77 \\
\textsc{Deterministic} $\pi_{\phi}^{p}$ & 0.92  &  0.79 \\
\textsc{No Adjustments} & 0.91  &  0.87 \\
\textsc{No} $\tau$ \textsc{Adjustment} & 0.95  &  0.93 \\
\textsc{No} $\beta^{o}$ \textsc{Adjustment} & 0.84  &  0.99 \\\midrule
\textsc{Base DAC} & 1.00  &  1.00
\end{tabular}}
\label{tab:design_ablation}
\end{wraptable}

\paragraph{Design decisions}

We explore how DAC performance is influenced by variations in its design. We conduct evaluations of various DAC modifications in $15$ tasks listed in Table \ref{tab:all_tasks}, where we train for $500k$ environment steps and consider both replay ratios. We assess the performance of the following variations: \textsc{Only} $\pi_{\eta}^{o}$ where we use the optimistic actor for both exploration and exploitation; \textsc{No KL Reg} where we do not use KL regularization for the optimistic actor; \textsc{Deterministic} $\pi_{\eta}^{o}$ where we set the optimistic actor to be deterministic; \textsc{Deterministic} $\pi_{\phi}^{p}$ where we set the pessimistic actor to be deterministic; \textsc{No Adjustments} where we disable the DAC adjustment mechanisms; \textsc{No} $\tau$ \textsc{Adjustment} where we disable $\tau$ adjustment; and \textsc{No} $\beta^{o}$ \textsc{Adjustment} where we disable $\beta^{o}$ adjustment. As shown in Table \ref{tab:design_ablation}, we find that all of these design choices bring significant gains to DAC performance. In particular, we observe that the dual policy setup allows for effective use of optimism, as evidenced by poor performance of the \textsc{Only} $\pi_{\eta}^{o}$ agent. Similarly, we find that KL regularization yields significant improvements, most likely due to limiting the tandem problem.

\paragraph{Hyperparameter Sensitivity} 

\begin{figure}[t!]
\begin{center}
\begin{minipage}[h]{1.0\linewidth}
    \begin{subfigure}{1.0\linewidth}
    \includegraphics[width=0.32\linewidth]{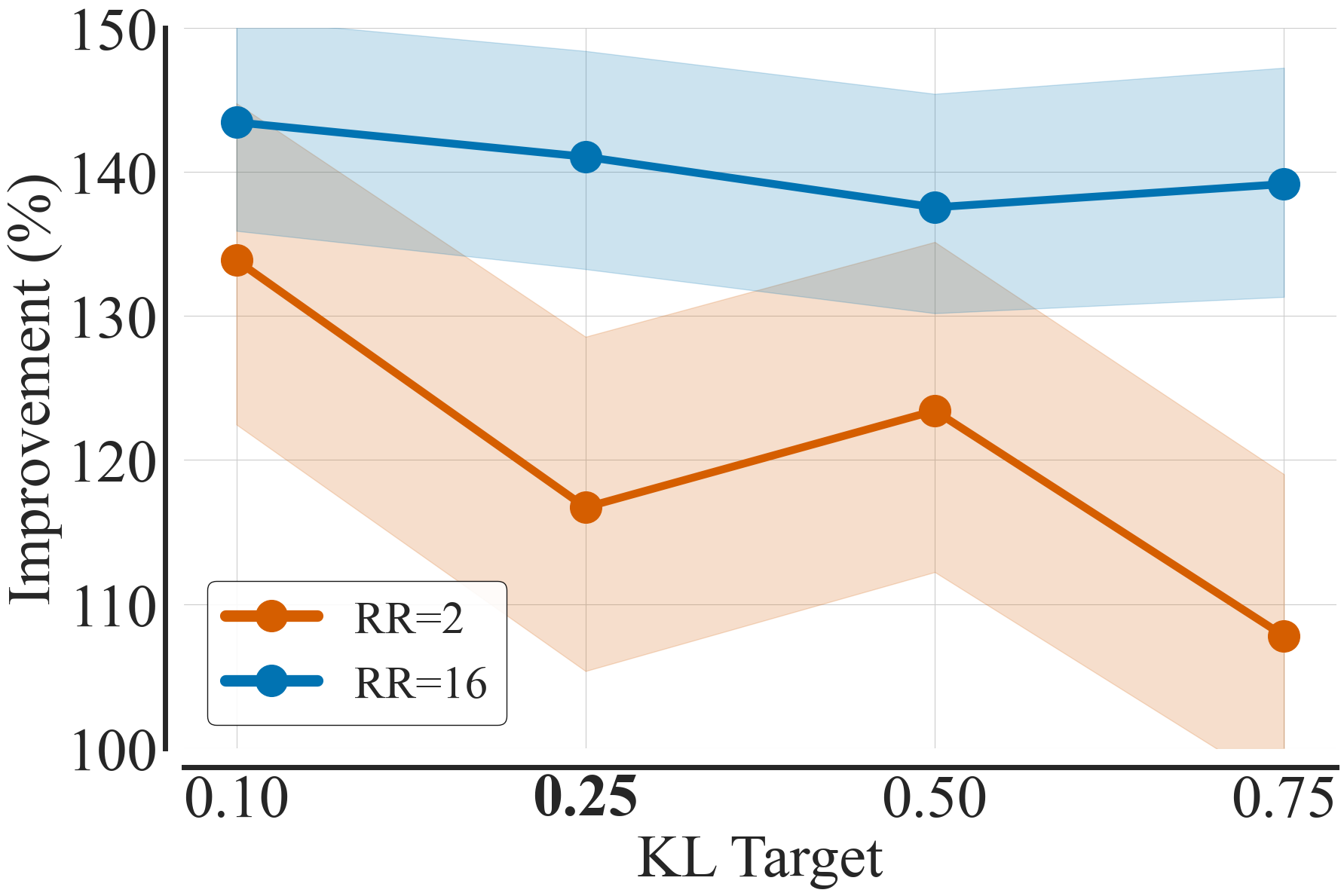}
    \hfill
    \includegraphics[width=0.32\linewidth]{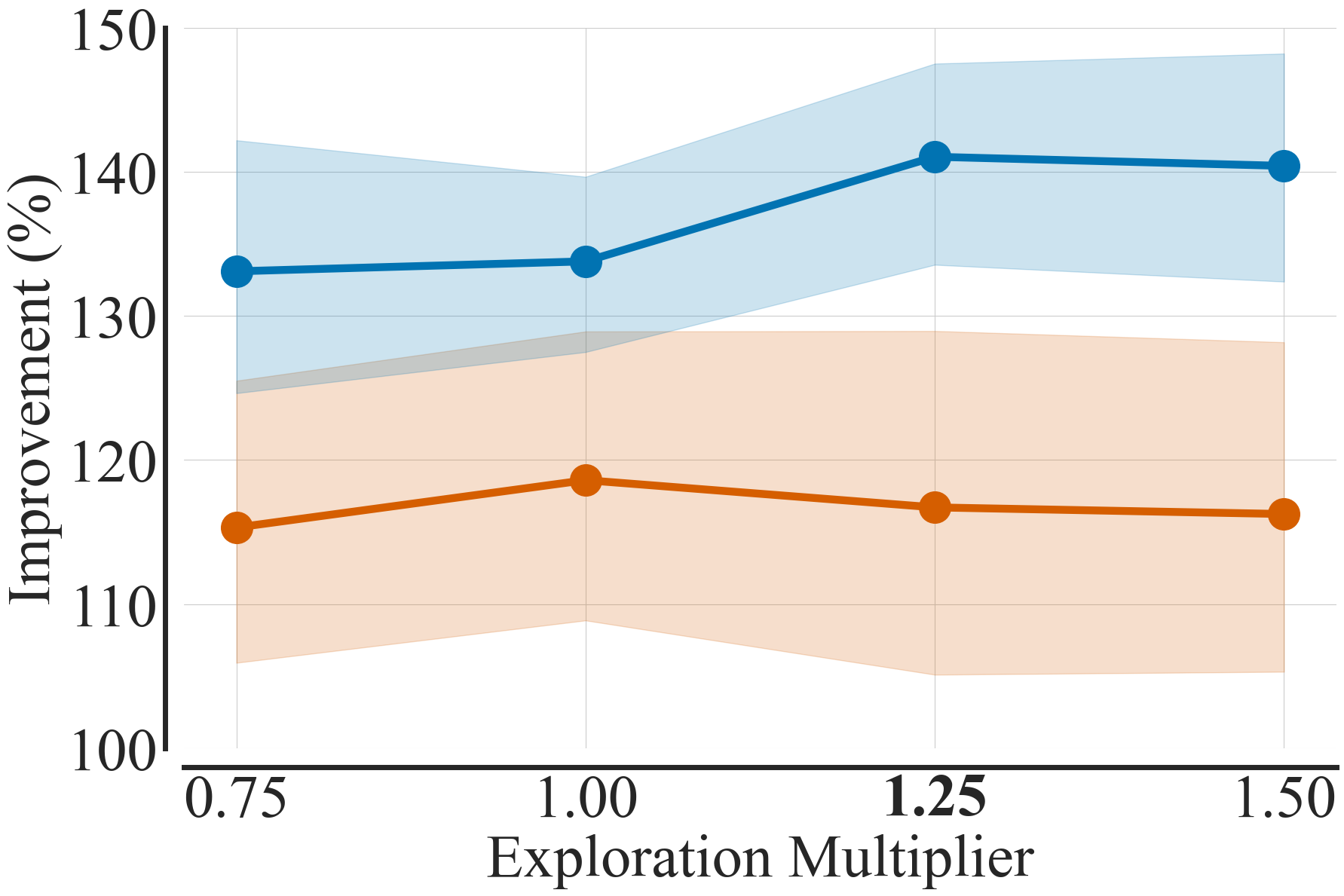}
    \hfill
    \includegraphics[width=0.32\linewidth]{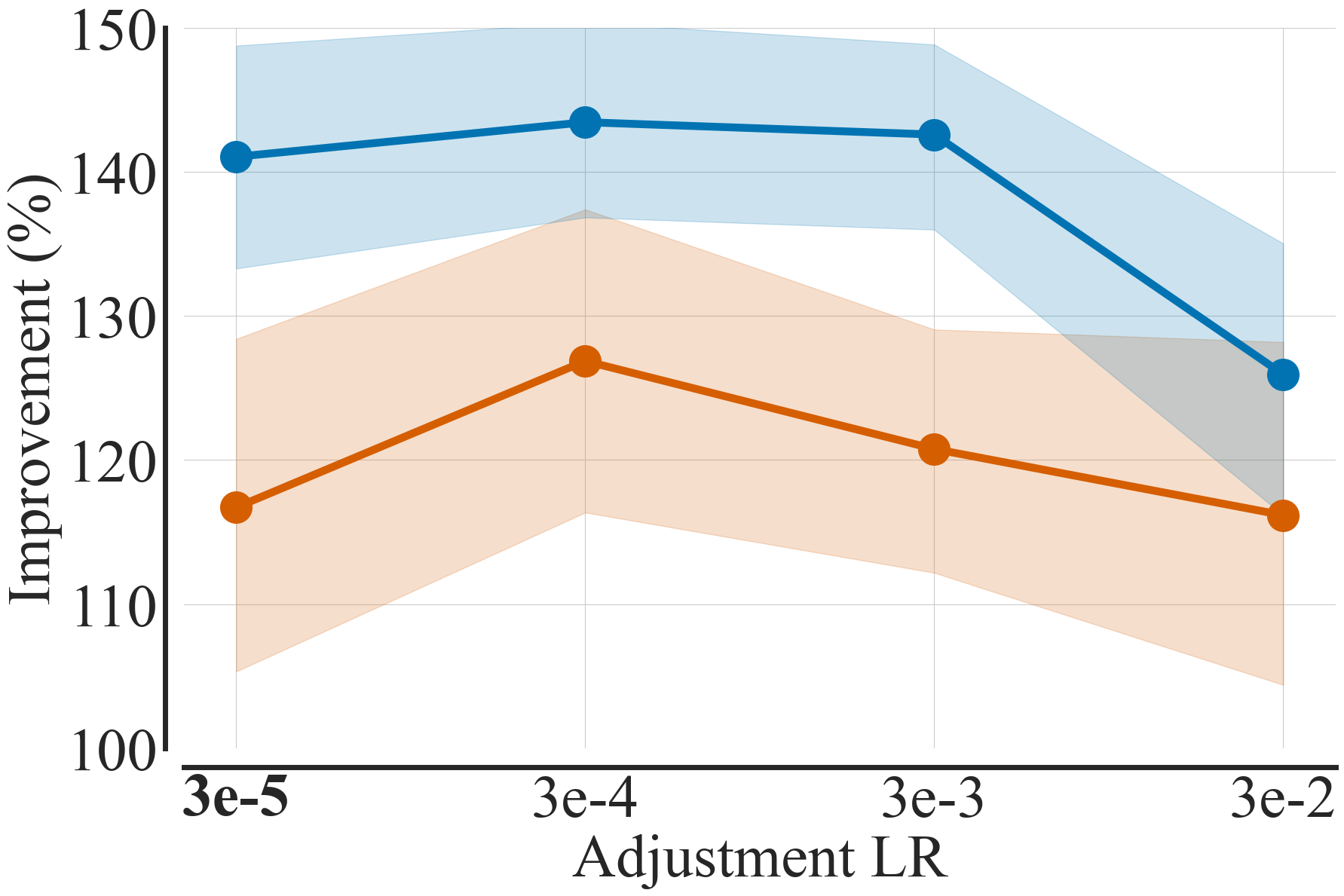}
    \end{subfigure}
\end{minipage}
\caption{We evaluate DAC performance when changing the values of its hyperparameters ($X$-axis) for two replay ratio regimes. The bold value denotes the value used in the main experiment. $Y$-axis reports the percentage improvement over tuned SAC. $15$ tasks, $500k$ steps, $10$ seeds per task.}
\label{fig:hyperparams}
\end{center}
\vspace{-0.1in} 
\end{figure}

As discussed in Section \ref{section:dac}, DAC introduces three new hyperparameters: KL divergence target, exploration multiplier, as well as the learning rates for the adjustment mechanism for $\beta^{o}$ and $\tau$. Here, we assess DAC sensitivity when changing the values of these hyperparameters. To this end, we train $12$ DAC agents (each with a different hyperparameter setting) on $500k$ environment steps, in both replay ratio regimes, and on tasks listed in Table \ref{tab:all_tasks}. We report these results In Figure \ref{fig:hyperparams}, where we compare the final performance of these agents against SAC with analogical replay ratio. As follows, we find that DAC performance is stable when changing the values of its hyperparameters, with all tested variations outperforming the tuned SAC baseline. 

\paragraph{Additional Experiments} In Appendix \ref{app:additional_exp}, we detail supplementary experiments conducted to further evaluate DAC. Firstly, we assess the effectiveness of layer normalization when used with DAC. This regularization technique was shown to improve the performance of RL agents on a variety tasks \citep{hiraoka2021dropout, li2022efficient, lyle2023understanding}. We find that using layer normalization further improves DAC performance on locomotion tasks, and detail the results in Figure \ref{fig:layernorm}. Furthermore, we evaluate the compute costs associated with running a Dual Actor setup proposed by DAC. We find that DAC requires around $15$\% additional wallclock time to complete 1mln environment steps training as compared to SAC. However, as shown in Figure \ref{fig:wallclock}, DAC performance improvements offset the additional compute costs, leading to a beneficial trade-off between compute and performance. In Figure \ref{fig:overestimation}, we find that the dual architecture indeed mitigates the overestimation associated with an optimistic TD update. Finally, in Figures \ref{fig:distributional1} \& \ref{fig:distributional2}, we investigate the performance of distributional DAC that calculates upper-bound with respect to different types of uncertainties. 
 
\section{Limitations}
\label{sec:limitations}

Main limitation of DAC is its dual-actor framework, which incurs slightly higher memory and computation costs relative to the standard SAC. In our implementation, DAC wallclock is around $15$\% longer than SAC/TD3. Furthermore, using the extra actor network requires additional memory. In our implementation, the usage of optimistic actor translates to a $20\%$ increased GPU memory dedicated to models as compared to SAC. Finally, DAC implementation requires setting three additional hyperparameters. Although tests showed DAC robust performance across a variety of hyperparameter values, it is uncertain whether this robustness translates to more complex environments.

\section{Conclusions}

In this work, we apply the expected utility theorem to reason about the empirical effectiveness of risk-aware actor-critic algorithms. In particular, we demonstrate that policies derived from the commonly used pessimistic objective are approximately utility-optimal when aligned with an exponential utility function. We think that the proposed framework is interesting as it revisits the foundational decision-theoretic perspective of RL — namely, optimization within the utility space rather than the space of utility inputs. Furthermore, we present DAC, a risk-aware actor-critic framework. DAC employs two actors with distinct risk appetites to optimize their respective expected utility functions. The pessimistic actor focuses on TD learning and evaluation, while the optimistic actor facilitates exploration. We evaluate DAC on various locomotion and manipulation tasks and compare it with over ten baseline algorithms. We find that DAC demonstrates significant performance improvements relative to other model-free methods and is competitive with leading model-based approaches. Finally, we investigate DAC performance robustness across diverse hyperparameter configurations and find that the experiments affirm its practical applicability.


\bibliography{bib.bib}
\bibliographystyle{bibstyle}

\appendix

\newpage

\subsubsection*{Broader Impact}

In this paper, we analyze the pessimistic RL algorithms through the expected utility theorem and introduce a sample-efficient RL algorithm inspired by microeconomic principles. This approach to integrating behavioral theories into RL presents significant opportunities to improve agents. However, it also introduces a reverse application: using RL insights to inform microeconomic understanding and human behavior. This latter brings important ethical considerations. Applying economic theories to RL can enhance machine decision-making to more closely resemble human judgment. However, the consequences of machines operating in roles traditionally occupied by humans must be carefully analyzed. Similarly, applying RL principles to microeconomics offers perspectives on human decision-making and economic models, but should be done cautiously to avoid oversimplifying the human behavior and societal interactions. 

\section{Derivations}
\label{appendix:derivations}

We consider a single step of policy evaluation and improvement. We assume the value approximations $V^{i}(s)$ to be derived from a distribution $\mathcal{V}(s)$. This distribution is characterized by finite moments and an expected value $V^{\mu}(s)$:

\begin{equation}
     V^{\mu}(s) = \underset{i \sim \mathcal{V}}{\mathrm{E}} V_{i}(s), \quad \text{and} \quad \epsilon_{i}(s) = V^{i}(s) - V^{\mu}(s).
\end{equation}

Above, $\epsilon_{i}(s)$ is the deviation of value approximation from the mean, which per definition has expected value of zero (ie. $\mathrm{E}_{i \sim \mathcal{V}} ~ \epsilon_{i}(s) = 0$). The certainty value is defined by:

\begin{equation}
    V^{c}(s) = \mathcal{I} ~ \underset{i \sim \mathcal{V}}{\mathrm{E}} ~ \mathcal{U} ~ V_{i}(s) = V^{\mu}(s) + \Upsilon(s).
\end{equation}

Where $\Upsilon(s)$ is the risk premium. The pessimistic value is defined as:

\begin{equation}
    V^{\beta}(s) = \underset{i \sim \mathcal{V}}{\mathrm{E}} V_{i}(s) + \beta ~\underset{i \sim \mathcal{V}}{\mathrm{E}} \bigl(V_{i}(s) - V^{\mu}(s) \bigr)^2
\end{equation}

Thus, the pessimistic value is equal to the certainty equivalent value if, for a given utility function, it is true that $\Upsilon(s) = \beta ~ V^{\sigma}(s)$. The exponential utility function is defined as:

\begin{equation}
    \mathcal{U}(V_{i}(s)) = e^{2\beta V_{i}(s)}
\end{equation}

We start the derivation by applying the exponential utility to the certainty equivalent value definition:

\begin{equation}
\begin{split}
     e^{2\beta V^{c}(s)} & = \underset{i \sim \mathcal{V}}{\mathrm{E}} ~ e^{2\beta V_{i}(s)} = \underset{i \sim \mathcal{V}}{\mathrm{E}} ~ e^{2\beta (V^{\mu}(s) + \epsilon_{i}(s))} = \underset{i \sim \mathcal{V}}{\mathrm{E}} ~ e^{2\beta V^{\mu}(s)} e^{2\beta \epsilon_{i}(s))} \\
     & = e^{2\beta V^{\mu}(s)} \underset{i \sim \mathcal{V}}{\mathrm{E}} ~ e^{2\beta \epsilon_{i}(s))}
\end{split}
\end{equation}

Inverting the utility function on the left-hand side (LHS) leads to:

\begin{equation}
\begin{split}
    V^{c}(s) = V^{\mu}(s) + \frac{1}{2\beta} \ln \underset{i \sim \mathcal{V}}{\mathrm{E}} ~ e^{2\beta \epsilon_{i}(s)} \implies \Upsilon(s) = \frac{1}{2\beta} \ln \underset{s \sim p} {\mathrm{E}} ~ \underset{i \sim \mathcal{V}}{\mathrm{E}} ~ e^{2\beta \epsilon_{i}(s)}
\end{split}
\end{equation}

Given that $\epsilon^{i}(s)$, by definition, centers around zero, we use the Maclaurin series which is convergent when applied to an exponential function. We use a second order approximation:

\begin{equation}
    \Upsilon(s) \approx \frac{1}{2\beta} \ln \underset{i \sim \mathcal{V}}{\mathrm{E}} ~ \bigl(1 + 2\beta \epsilon_{i}(s) + 2\beta^{2} \epsilon_{i}(s)^{2} \bigr) = \frac{1}{2\beta} \ln \underset{i \sim \mathcal{V}}{\mathrm{E}} ~ \bigl(1 + 2\beta^{2} \epsilon_{i}(s)^{2} \bigr) 
\end{equation}

Here, $\epsilon^{i}(s)^{2}$ represents the variance of the model distribution at state $s$. Thus, for optimization purposes, the approximate equality $\nabla V^{c}(s) \approx \nabla V^{\beta}(s)$ is practically established. To calculate the values, not just their gradients, we expand the logarithm function using first-order Maclaurin series. Again, the series are convergent since $1 + 2\beta^{2} \epsilon_{i}(s)^{2} \geq 1$:

\begin{equation}
     \Upsilon(s) \approx \frac{1}{2\beta} \underset{i \sim \mathcal{V}}{\mathrm{E}} ~ 2\beta^{2} \epsilon_{i}(s)^{2} = \beta \underset{i \sim \mathcal{V}}{\mathrm{E}} (V_{i}(s) - V^{\mu}(s))^{2}
\end{equation}

Which concludes the derivation. Thus, the pessimistic correction arises from the second-order Maclaurin approximation of the model risk's value risk premium. For completeness, we calculate the approximation error for the approximation of the exponential function, which we denote as $\mathcal{E}$:

\begin{equation}
    \mathcal{E} = \sum_{n=3}^{\infty} \underset{i \sim \mathcal{V}}{\mathrm{E}} \frac{(2\beta \epsilon_{i}(s))^{n}}{n!}
\end{equation}

\paragraph{Additional comment} We note that the approach outlined in this Section can be adapted to address uncertainties inherent to a range of RL problems. An interesting example is the estimation of values through single evaluations of the critic. In such scenarios, adopting a risk-loving utility function can lead to objectives resembling Maximum Entropy (MaxEnt) principles. This suggests a potential avenue for integrating other risk preferences into actor-critic like setting. Moreover, the proof for the result, as detailed, can be considerably streamlined when focusing on gradients with respect to value functions in the context of policy optimization. Specifically, the approximation of the logarithm can be abandoned when the interest lies in optimizing policies rather than the values themselves. 

\section{Related Work}
\label{appendix:related}

\subsection{Risk-Awareness Exploration}

The exploration-exploitation dilemma has been the subject of extensive research. One prominent principle that has emerged in addressing this dilemma is Optimism in the Face of Uncertainty (OFU) \citep{auer2002finite, filippi2010optimism, ciosek2019better}, which prioritizes actions with a balance of high expected rewards and uncertainty. Whereas OFU has been extensively studied in the tabular and bandit RL setting \citep{auer2002finite, garivier2011upper, kaufmann2012bayesian}, it has not yet become as standard in deep RL. However, it has been shown that DQN ensembles used for uncertainty-driven updates can provide performance improvements \citep{osband2016deep, chen2017ucb, osband2018randomized, lee2021sunrise}. Similarly, OAC \citep{ciosek2019better} and TOP \citep{moskovitz2021tactical} leverage uncertainty estimates over the state-action value function for exploration, albeit still using only pessimistic actor. On that note, \cite{schafer2022decoupled} propose to use two actors, with one actor using some form of novelty exploration objective. 

The optimism-driven exploration was also considered for model-based agents. \citet{sekar2020planning} and \citet{seyde2022learning} consider exploration driven by reward model ensemble. Similarly to DAC, \citet{seyde2022learning} considers using an optimistic upper bound exploration policy and a distinct exploitation policy. Furthermore, agents like RP1 \citep{ball2020ready} leverage reward uncertainty despite access to the nominal rewards. Finally, a variety of agents that leverage MCTS have been proposed \citep{silver2017mastering, schrittwieser2020mastering, zawalski2022fast}. 

\subsection{Risk-Awareness for Overestimation}

Overestimation of values is a long standing problem associated with Q-learning \citep{hasselt2010double}, which is particularly pronounced in the continuous-action deep RL setting \citep{fujimoto2018addressing, hiraoka2021dropout, cetin2023learning}. Undoubtedly, the most popular approach in dealing with overestimation is CDQL \citep{fujimoto2018addressing, haarnoja2018soft2, ciosek2019better, bhatt2019crossq, hafner2019dream, ciosek2020expected, hiraoka2021dropout, hansen2022temporal}. However, it was noticed that overly pessimistic target can also negatively affect the agent performance \citep{moskovitz2021tactical}. We believe that this observation is particularly sensible in light of our proposed theory - since the agents are evaluated in expected return regime, training them with a distorted, risk-aware objective might not be optimal. To this end, a variety of works considered tuning the pessimism online \citep{moskovitz2021tactical, kuznetsov2020controlling, cetin2023learning}.

\subsection{Risk-Awareness for Safety}

On a different note, multiple works considered utility objectives for control that optimizes for worst-case \citep{jaquette1976utility, chung1987discounted, chow2018risk}. In deep RL,
there is a substantial body of work on utilizing distributional RL \citep{bellemare2017distributional, dabney2018distributional, dabney2018implicit} for uncertainty estimation, which in turn is used for building risk-aware policies. For example, \citet{dabney2018implicit} uses a quantile Q-network for estimation of the aleatoric uncertainty, and considers a variety of risk measures calculated over this distribution. Similarly, \citet{stachowicz2024racer} uses an ensemble of quantile networks to approximate the epistemic uncertainty and builds a CVaR policy that is sensitive with respect to that uncertainty. Usually, these works focus on optimizing some sort of "worst-case" objective \citep{tang2019worst}. In contrast to these works, we focus on the generalizations of CDQL objective \citep{fujimoto2018addressing}, which is usually used by agents that are evaluated via risk-neutral maximum returns. In fact, our work focuses on showing that these objectives are in fact risk-aware. 

\begin{figure}[ht!]
\begin{center}
\begin{minipage}[h]{1.0\linewidth}
    \begin{subfigure}{1.0\linewidth}
    \includegraphics[width=0.325\linewidth]{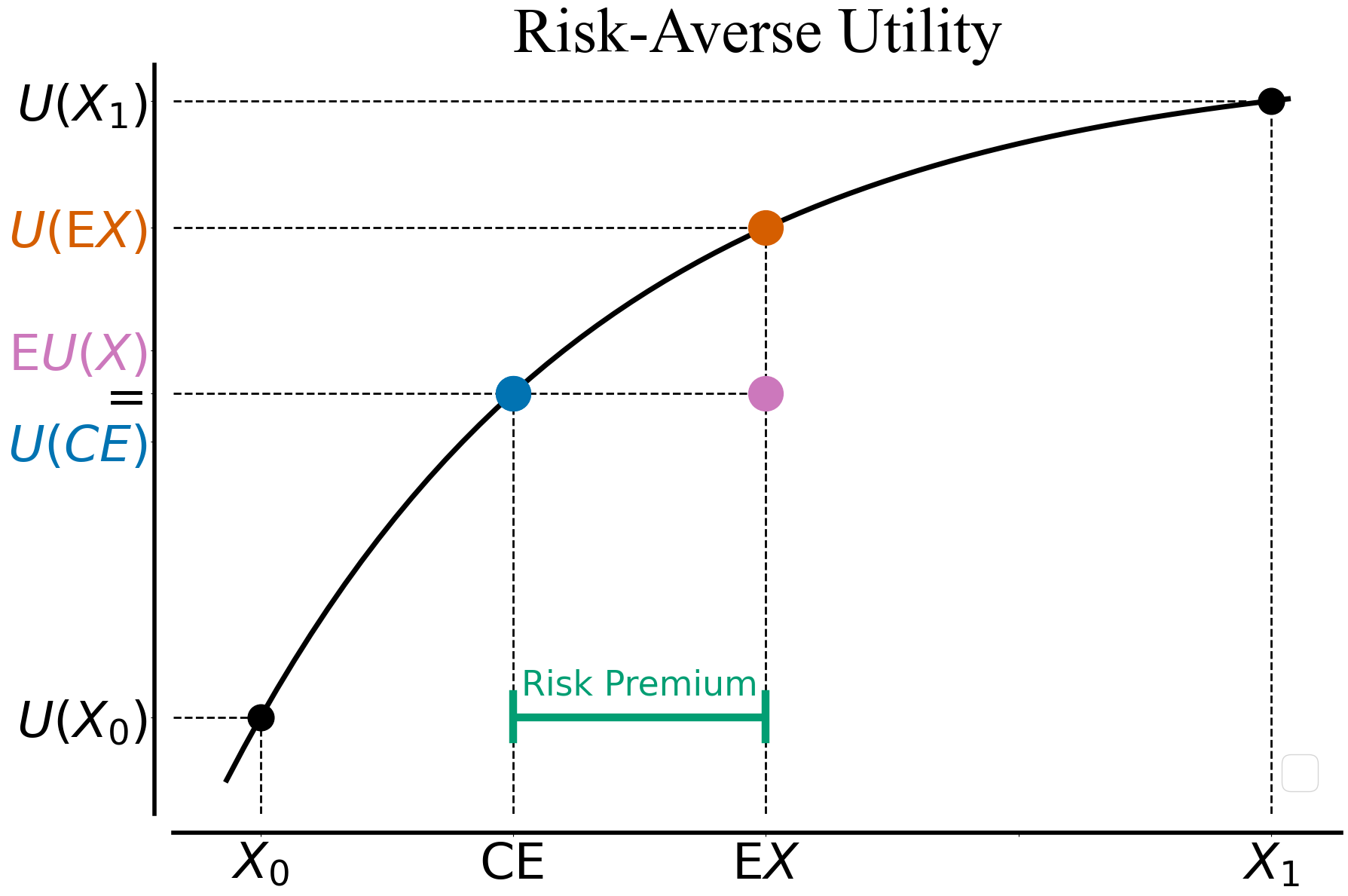}
    \hfill
    \includegraphics[width=0.325\linewidth]{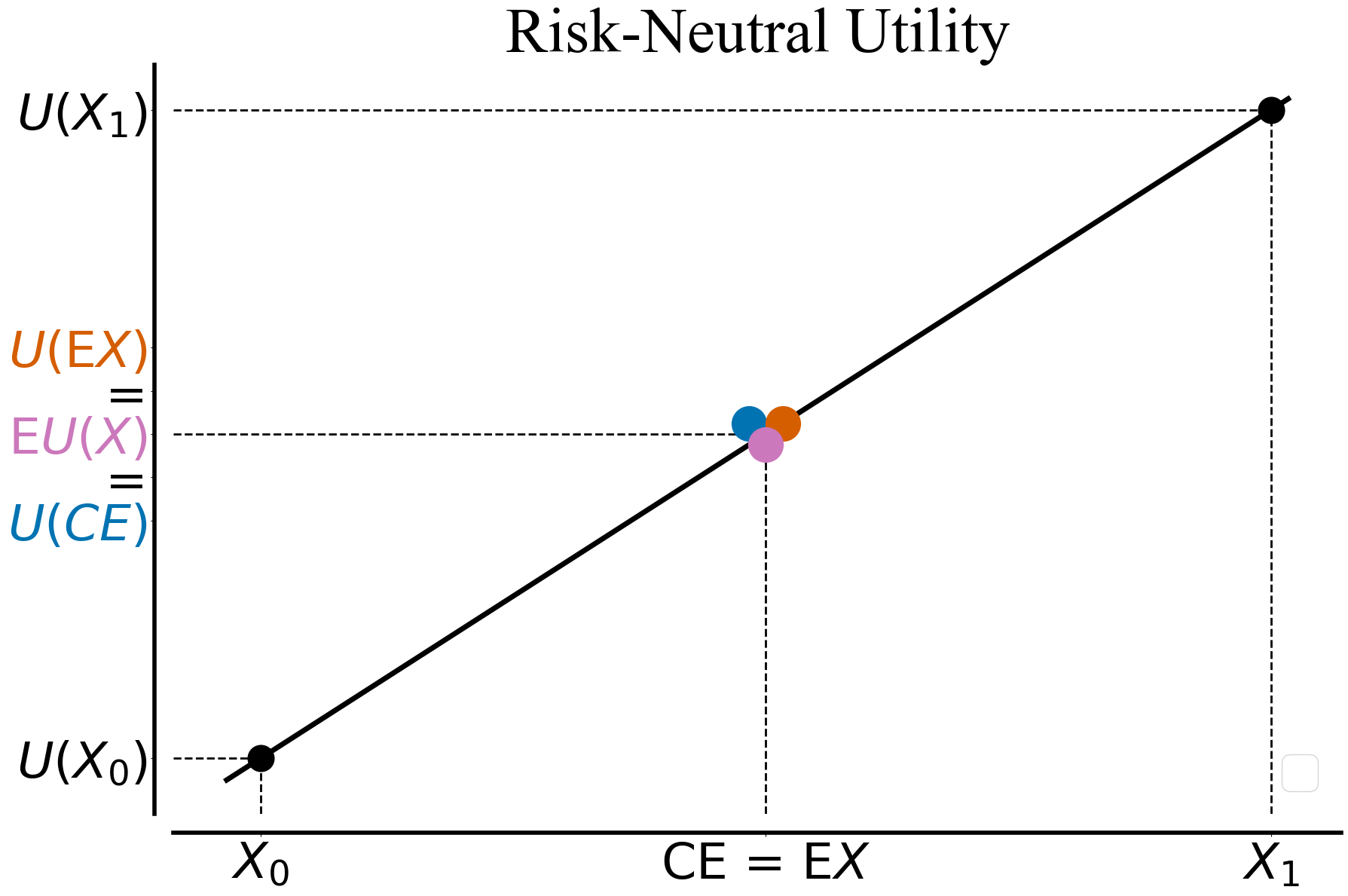}
    \hfill
    \includegraphics[width=0.325\linewidth]{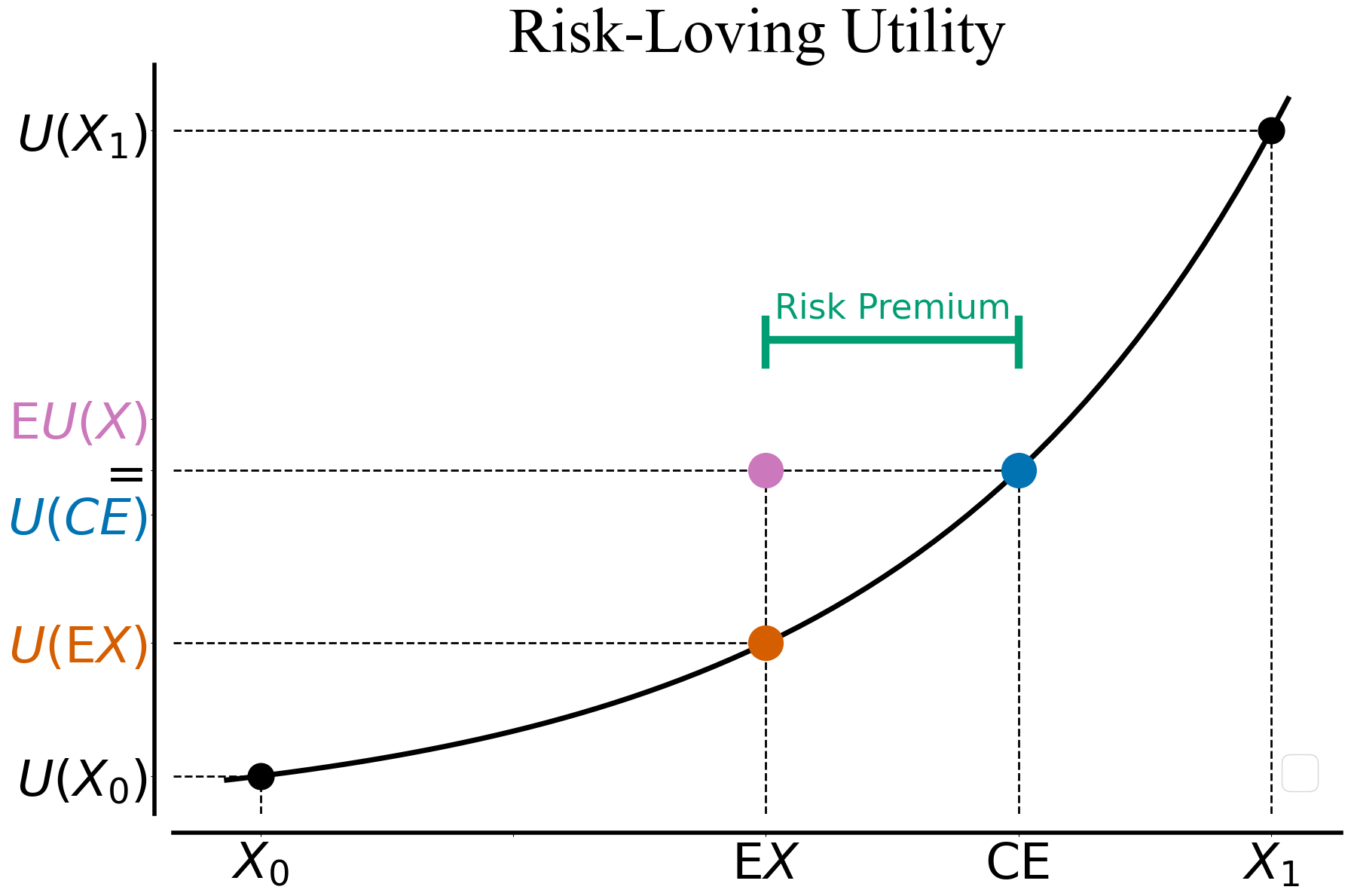}
    \end{subfigure}
\end{minipage}
\vspace{-0.1in} 
\caption{The risk appetite follows from the Jensen inequality: a convex utility indicates risk-seeking behavior, as evidenced by a CE exceeding $\mathrm{E}X$; conversely, a concave utility signifies risk-aversion, reflected in a CE that is smaller than $\mathrm{E}X$.}
\label{fig:risks}
\end{center}
\vspace{-0.1in} 
\end{figure}

\subsection{Comparison to OAC}
\label{appendix:oac}

DAC leverages two policies: a pessimistic one used for sampling the temporal-difference target and evaluation; and an optimistic one used for sampling transitions added to the experience buffer. Similarly to DAC, OAC performs evaluation and Bellman backups according to a pessimistic lower bound policy. However, DAC differs from OAC on three main design choices: \textit{how to model the optimistic policy}; \textit{how to constraint the optimistic policy}; and \textit{how to set the level of optimism $\beta^{o}$}. 

Firstly, OAC models the optimistic policy by combining pessimistic policy with the linear approximation of Q-value upper bound, and as such uses one actor network. The linear approximation combined with constrained optimization results in simplistic solutions along the constraint boundary \citep{protter2012first}. As such, OAC's applicability is limited to small $\delta$ values due to the Taylor theorem. In contrast to that, DAC uses two actors. Modeling the second policy via an actor network allows for exploration policies that are far more complex than a linear approximator. Whereas this introduces a computational cost, employing techniques like delayed policy updates can result in costs smaller than that of OAC. 

Furthermore, OAC enforces a hard KL constraint by directly solving a Lagrangian. Since the Q-value upper bound is approximated via a linear function, the solution is placed on the constraint boundary unless the slope is zero \citep{protter2012first}. In contrast, DAC imposes KL as a soft constraint. Paired with the neural network approximator, this allows DAC to balance the KL with potential gains to the upper bound and generate complex exploration policies. 

Finally, OAC treats $\beta^{o}$ as a hyperparameter which is fixed during the training. Since values of $Q^{\mu}$ and $Q^{\sigma}$ depend on reward scales, as well as aleatoric and epistemic uncertainty of the environment, the value of $\beta^{o}$ has to be searched per task. Furthermore, due to decreasing critic disagreement as the training progresses, fixed levels of $\beta^{o}$ yield decreasing the impact of uncertainty on the optimistic policy. DAC leverages that the desired level of optimism can be defined through divergence between the pessimistic baseline policy and the optimistic policy optimizing objective related to $\beta^{o}$. Such definition allows for dynamics adjustment of both $\beta^{o}$ and the KL penalty weight $\tau$.

\section{Further Description of DAC}
\label{appendix:dac}

\subsection{Pessimistic Actor}

The pessimistic actor, denoted as $\pi^{p}_{\phi}$, employs a soft policy target for optimization, as outlined in Equation \ref{eq:upd_pessactor}. This actor operates entirely off-policy, utilizing data exclusively from the optimistic actor. The soft policy approach not only promotes state-dependent exploration \citep{haarnoja2018soft} but also helps in maintaining the hyperbolic tangent (Tanh) output in a non-saturated state \citep{wang2020striving}. Moreover, the pessimistic actor's inherent non-zero variance contributes to stabilizing the TD learning for the critic. While separating exploration and exploitation theoretically allows for zero variance in TD target sampling, our empirical evidence suggests that introducing noise regularizes the critic (see Table \ref{tab:design_ablation}). 

\begin{figure}[h]
\begin{center}
\hfill
\begin{subfigure}{0.225\linewidth}
    \includegraphics[width=\textwidth]{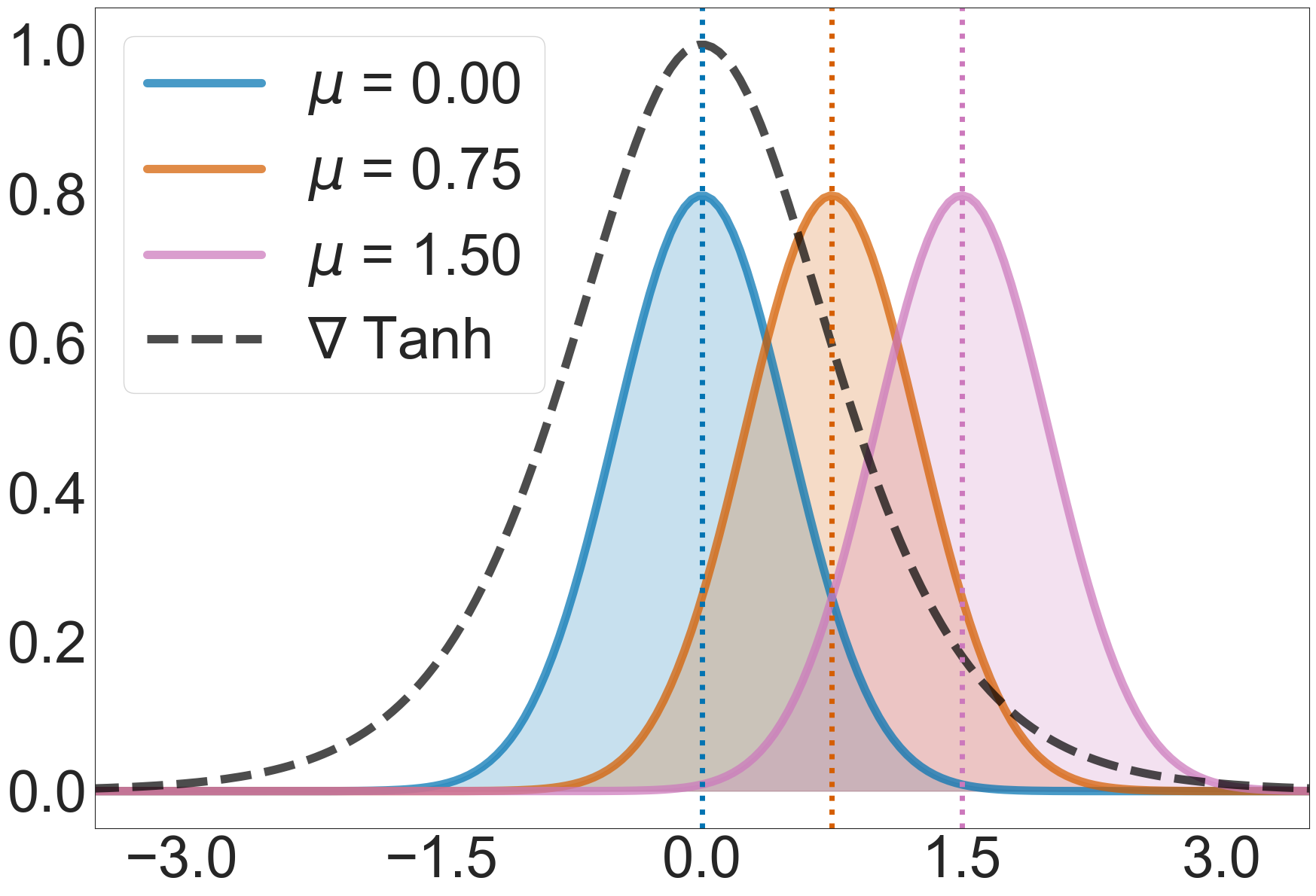}
    \caption{Normal}
    \label{fig:satur1}
\end{subfigure}
\hfill
\begin{subfigure}{0.225\linewidth}
    \includegraphics[width=\textwidth]{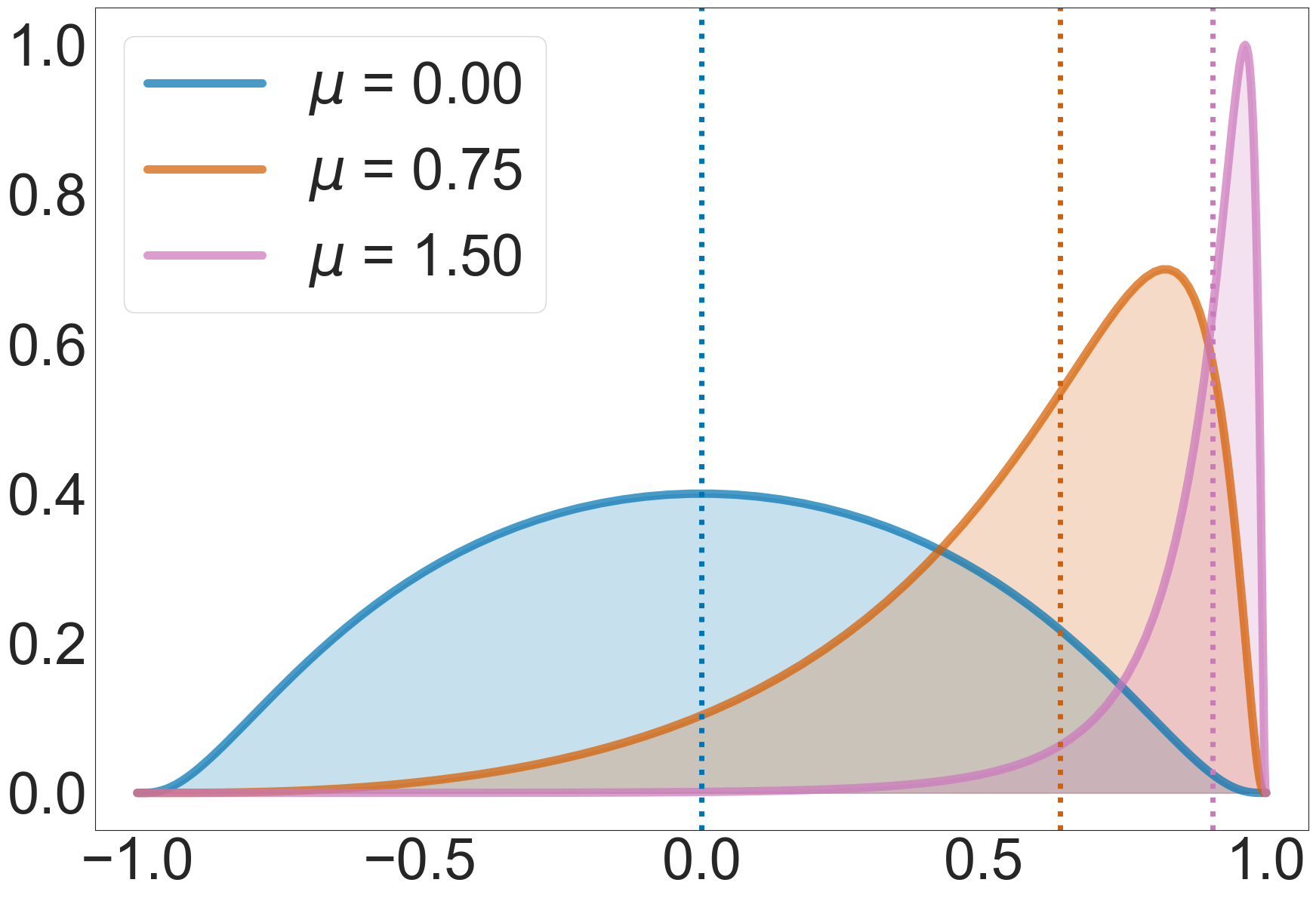}
    \caption{Tanh-Normal}
    \label{fig:satur2}
\end{subfigure}
\hfill
\begin{subfigure}{0.225\linewidth}
    \includegraphics[width=\textwidth]{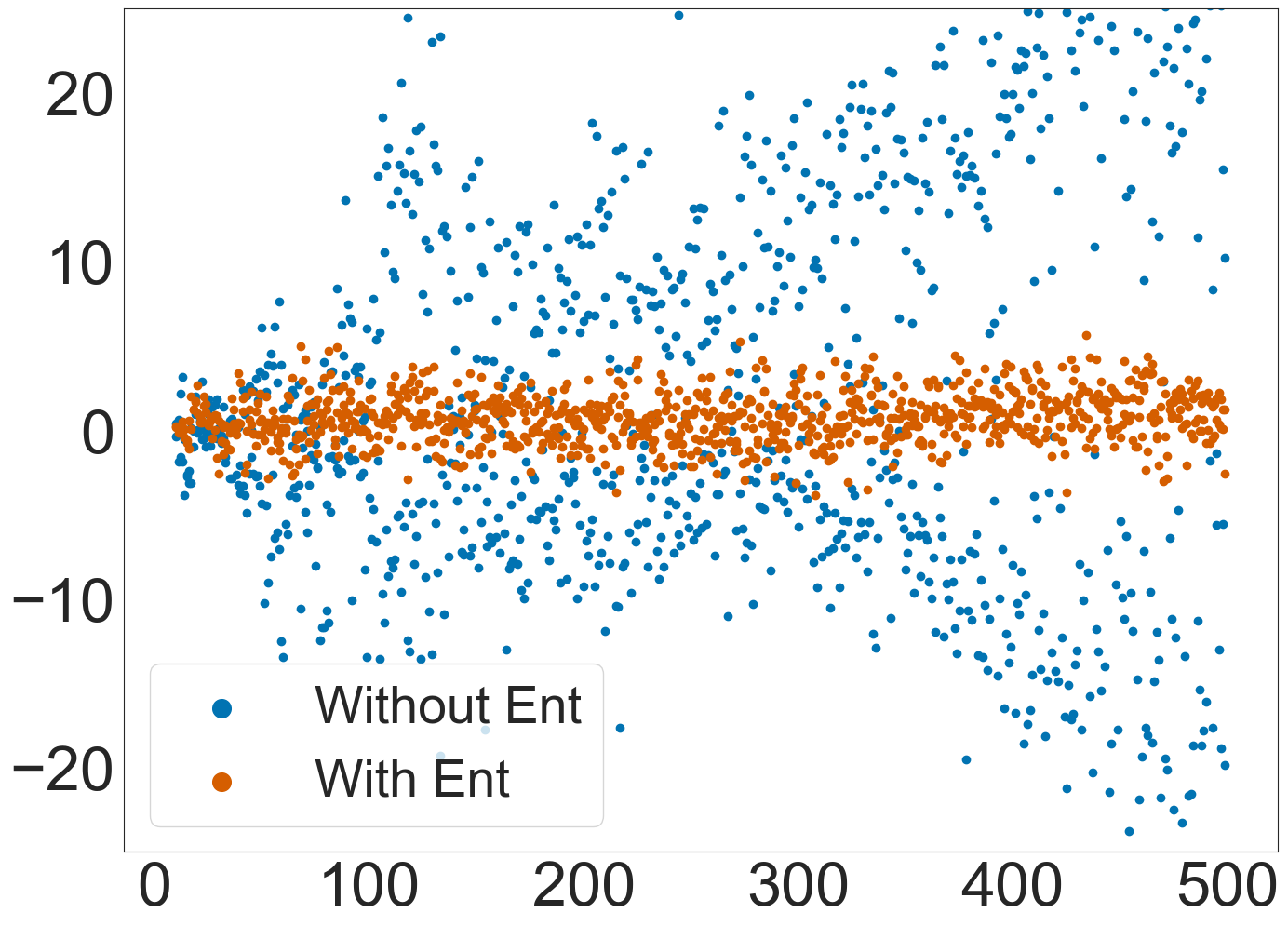}
    \caption{Pre-Tanh $\mu$}
    \label{fig:satur3}
\end{subfigure}
\hfill
\begin{subfigure}{0.225\linewidth}
    \includegraphics[width=\textwidth]{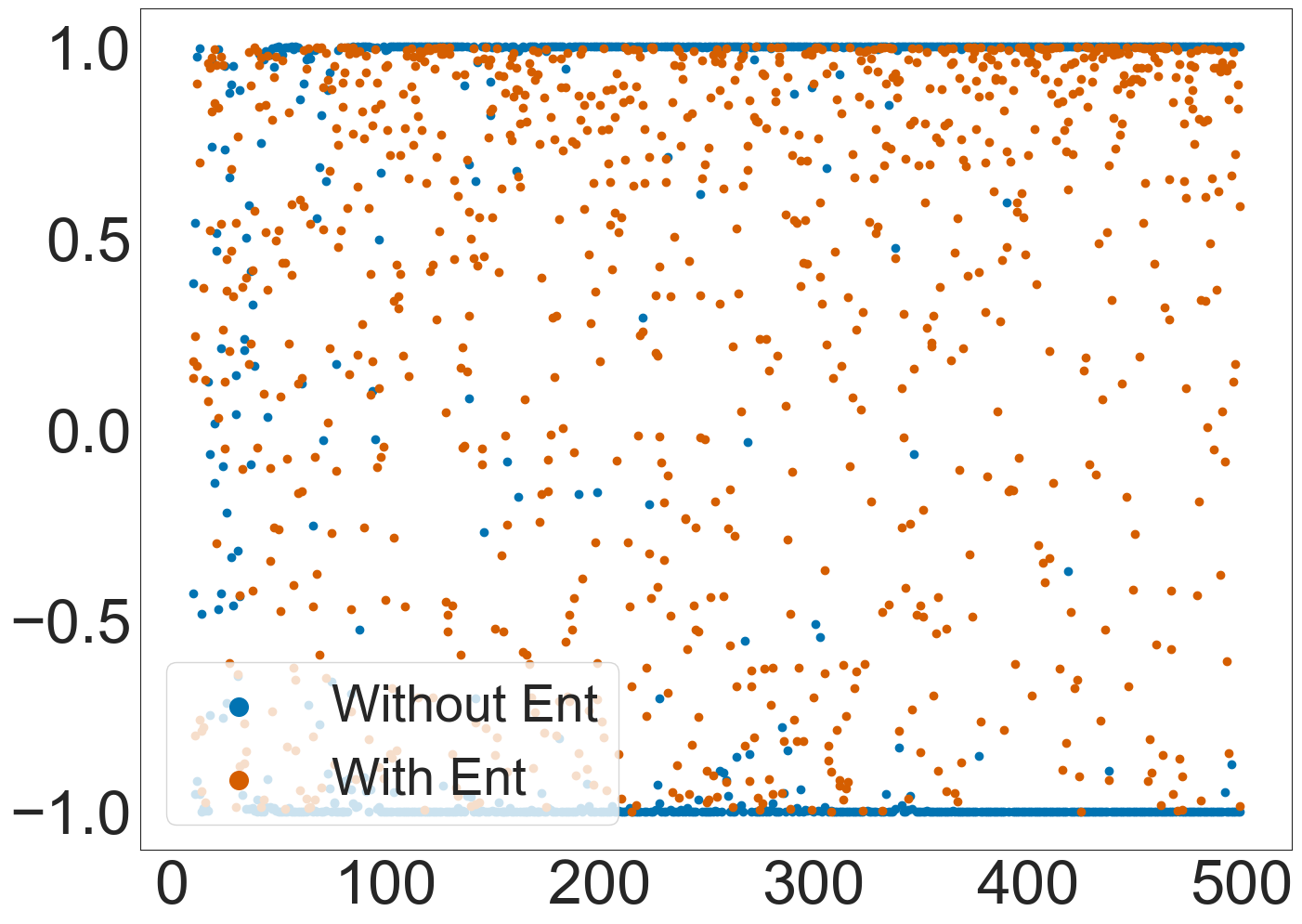}
    \caption{Post-Tanh Actions}
    \label{fig:satur4}
\end{subfigure}
\hfill
\caption{Soft policy learning prevents policy saturation. Figures \ref{fig:satur1} and \ref{fig:satur2} show three Gaussians and the Tanh transformed counterparts. Before Tanh transformation, all policies have equal variance (\ref{fig:satur1}). However, since Tanh is nonlinear, applying the transformation changes the policy variance depending on the location of the mean (\ref{fig:satur2}). Because of this effect, when using soft policy learning, the agent incurs loss when moving along the saturated portions of the Tanh activation function. Figures \ref{fig:satur3} and \ref{fig:satur4} show policy means and executed actions for SAC with and without soft policy learning on a Humanoid-Stand task. As follows, SAC without the entropy objectives follows a bang-bang policy, whereas regular SAC anchors the policy within the non-saturated portion of Tanh. To this end, soft policy learning reduces the risk of a bang-bang policy.}
\label{fig:satur}
\end{center}
\end{figure}

In the context of representing a policy with a Tanh-Normal distribution, soft policy learning serves two critical functions: promoting entropy and preventing saturation. Firstly, soft policy learning guarantees a certain degree of policy entropy, essential for exploration during the learning process \citep{haarnoja2018soft}. This approach contrasts with methods like TD3 or TOP, where constant entropy is maintained across various states. By optimizing an objective function that balances entropy maximization with Q-values, our policy exhibits variable entropy depending on the state, encouraging exploration particularly in states with lower Q-values. Furthermore, the Tanh function's inherent saturation characteristic impacts the exploration. The further the unsquashed distribution's mean is from zero, the more the variance reduces post-squashing, as depicted in Figures \ref{fig:satur1} and \ref{fig:satur2}. Implementing soft policy learning thus incurs a loss when the policy deviates from zero. Figures \ref{fig:satur3} and \ref{fig:satur4} illustrate how this approach results in policies predominantly residing within the non-saturated Tanh region \citep{wang2020striving}, effectively avoiding extreme bang-bang policies \citep{seyde2021bang}.

\subsection{Optimistic Actor}
\label{section:method_optimistic_actor}

The optimistic actor $\pi^{o}_{\eta}$, follows the optimistic certainty equivalence objective, detailed in Equation \ref{eq:upd_optiactor}. This objective prioritizes the upper bound of Q-values, fostering a policy that embraces uncertainty and encourages actions leading to critic disagreement. Such a policy approach is known to generate more diverse samples and consequently, ensures better coverage of the state-action space \citep{pathak2019self, lee2021sunrise}. This is significant as ensemble disagreement is often a surrogate for sample novelty \citep{yahaya2019consensus, han2021gan}. Although traditional RL does not explicitly focus on coverage, recent research highlights the critical role of data diversity in RL \citep{xie2022role, foster2022offline, zhan2022offline}. 

\begin{figure}[ht]
\begin{center}
\hfill
    \begin{subfigure}{0.3\linewidth}
    \includegraphics[width=\textwidth]{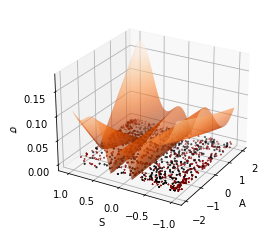}
    \caption{Critic Variance}
    \label{fig:intro21}
\end{subfigure}
\hfill
\begin{subfigure}{0.3\linewidth}
    \includegraphics[width=\textwidth]{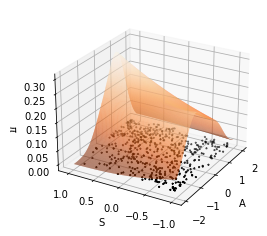}
    \caption{Pessimistic Actor}
    \label{fig:intro22}
\end{subfigure}
\hfill
\begin{subfigure}{0.3\linewidth}
    \includegraphics[width=\textwidth]{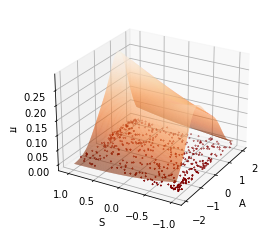}
    \caption{Optimistic Actor}
    \label{fig:intro23}
\end{subfigure}
\hfill
\caption{Pessimistic underexploration and state-action space coverage on the Pendulum task with state representation embedded into $1$ dimension. The dots represent $500$ state-action samples gathered using the latest policy (pessimistic (black) or optimistic (red)). Figure \ref{fig:intro21} displays the standard deviation ($\sigma$) of the two critics, with smaller values observed in well-explored state-action regions. In Figure \ref{fig:intro22}, we depict pessimistic policy probabilities. Due to lower bound optimization, the actor prioritizes state-action subspaces that have already been explored and do not yield critic disagreement. Figure \ref{fig:intro23} illustrates optimistic policy probabilities. Despite having similar entropy levels, following the upper bound policy results in better coverage within critic disagreement regions.}
\label{fig:intro2}
\end{center}
\end{figure}

In the development of DAC, our aim was to enforce entropy and non-saturation within both the pessimistic and optimistic actors. DAC accomplishes this through distinct yet complementary methods for each actor. For the pessimistic actor, soft policy learning is employed to enforce entropy and prevent policy saturation. In contrast, the optimistic actor uses KL divergence as a means of maintaining similarity with the pessimistic actor and adhering to certain policy constraints. The incorporation of KL divergence in the optimistic actor's objective serves multiple purposes:

\begin{enumerate}
    \item Reducing the degree of off-policy learning - Given that exploration is conducted solely by the optimistic actor, the pessimistic actor updates entirely off-policy, using transitions sampled from the optimistic policy. This can potentially lead to unstable learning, a phenomenon known as the "deadly triad" \citep{sutton2018reinforcement}. Integrating KL divergence into the optimistic objective aligns the transitions sampled from the optimistic policy with the expectations under the pessimistic policy, thereby mitigating the extent of off-policy learning.
    \item Anchoring the adjustment mechanisms - As discussed in Section \ref{section:dac}, the automatic adjustment of parameters $\beta^{o}$ (optimism) and $\tau$ (KL penalty weight) depends on a target divergence value. Without the divergence penalty, these adjustment mechanisms would have to be designed in a different way. 
    \item Preventing saturation of the optimistic policy - By minimizing the KL divergence between the optimistic and pessimistic actors, the optimistic actor is encouraged to emulate a policy trained via soft policy learning. This discourages the optimistic actor from policy saturation, maintaining the integrity of its actions.
\end{enumerate}

In DAC, both the pessimistic and optimistic actors are represented as simple diagonal normal distributions, further transformed by the hyperbolic tangent (Tanh) activation. The KL divergence between these actors is computed in a closed form, utilizing the change of variables technique. Though this is a well-established result, we include it here for completeness. Denote $x$ as samples from the policy distributions before the Tanh activation, $p^p$ and $p^o$ as the distributions of the pessimistic and optimistic actors on $x$, respectively. The application of Tanh is expressed as $y = h(x)$, resulting in the transformed distributions $\pi^{p}, \pi^{o}$. Using the change of variables formula, the KL divergence is derived as follows:

\begin{equation}
\begin{split}
    KL \bigl(p_c | p_o\bigr) & = \int_{-\infty}^{\infty} p^p(x) \log \frac{p^p(x)}{p^o(x)} dx \\
    & = \int_{-\infty}^{\infty} \pi^{p}(h(x)) ~ |\frac{dy}{dx}(x)| ~ \log \frac{\pi^{p}(h(x)) ~ |\frac{dy}{dx}(x)|}{\pi^{o}(h(x)) ~ |\frac{dy}{dx}(x)|} dx \\
    & = \int_{-\infty}^{\infty} \pi^{p}(y) \log \frac{\pi^{p}(y)}{\pi^{o}(y} dy = KL \bigl(p^p | p^o\bigr)
\end{split}
\end{equation}

This relationship holds for any distributions $p_c$ and $p_o$. With the assumption of diagonal Gaussian distributions with dimensionality $|A|$, the KL divergence simplifies to:

\begin{equation}
    KL \bigl(p_p | p_o\bigr) = KL \bigl(\pi_{c} | \pi_{o}\bigr) = \sum_{i=1}^{|A|}\biggl( \log \frac{\sigma^{p}_{i}}{\sigma^{o}_{i}} + \frac{(\sigma^{o}_{i})^{2} + (\mu^{o}_{i} - \mu^{p}_{i})^{2}}{2 ~(\sigma^{p}_{i})^{2}} - \frac{1}{2} \biggr)
\end{equation}

Using KL guarantees that the optimistic policy optimizes for a specified level of variance, which can be distinct from $\pi^{p}_{\phi}$. To this end, we define the exploration variance multiplier $m$, which is a simple standard deviation multiplier. This leads to the following formulation of KL applied to DAC:

\begin{equation}
\label{eq:KL_sigma}
    KL \bigl(\pi_{c} | \pi_{o}\bigr) = \sum_{i=1}^{|A|}\biggl( \log \frac{\sigma^{p}_{i}}{\Bar{\sigma}^{o}_{i}} + \frac{(\Bar{\sigma}^{o}_{i})^{2} + (\mu^{o}_{i} - \mu^{p}_{i})^{2}}{2 ~(\sigma^{p}_{i})^{2}} - \frac{1}{2} \biggr), \quad \Bar{\sigma}^{o}_{i} = \frac{\sigma^{o}_{i}}{m}
\end{equation}

This approach ensures distinct entropy levels for TD learning and exploration while maintaining the standard convergence properties of Actor-Critic (AC) algorithms. In fact, if $\lim_{\mathcal{D}\to\infty} Q^{\sigma}_{\theta}(s, a) = 0$ it follows that in the limit both actors recover a policy that differs only by $m$. While other differentiable divergence or distance functions could be used, the appeal of KL divergence lies in its closed-form solution for any invertible and differentiable transformation of the given distributions.

\subsection{PseudoCode}

\begin{figure}[ht!]
    \begin{algorithmic}[1]
       \STATE {\bfseries Input:} $\pi^{p}_{\phi}$ - pessimistic actor; $\pi^{o}_{\eta}$ - optimistic actor; $Q_{\theta}$ - double critic; $Q_{\Bar{\theta}}$ - target critic; $\alpha$ - temperature; $\beta^{o}$ - optimism; $\tau$ - KL weight;
       \STATE {\bfseries Hyperparameters:}   
       $\beta^{p}$ - pessimism; $\mathcal{KL}^{*}$ - target KL; 
       \\\hrulefill
       \STATE\textcolor{purple}{\textit{Sample action from the optimistic actor}} \\
       \textcolor{purple}{$s', r = \textsc{env.step}(a) \quad a \sim \pi^{o}_{\eta}$}
       \STATE \textit{Add transition to the replay buffer} \\
       $\textsc{buffer.add}(s,a,r,s')$
       \FOR{$i=1$ {\bfseries to} ReplayRatio}
       \STATE \textit{Sample batch of transitions} \\
       $s, a, r, s' \sim \textsc{buffer.sample}$ 
       \STATE \textit{Update critic using pessimistic actor (Eq. \ref{eq:upd_critic})} \\
       $\theta \leftarrow \theta - \nabla_{\theta} \bigl(Q_{\theta} (s,a) - (r + \gamma ($\textcolor{purple}{$Q^{\mu}_{\Bar{\theta}}(s', a') + \beta^{p} Q^{\sigma}_{\Bar{\theta}}(s', a')$}$ - \alpha \log\pi^{p}_{\phi}(a'|s') \bigr)^{2}, ~ a' \sim \pi^{p}_{\phi}(s')$
       \STATE \textit{Update pessimistic actor (Eq. \ref{eq:upd_pessactor})} \\
       $\phi \leftarrow \phi + \nabla_{\phi} \bigl($\textcolor{purple}{$Q^{\mu}_{\theta}(s, a) + \beta^{p} Q^{\sigma}_{\theta}(s, a)$}$ - \alpha \log\pi^{p}_{\phi}(a|s)\bigr) ~~~~ \text{with} ~~~~ a \sim \pi^{p}_{\phi}(s)$
       \STATE \textcolor{purple}{\textit{Update optimistic actor (Eq. \ref{eq:upd_optiactor})}} \\
       \textcolor{purple}{$\eta \leftarrow \eta + \nabla_{\eta} \bigl( Q^{\mu}_{\theta}(s, a) + \beta^{o} Q^{\sigma}_{\theta}(s, a) - \tau KL\bigl(\pi_{\phi}^{p}(s)|\Bar{\pi}_{\eta}^{o}(s)\bigr) \bigr) \quad \text{with} \quad a \sim \pi^{o}_{\eta}(s)$}
       \STATE \textit{Update entropy temperature} \\
       $\alpha \leftarrow \alpha - \nabla_{\alpha} \alpha \bigl(\mathcal{H}^{*} - \mathcal{H}(s)\bigr)$
       \STATE \textcolor{purple}{\textit{Update optimism (Eq. \ref{eq:optimism})}} \\
       \textcolor{purple}{$\beta^{o} \leftarrow \beta^{o} -\nabla_{\beta^{o}} (\beta^{o} - \beta^{p} ) (\frac{1}{|A|}KL(\pi_{\phi}^{p}|\pi_{\eta}^{o}) - \mathcal{KL}^{*})$}
       \STATE \textcolor{purple}{\textit{Update KL weight (Eq. \ref{eq:optimism})}} \\
       \textcolor{purple}{$\tau \leftarrow \tau + \nabla_{\tau} \tau (\frac{1}{|A|}KL(\pi_{\phi}^{p}|\pi_{\eta}^{o}) - \mathcal{KL}^{*})$}
       \STATE \textit{Update target network} \\
       $\Bar{\theta} \leftarrow \textsc{Polyak}(\theta, \Bar{\theta})$
       \ENDFOR \\
       \hrulefill
    \end{algorithmic}
    \vspace{-0.1in}
    \caption{Pseudo-code shows DAC training step, where changes with respect to SAC are colored.}
    \label{fig:pseudocode}
\end{figure}

\subsection{Optimistic Actor Implementation}

Both optimistic and pessimistic policies are modelled using the Tanh-transformed Gaussian distribution denoted as $\mathcal{T}(\mu, \sigma)$, where $\mu$ and $\sigma$ denote the parameters of the underlying Gaussian \citep{haarnoja2018soft, wang2020striving}. We implement the optimistic actor as a perturbation to the pessimistic policy parameters:

\begin{equation}
\begin{split}
    & \pi_{\eta}^{o}(s) = \mathcal{T}((\mu^{p}_{\phi}(s) + \mu^{o}_{\eta}(s)), (\sigma^{p}_{\phi}(s) * \sigma^{o}_{\eta}(s))).
\end{split}
\end{equation}

Above, $\mu^{p}_{\phi}(s)$ and $\sigma^{p}_{\phi}(s)$ denote mean and standard deviation given by the pessimistic actor network, $\mu^{o}_{\eta}(s)$ and $\sigma^{o}_{\eta}(s)$ denote the optimistic perturbation to the pessimistic policy. Such design allows the optimistic actor network to trivially reduce the KL divergence - to minimize divergence, the optimistic actor network needs to adjust its output towards $0$, independently of the values of the pessimistic policy parameters. This design allows the model to keep the KL low, which as we show in Table \ref{tab:design_ablation} is crucial for good performance. 

\subsection{Adjustment of $\beta^{o}$ and $\tau$}
\label{section:method_adjustment}

Since values of $Q^{\mu}_{\theta}$ and $Q^{\sigma}_{\theta}$ depend on reward scales, as well as aleatoric and epistemic uncertainty of the environment, the value of $\beta^{o}$ cannot be easily set. DAC leverages an observation that for $\beta^{o} = -\beta^{p}$ the optimistic actor recovers the objective of the pessimistic actor. Then, $\beta^{o}$ can be defined such that the divergence between the pessimistic baseline policy and the optimistic policy reaches a desired level, as shown in Equation \ref{eq:optimism}. This adaptive approach, as illustrated in Figure \ref{fig:adjustments}, accommodates different scales of Q-values and contrasts with setups like OAC, where optimism is predefined by fixing $\beta^{o}$ at a specific value. We find that the optimism and KL weight used by DAC are highly variable between tasks and even phase of the training, showing the advantage of the automatic adjustment mechanism as opposed to using predefined levels. Furthermore, we find that DAC uses higher levels of optimism than values that were found optimal for OAC \citep{ciosek2019better}.

\section{Additional Experiments}
\label{app:additional_exp}

\subsection{Wallclock performance}

Firstly, we assess the wallclock efficiency of DAC as compared to SAC. Figure below investigates the performance of DAC and SAC in both low and high replay regimes.

\begin{figure}[ht!]
\begin{center}
\begin{minipage}[h]{1.0\linewidth}
    \begin{subfigure}{1.0\linewidth}
    \includegraphics[width=0.49\linewidth]{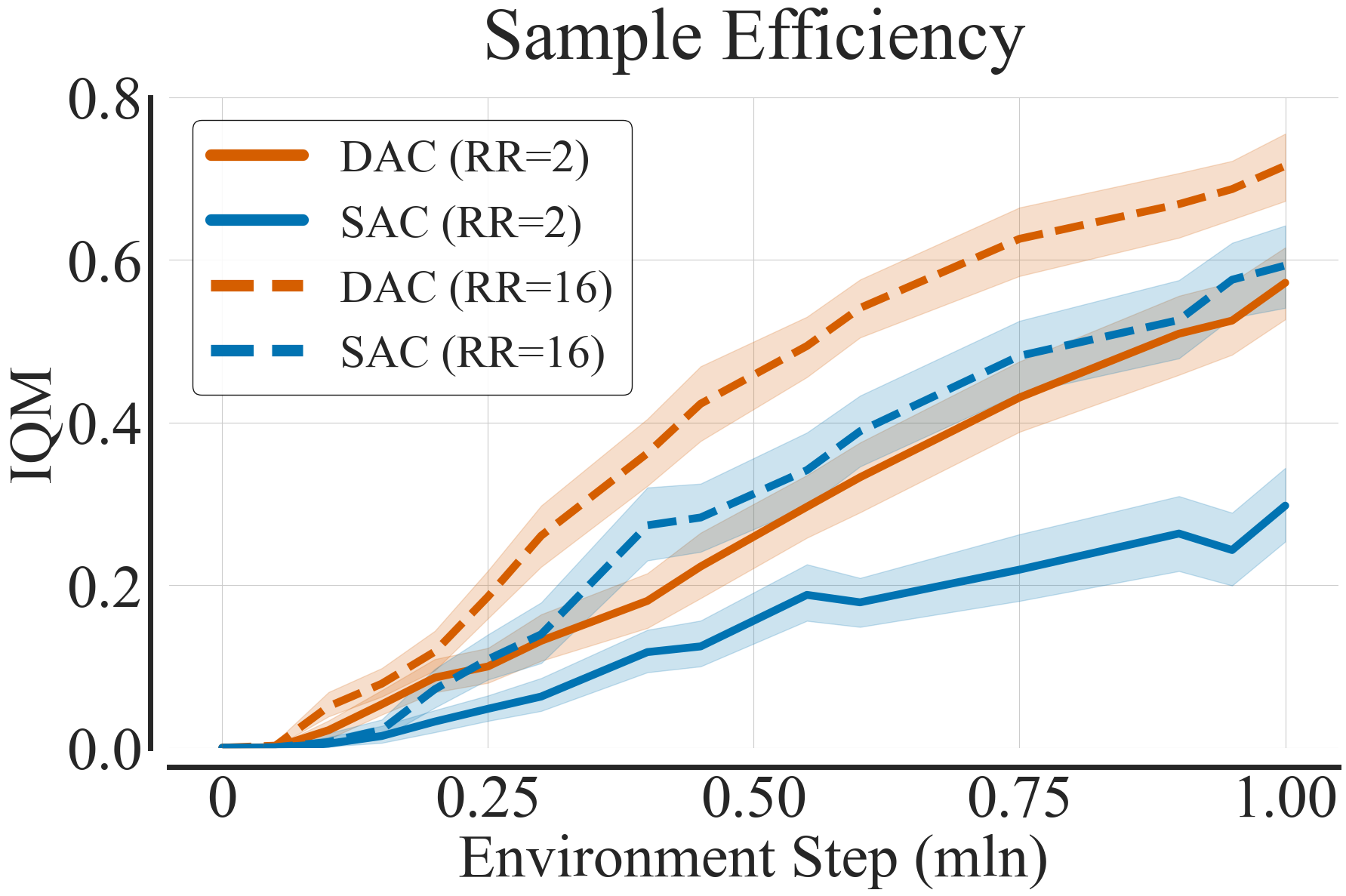}
    \hfill
    \includegraphics[width=0.49\linewidth]{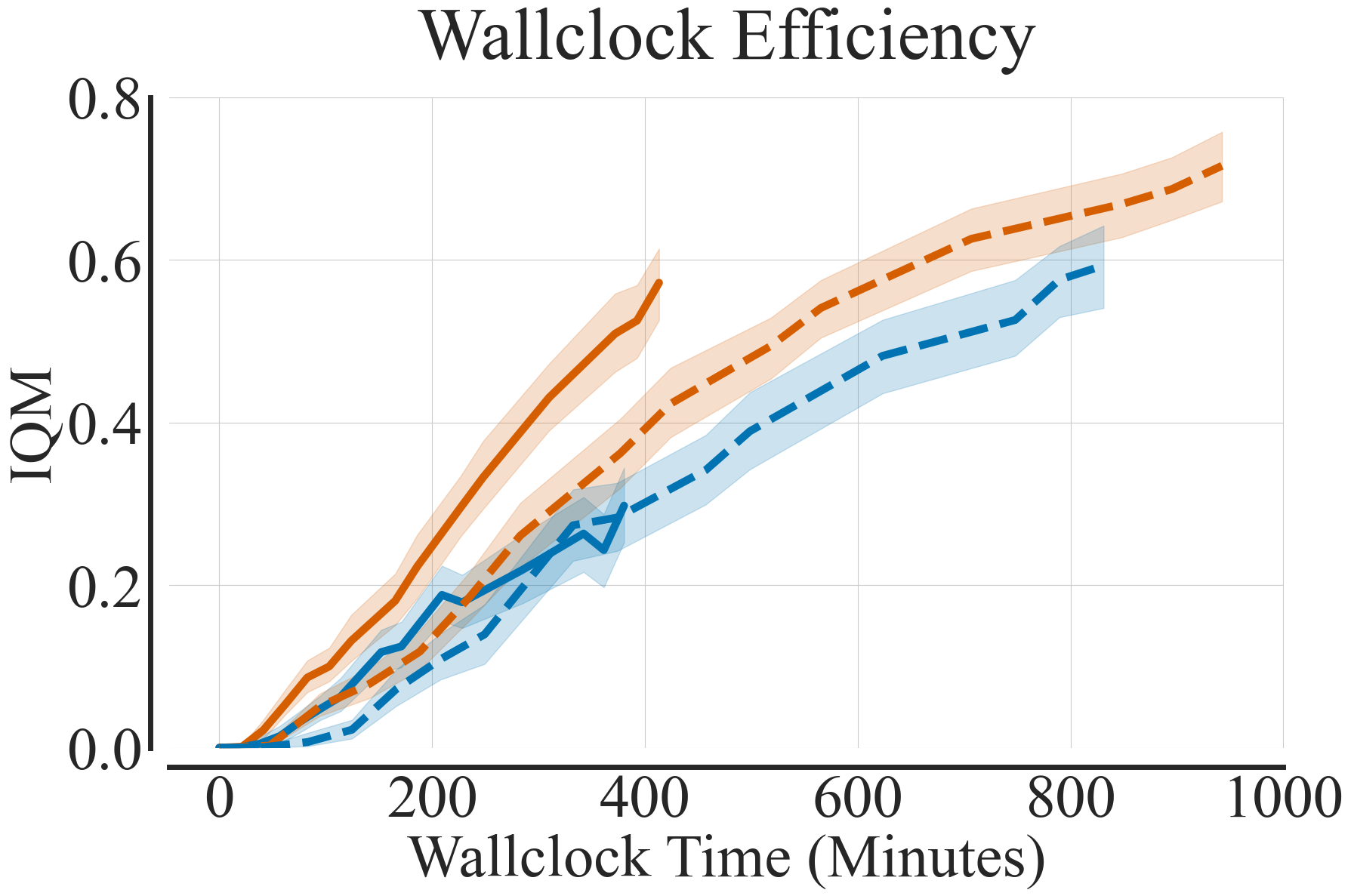}
    \end{subfigure}
\end{minipage}
\caption{DAC wall-clock runtime for $1$mln steps is around $15\%$ longer than SAC. Despite additional computational requirements associated with the dual actor, the improvements stemming from DAC lead to a beneficial compute/performance tradeoff. The $Y$-axis denotes IQM with $95$\% CI. The $X$-axis denotes the environment steps (left) and wallclock (right) for $10$ seeds. $30$ tasks, $1$mln steps and $10$ seeds. Experiments were conducted on an NVIDIA A100 GPU with 40GB of RAM.}
\label{fig:wallclock}
\end{center}
\vspace{-0.1in} 
\end{figure}

\subsection{Layer Normalization}

Recently, there has been a surge of works exploring the importance of critic regularization in high replay regimes \citep{liu2020regularization, laskin2020reinforcement, hiraoka2021dropout, gogianu2021spectral, li2022efficient, d2022sample}. Whereas there is still a lot to understand about the interplay of TD learning and network regularization, it is clear that a regularized critic allows a higher replay ratio to be used \citep{hiraoka2021dropout, li2022efficient, d2022sample}. In this paper, we explore only one method that could be considered regularization, that is full-parameter resets \citep{nikishin2022primacy, d2022sample}. Given the effectiveness of the only regularization we tested, we hypothesize that experimenting with methods like layer normalization \citep{ba2016layer}, weight decay \citep{krogh1991simple}, spectral normalization \citep{gogianu2021spectral} or dropout \citep{srivastava2014dropout} could further improve DAC performance. 

\begin{figure}[ht!]
\begin{center}
\begin{minipage}[h]{1.0\linewidth}
    \begin{subfigure}{1.0\linewidth}
    \includegraphics[width=0.24\linewidth]{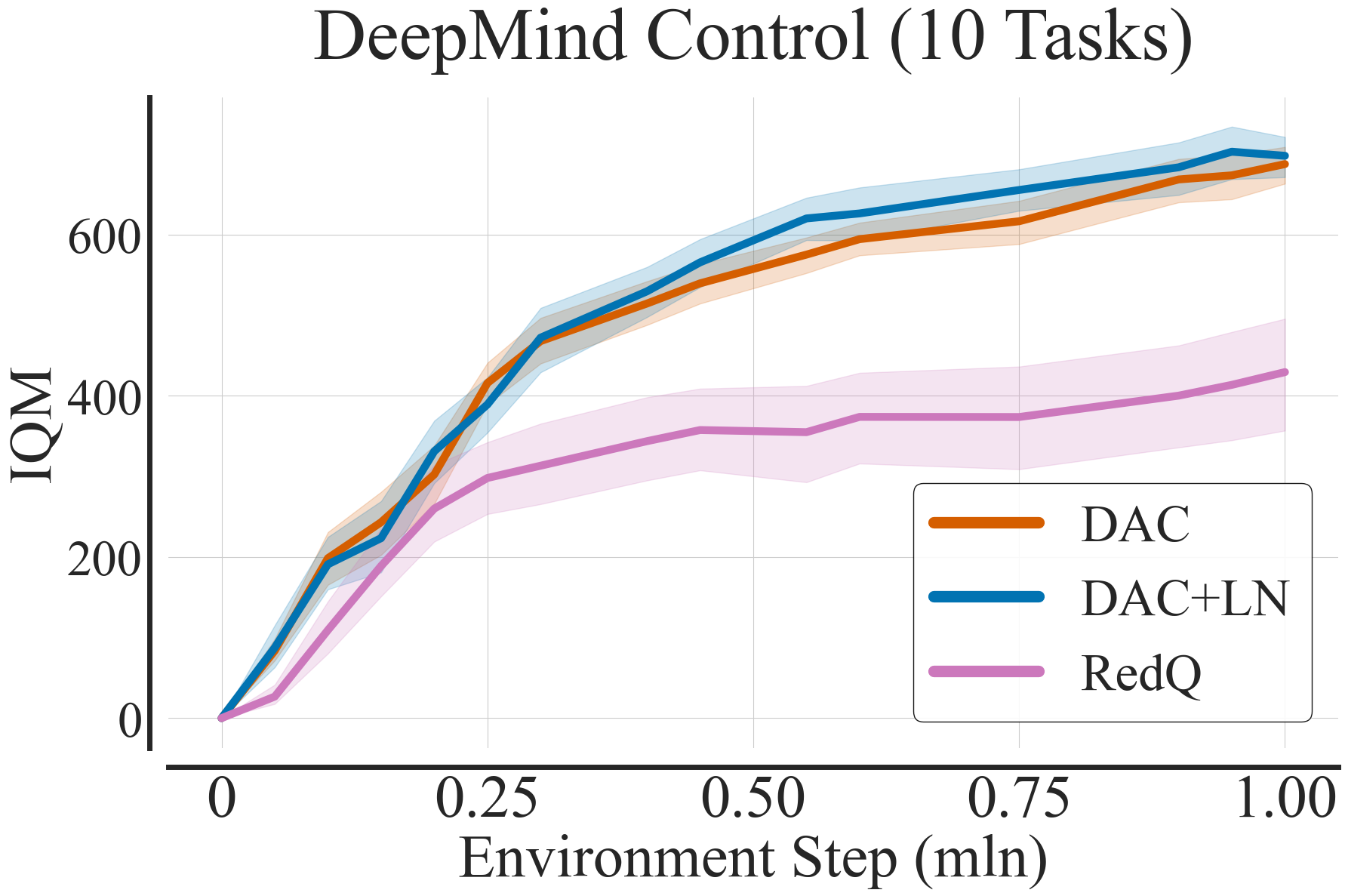}
    \hfill
    \includegraphics[width=0.24\linewidth]{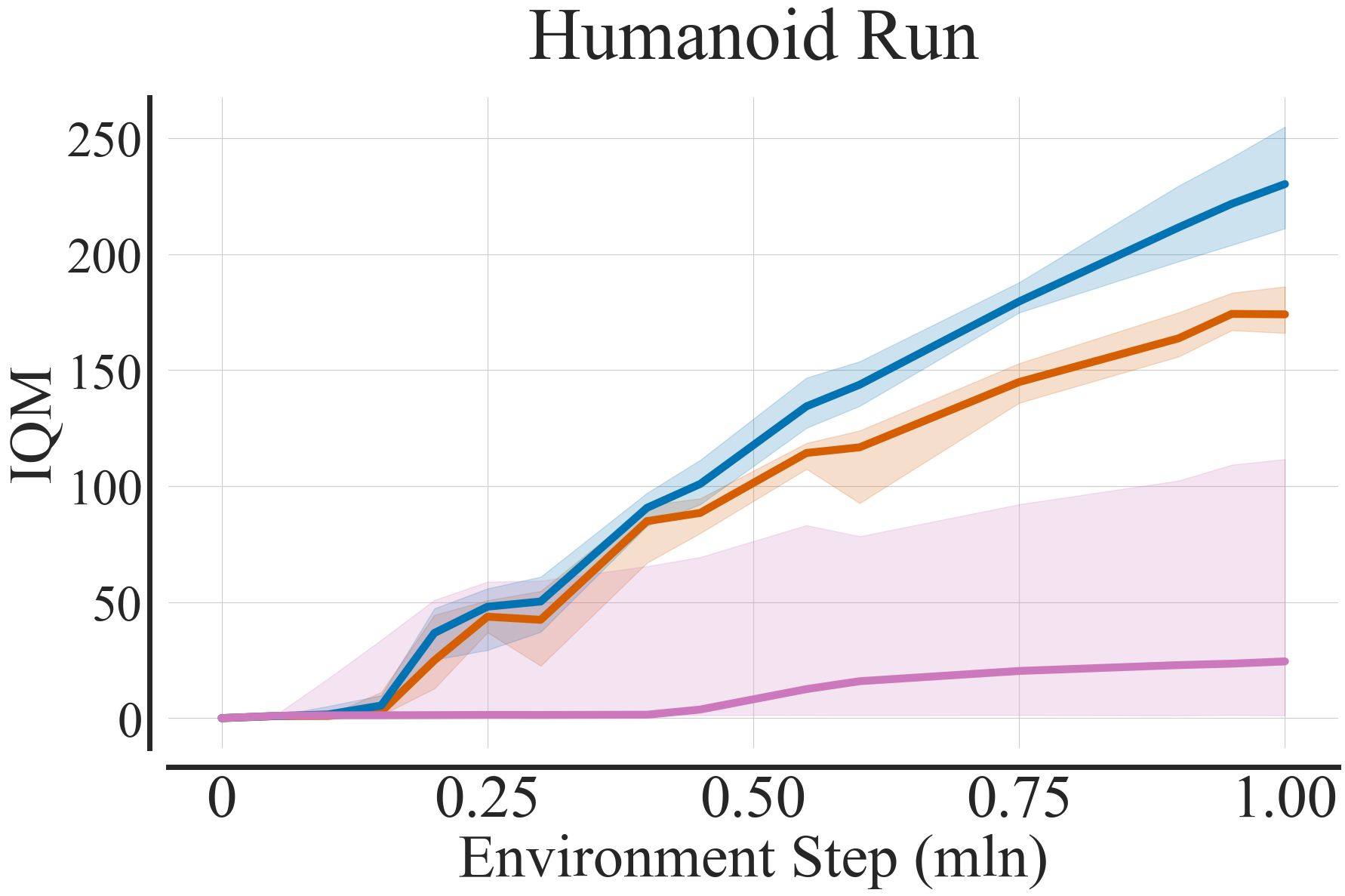}
    \hfill
    \includegraphics[width=0.24\linewidth]{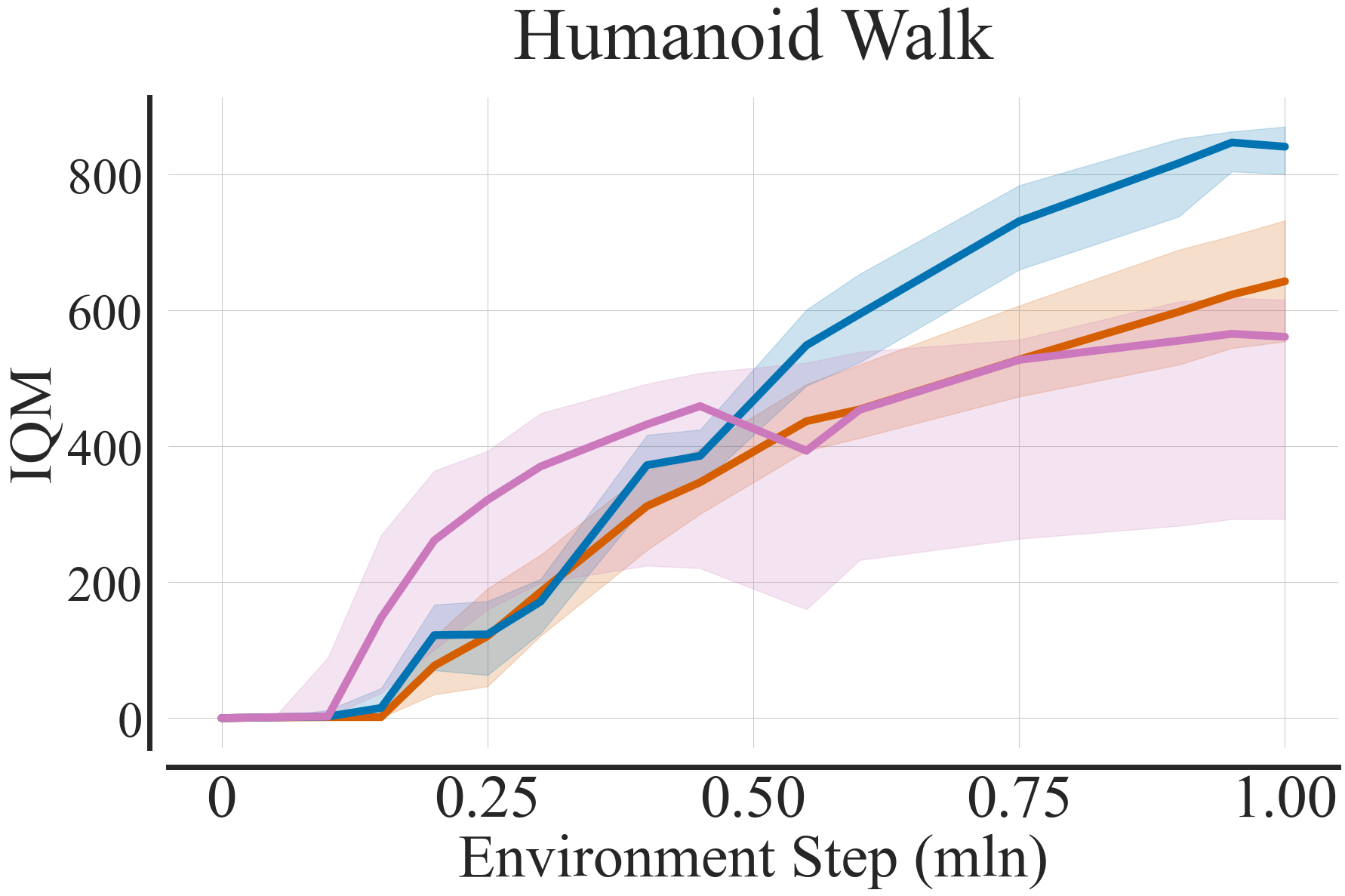}
    \hfill
    \includegraphics[width=0.24\linewidth]{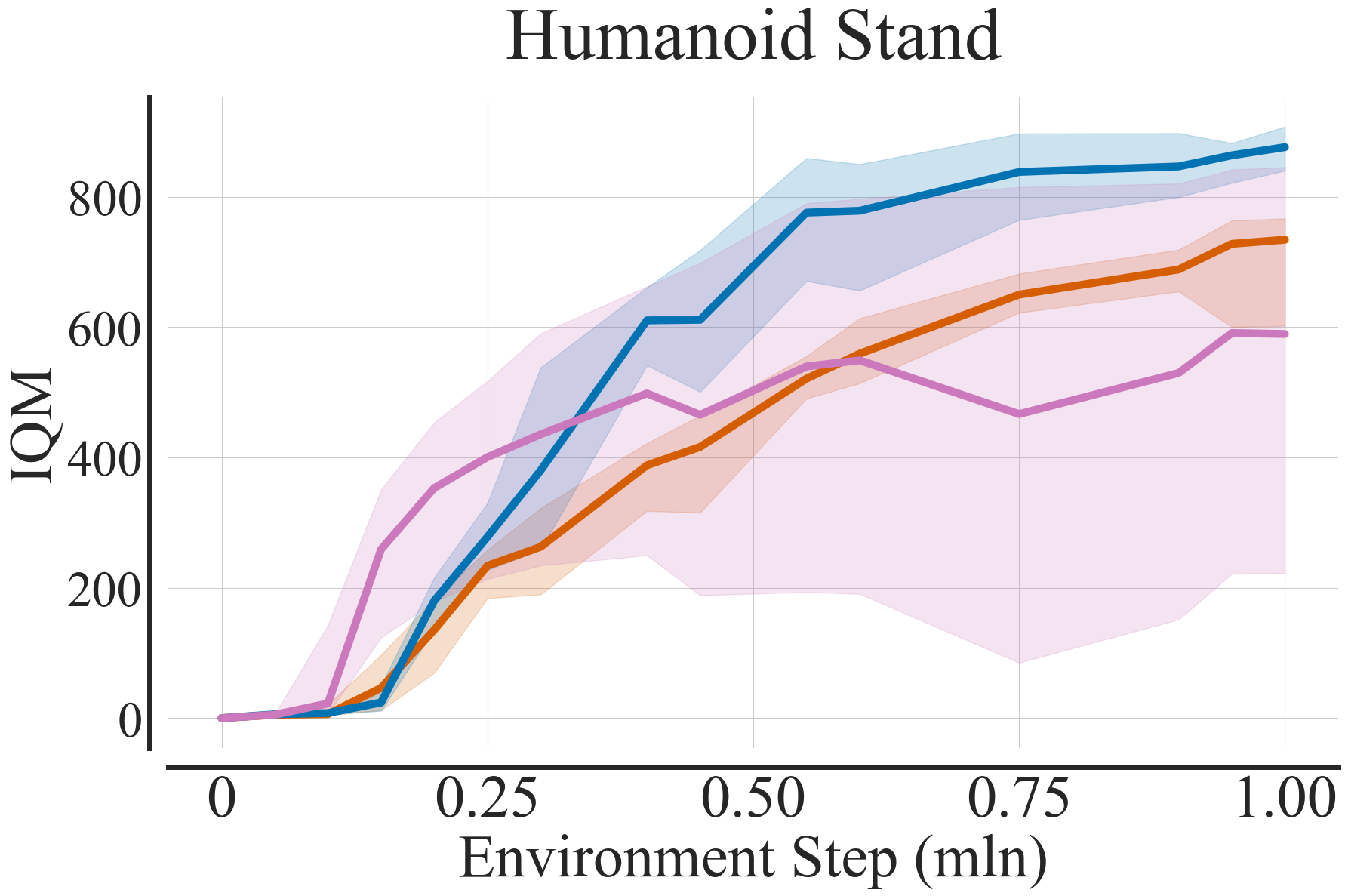}
    \end{subfigure}
\end{minipage}
\caption{We compare high-replay DAC performance with and without layer normalization, as well as the ensemble based, high-replay RedQ \citep{hiraoka2021dropout}. We use $10$ DMC tasks listed in Table \ref{tab:all_tasks} and run $10$ seeds for $1$mln environment steps.}
\label{fig:layernorm}
\end{center}
\vspace{-0.1in} 
\end{figure}

As shown in Figure \ref{fig:layernorm}, we find that using layer normalization on critic within the DAC design significantly improves its performance on locomotion tasks. This results is in line with previous work that augmented other RL algorithms with layer normalization \citep{li2022efficient, hiraoka2021dropout, lyle2023understanding, ball2023efficient, cetin2023learning}

\subsection{Critic Overestimation}

Here, we investigate whether the decoupled architecture (using a conservative policy for TD updates and optimistic policy for exploration) indeed mitigates Q-value overestimation characteristic for non-conservative updates \citep{fujimoto2018addressing, kuznetsov2020controlling, moskovitz2021tactical, cetin2023learning}. As such, we compare the environment returns to the returns implied by the critic, according to the following:

\begin{equation}
    O_{\theta} = \frac{|Q_{\theta}(s,a)|}{|G^{\pi}(s,a)|} 
\end{equation}

\begin{figure}[ht!]
\begin{center}
\begin{minipage}[h]{1.0\linewidth}
    \begin{subfigure}{1.0\linewidth}
    \includegraphics[width=0.31\linewidth]{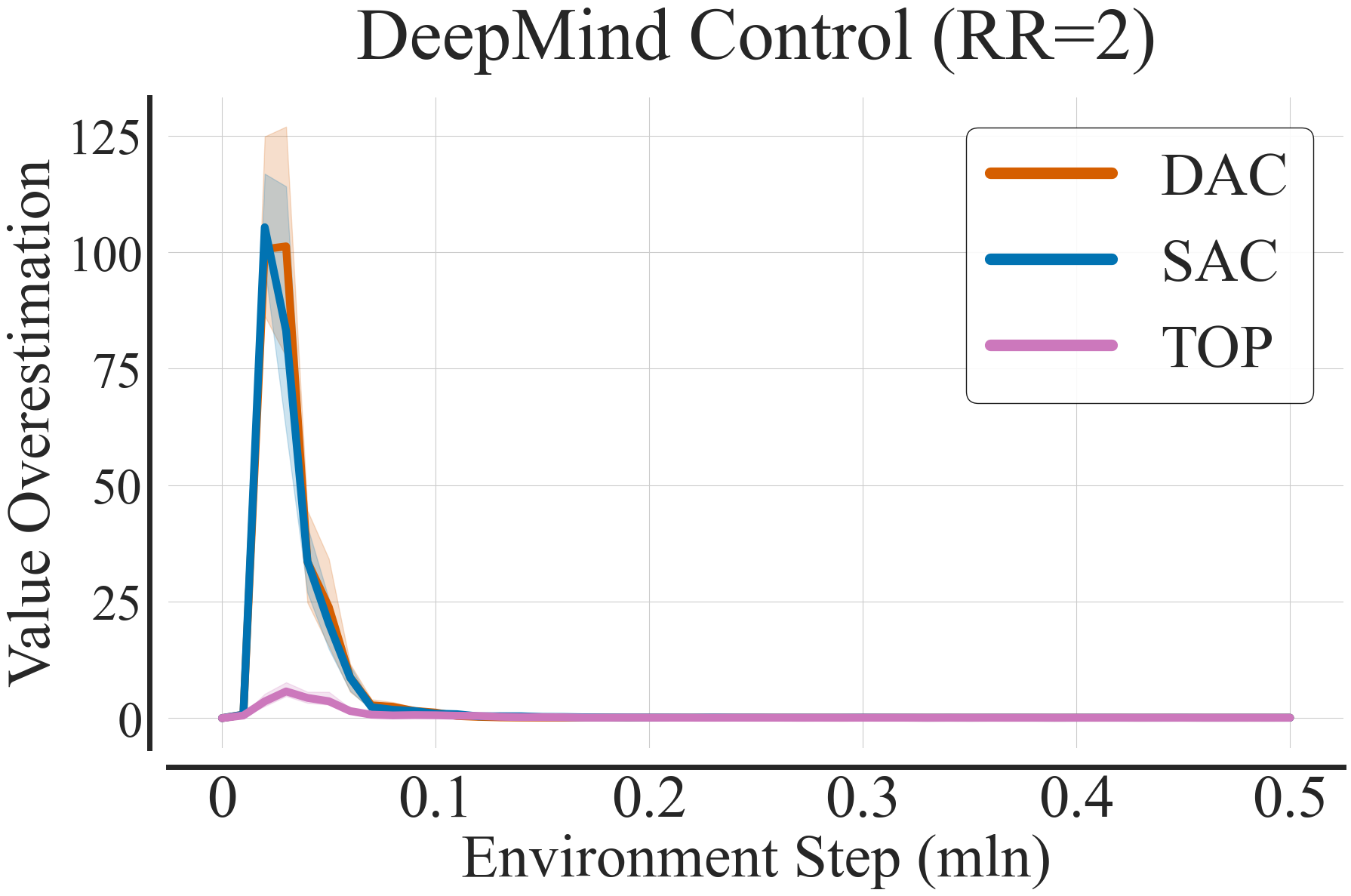}
    \hfill
    \includegraphics[width=0.31\linewidth]{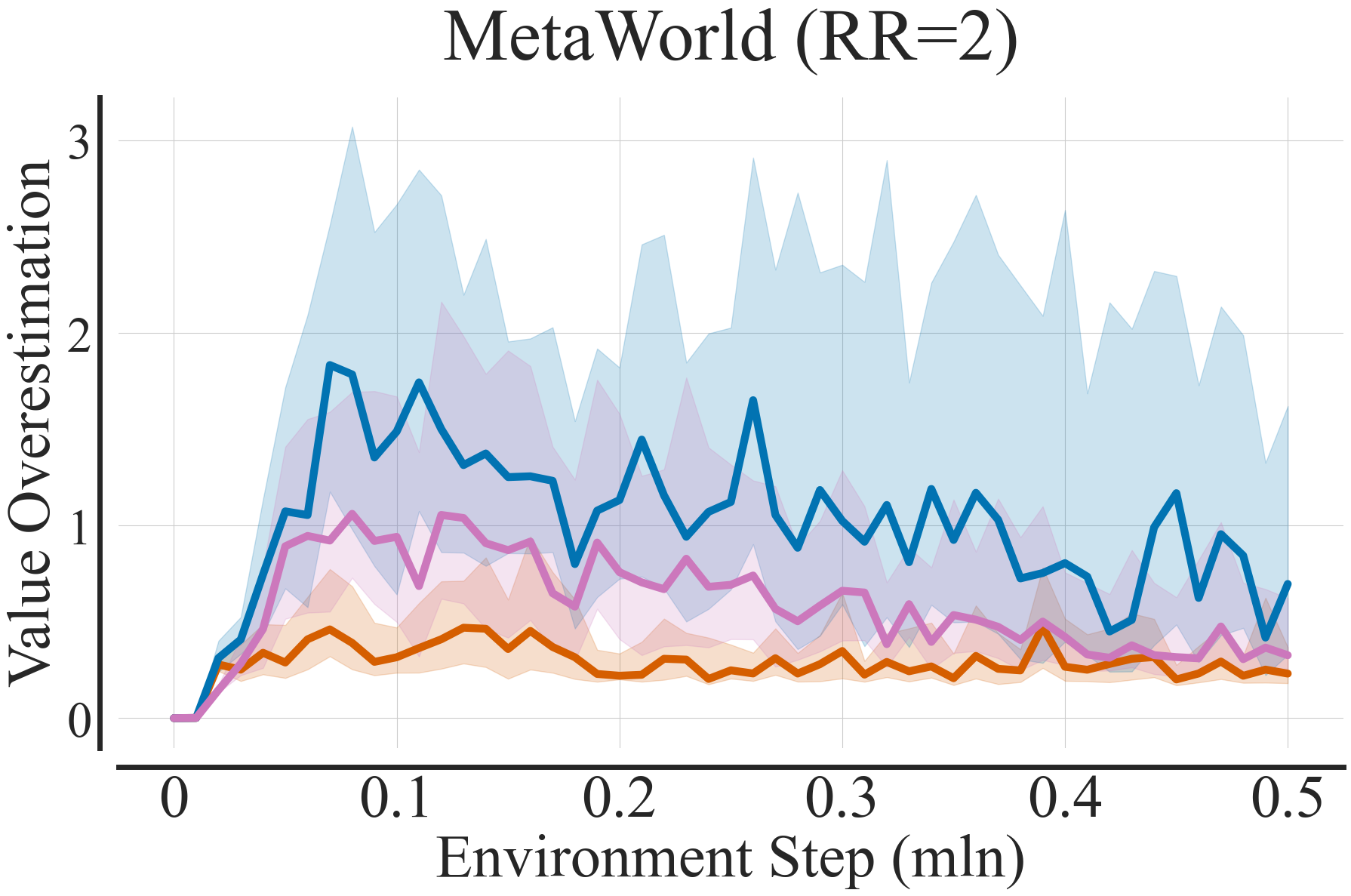}
    \hfill
    \includegraphics[width=0.31\linewidth]{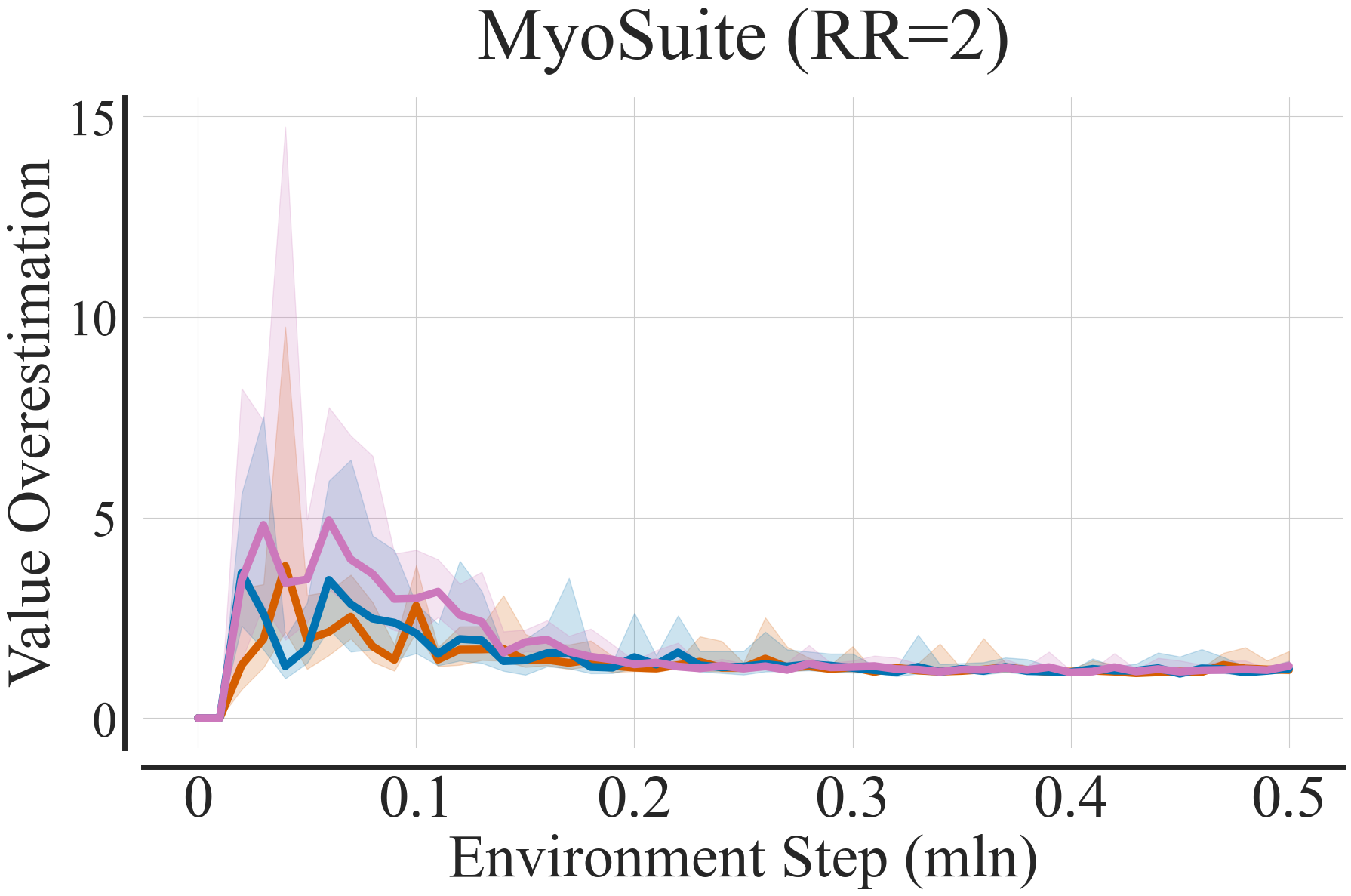}
    \end{subfigure}
\end{minipage}
\begin{minipage}[h]{1.0\linewidth}
    \begin{subfigure}{1.0\linewidth}
    \includegraphics[width=0.31\linewidth]{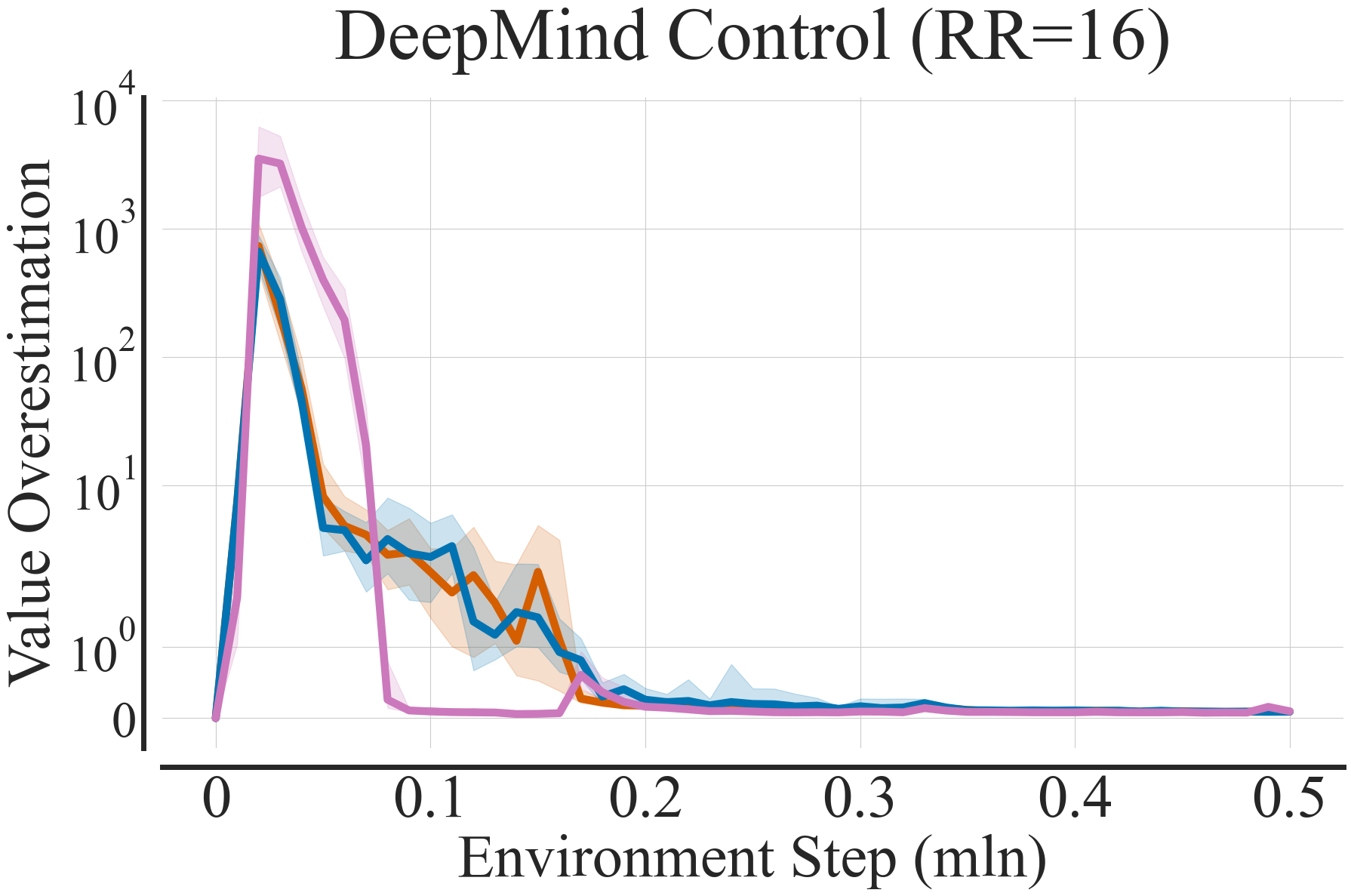}
    \hfill
    \includegraphics[width=0.31\linewidth]{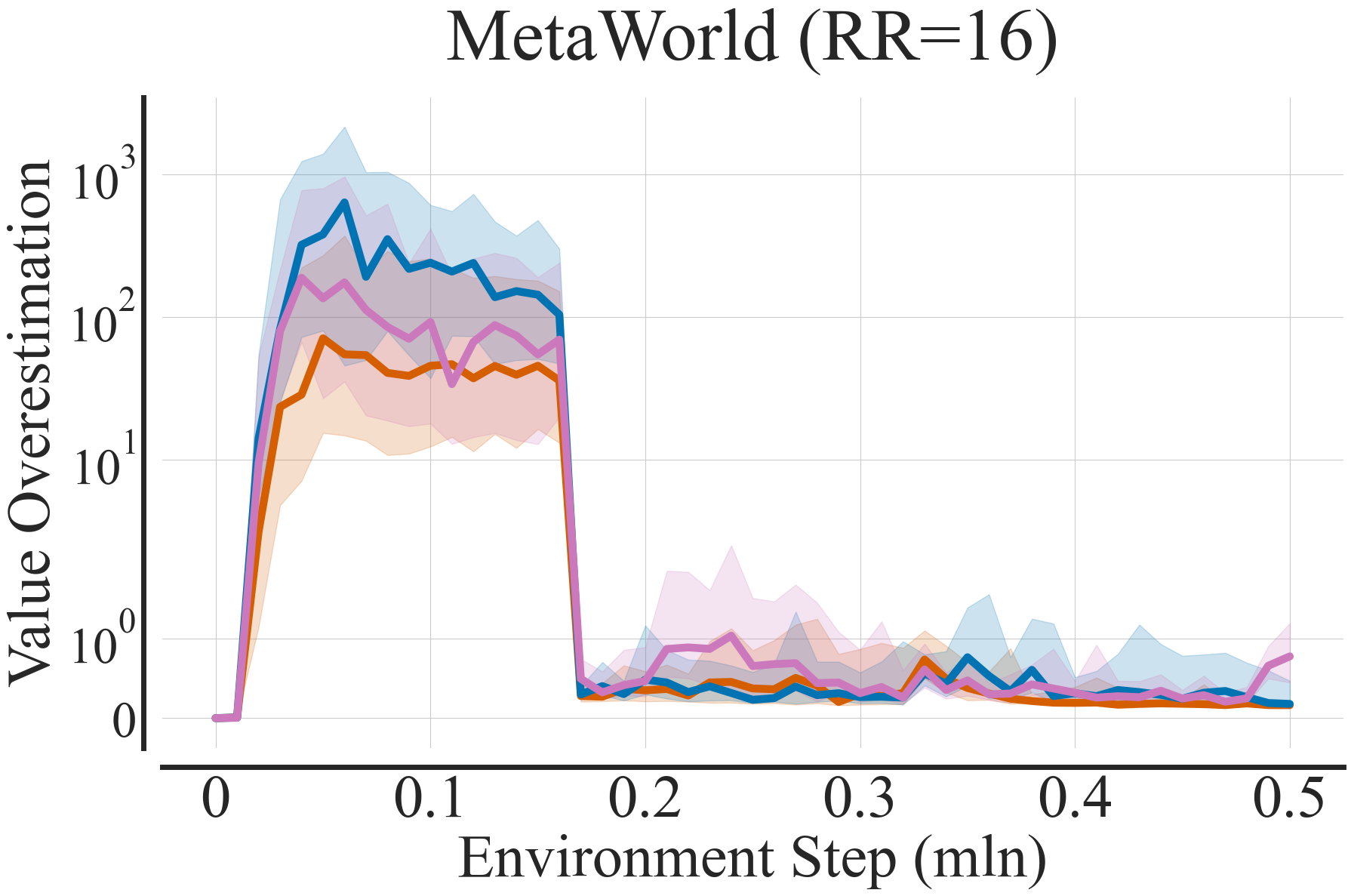}
    \hfill
    \includegraphics[width=0.31\linewidth]{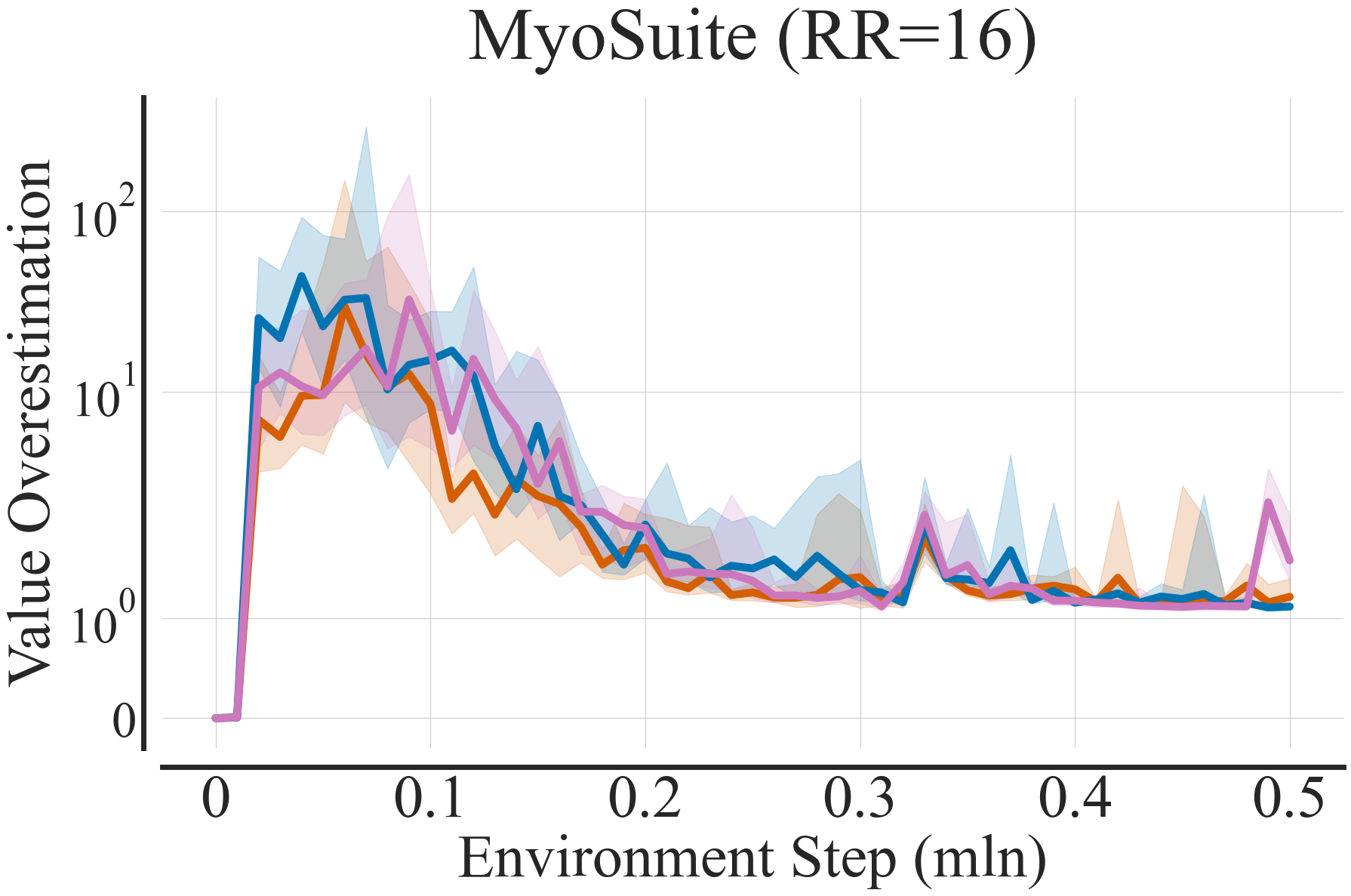}
    \end{subfigure}
\end{minipage}
\caption{We investigate overestimation in DAC, SAC and TOP. We use 15 tasks listed in Table \ref{tab:all_tasks} and run $10$ seeds for $500$k environment steps.}
\label{fig:overestimation}
\end{center}
\vspace{-0.1in} 
\end{figure}

Where $O_{\theta}$ denotes the overestimation metric, $Q_{\theta}(s,a)$ is the critic output and $G^{\pi}(s,a)$ stands for the observed empirical returns. Whereas the metric does not perfectly measure the overestimation, it allows to investigate the relative overestimation in the group of considered algorithms. As shown in Figure \ref{fig:overestimation}, we find that the dual architecture prevents the overestimation associated with optimistic TD updates and yields measurements similar to other risk-aware algorithms.  

\subsection{Distributional DAC}

As discussed in Section \ref{section:dac}, our algorithmic implementation follows SAC/TD3 in that it uses a non-distributional double critic, where each critic is taught to output the expected value of returns \citep{haarnoja2018soft, fujimoto2018addressing, chen2020randomized, kostrikov2020image}. As such, the critic ensemble disagreement measures the entire uncertainty associated with Q-value approximation. Alternatively, the critic can be implemented to learn the full distribution of returns in an approach referred to as distributional RL \cite{bellemare2017distributional, dabney2018distributional, dabney2018implicit, moskovitz2021tactical}. In this section, we investigate DAC implementation where the exploration policy is optimistic with respect to aleatoric, epistemic or total uncertainty. Following TOP \citep{moskovitz2021tactical} we implement DAC with double quantile distributional critics, where each critic outputs $K$ quantiles of the return distribution \citep{dabney2018implicit}. In such setup, the $kth$ quantile of the distribution at state-action $(s,a)$, denoted as $Q^\pi_{k}(s,a)$, is learned via optimization of Huber loss of the distributional Bellman error \cite{dabney2018implicit} defined as $\delta^{\pi}_k = Z^{\pi}(s, a) - Q^\pi_{k}(s,a)$ with $Z^{\pi}(s, a)$ representing a sample from the return distribution. We note, that following the Maximum Entropy formulation \citep{haarnoja2018soft}, we define the returns $Z^{\pi}(s, a)$ as a sum of discounted rewards and policy entropy. Furthermore, the point estimate of Q-value is calculated by $Q^{\pi}(s,a) = \frac{1}{K}\sum_{k=1}^{K} Q^\pi_{k}(s,a)$. Then, an ensemble of quantile critics can be used to explicitly model both aleatoric and epistemic uncertainties \citep{moskovitz2021tactical, lockwood2022review}. To model aleatoric and epistemic uncertainties, we use the procedure outlined in TOP \citep{moskovitz2021tactical}. As such, we model aleatoric uncertainty by learning the quantiles of the state-action return distribution, which we denote as $\mathcal{Z}^{\pi}(s,a)$. Thus, we define Q-value as $Q^{\pi} = \mathrm{E} ~ Z^{\pi} (s,a)$ with $Z^{\pi} (s,a) \sim \mathcal{Z}^{\pi}(s,a)$. The spread of the learned distribution $\mathcal{Z}^{\pi}(s,a)$ models the aleatoric uncertainty associated with the policy and the environment. Similarly, we model epistemic uncertainty by the ensemble disagreement when approximating a particular quantile of the return distribution. We consider three implementations of the optimistic policy in a distributional quantile DAC. All approaches apply the optimistic policy optimization as defined in Equation \ref{eq:upd_optiactor}, but differ in how the point estimate of Q-value upper-bound is calculated:

Optimistic policy wrt. epistemic uncertainty - In this setup, for a fixed policy $\pi$, we define the Q-value upper bound as $Q^{\beta^{o}}(s,a) = \frac{1}{K}\sum_{k=1}^{K} Q^{o}_{k}(s,a)$ and $Q^{o}_{k}(s,a) = Q^{\mu}_{k}(s,a) + \beta^{o} Q^{\sigma}_{k}(s,a)$, where $Q^{\mu}_{k}$ and $Q^{\sigma}$ represent the ensemble mean and standard deviation in $kth$ quantile. As such, the optimistic policy promotes actions with high epistemic uncertainty measured as critic ensemble disagreement in approximation of the $kth$ quantile.

Optimistic policy wrt. aleatoric uncertainty - Here, for a given policy, we calculate the Q-value upper bound as $Q^{\beta^{o}}(s,a) = \frac{1}{K}\sum_{k=1}^{K} Q^{\mu}_{k}(s,a) + \beta^{o} \sigma_{Z}(s,a)$ where $\sigma_{Z}(s,a)$ is the standard deviation of $K$ quantile outputs, averaged over the critic ensemble. As such, the optimistic policy promotes actions with high aleatoric uncertainty measured as the spread of the return distribution.

Optimistic policy wrt. both uncertainties - Finally, we consider an optimistic policy that promotes both aleatoric and epistemic uncertainties, albeit still explicitly modeled via an ensemble of quantile critics. Here, for a given policy we calculate the upper-bound Q-value by $Q^{\beta^{o}}(s,a) = \frac{1}{K}\sum_{k=1}^{K} Q^{o}_{k}(s,a) + \beta^{o} \sigma_{Z}(s,a)$, with $Q^{o}_{k}(s,a) = Q^{\mu}_{k}(s,a) + \beta^{o} Q^{\sigma}_{k}(s,a)$. As such, the optimistic policy promotes actions that are associated with both epistemic and aleatoric uncertainty.

\begin{figure}[ht!]
\begin{center}
\begin{minipage}[h]{1.0\linewidth}
\centering
    \begin{subfigure}{0.37\linewidth}
    \includegraphics[width=\textwidth]{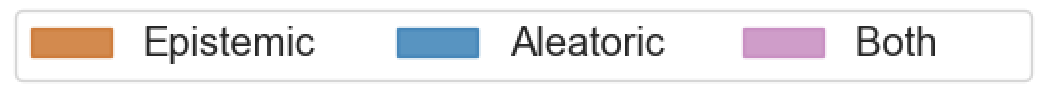}
    \end{subfigure}
\end{minipage}
\begin{minipage}[h]{1.0\linewidth}
    \begin{subfigure}{1.0\linewidth}
    \includegraphics[width=0.195\linewidth]{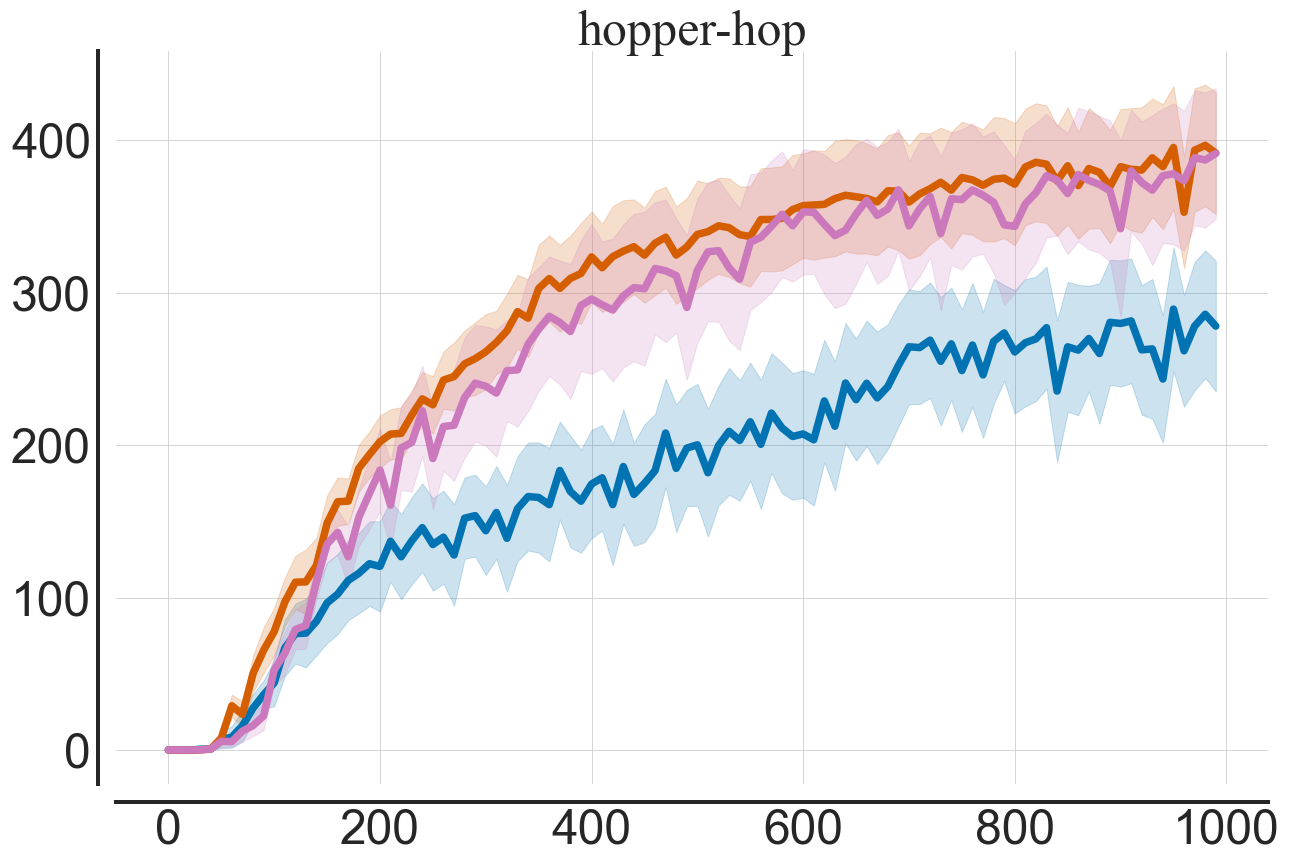}
    \hfill
    \includegraphics[width=0.195\linewidth]{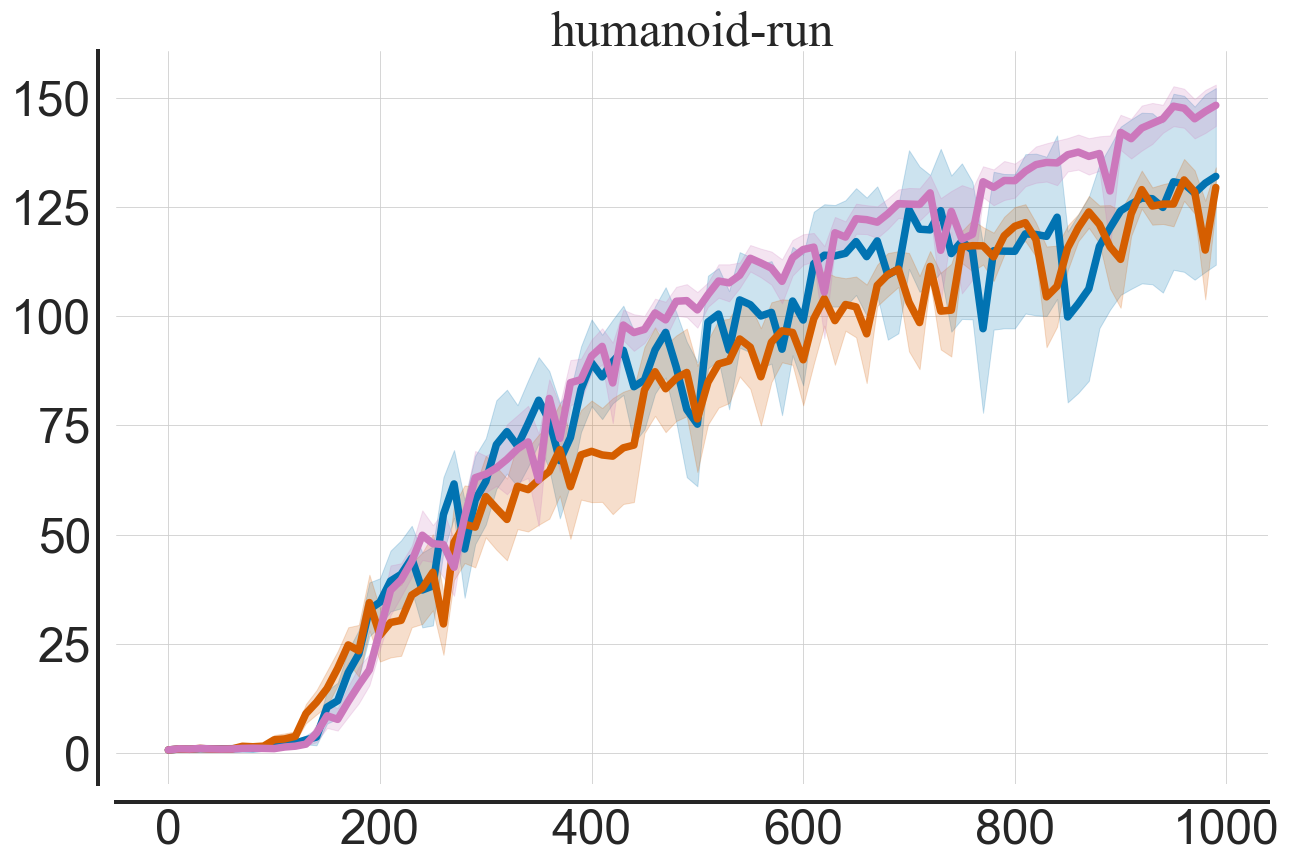}
    \hfill
    \includegraphics[width=0.195\linewidth]{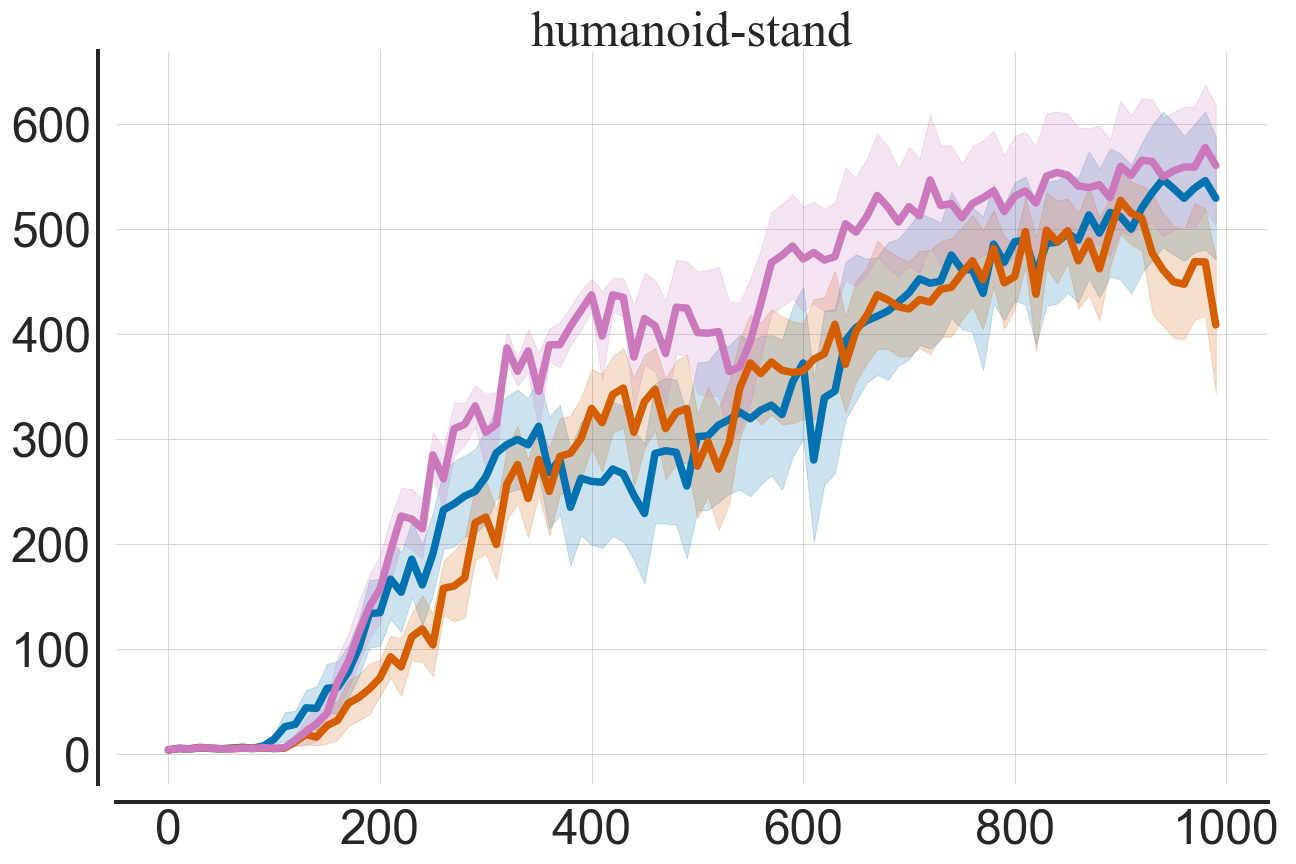}
    \hfill
    \includegraphics[width=0.195\linewidth]{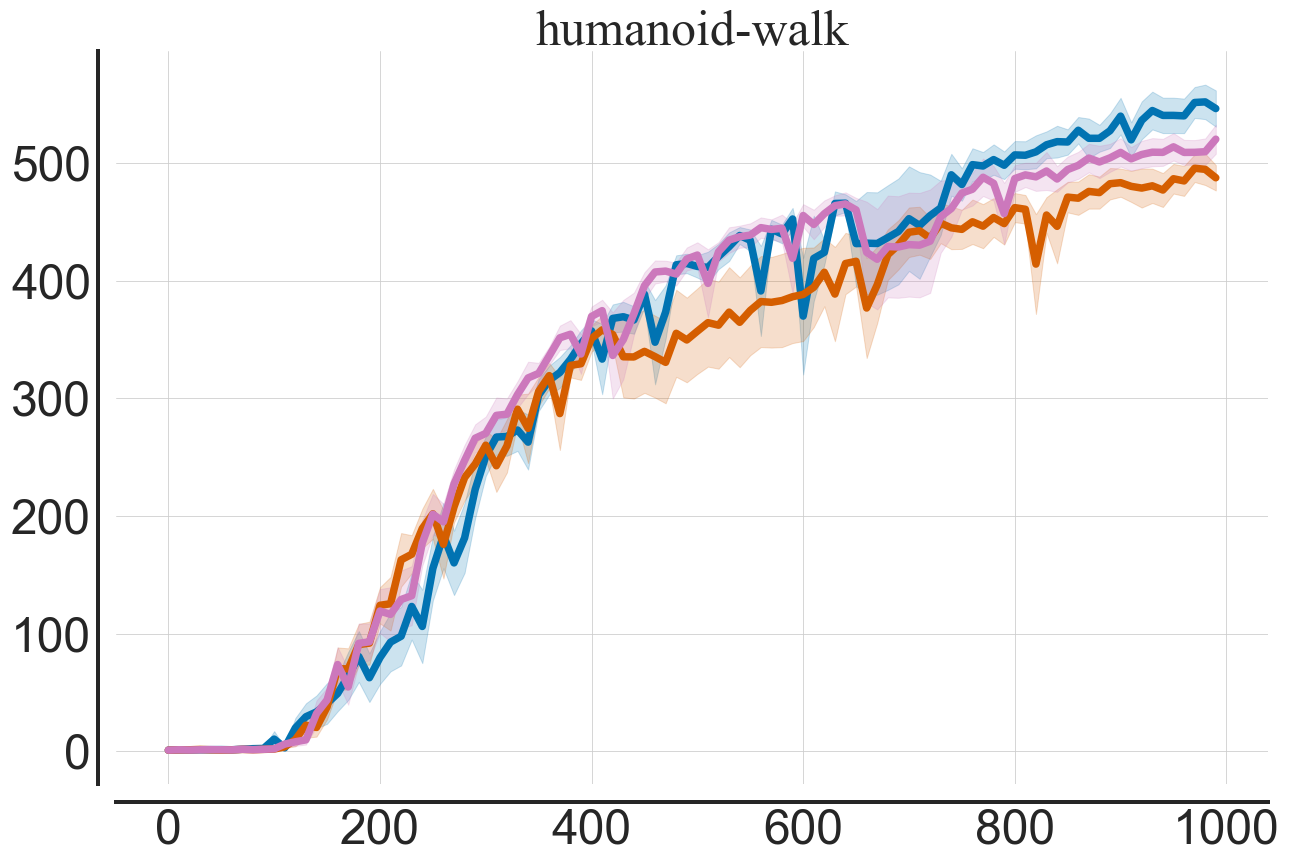}
    \hfill
    \includegraphics[width=0.195\linewidth]{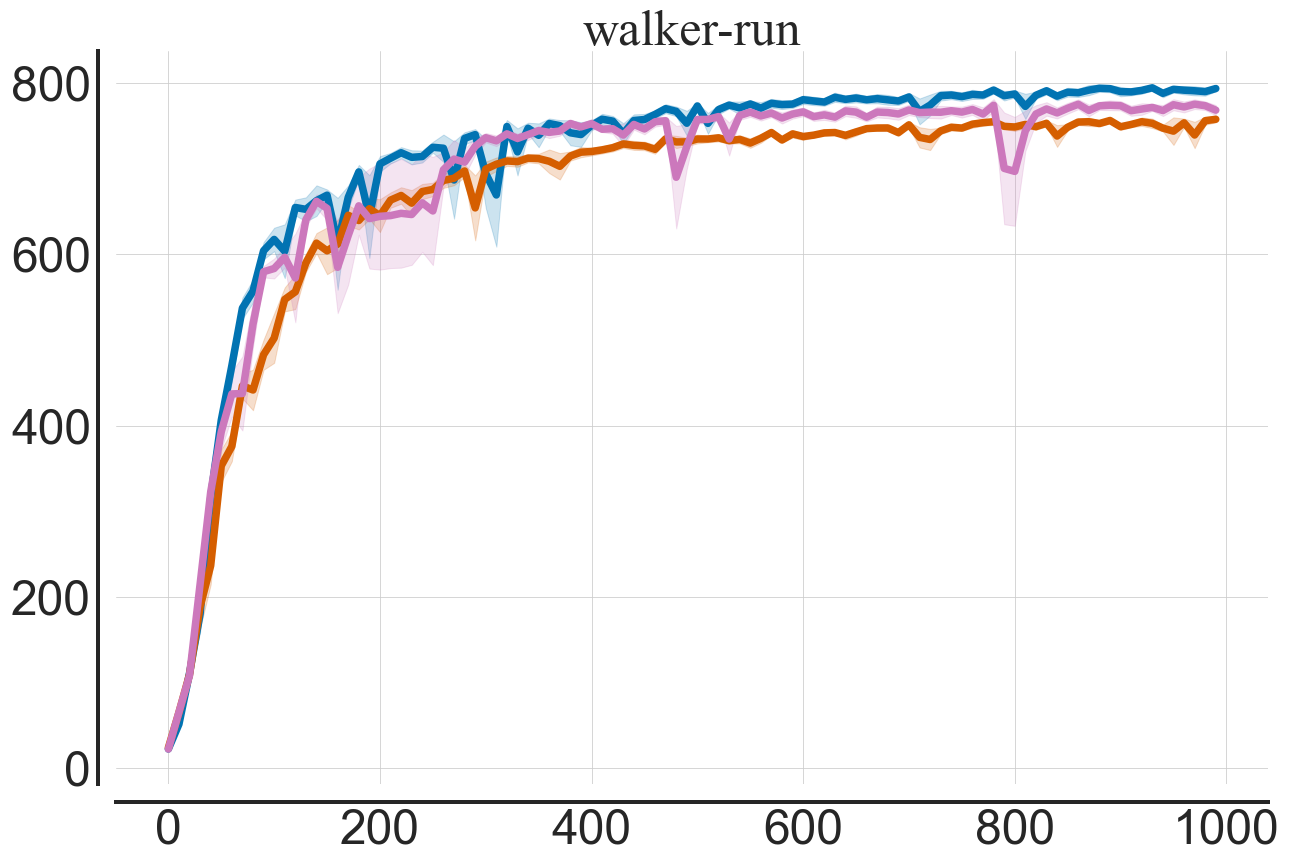}
    \end{subfigure}
\end{minipage}
\begin{minipage}[h]{1.0\linewidth}
    \begin{subfigure}{1.0\linewidth}
    \includegraphics[width=0.195\linewidth]{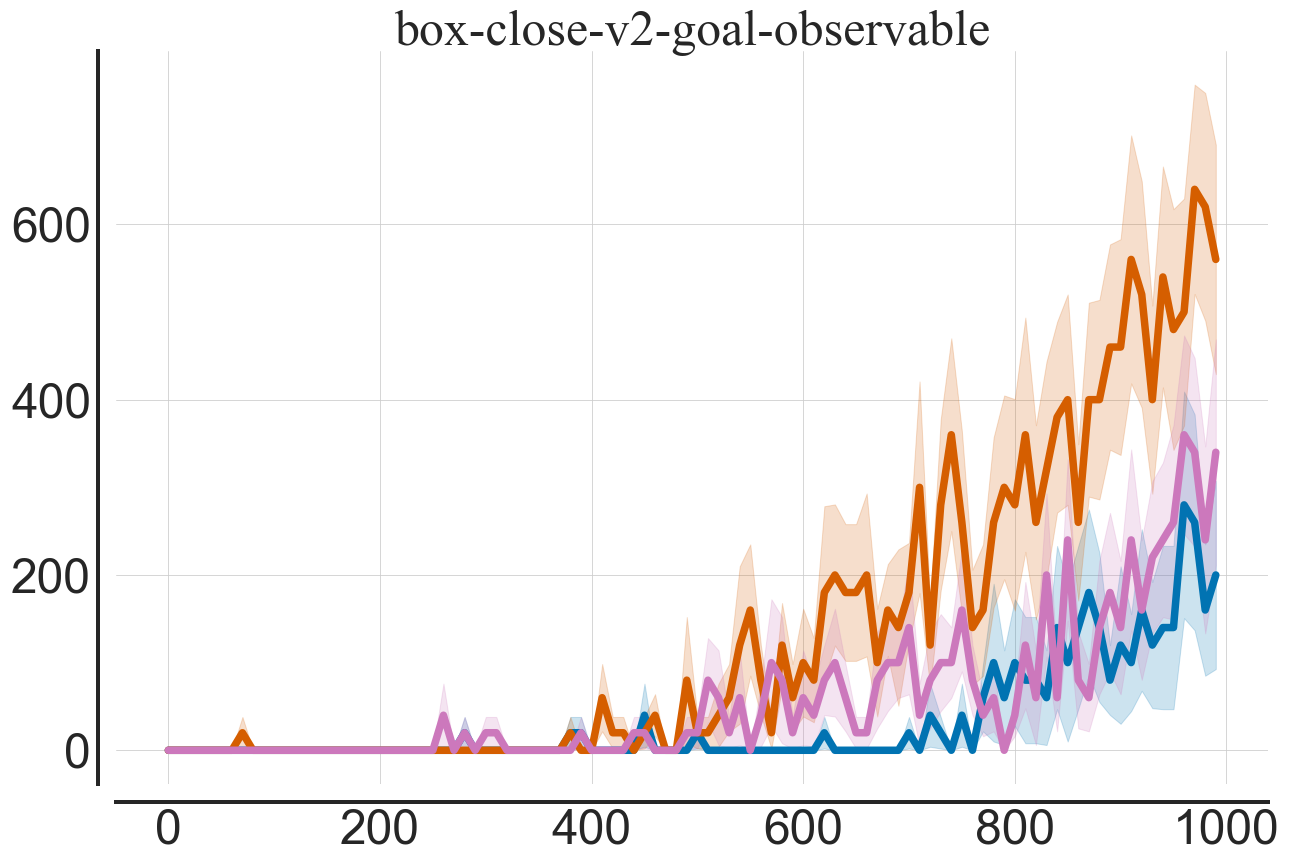}
    \hfill
    \includegraphics[width=0.195\linewidth]{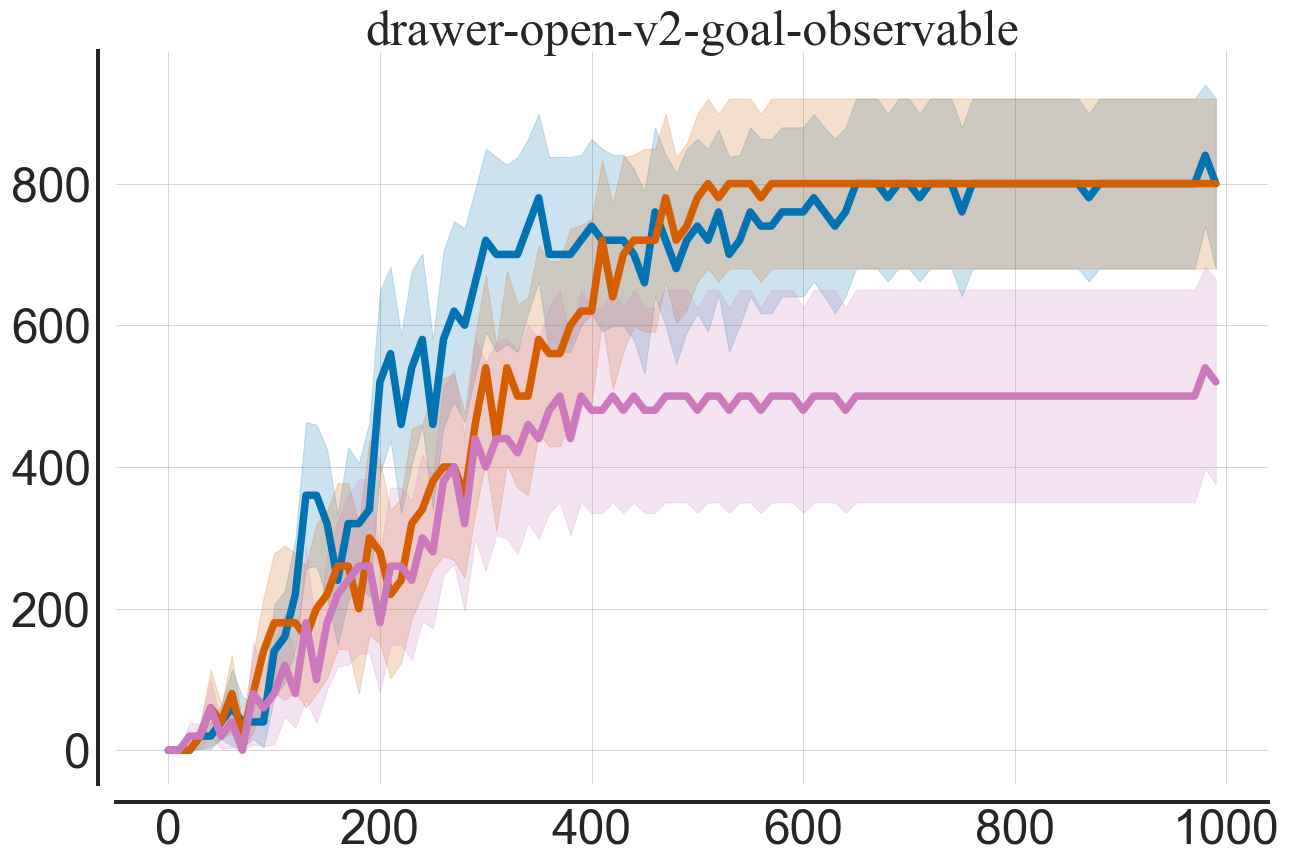}
    \hfill
    \includegraphics[width=0.195\linewidth]{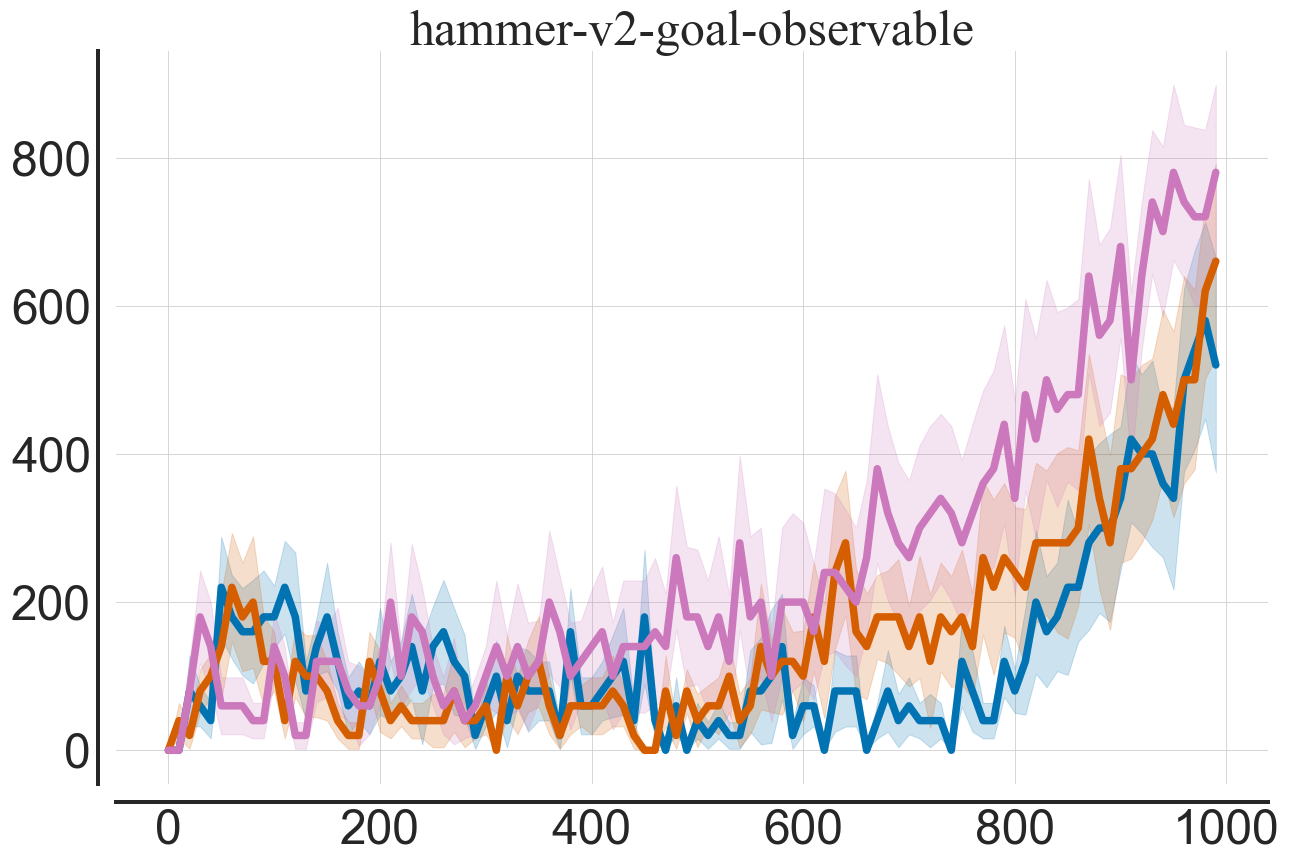}
    \hfill
    \includegraphics[width=0.195\linewidth]{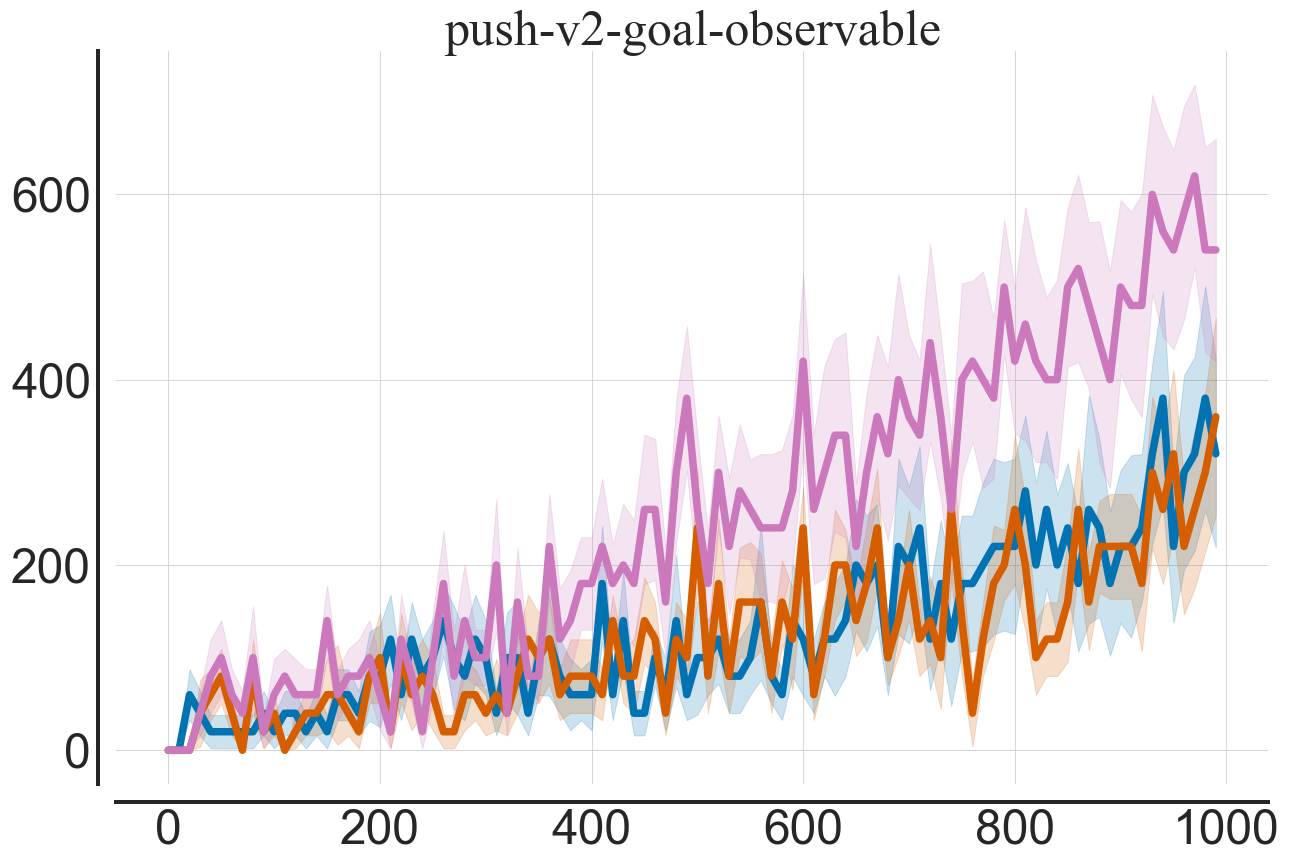}
    \hfill
    \includegraphics[width=0.195\linewidth]{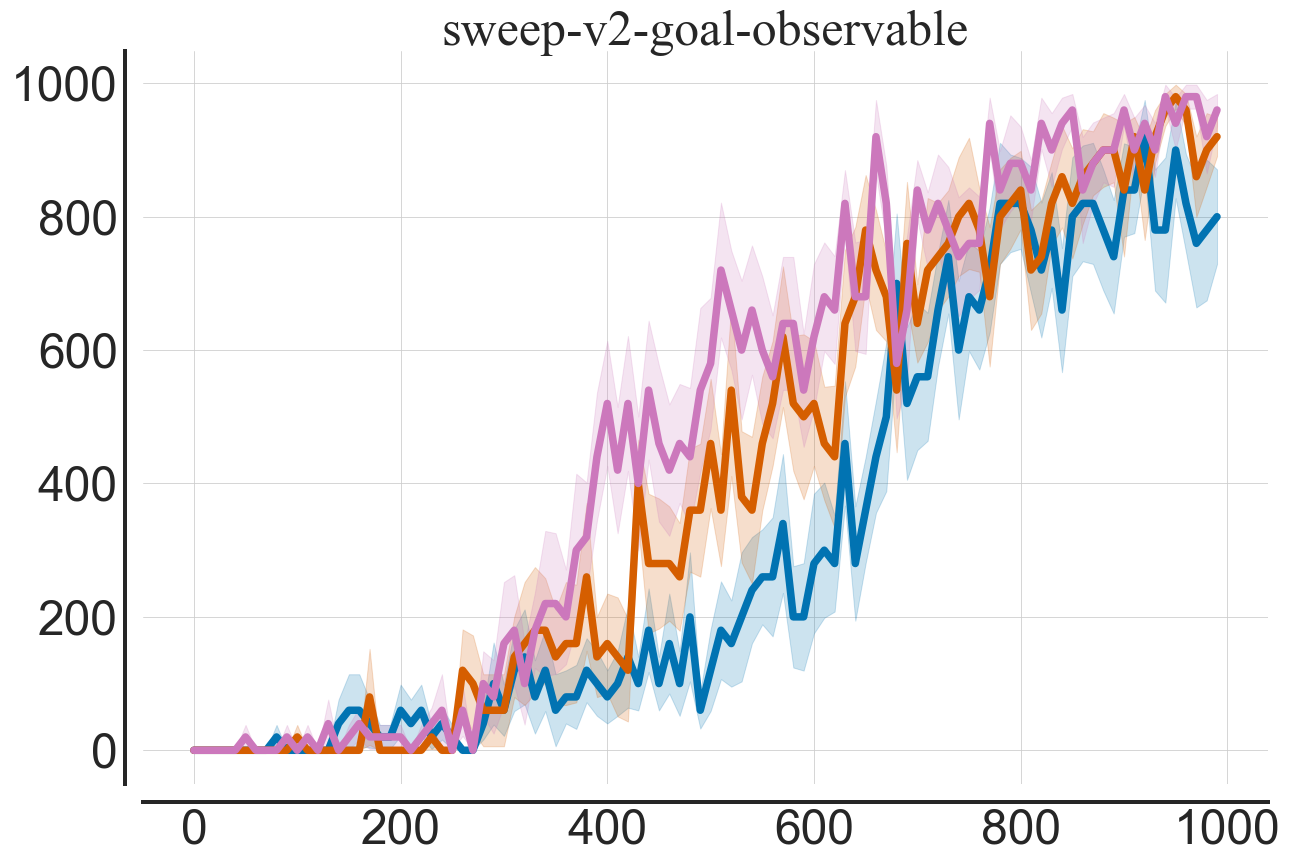}
    \end{subfigure}
\end{minipage}
\caption{We run three variations of distributional DAC ($RR=2$) on $10$ tasks from DMC and MW benchmarks. $Y$-axis reports IQM and $X$-axis reports environment steps. 10 seeds per task.}
\label{fig:distributional1}
\end{center}
\vspace{-0.1in} 
\end{figure}

\begin{figure}[ht!]
\begin{center}
\begin{minipage}[h]{1.0\linewidth}
\centering
    \begin{subfigure}{0.37\linewidth}
    \includegraphics[width=\textwidth]{images/appendix/distributional/rebuttal_legend.png}
    \end{subfigure}
\end{minipage}
\begin{minipage}[h]{1.0\linewidth}
    \begin{subfigure}{1.0\linewidth}
    \includegraphics[width=0.195\linewidth]{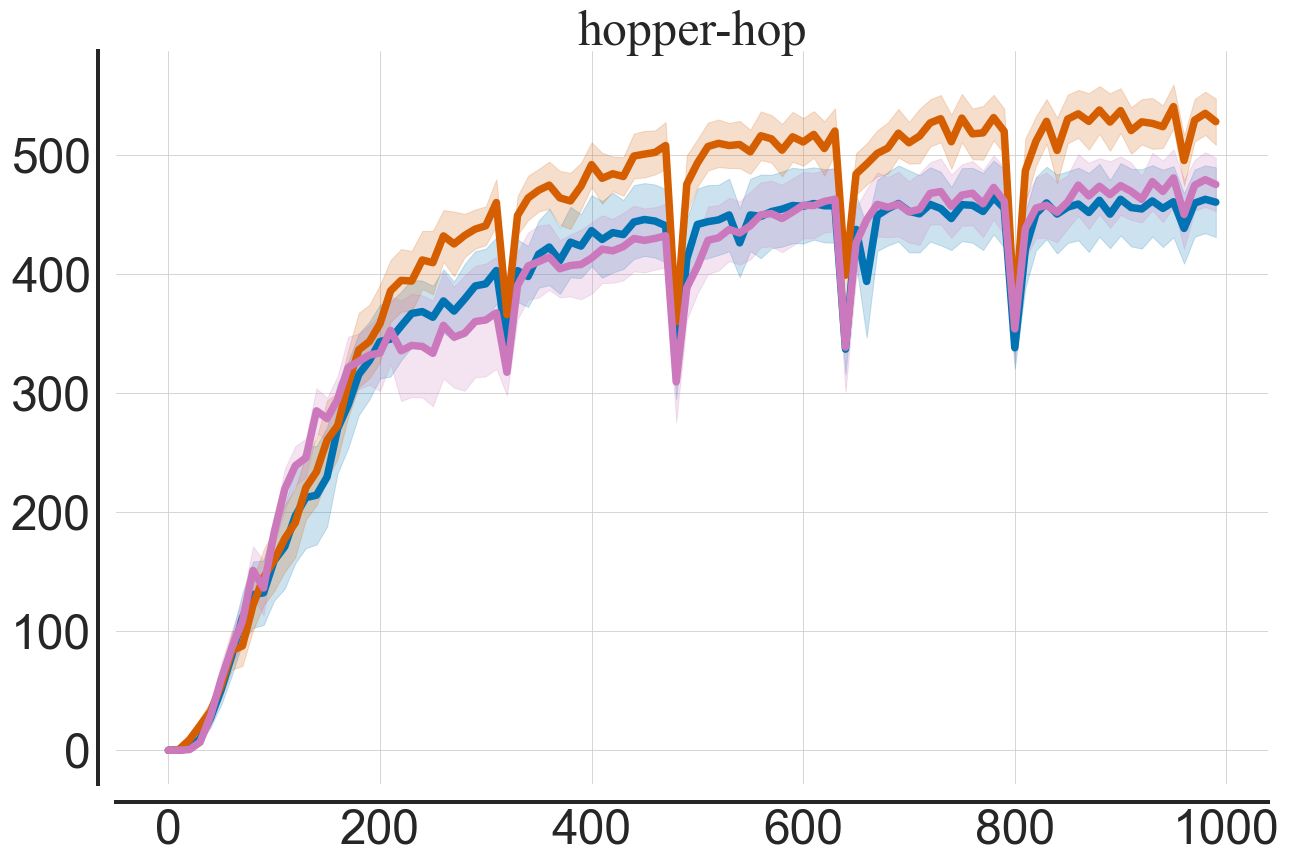}
    \hfill
    \includegraphics[width=0.195\linewidth]{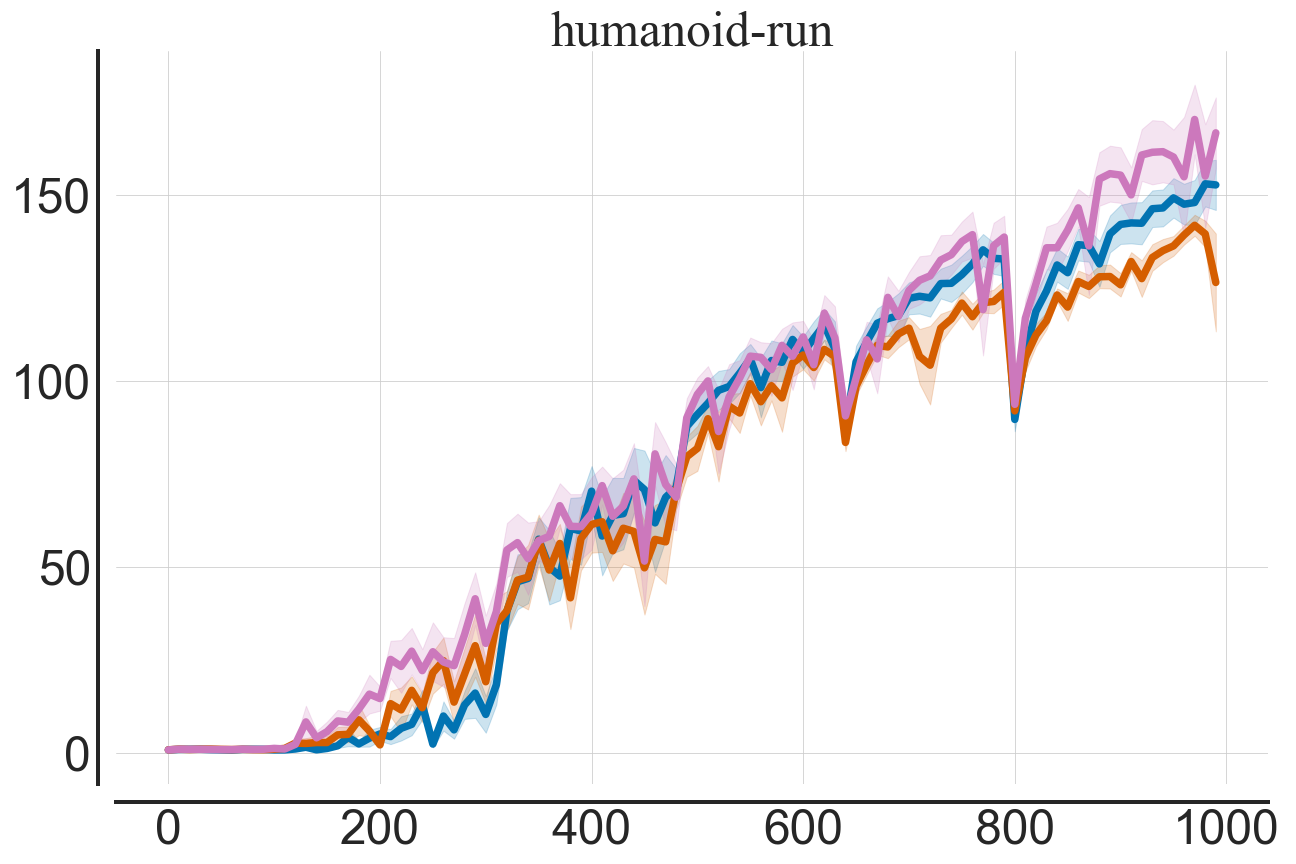}
    \hfill
    \includegraphics[width=0.195\linewidth]{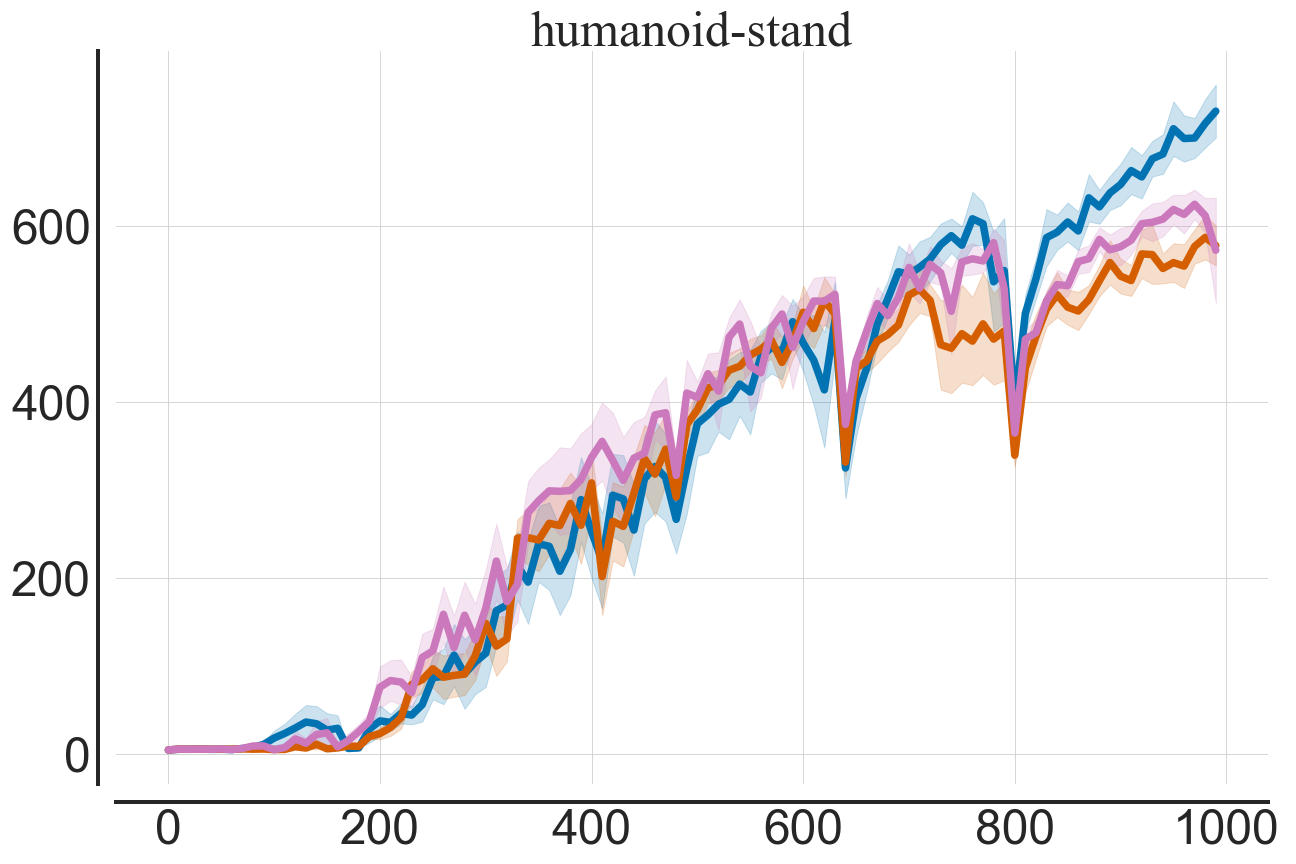}
    \hfill
    \includegraphics[width=0.195\linewidth]{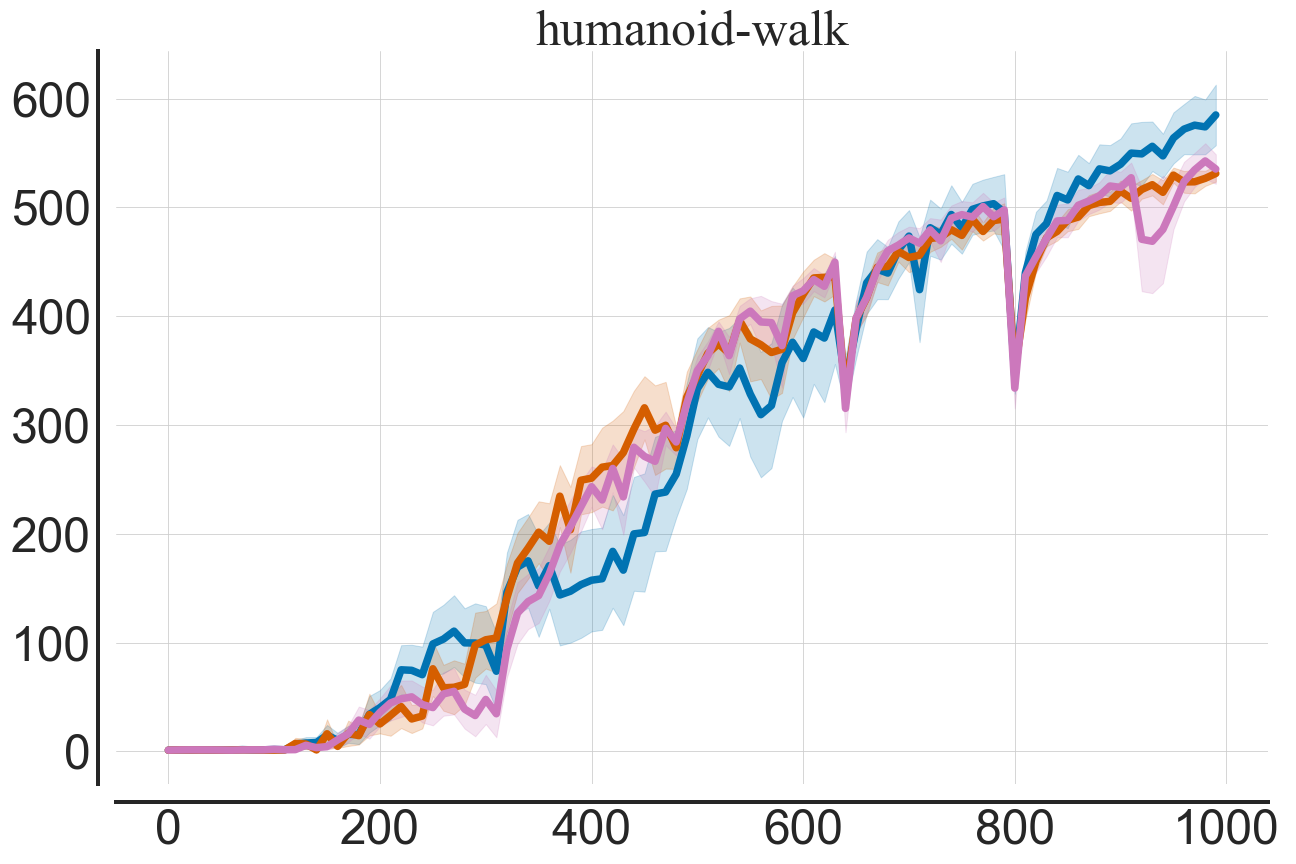}
    \hfill
    \includegraphics[width=0.195\linewidth]{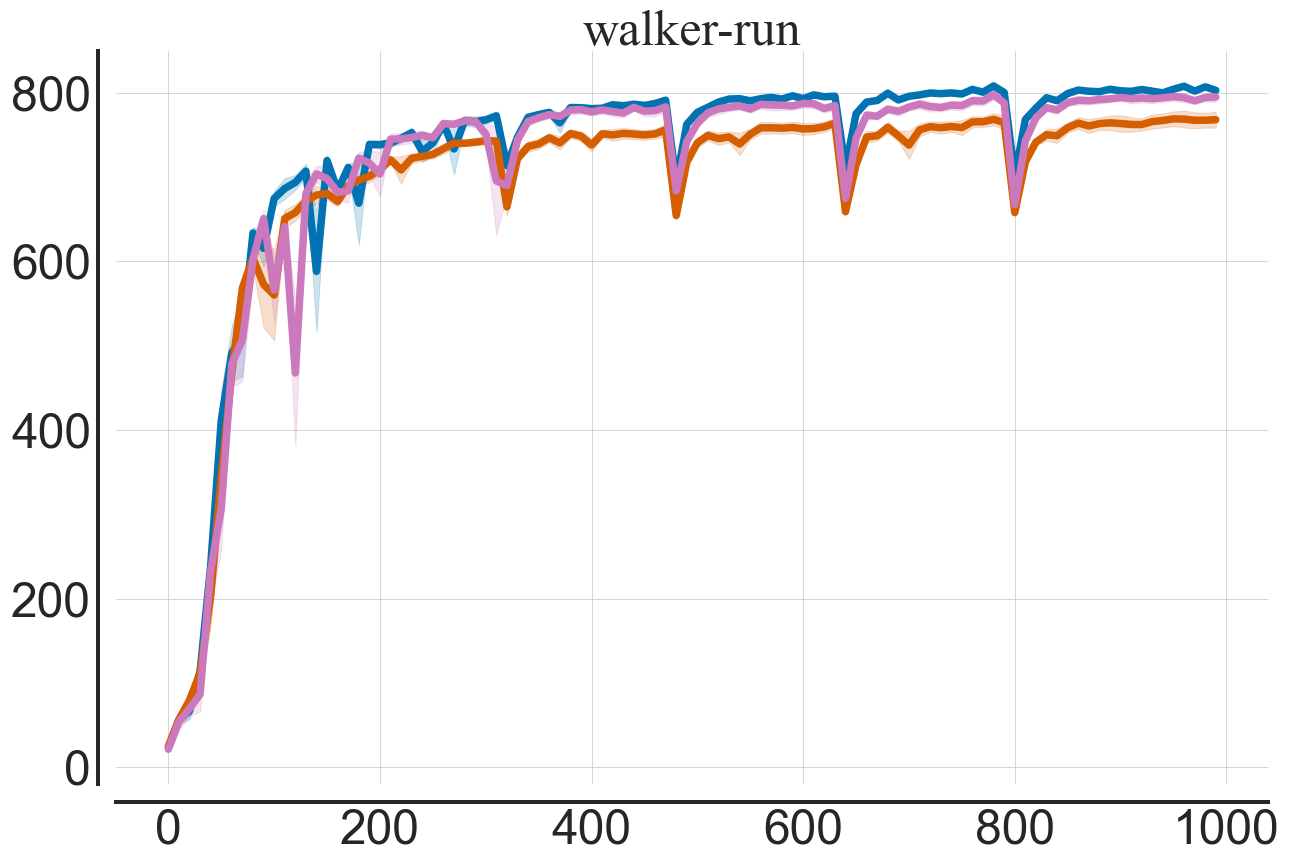}
    \end{subfigure}
\end{minipage}
\begin{minipage}[h]{1.0\linewidth}
    \begin{subfigure}{1.0\linewidth}
    \includegraphics[width=0.195\linewidth]{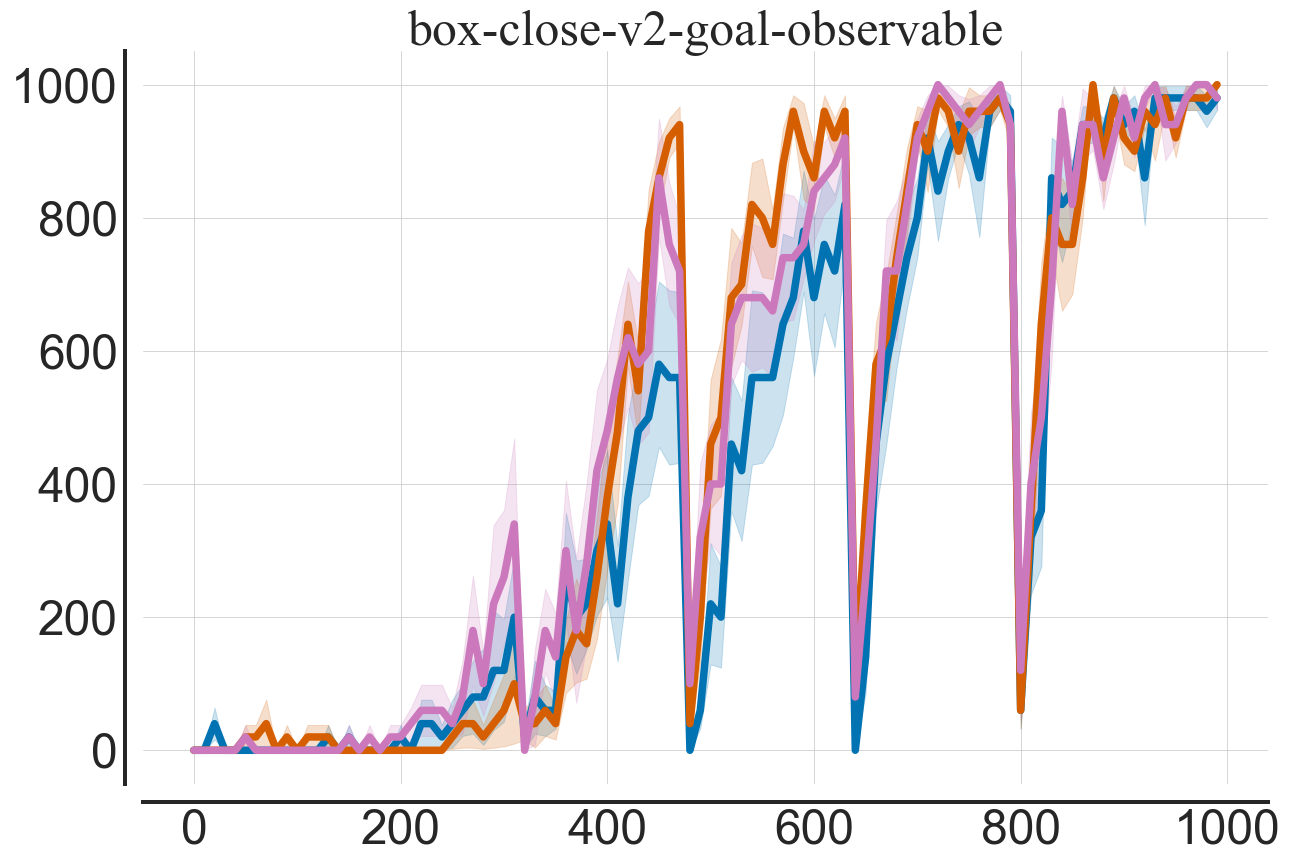}
    \hfill
    \includegraphics[width=0.195\linewidth]{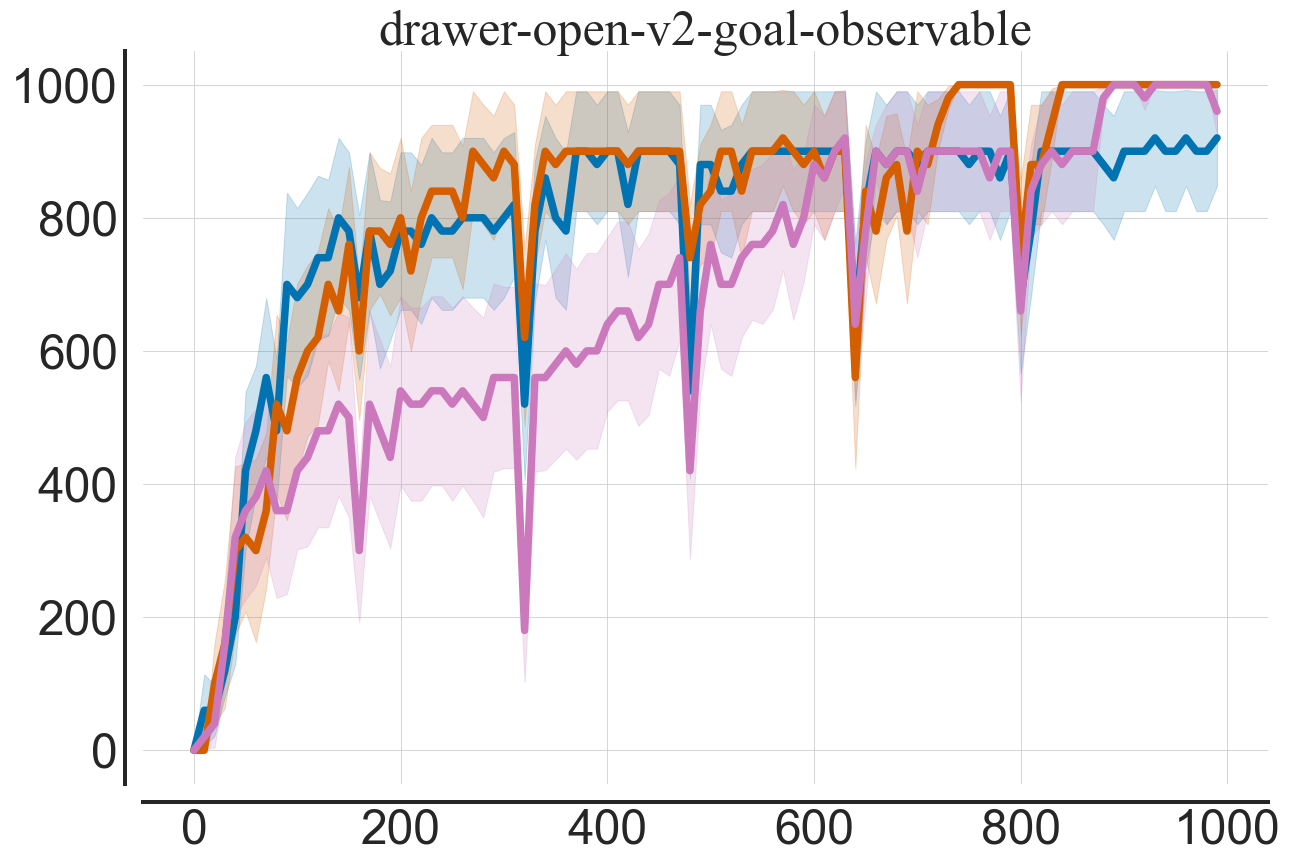}
    \hfill
    \includegraphics[width=0.195\linewidth]{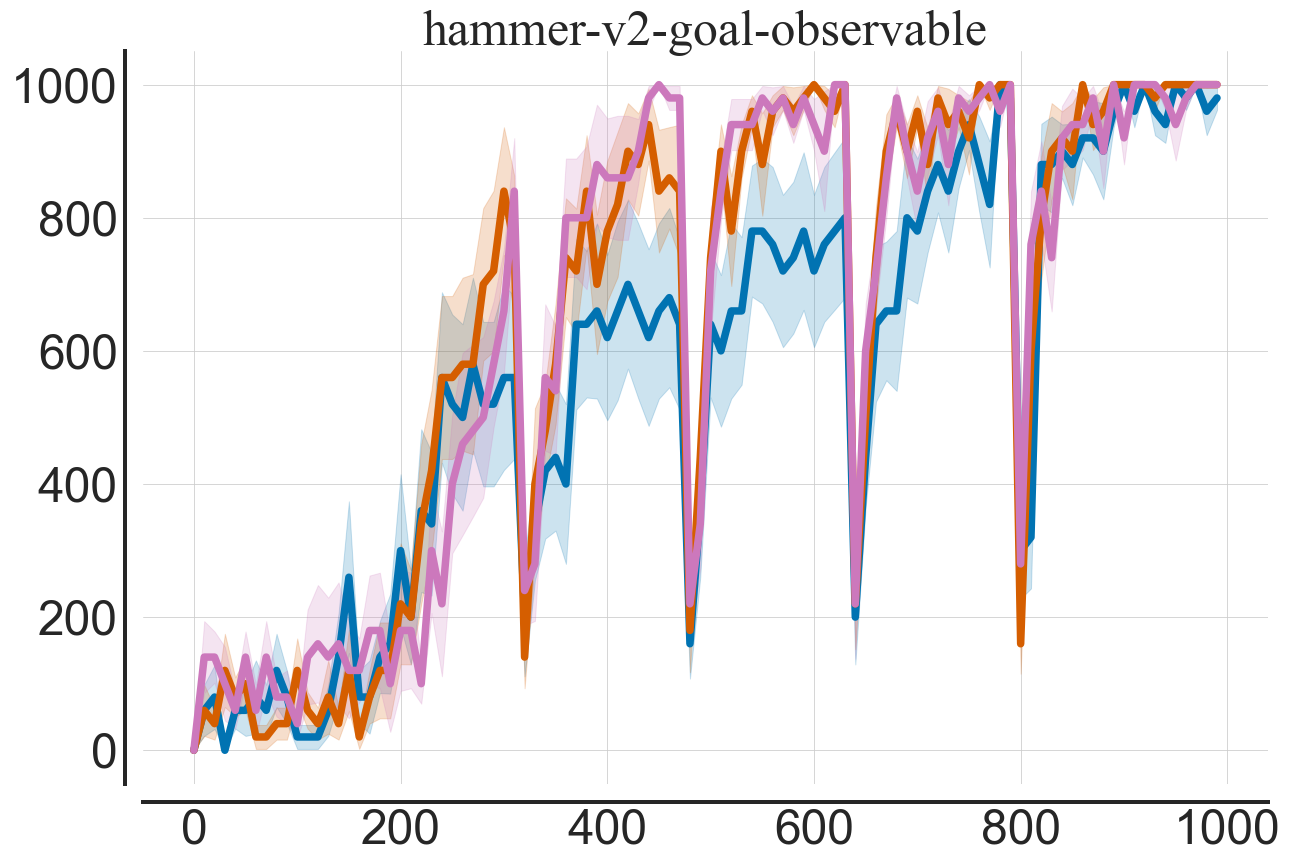}
    \hfill
    \includegraphics[width=0.195\linewidth]{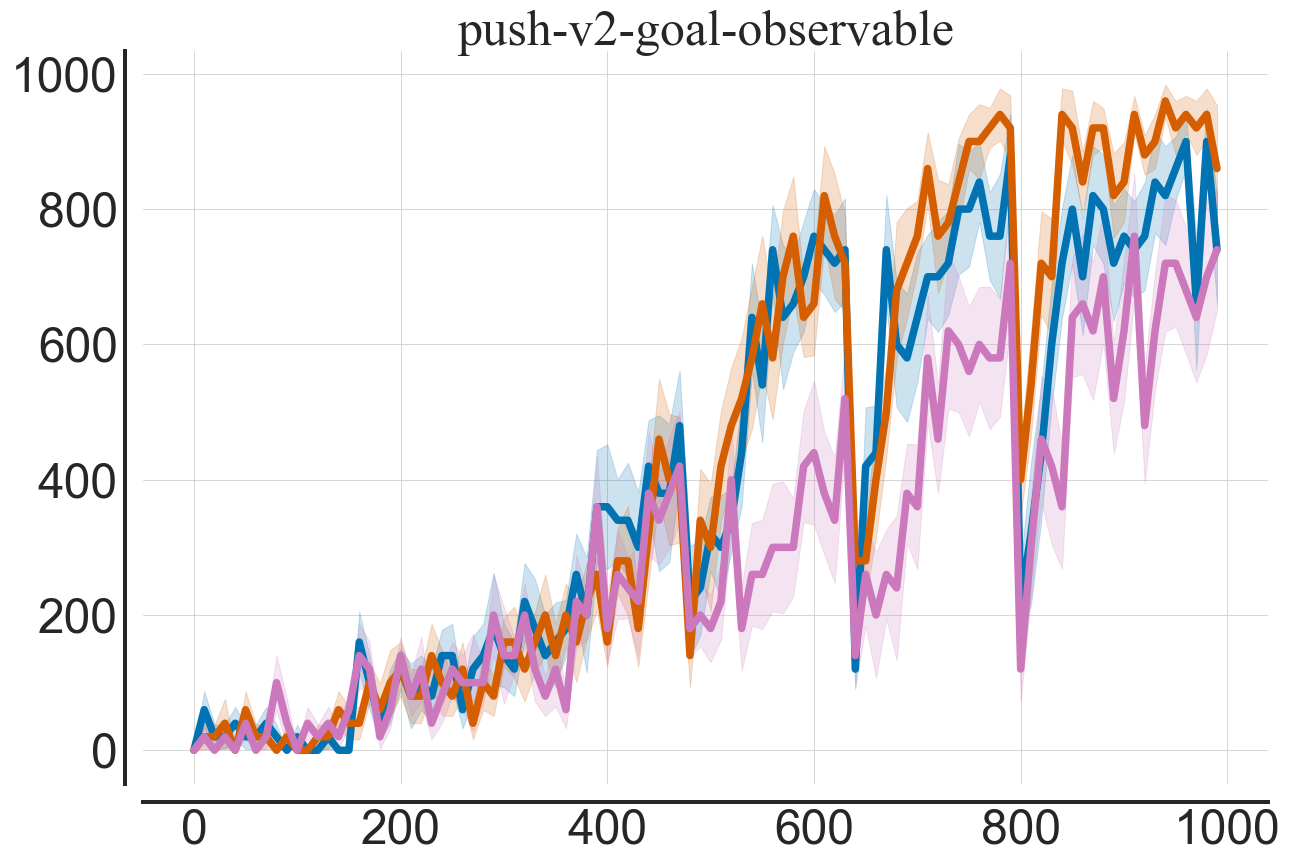}
    \hfill
    \includegraphics[width=0.195\linewidth]{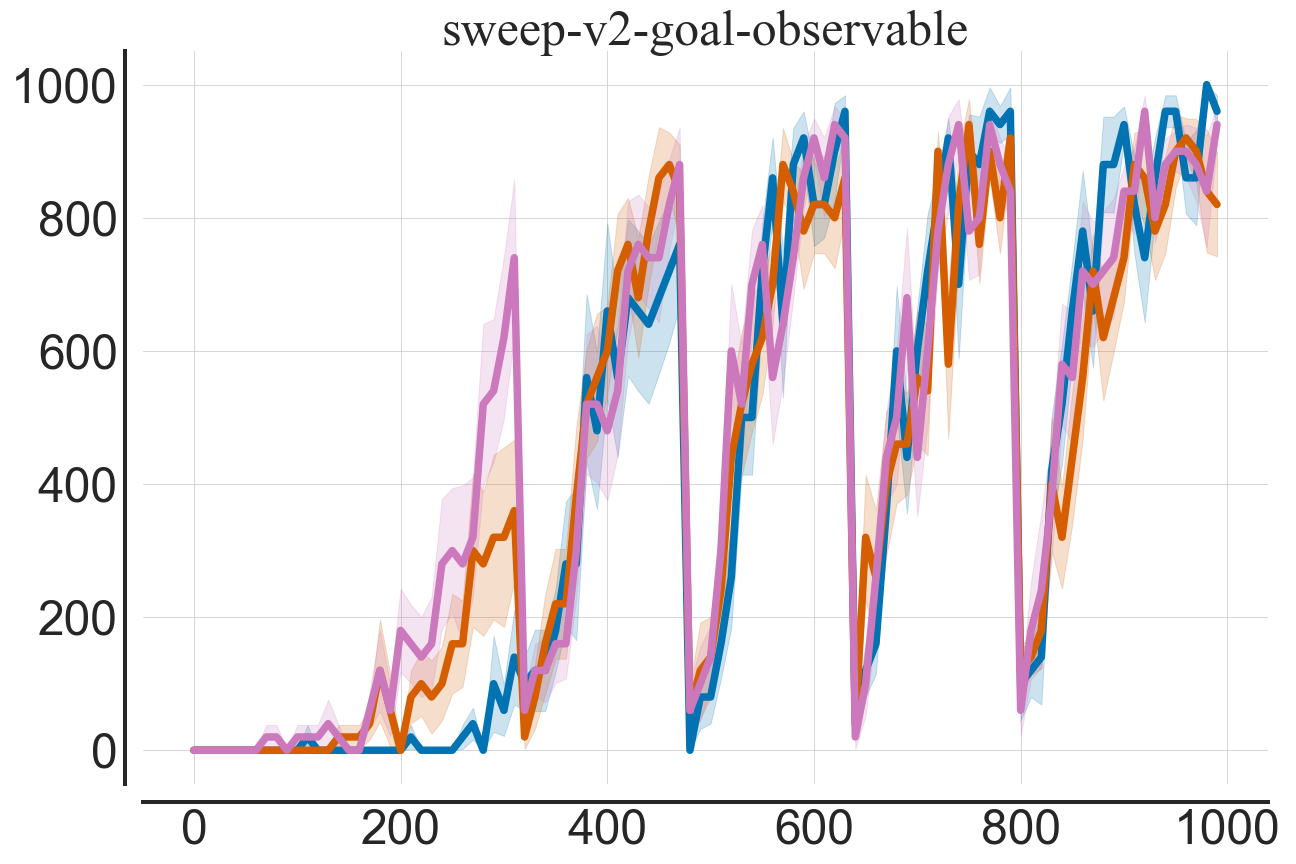}
    \end{subfigure}
\end{minipage}
\caption{We run three variations of distributional DAC ($RR=16$) on $10$ tasks from DMC and MW benchmarks. $Y$-axis reports IQM and $X$-axis reports environment steps. 10 seeds per task.}
\label{fig:distributional2}
\end{center}
\vspace{-0.1in} 
\end{figure}

We run the three variations of Distributional Quantile DAC for $1$mln environment steps and both replay regimes. The results are presented in the Figures below. We find that all variations achieve similar performance, with the standard implementation promoting the epistemic uncertainty slightly outperforming other variations.

\section{Experimental Details}
\label{appendix:experimental_setting}

\subsection{Environments}

The table below lists task used in the main experiment.

\begin{table*}[ht!]
\caption{$30$ tasks used for the evaluation of model-free algorithms in low and high replay ration regimes. The $15$ bold tasks are used for the hyperparameter sensitivity and design choices ablation.}
\vspace{0.1in}
\label{tab:all_tasks}
\begin{center}
{\renewcommand{\arraystretch}{1.05}%
 \begin{tabular}{||c|c|c||} 
 \hline \hline
 \textsc{DeepMind Control} & \textsc{MetaWorld} & \textsc{MyoSuite} \\
 \hline \hline
 \textsc{\textbf{Acrobot Swingup}} & \textsc{Assembly} & \textsc{\textbf{Reach Easy}} \\
 \textsc{\textbf{Cheetah Run}} & \textsc{\textbf{Box Close}} & \textsc{Reach Hard} \\
 \textsc{\textbf{Hopper Hop}} & \textsc{Coffee Pull} & \textsc{\textbf{Pose Easy}} \\
 \textsc{Humanoid Run} & \textsc{\textbf{Drawer Open}} & \textsc{Pose Hard} \\
 \textsc{\textbf{Humanoid Stand}} & \textsc{\textbf{Hammer}} & \textsc{\textbf{Pen Twirl Easy}} \\
 \textsc{\textbf{Humanoid Walk}} & \textsc{Lever Pull} & \textsc{Pen Twirl Hard} \\
 \textsc{Swimmer Swimmer6} & \textsc{\textbf{Push}} & \textsc{\textbf{Object Hold Easy}} \\
 \textsc{Pendulum Swingup} & \textsc{Stick Pull} & \textsc{Object Hold Hard} \\
 \textsc{Quadruped Run} & \textsc{Stick Push} & \textsc{\textbf{Key Turn Easy}} \\
 \textsc{Walker Run} & \textsc{\textbf{Sweep}} & \textsc{Key Turn Hard} \\
 \hline \hline
\end{tabular}}
\end{center}
\end{table*}

\subsection{Baseline Algorithms}

We standardize the common hyperparameters between all algorithms to minimize the differences between all implementations. We align the basic values with the state-of-the-art SAC implementation \citep{d2022sample}, thus limiting the requirement for extensive hyperparameter search across algorithms. We employ uniform network architectures and a standard ensemble of two critics \citep{fujimoto2018addressing, haarnoja2018soft, ciosek2019better, moskovitz2021tactical, cetin2023learning}. The set of tested algorithms includes:

\begin{itemize}
    \item DAC - Dual Actor-Critic. The approach proposed in this paper, described in detail in Section \ref{section:dac} and Appendix \ref{appendix:dac}.
    \item SAC - Soft Actor-Critic \citep{haarnoja2018soft} builds on DDPG \citep{silver2014deterministic}. SAC extends the DDPG algorithm with features such as a stochastic policy using the reparametrization trick, automatic entropy temperature adjustment, Clipped Double Q-learning, and maximum entropy updates for actor and critic networks.
    \item TD3 - Twin Deep Deterministic Policy Gradient \citep{fujimoto2018addressing} is an extension of the DDPG algorithm, including a target actor network and Clipped Double Q-learning. 
    \item OAC - Optimistic Actor-Critic \citep{ciosek2019better} extends SAC with an optimistic exploration policy approximating the Q-value upper bound.
    \item TOP - Tactical Optimism and Pessimism \citep{moskovitz2021tactical} addresses the optimism-pessimism dilemma with an external discrete bandit learning $\beta^{p}$ from predefined values. Based on TD3. 
    \item GPL - Generalized Pessimism Learning \citep{cetin2023learning} proposes updating pessimism in the pessimistic value approximation $V^{\pi}_{\beta}$ through a dual optimization objective. Based on SAC.
    \item TD-MPC - Temporal Difference Learning for Model Predictive Control \citep{hansen2022temporal} a state-of-the-art model-based algorithm leveraging model of the environment for critic-bootstrapped model-predictive control.
\end{itemize}

In high replay ratio, all algorithms undergo full-parameter resets according to the schedule proposed by \citet{d2022sample}. As such, given replay ratio of 16 gradient steps per environment step, the agents are resetted in $(160k, 320k, 480k, 640k, 800k)$ steps. ma

\begin{table}[ht!]
\centering
\caption{For non-shared hyperparameters we take values which were found to perform best by the authors of the respective method.}
\medskip
\label{tab:hyperparameters}
{\renewcommand{\arraystretch}{1.02}%
 \begin{tabular}{|| c | c | c ||} 
 \hline
 \textsc{Method} & \textsc{Hyperparameter} & \textsc{Value} \\
 \hline \hline
 \multirow{11}*{\textsc{SAC / Shared}} & \textsc{Network Size} & $(256, 256)$ \\
  & \textsc{Activations}& \textsc{ReLU} \\
  & \textsc{Critic Ensemble Size}& $2$ \\
  & \textsc{Action Repeat}& $1$ \\
  & \textsc{Optimizer} & \textsc{Adam} \\
  & \textsc{Learning Rate} & $3e-4$ \\
  & \textsc{Batch Size} & $256$ \\
  & \textsc{Discount} & $0.99$ \\
  & \textsc{Initial Temperature} & $1.0$ \\
  & \textsc{Initial Steps} & $10000$ \\
  & \textsc{Target Entropy} & $|\mathcal{A}|/2$ \\
  & \textsc{Polyak Weight} & $0.005$ \\
 \hline\hline
 \multirow{4}*{\textsc{TD3}} & \textsc{Policy Update Delay} & $2$ \\
  & \textsc{Exploration $\sigma$} & $0.5$ \\
  & \textsc{Target Policy $\sigma$} & $0.2$ \\
  & \textsc{Policy Delay} & $2$ \\
  & \textsc{Exploration Noise Clip} & $\pm ~0.5$ \\
 \hline\hline
 \multirow{3}*{\textsc{OAC}} & \textsc{Optimism} & $4.36$ \\
  & \textsc{Pessimism} & $-0.75$ \\
  & \textsc{KL Divergence Constraint} & $3.69$ \\
 \hline\hline
 \multirow{4}*{\textsc{TOP}} & \textsc{Optimistic Arm} & $0.0$ \\
  & \textsc{Pessimistic Arm} & $-1.0$ \\
  & \textsc{Bandit Learning Rate} & $0.1$ \\
  & \textsc{Bandit Decay} & $0.9$ \\
 \hline\hline
 \multirow{2}*{\textsc{GPL}} & \textsc{Initial
 Pessimism} & $-0.5$ \\
  & \textsc{Pessimistic Learning Rate} & $1e-4$ \\
 \hline\hline
 \multirow{6}*{\textsc{DAC}} & \textsc{Pessimism} & $-0.2$ \\
  & \textsc{Initial Optimism} & $1.0$ \\
  & \textsc{Initial KL Weight} & $0.25$ \\
  & \textsc{Target KL Divergence} & $0.25$ \\
  & \textsc{Standard Deviation Multiplier} & $1.25$ \\
  & \textsc{Adjustment Learning Rate} & $3e-5$ \\
 \hline \hline
\end{tabular}}
\end{table}

\subsection{Training Curves}

\begin{figure}[ht!]
\begin{center}
\begin{minipage}[h]{1.0\linewidth}
\centering
    \begin{subfigure}{0.95\linewidth}
    \includegraphics[width=\textwidth]{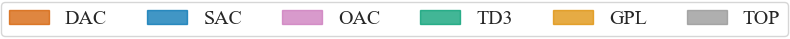}
    \end{subfigure}
\end{minipage}
\begin{minipage}[h]{1.0\linewidth}
    \begin{subfigure}{1.0\linewidth}
    \includegraphics[width=0.195\linewidth]{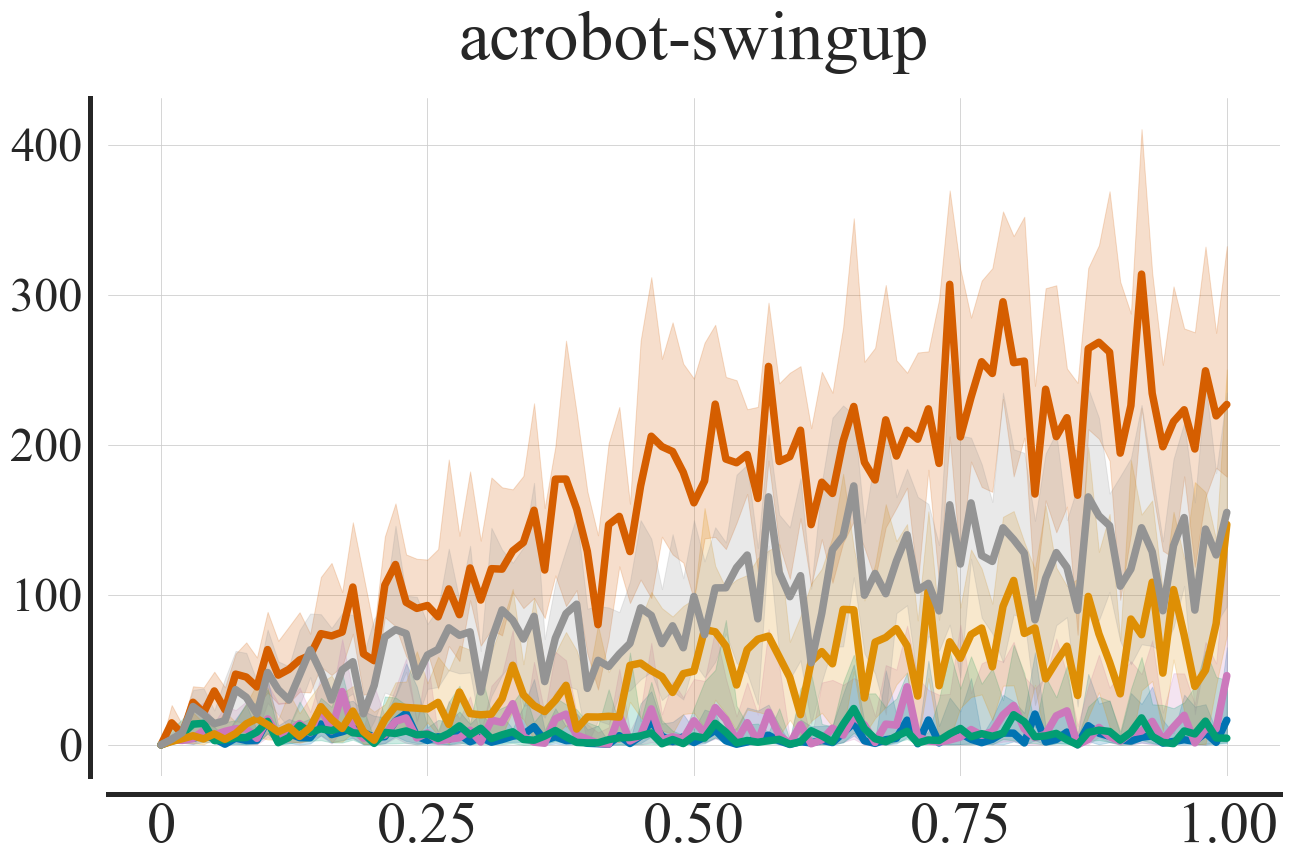}
    \hfill
    \includegraphics[width=0.195\linewidth]{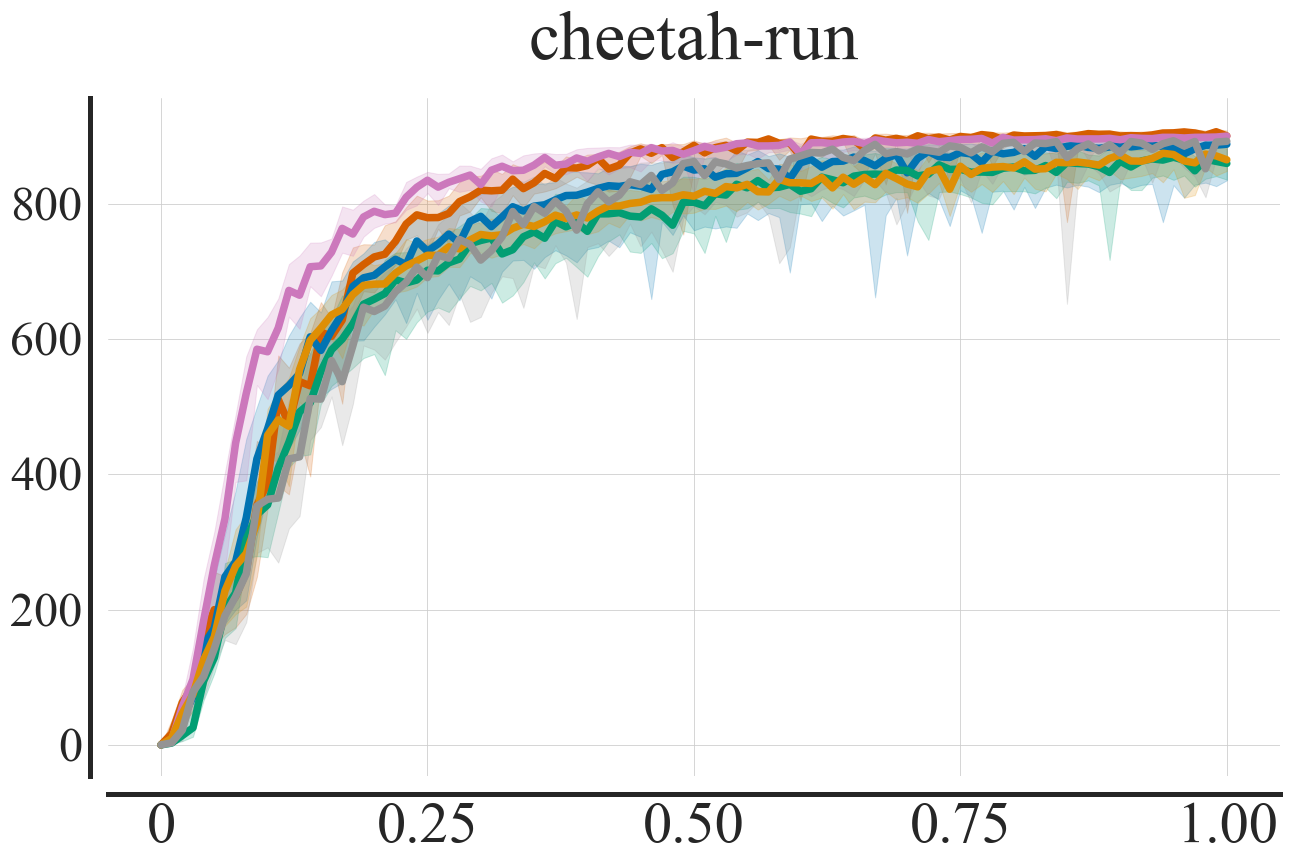}
    \hfill
    \includegraphics[width=0.195\linewidth]{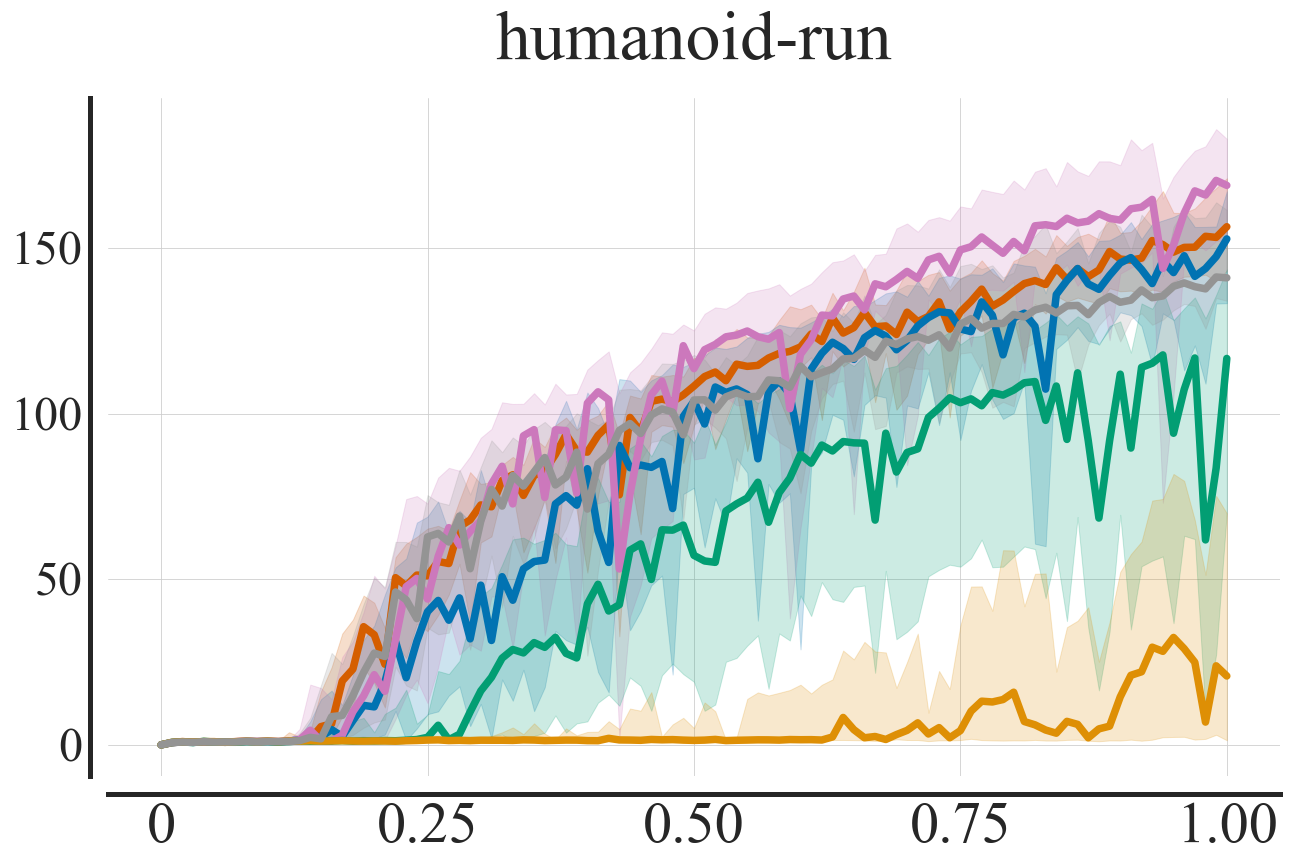}
    \hfill
    \includegraphics[width=0.195\linewidth]{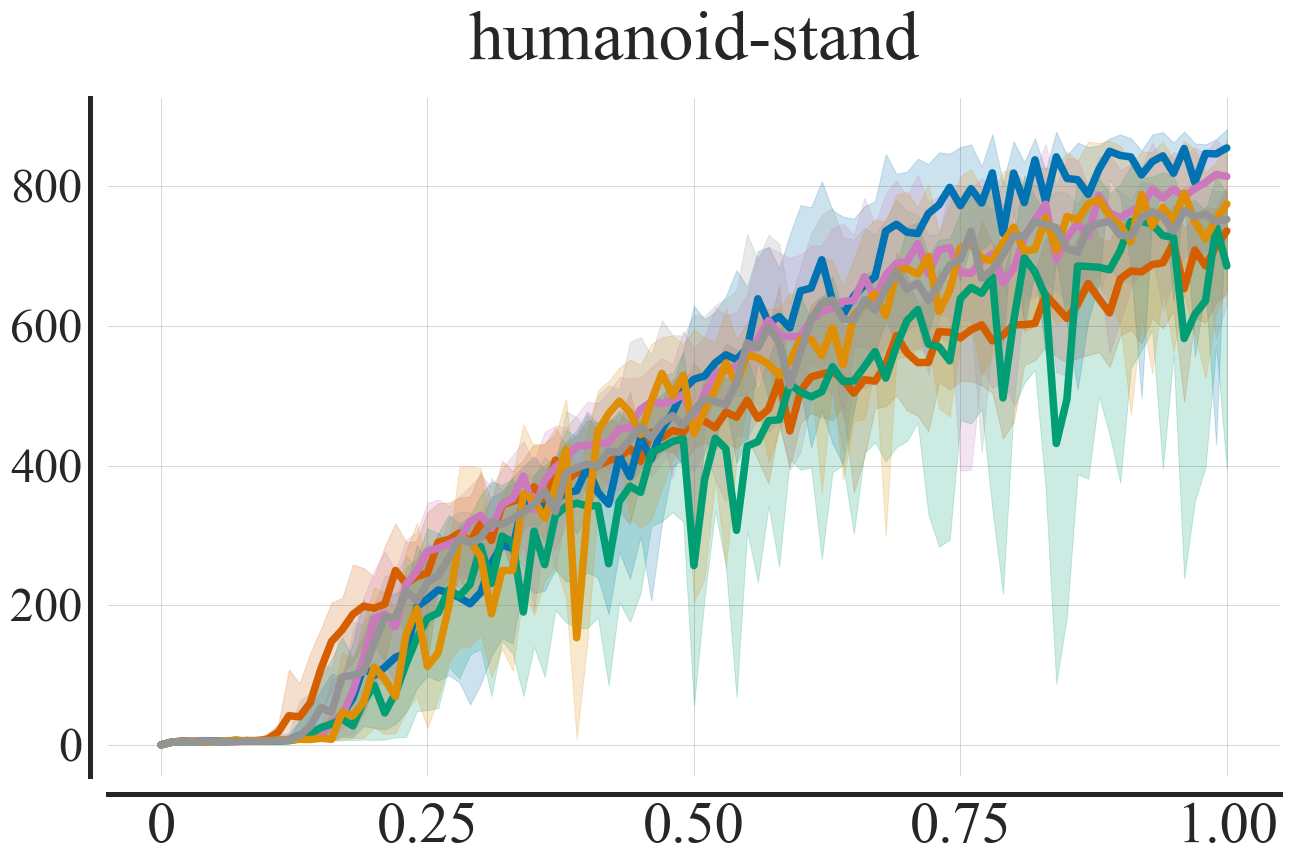}
    \hfill
    \includegraphics[width=0.195\linewidth]{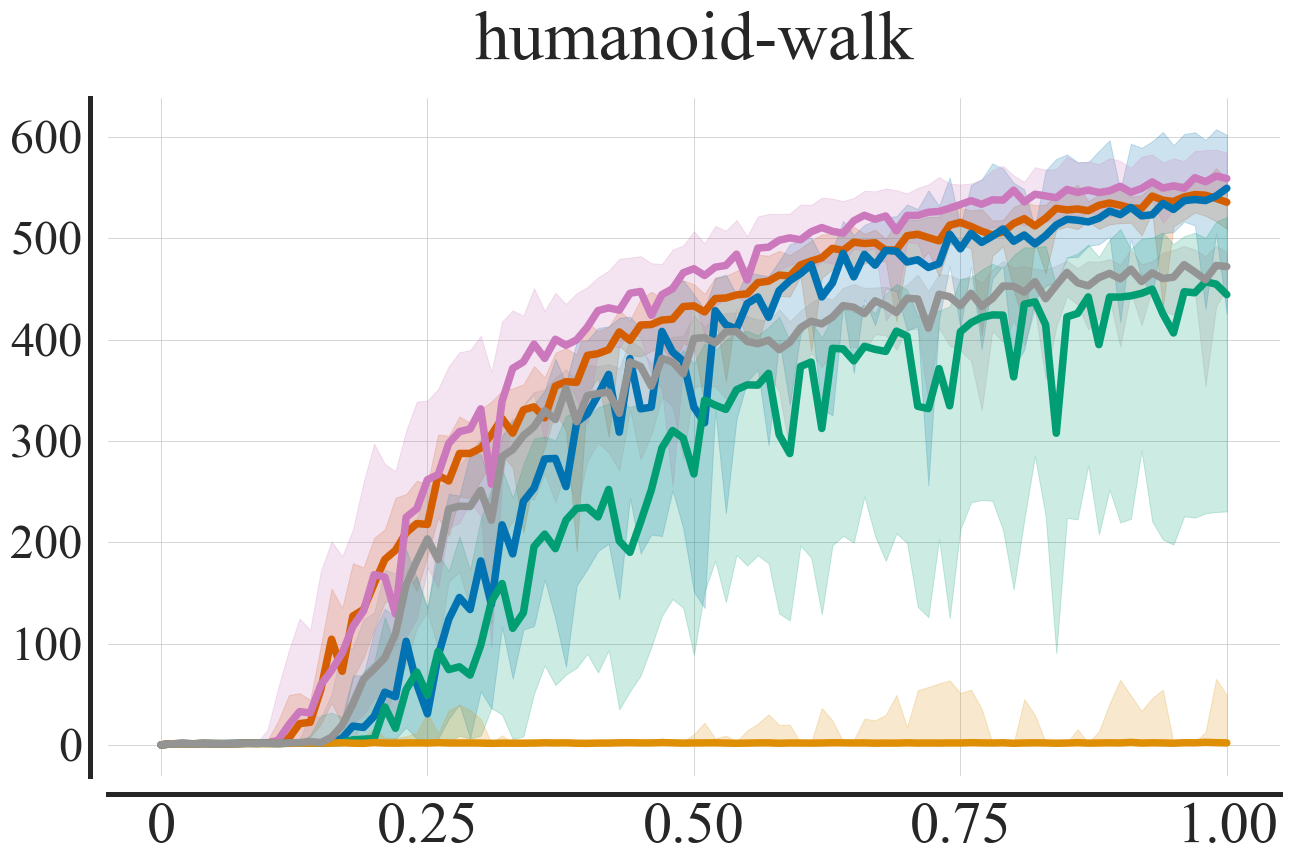}
    \end{subfigure}
\end{minipage}
\begin{minipage}[h]{1.0\linewidth}
    \begin{subfigure}{1.0\linewidth}
    \includegraphics[width=0.195\linewidth]{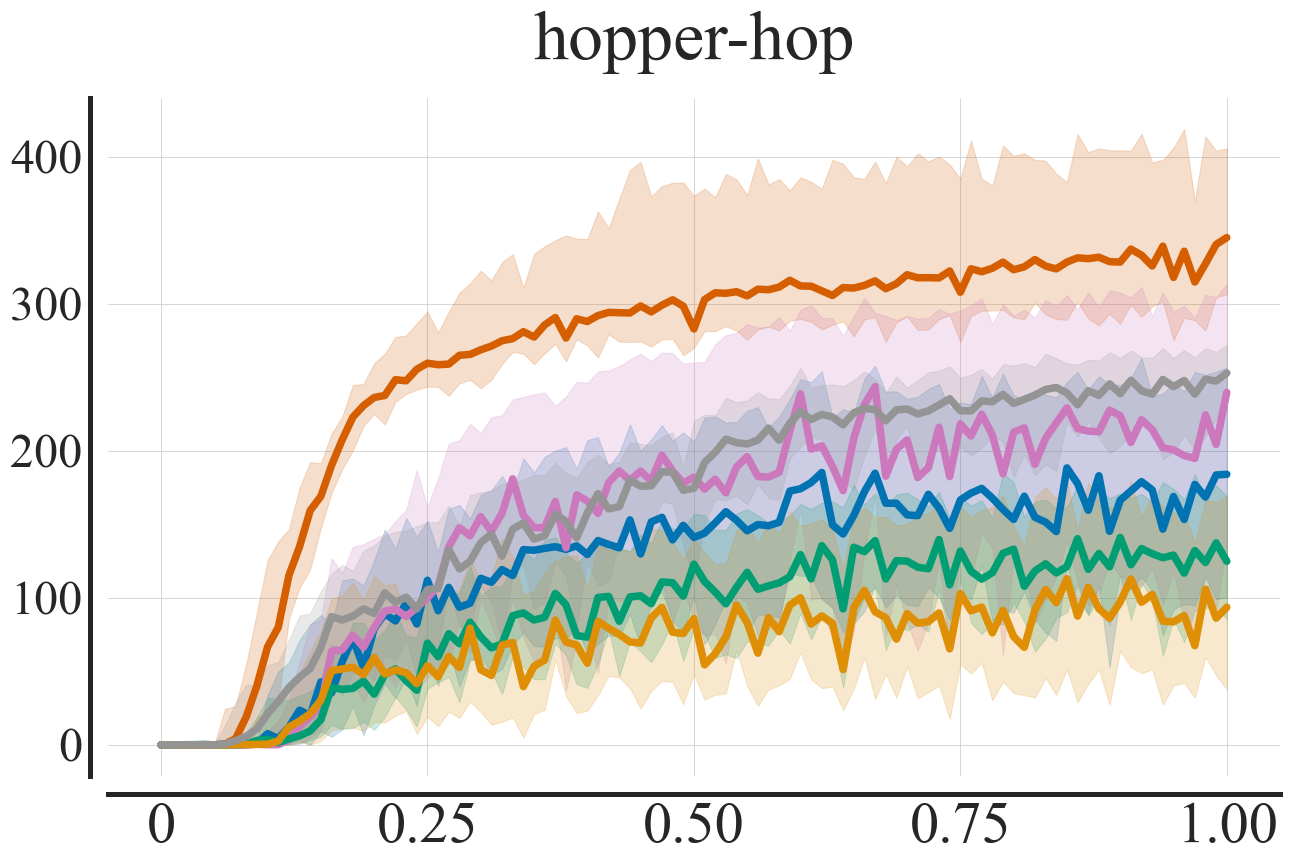}
    \hfill
    \includegraphics[width=0.195\linewidth]{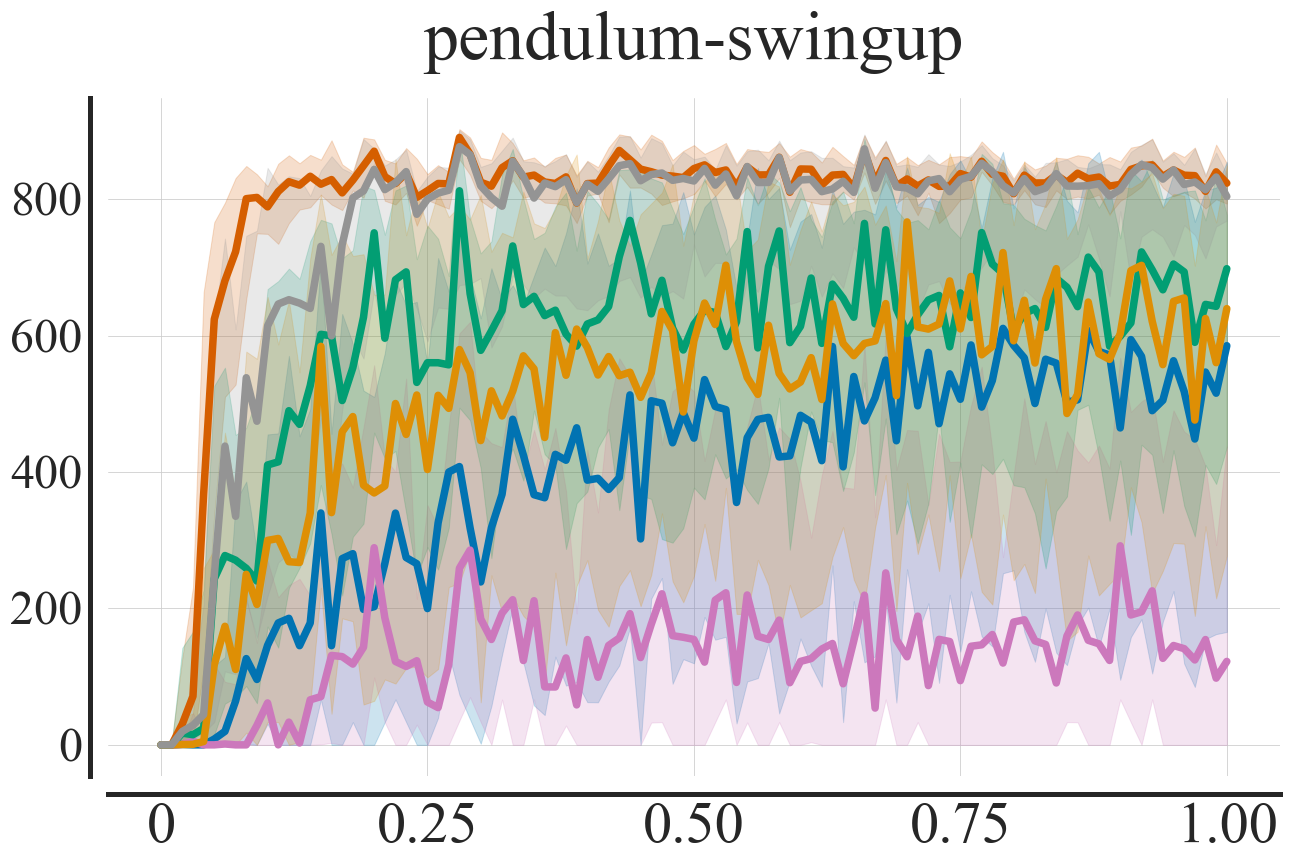}
    \hfill
    \includegraphics[width=0.195\linewidth]{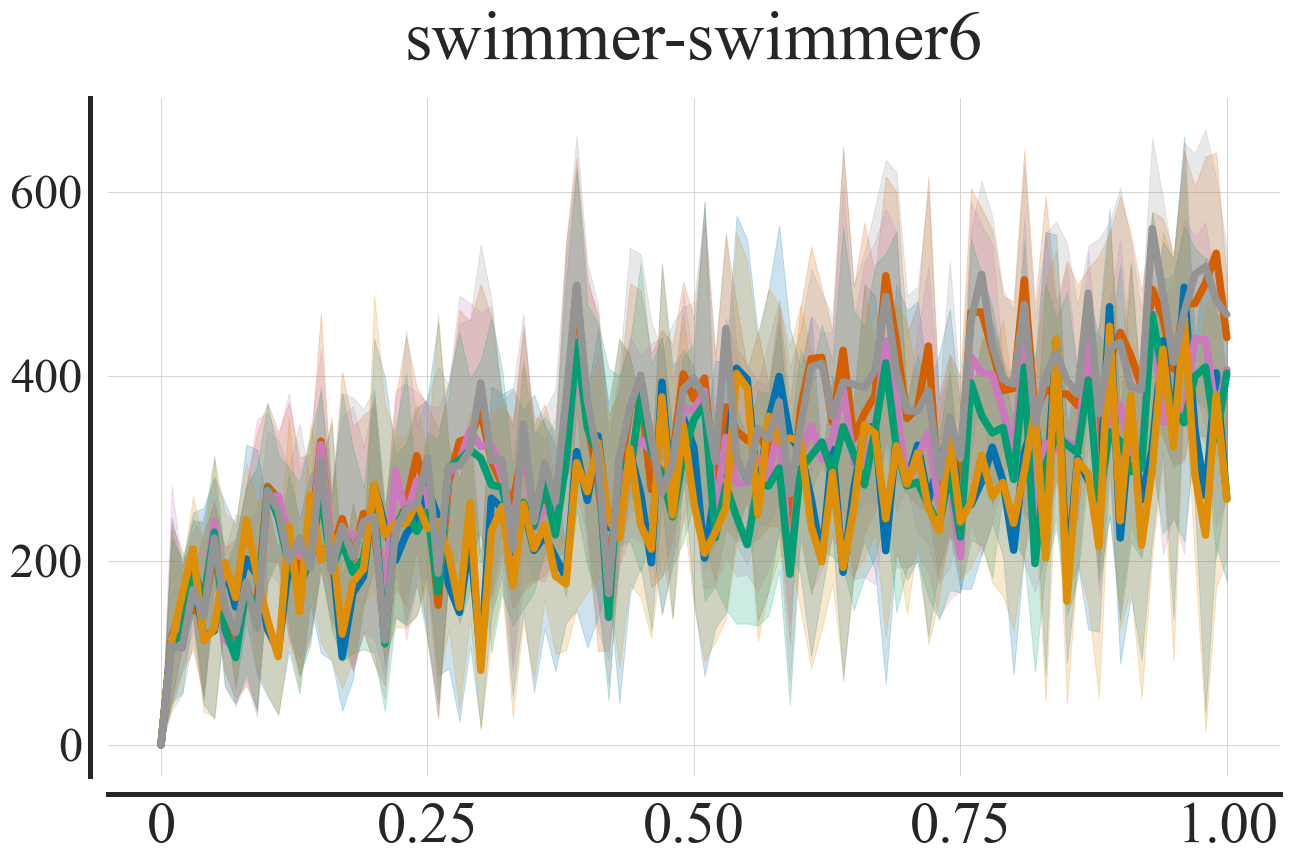}
    \hfill
    \includegraphics[width=0.195\linewidth]{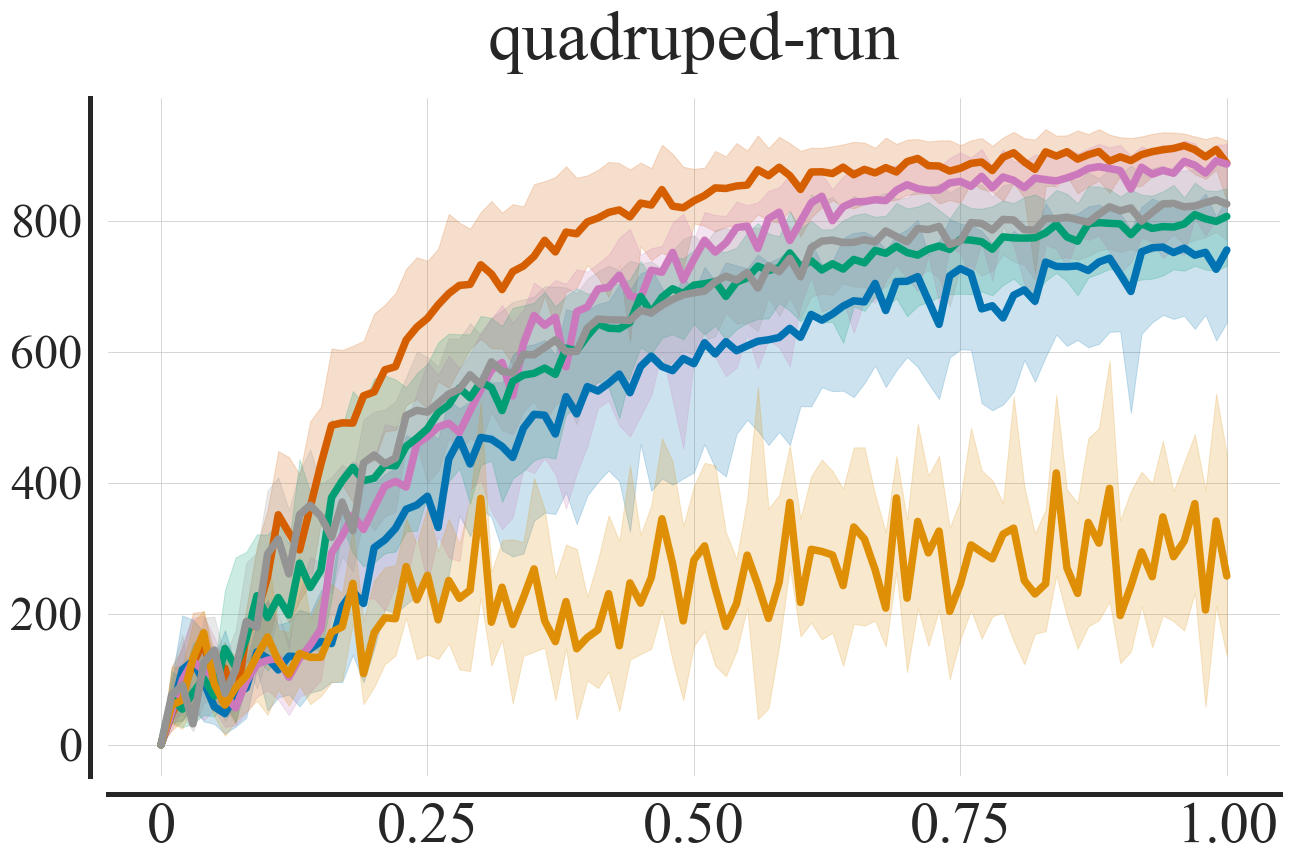}
    \hfill
    \includegraphics[width=0.195\linewidth]{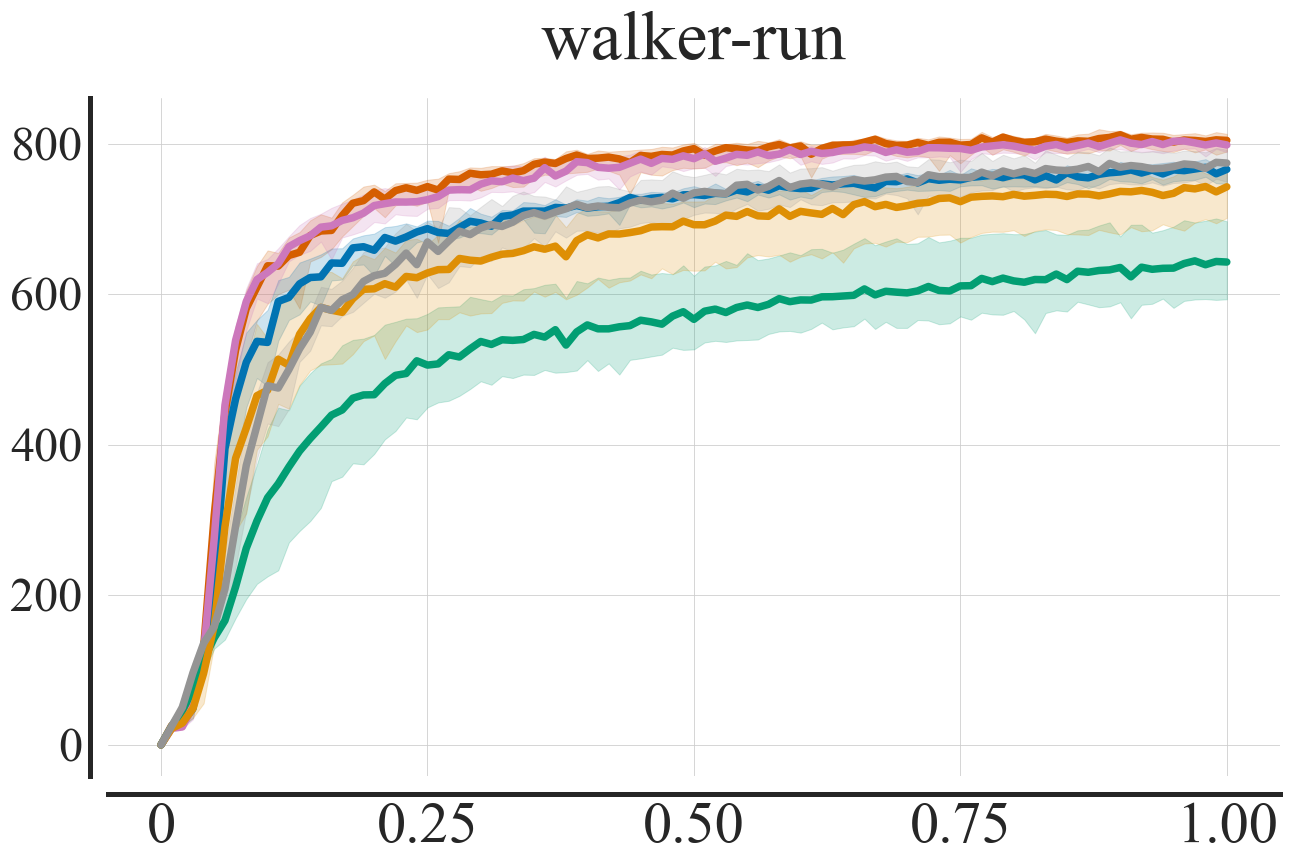}
    \end{subfigure}
\end{minipage}
\begin{minipage}[h]{1.0\linewidth}
    \begin{subfigure}{1.0\linewidth}
    \includegraphics[width=0.195\linewidth]{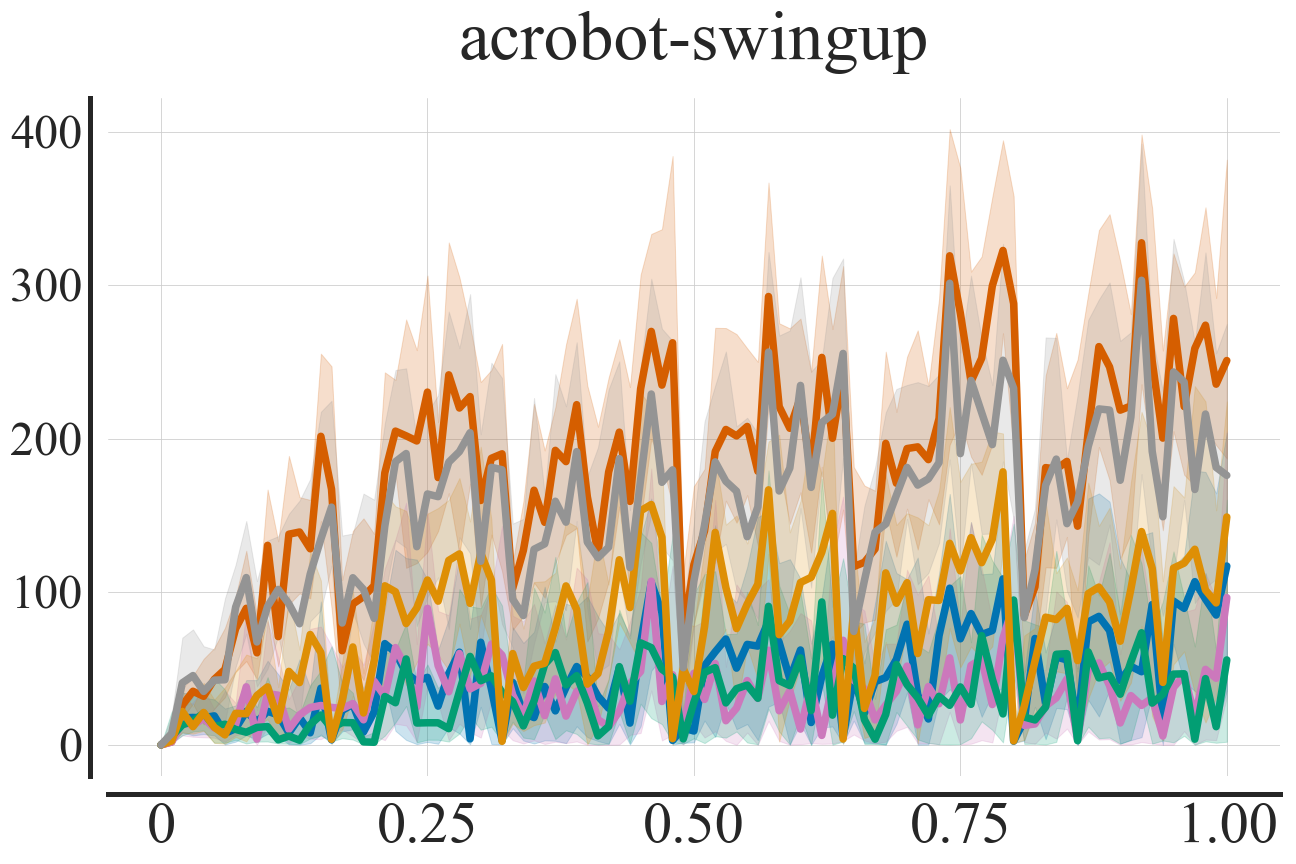}
    \hfill
    \includegraphics[width=0.195\linewidth]{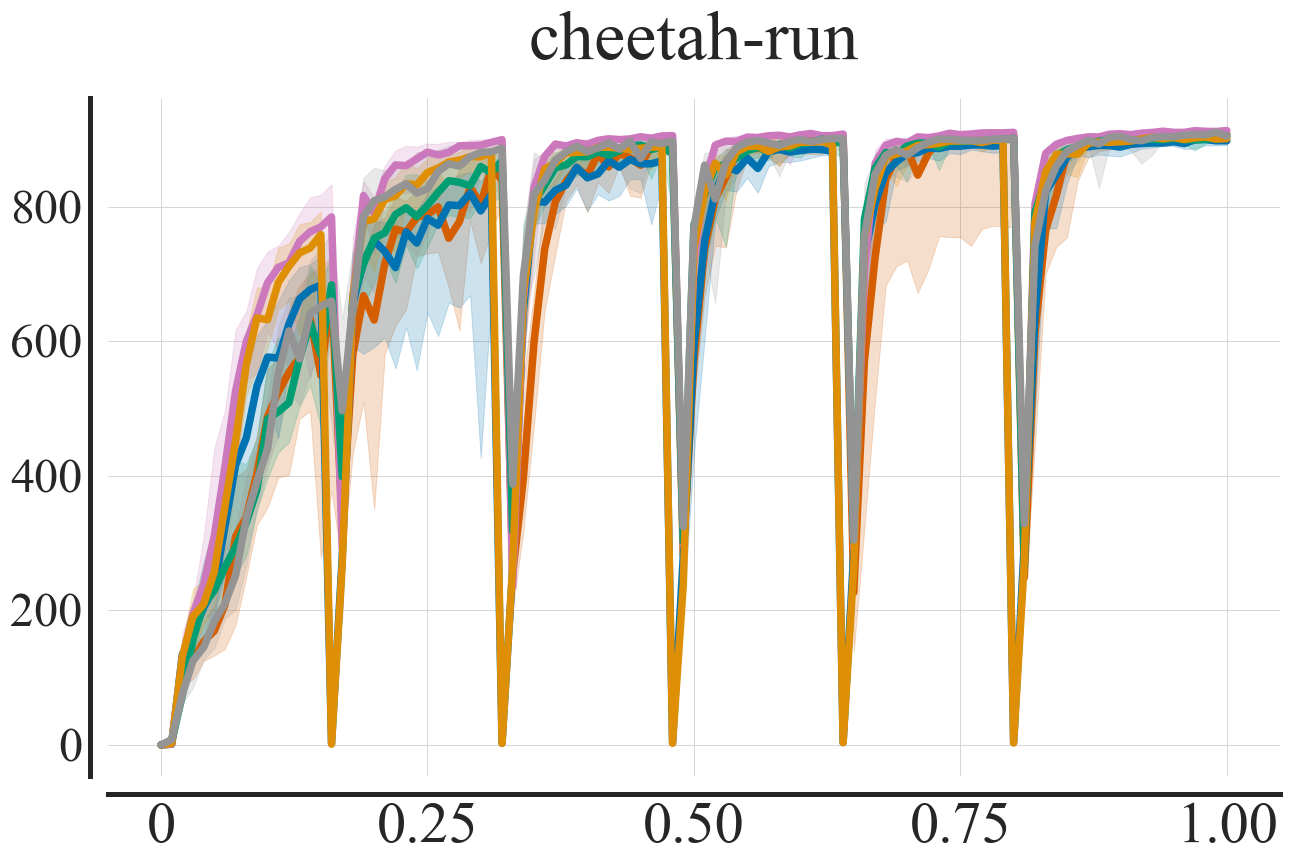}
    \hfill
    \includegraphics[width=0.195\linewidth]{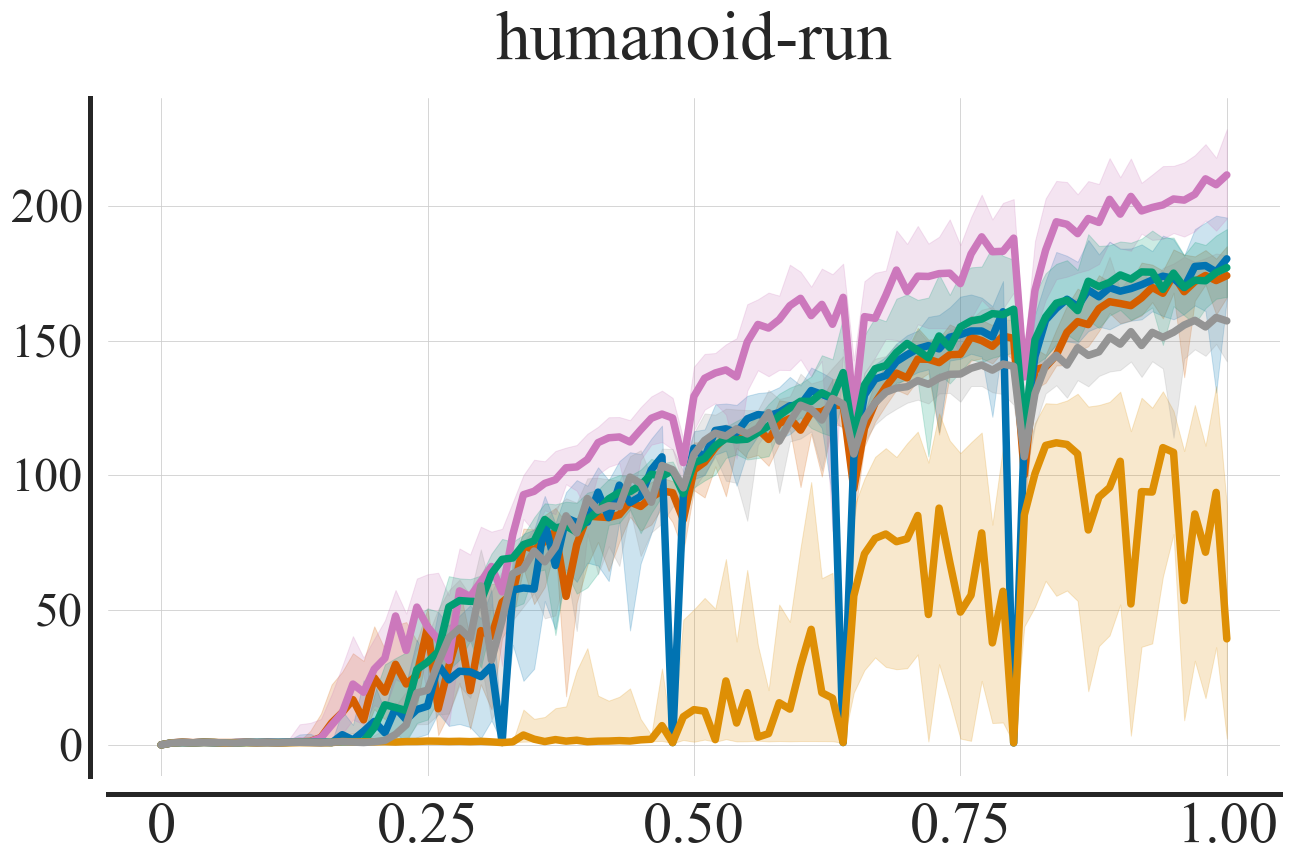}
    \hfill
    \includegraphics[width=0.195\linewidth]{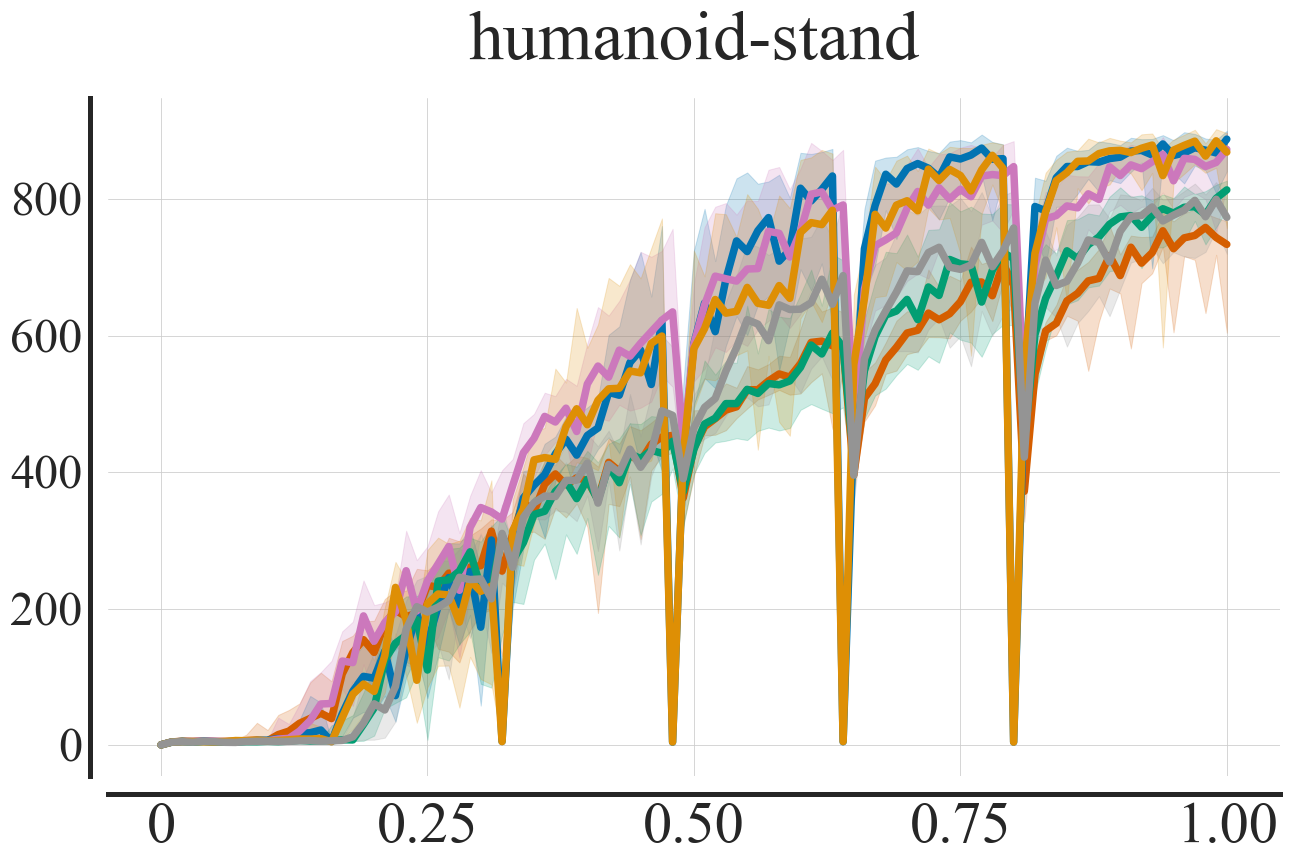}
    \hfill
    \includegraphics[width=0.195\linewidth]{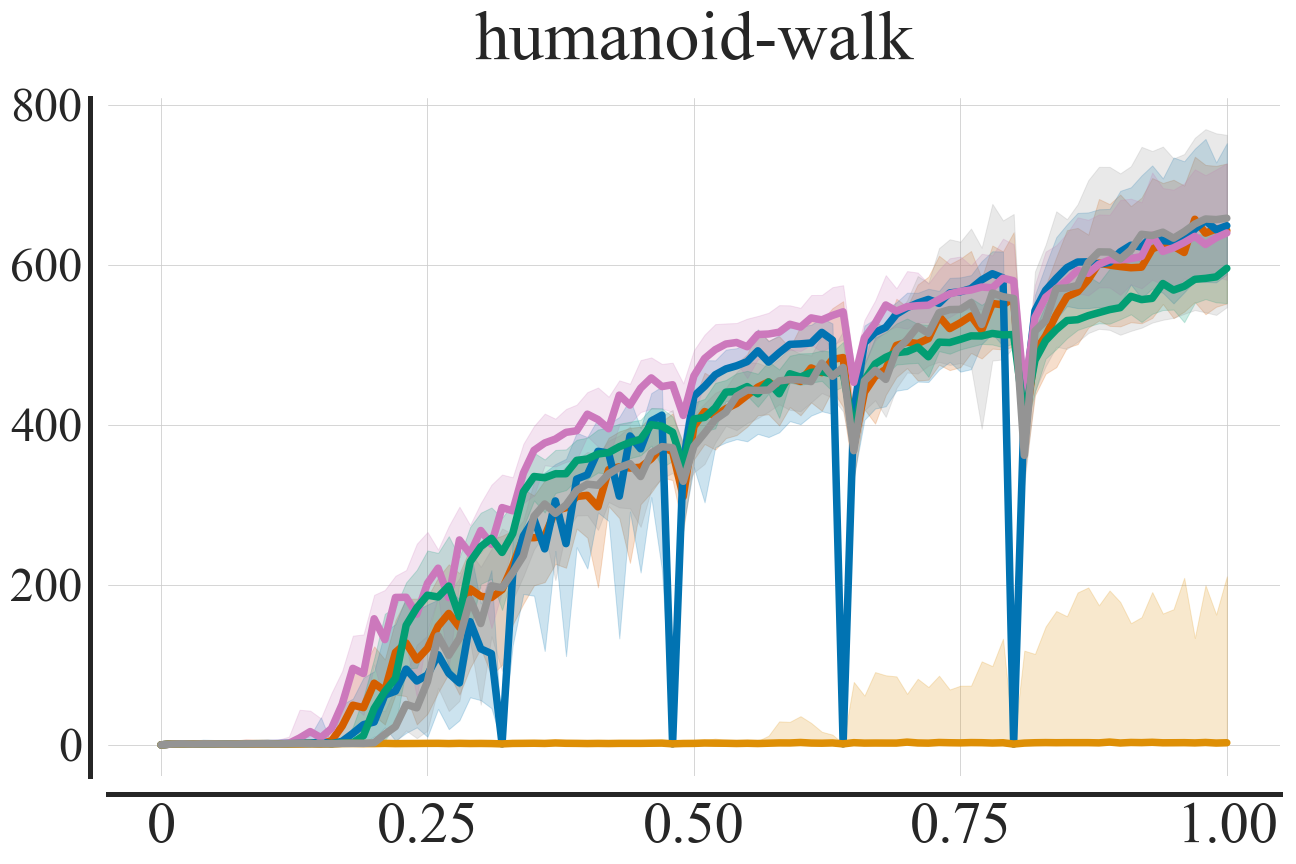}
    \end{subfigure}
\end{minipage}
\begin{minipage}[h]{1.0\linewidth}
    \begin{subfigure}{1.0\linewidth}
    \includegraphics[width=0.195\linewidth]{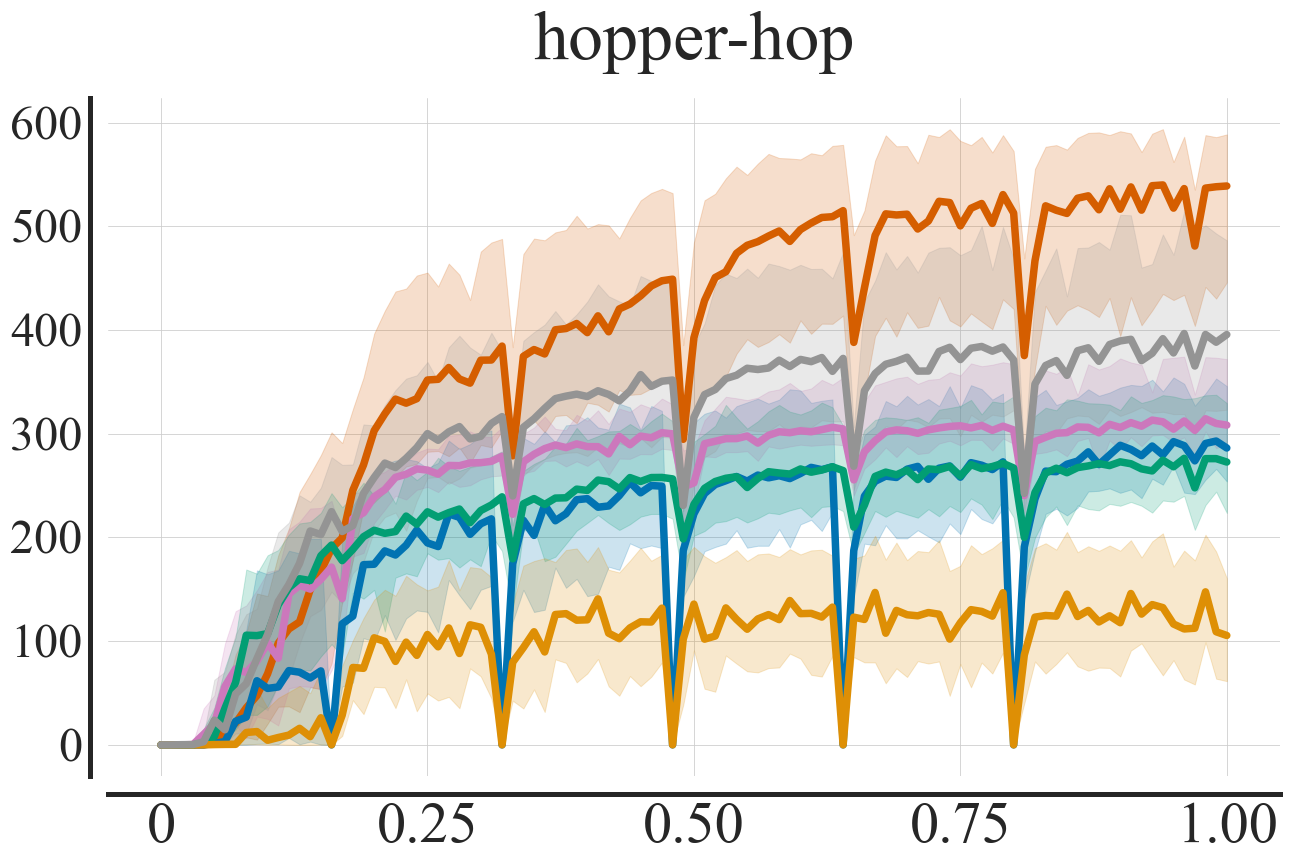}
    \hfill
    \includegraphics[width=0.195\linewidth]{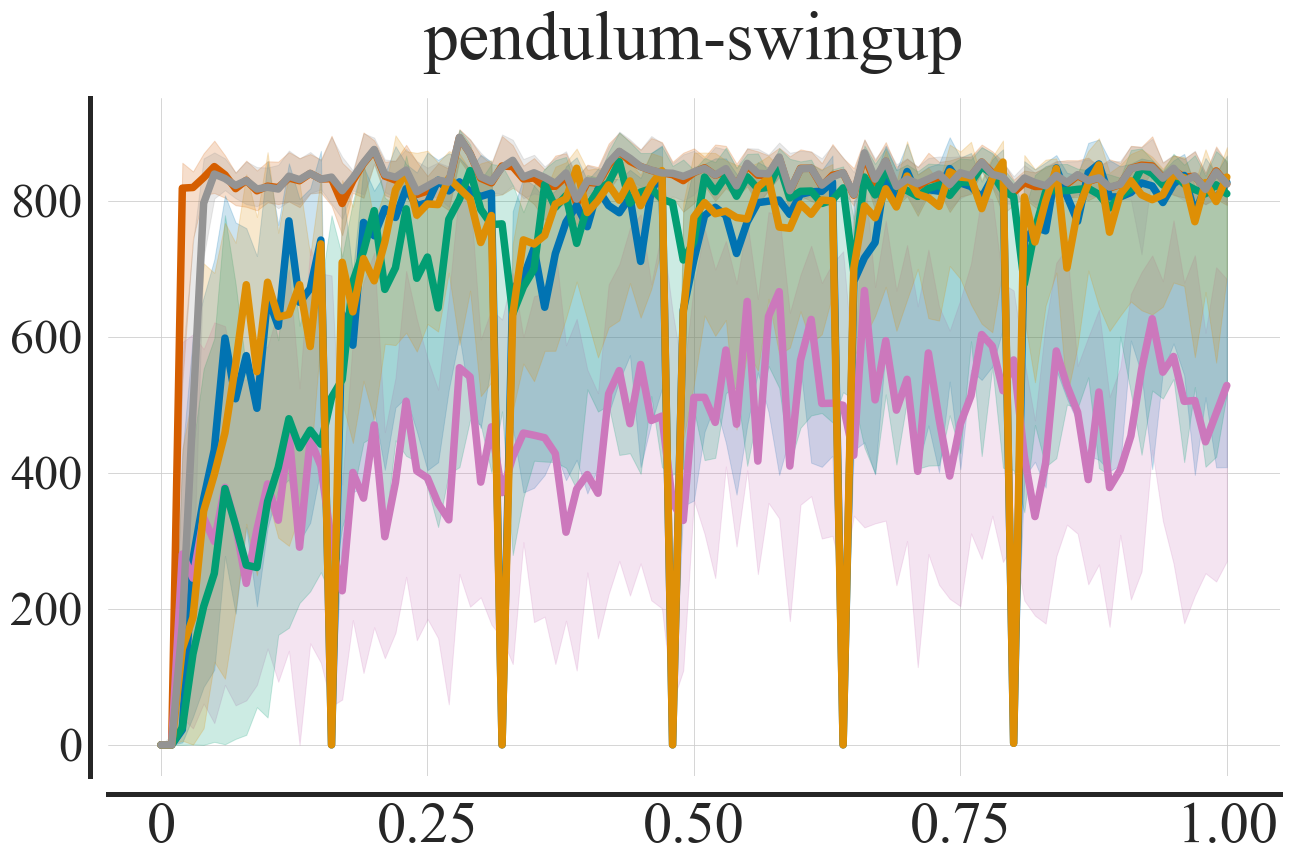}
    \hfill
    \includegraphics[width=0.195\linewidth]{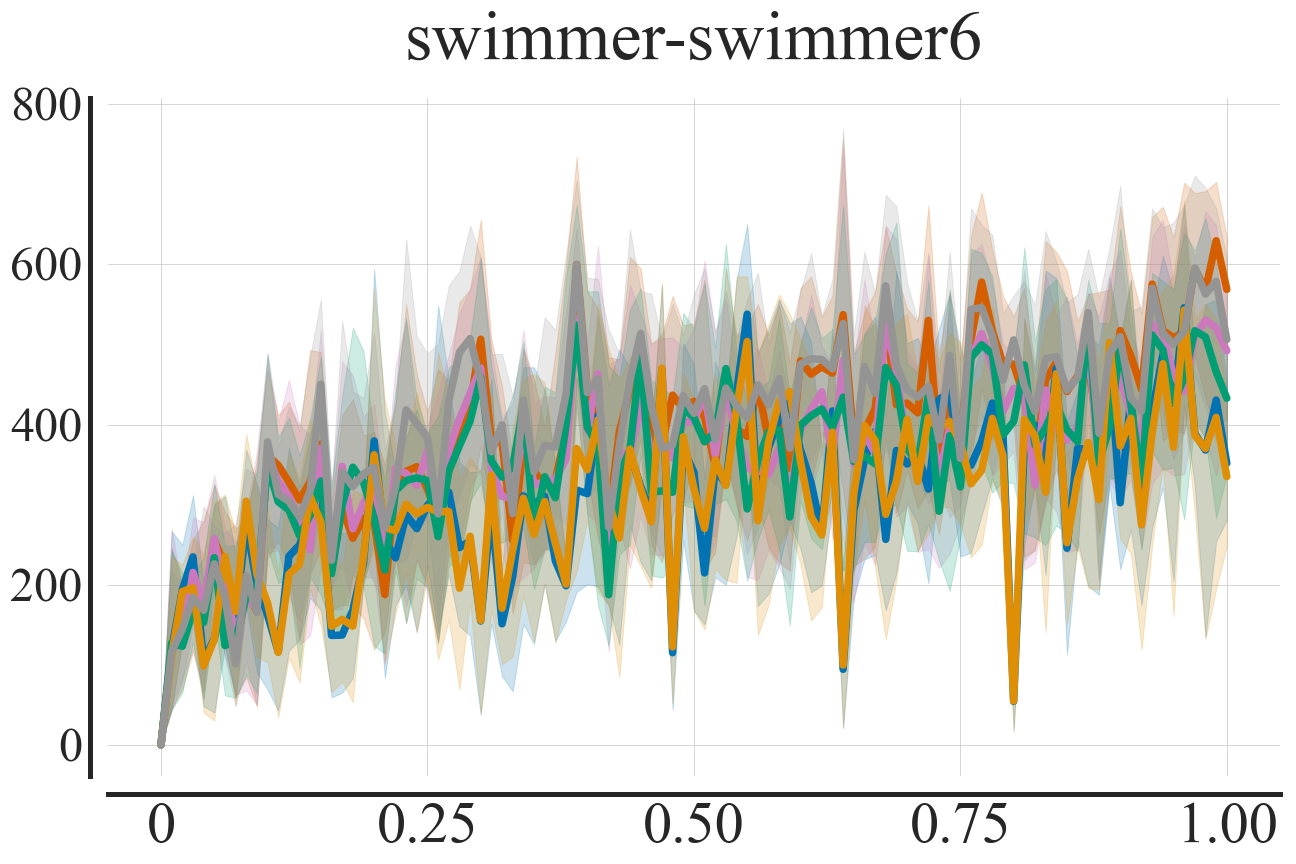}
    \hfill
    \includegraphics[width=0.195\linewidth]{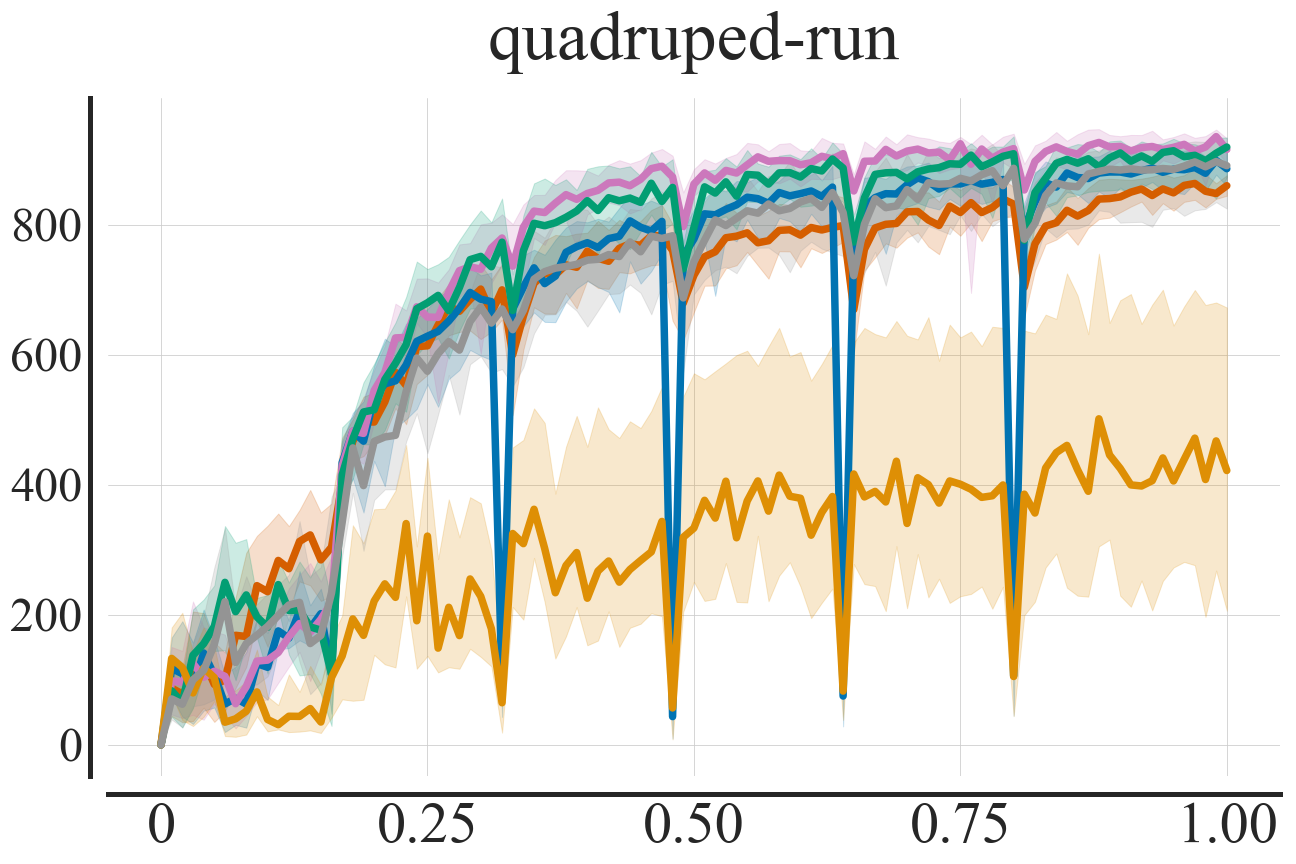}
    \hfill
    \includegraphics[width=0.195\linewidth]{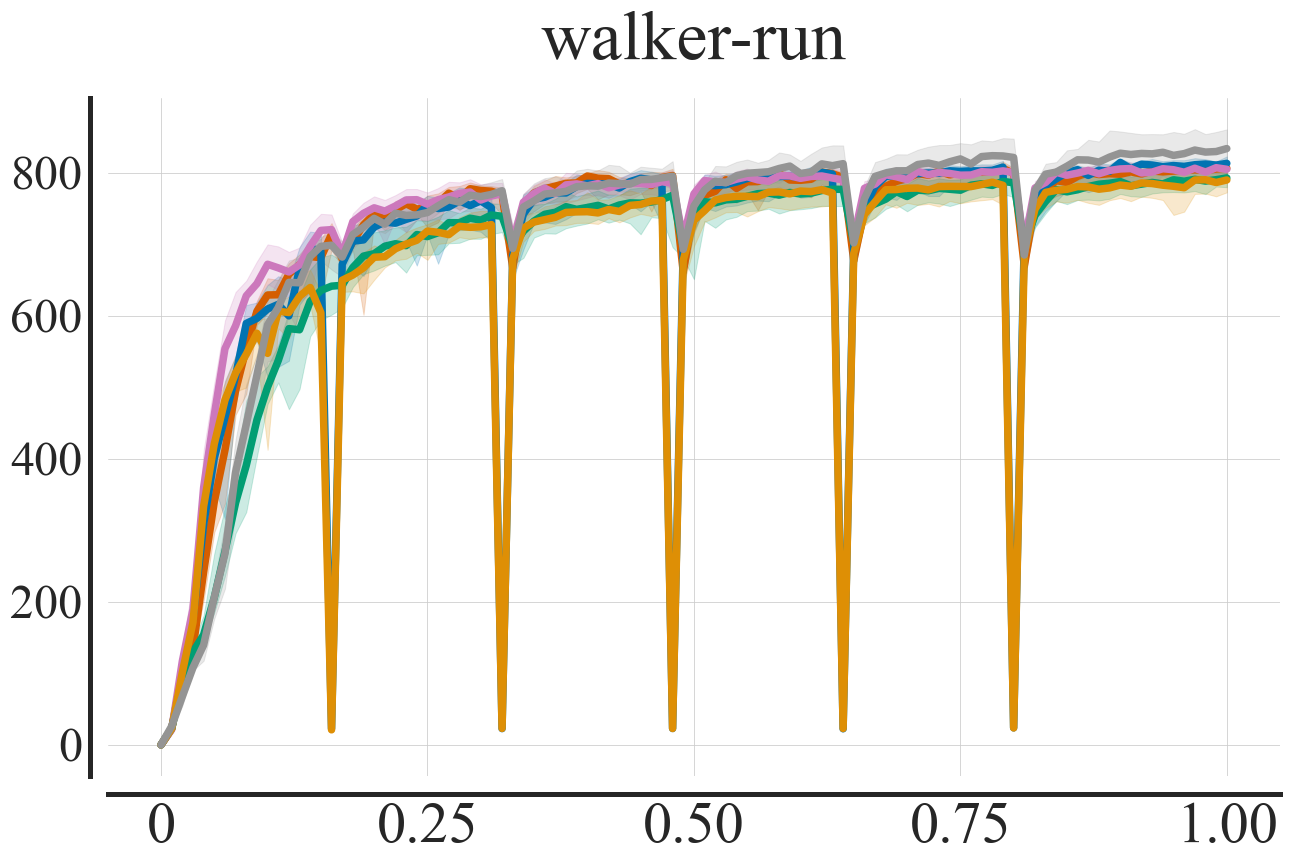}
    \end{subfigure}
\end{minipage}
\caption{Training curves for DMC ($RR=2$ rows $1$ \& $2$, $RR=16$ row $3$ \& $4$). $Y$-axis denotes IQM and $X$-axis denotes environment steps. 10 seeds per task.}
\label{fig:training_rr2_dmc}
\end{center}
\vspace{-0.1in} 
\end{figure}

\begin{figure}[ht!]
\begin{center}
\begin{minipage}[h]{1.0\linewidth}
\centering
    \begin{subfigure}{0.95\linewidth}
    \includegraphics[width=\textwidth]{images/appendix/legend_training.png}
    \end{subfigure}
\end{minipage}
\begin{minipage}[h]{1.0\linewidth}
    \begin{subfigure}{1.0\linewidth}
    \includegraphics[width=0.195\linewidth]{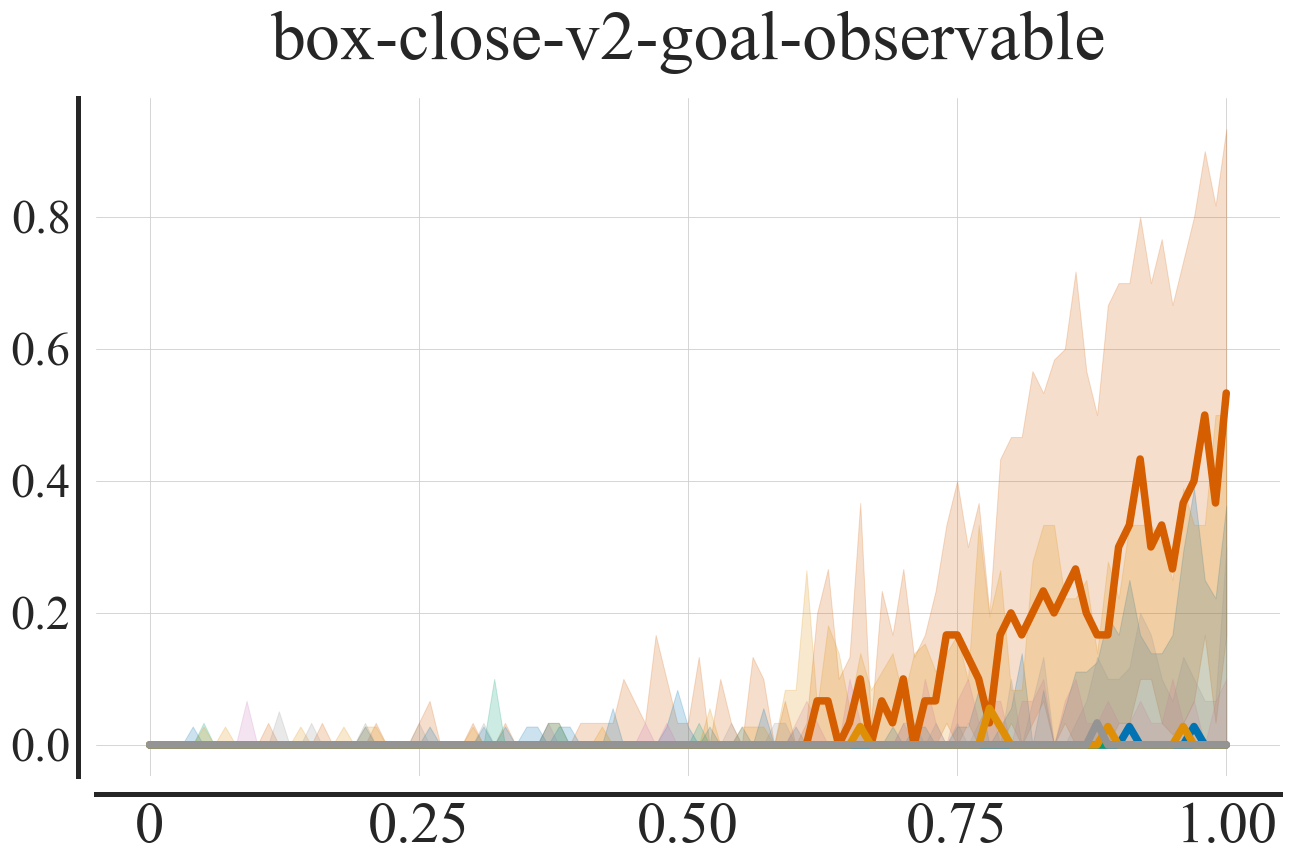}
    \hfill
    \includegraphics[width=0.195\linewidth]{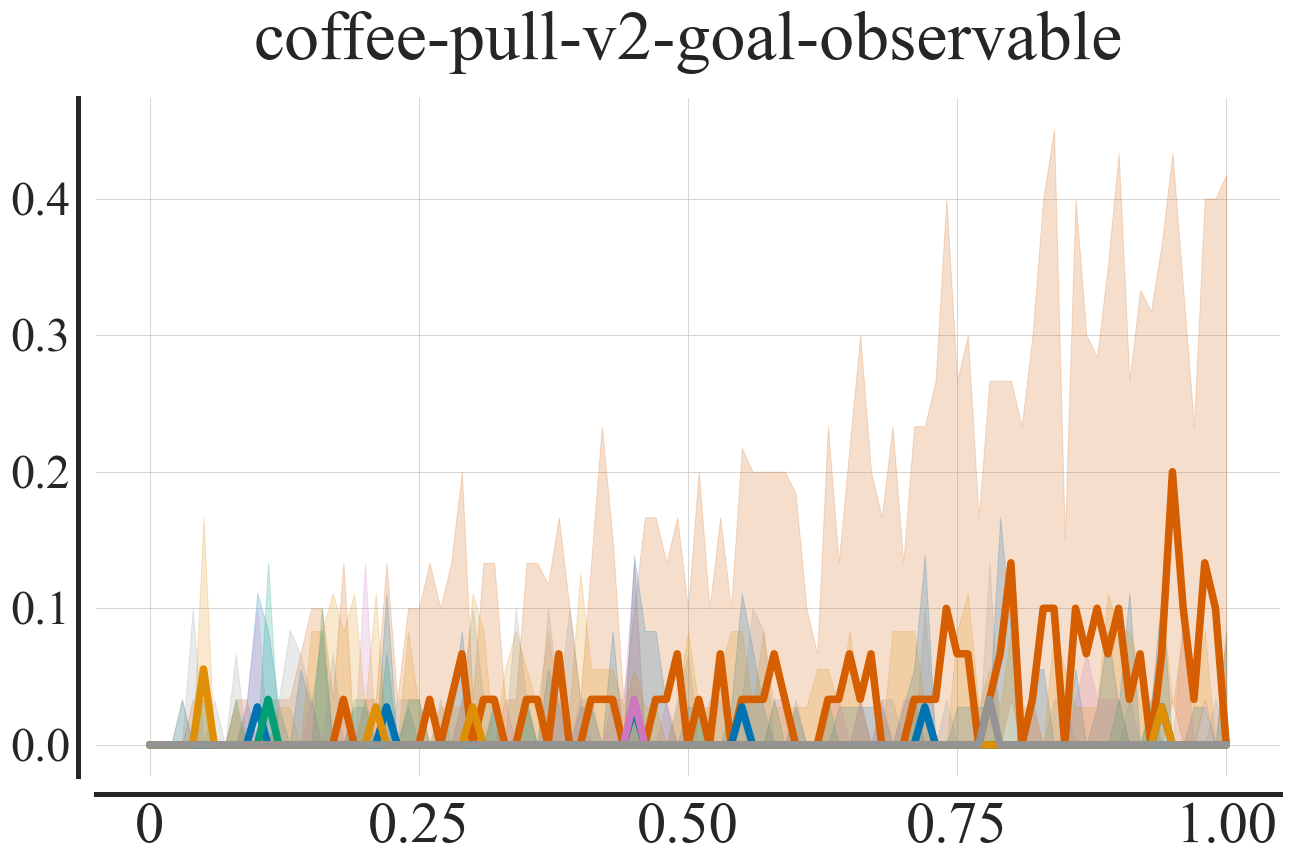}
    \hfill
    \includegraphics[width=0.195\linewidth]{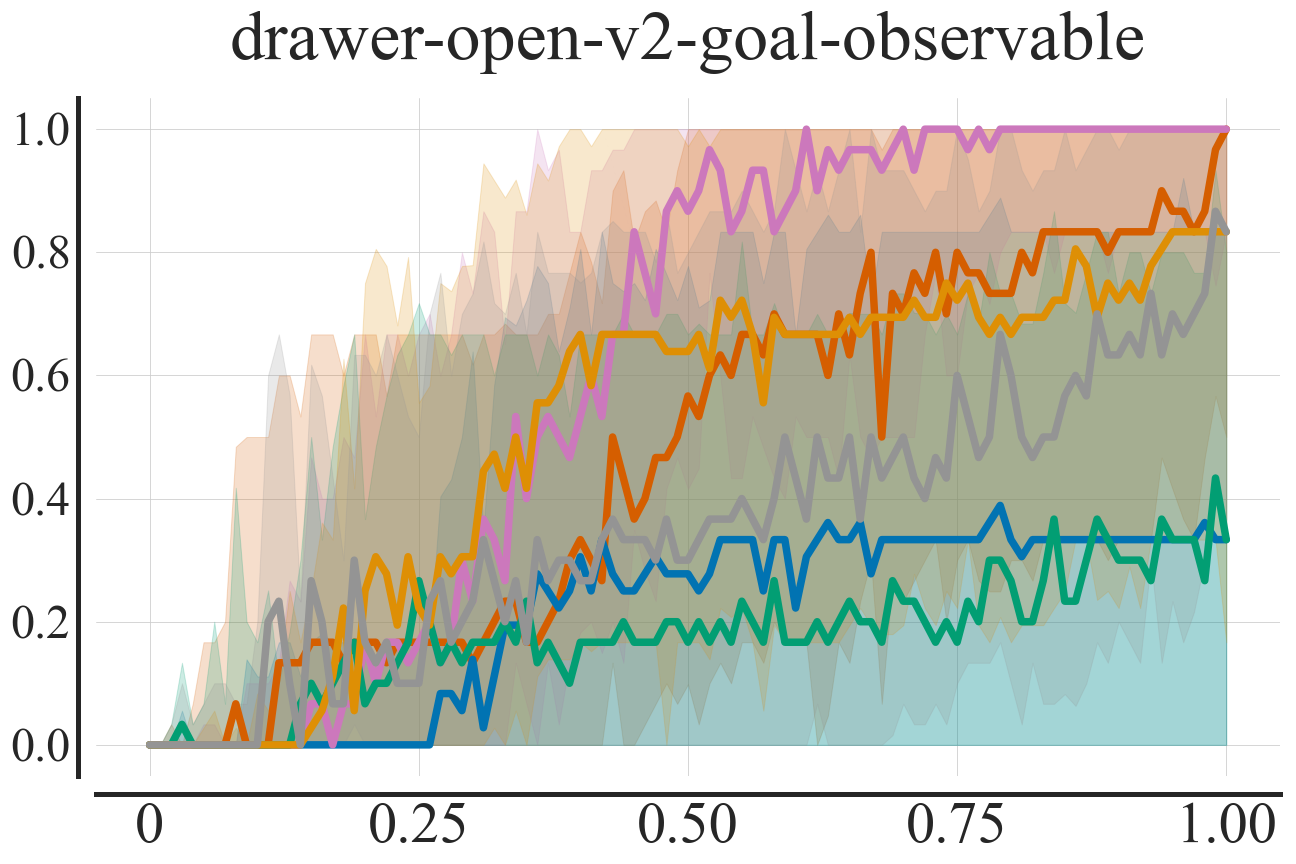}
    \hfill
    \includegraphics[width=0.195\linewidth]{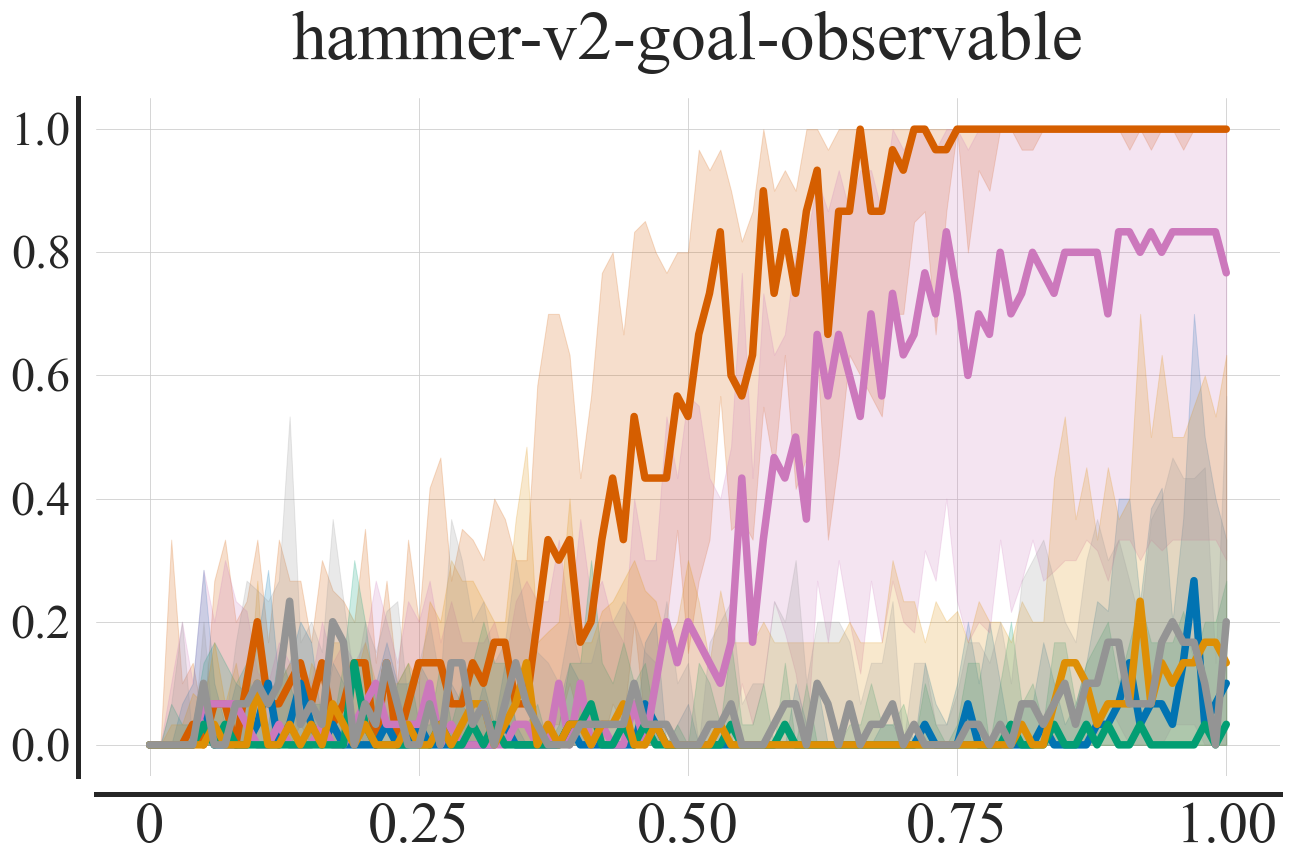}
    \hfill
    \includegraphics[width=0.195\linewidth]{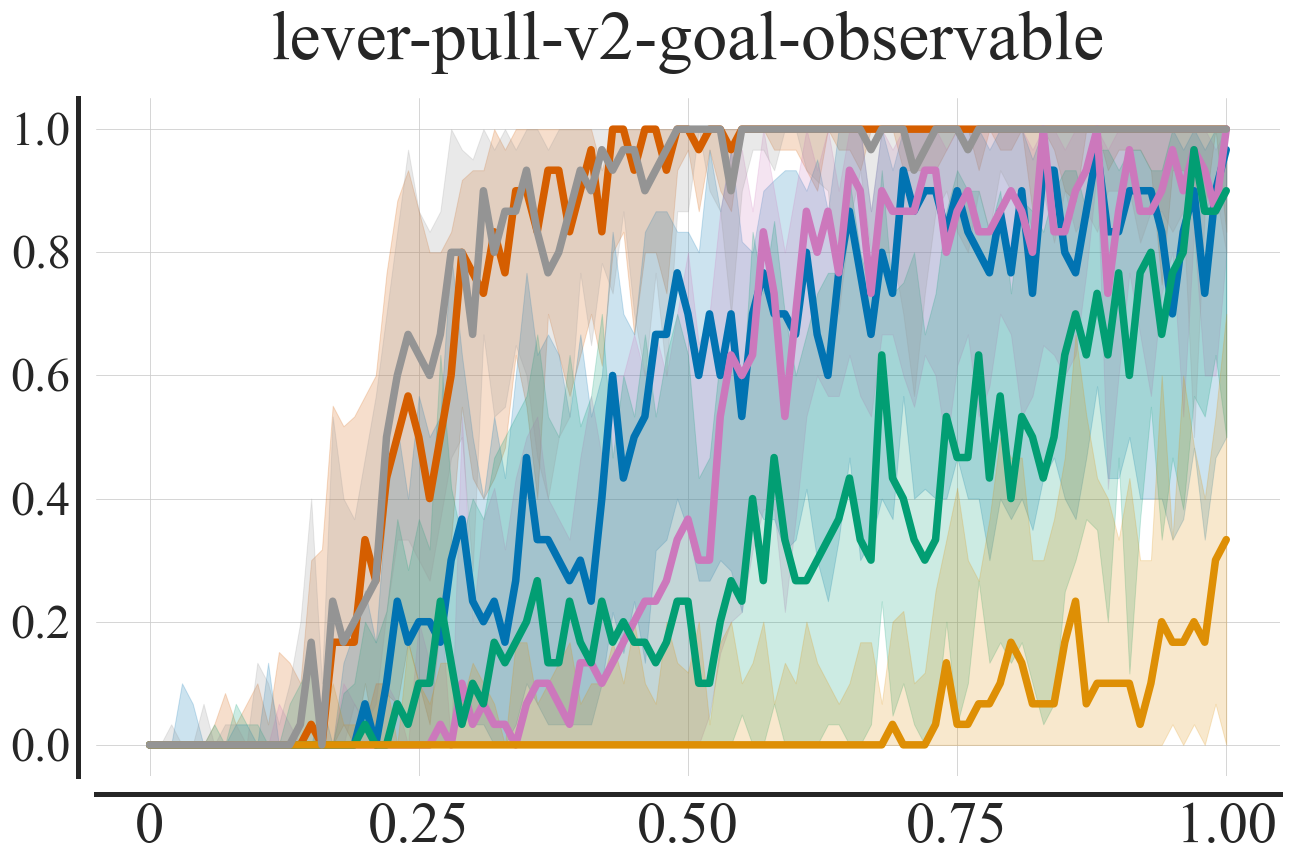}
    \end{subfigure}
\end{minipage}
\begin{minipage}[h]{1.0\linewidth}
    \begin{subfigure}{1.0\linewidth}
    \includegraphics[width=0.195\linewidth]{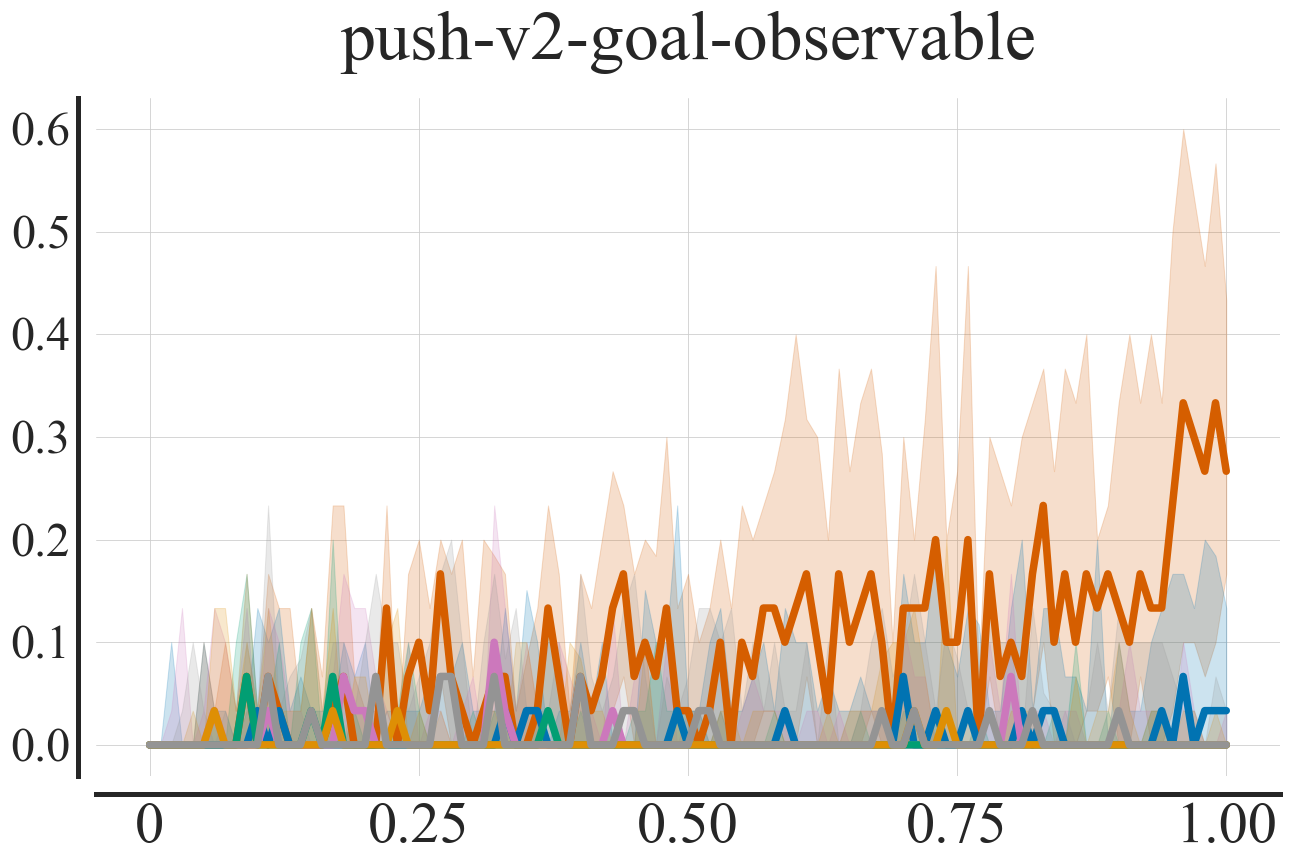}
    \hfill
    \includegraphics[width=0.195\linewidth]{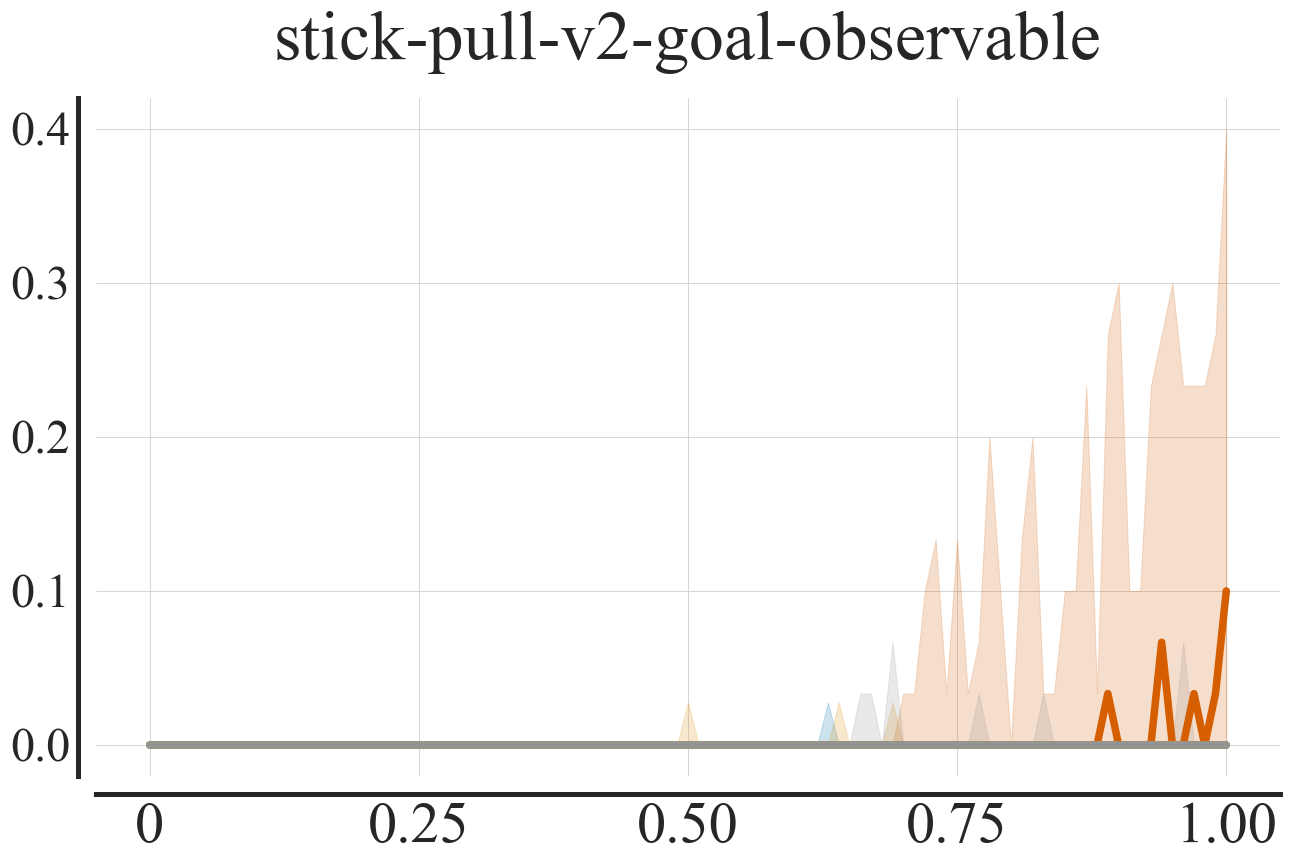}
    \hfill
    \includegraphics[width=0.195\linewidth]{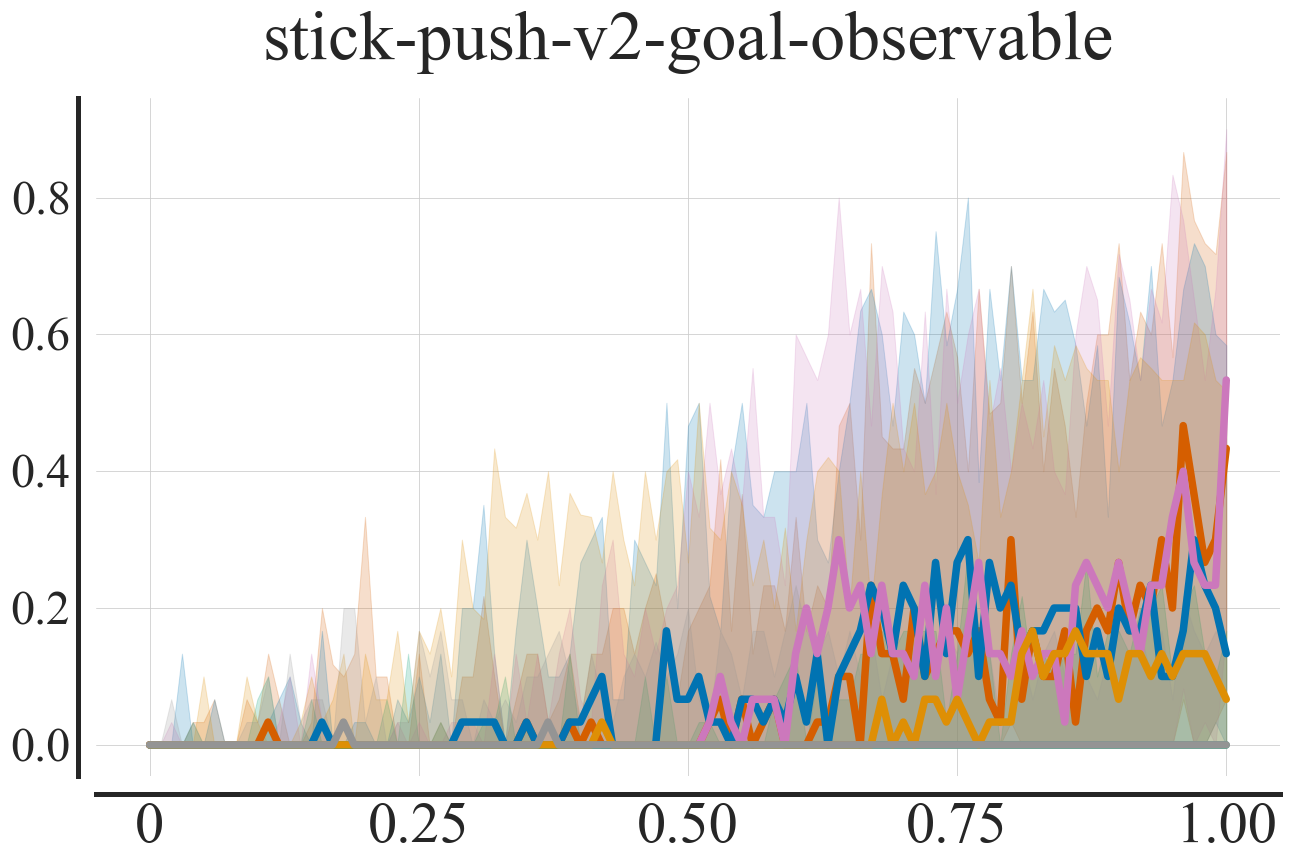}
    \hfill
    \includegraphics[width=0.195\linewidth]{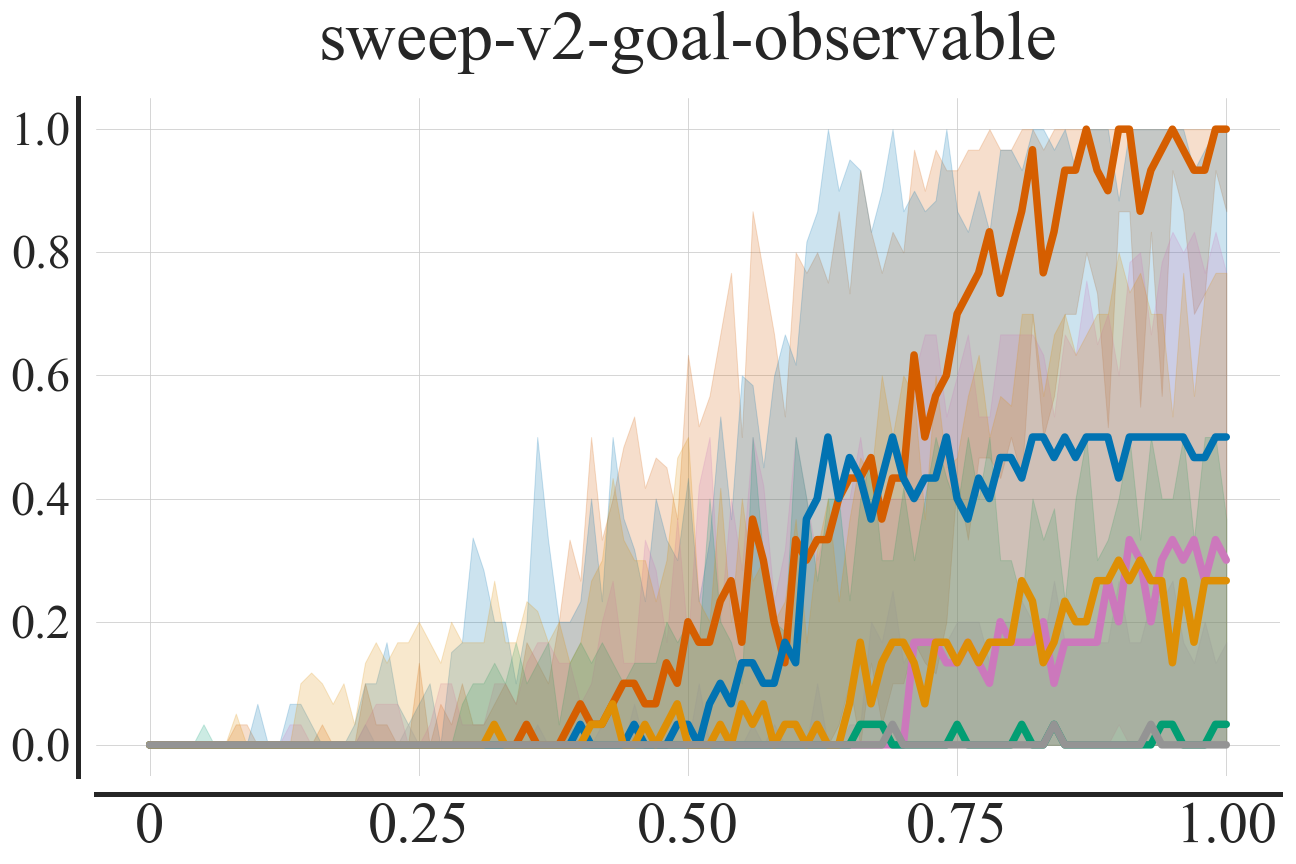}
    \hfill
    \includegraphics[width=0.195\linewidth]{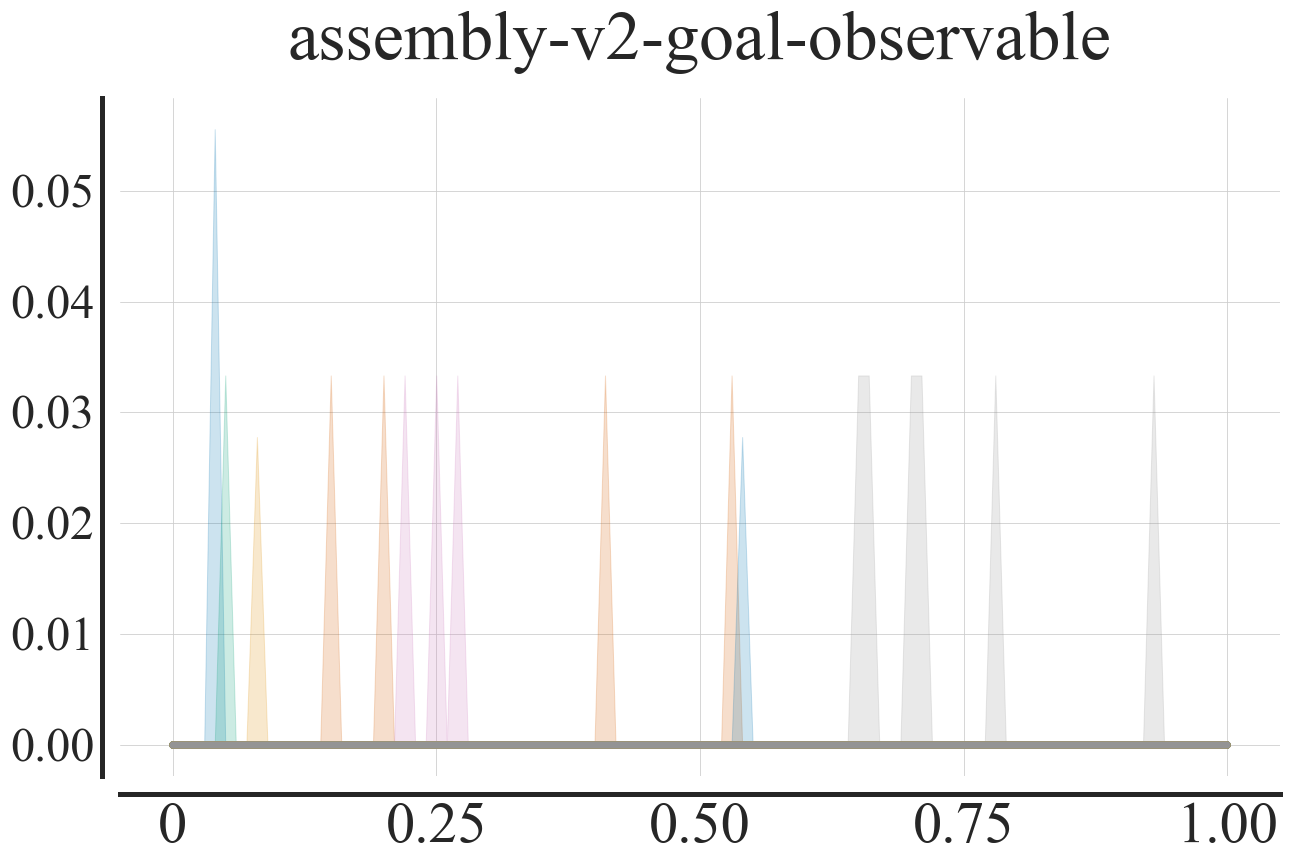}
    \end{subfigure}
\end{minipage}
\begin{minipage}[h]{1.0\linewidth}
    \begin{subfigure}{1.0\linewidth}
    \includegraphics[width=0.195\linewidth]{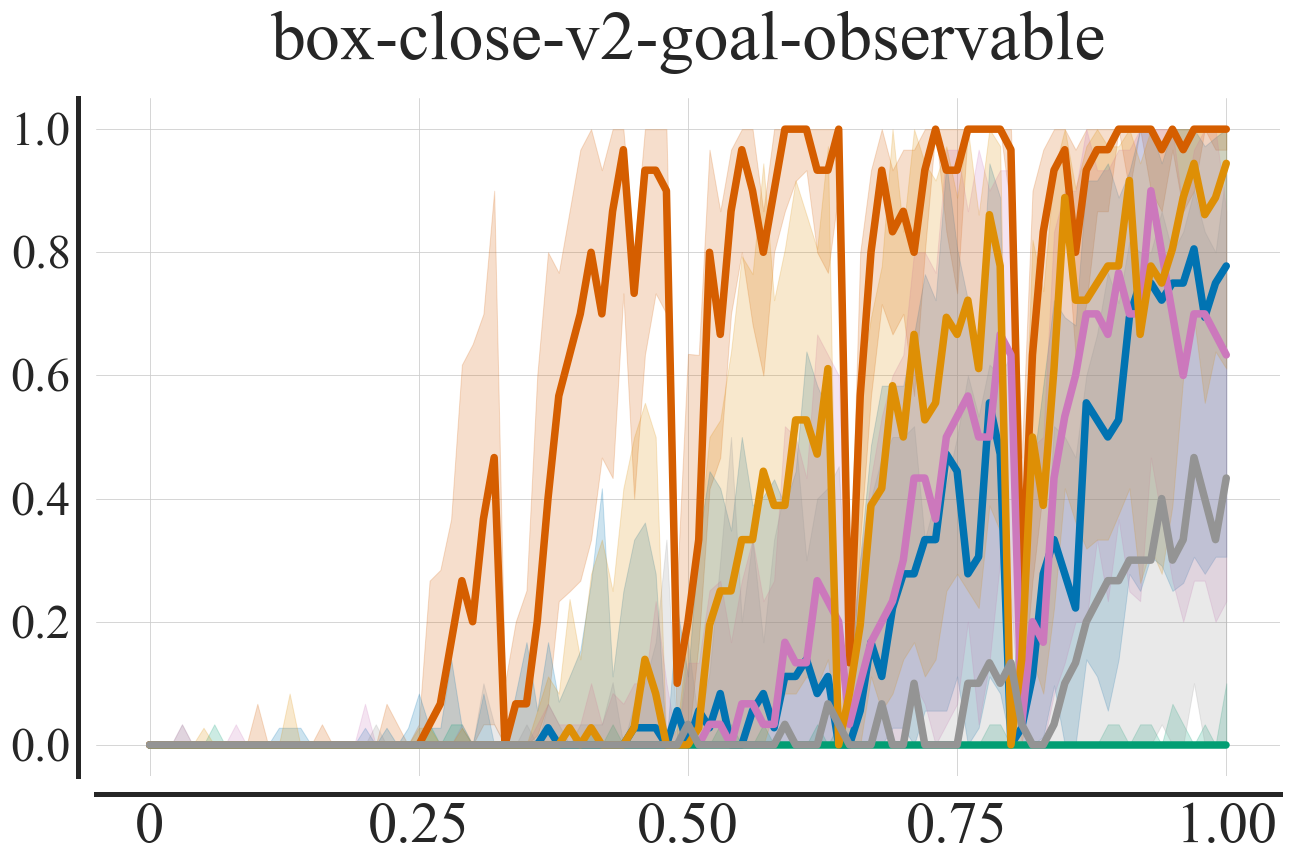}
    \hfill
    \includegraphics[width=0.195\linewidth]{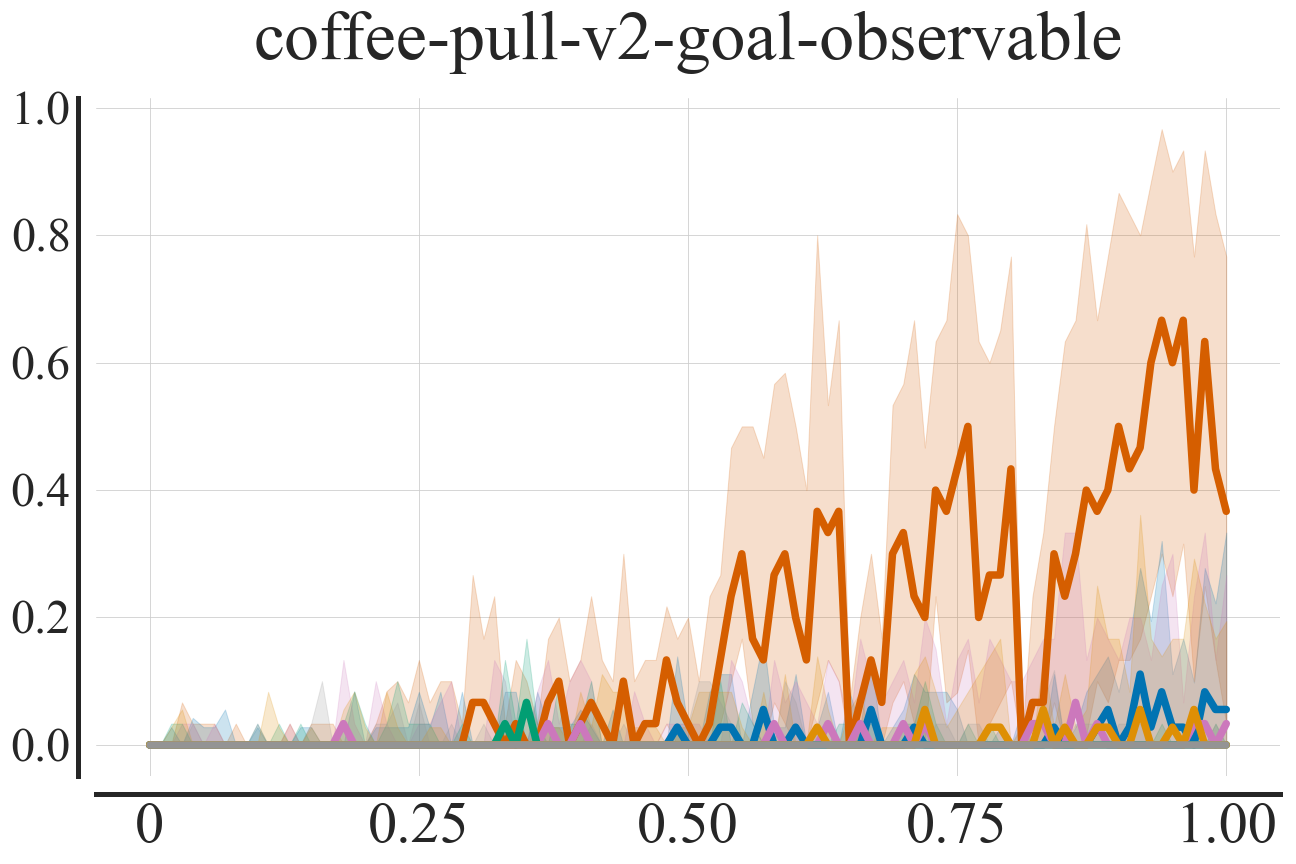}
    \hfill
    \includegraphics[width=0.195\linewidth]{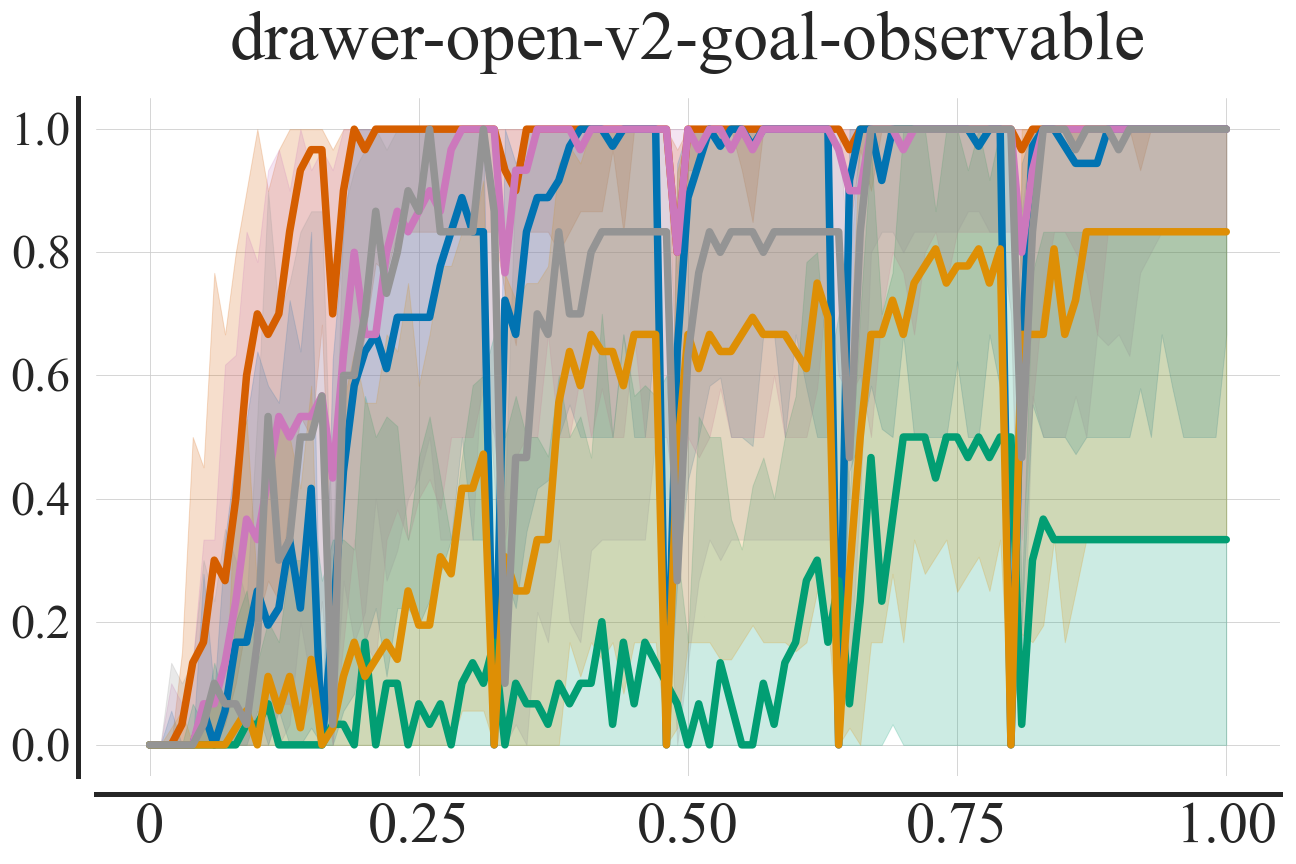}
    \hfill
    \includegraphics[width=0.195\linewidth]{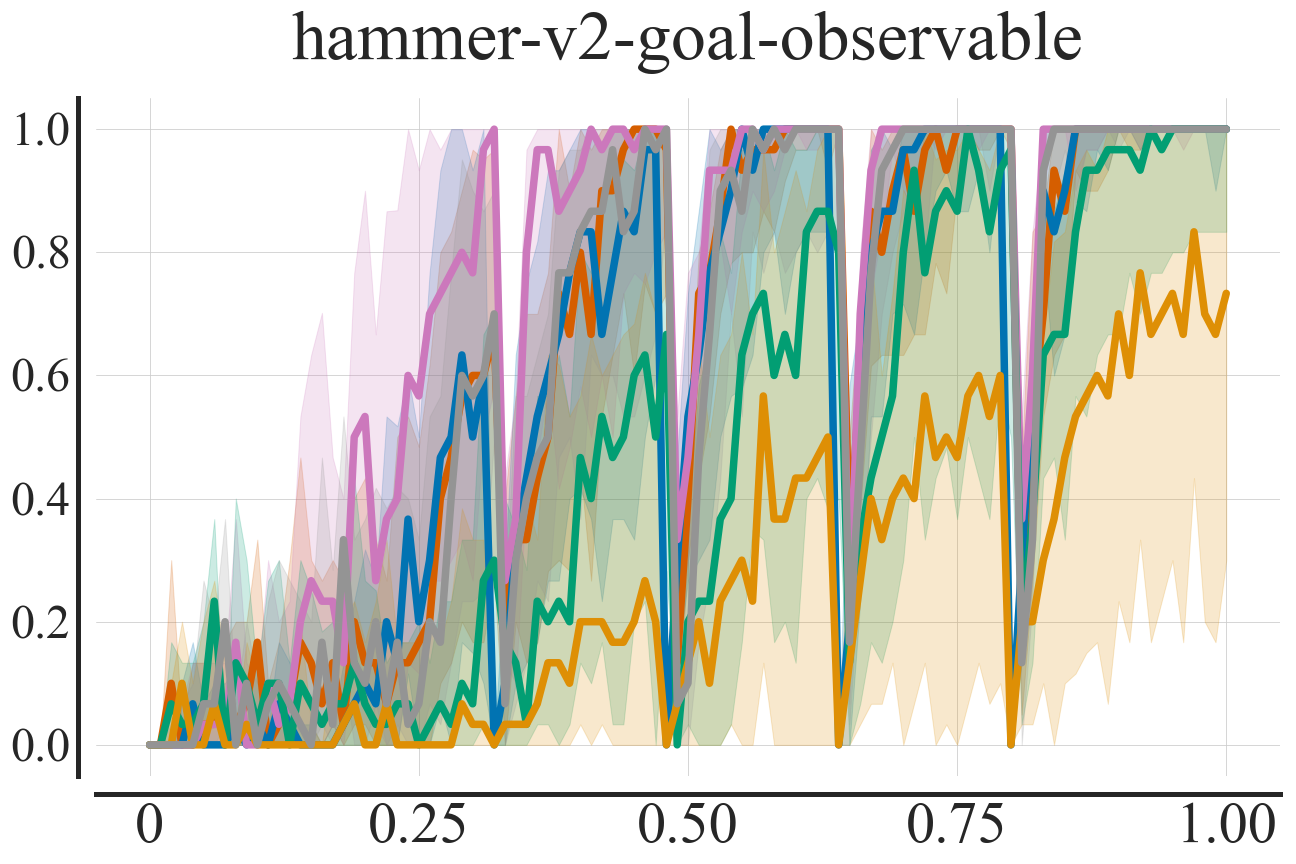}
    \hfill
    \includegraphics[width=0.195\linewidth]{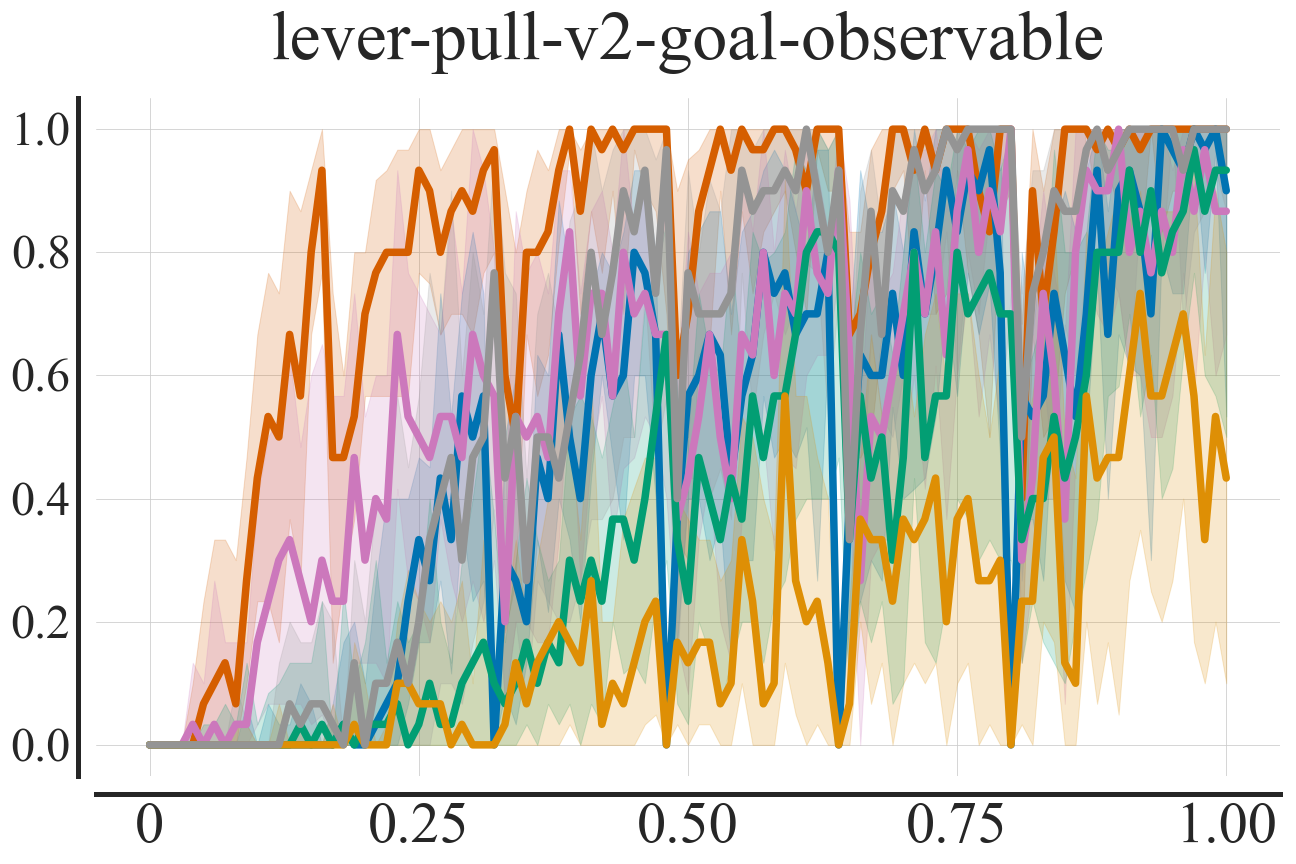}
    \end{subfigure}
\end{minipage}
\begin{minipage}[h]{1.0\linewidth}
    \begin{subfigure}{1.0\linewidth}
    \includegraphics[width=0.195\linewidth]{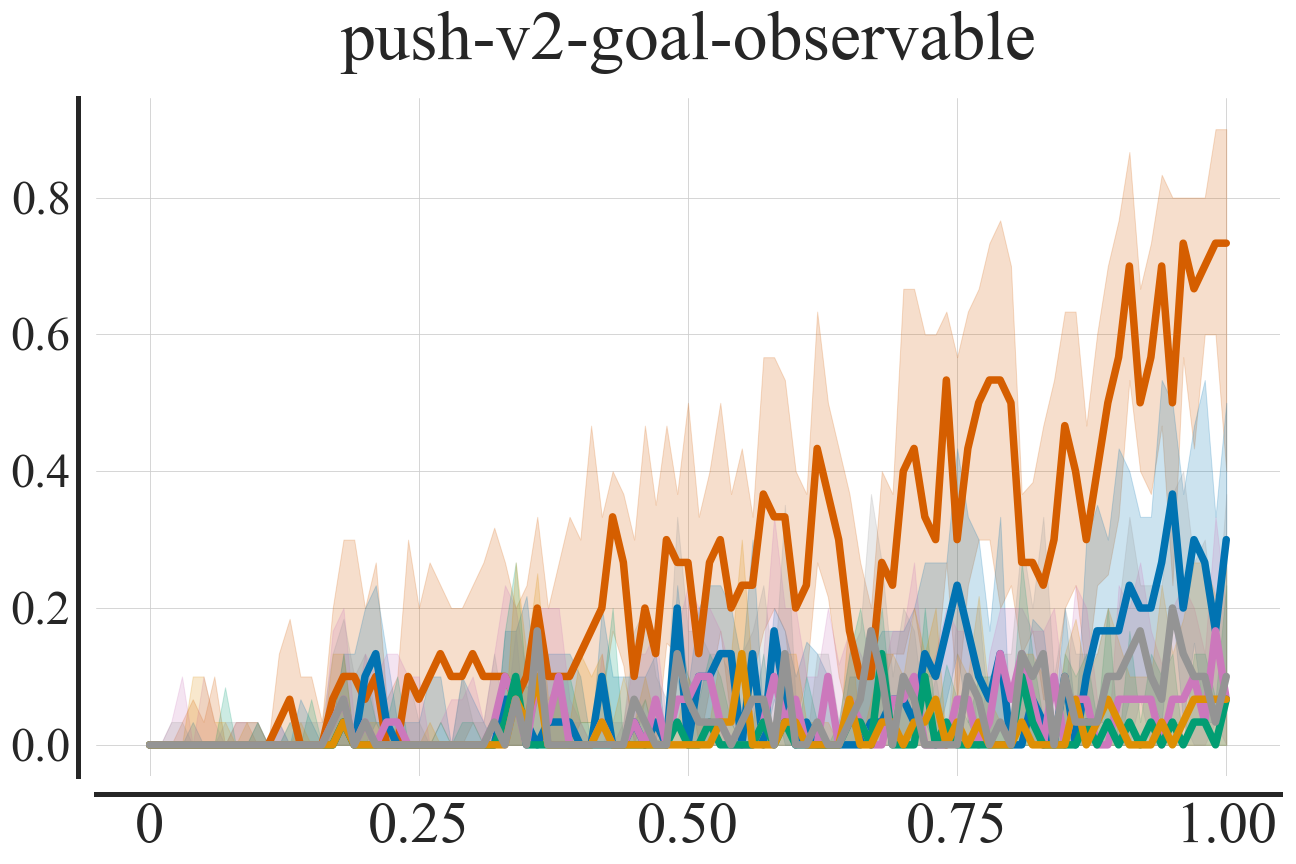}
    \hfill
    \includegraphics[width=0.195\linewidth]{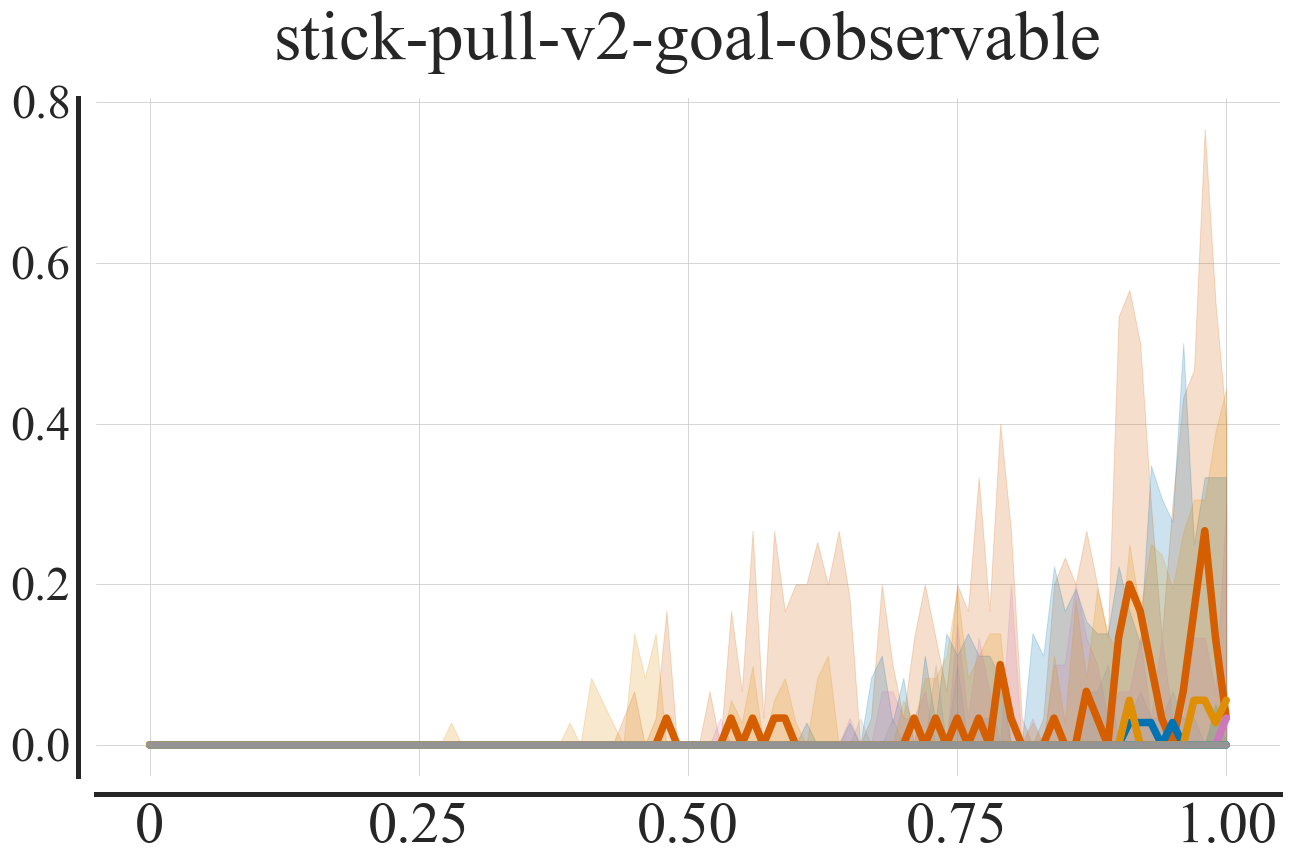}
    \hfill
    \includegraphics[width=0.195\linewidth]{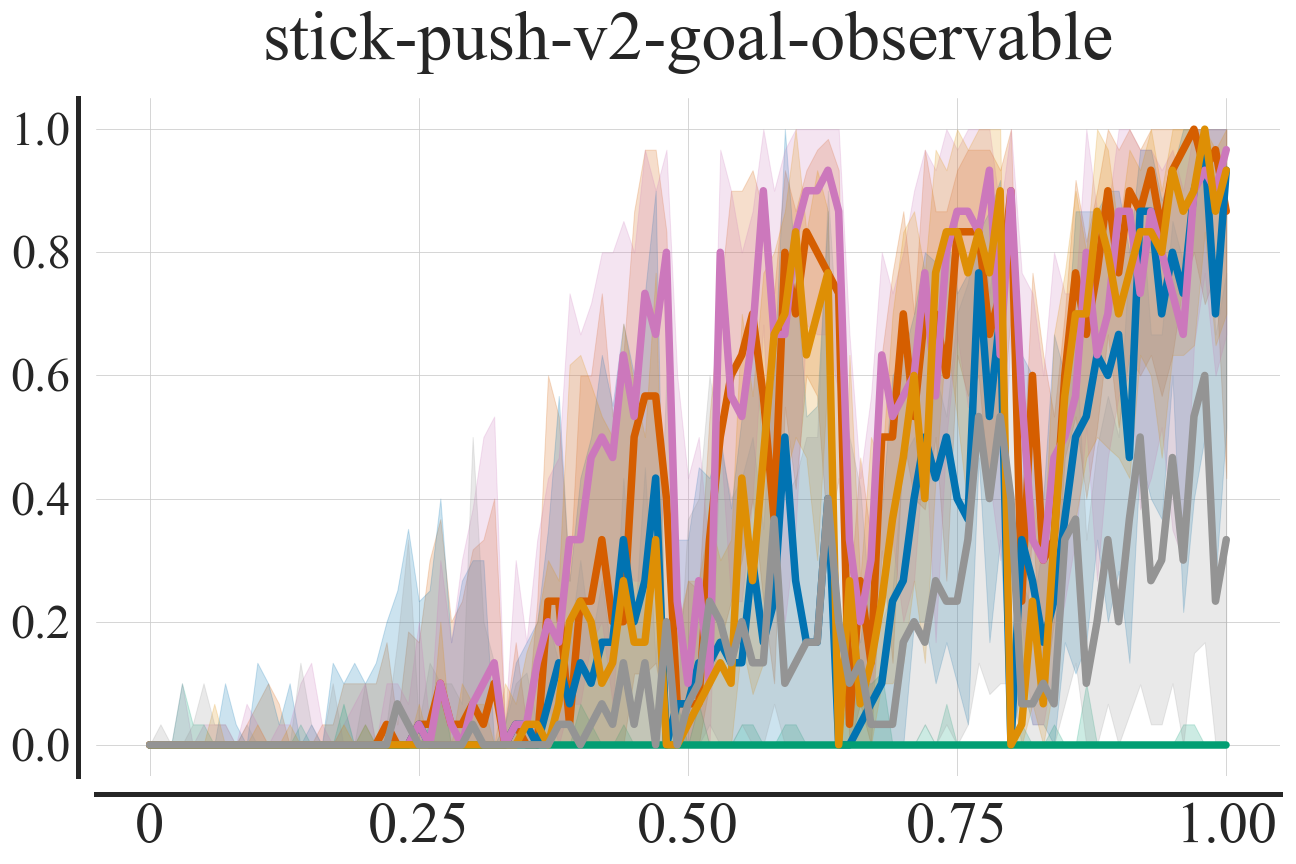}
    \hfill
    \includegraphics[width=0.195\linewidth]{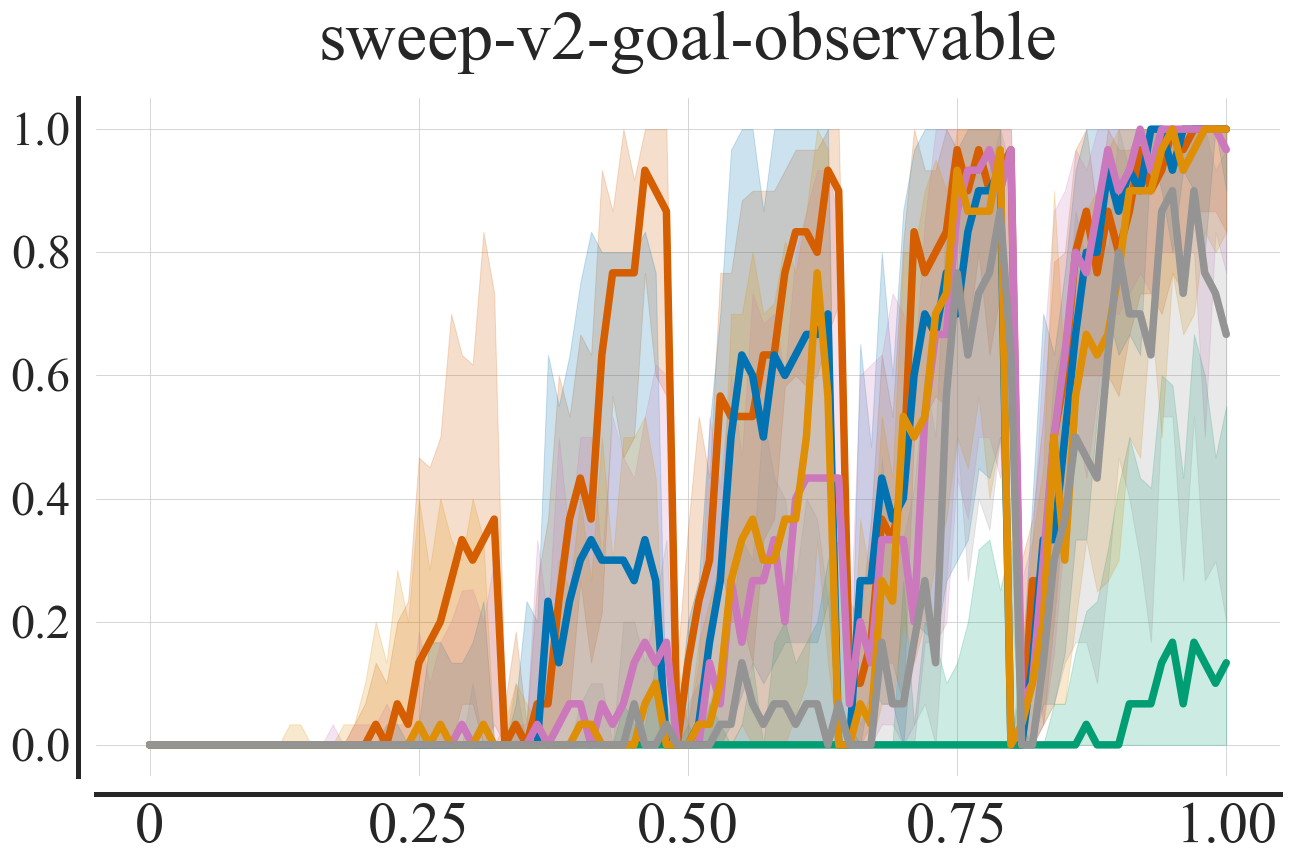}
    \hfill
    \includegraphics[width=0.195\linewidth]{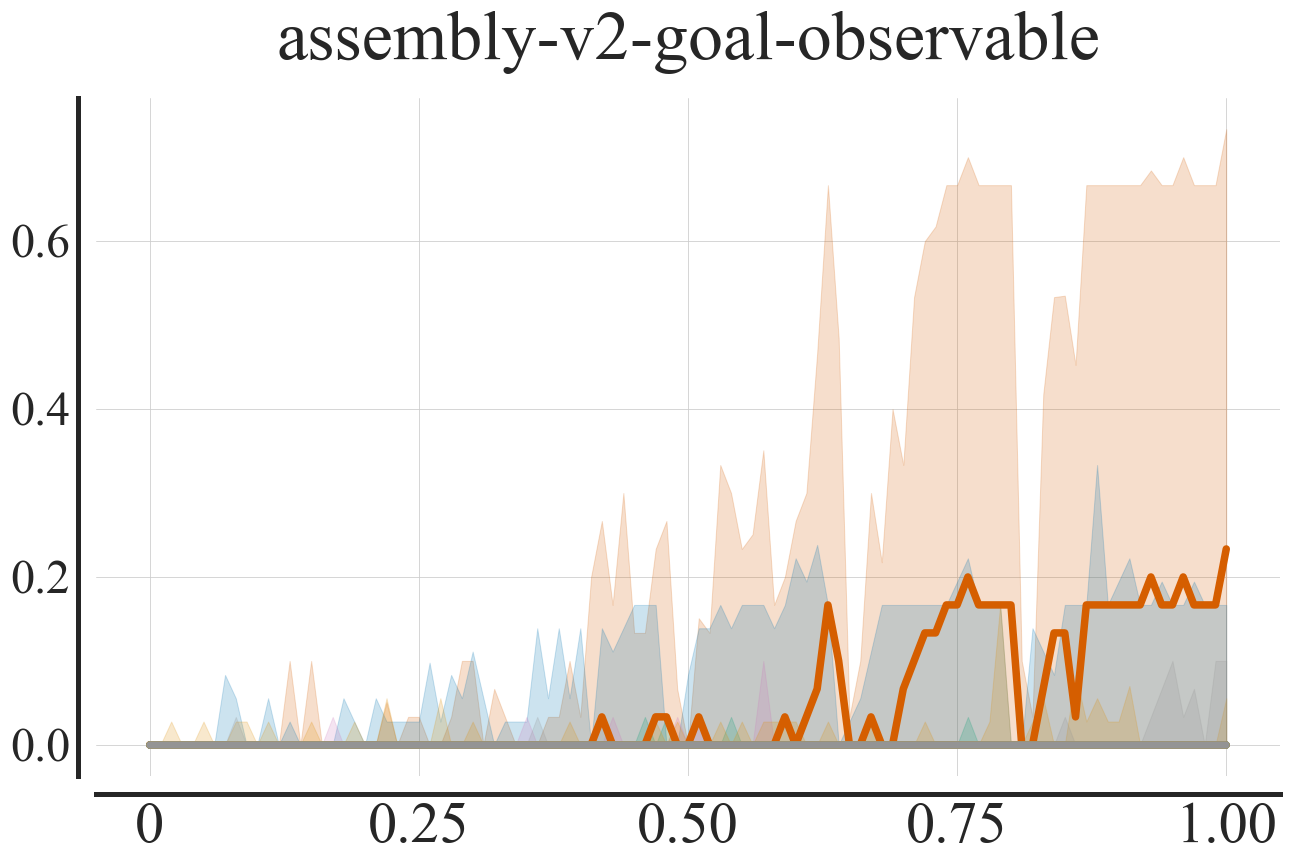}
    \end{subfigure}
\end{minipage}
\caption{Training curves for MW ($RR=2$ rows $1$ \& $2$, $RR=16$ rows $3$ \& $4$). $Y$-axis denotes IQM success rates and $X$-axis denotes environment steps. 10 seeds per task.}
\label{fig:training_rr2_mw}
\end{center}
\vspace{-0.1in} 
\end{figure}

\begin{figure}[ht!]
\begin{center}
\begin{minipage}[h]{1.0\linewidth}
\centering
    \begin{subfigure}{0.95\linewidth}
    \includegraphics[width=\textwidth]{images/appendix/legend_training.png}
    \end{subfigure}
\end{minipage}
\begin{minipage}[h]{1.0\linewidth}
    \begin{subfigure}{1.0\linewidth}
    \includegraphics[width=0.195\linewidth]{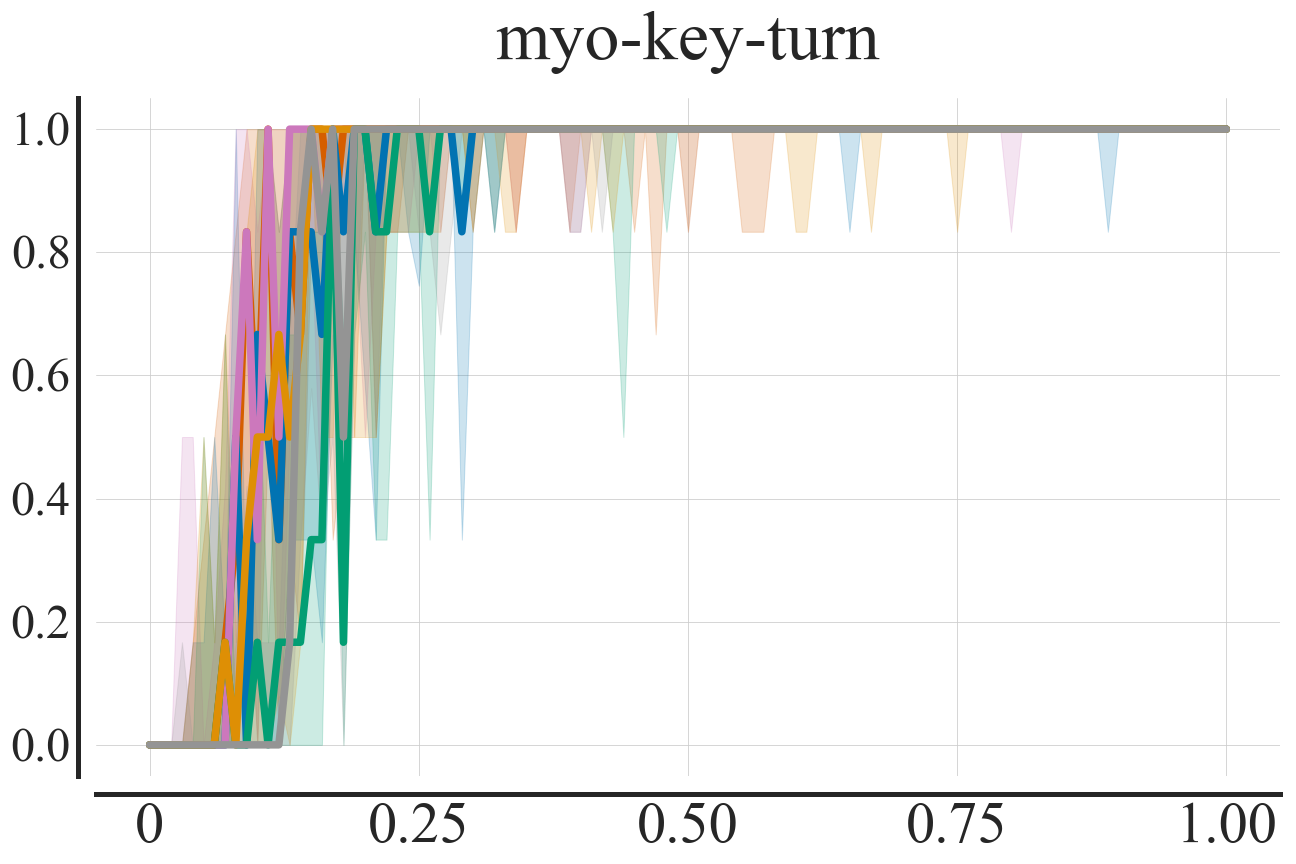}
    \hfill
    \includegraphics[width=0.195\linewidth]{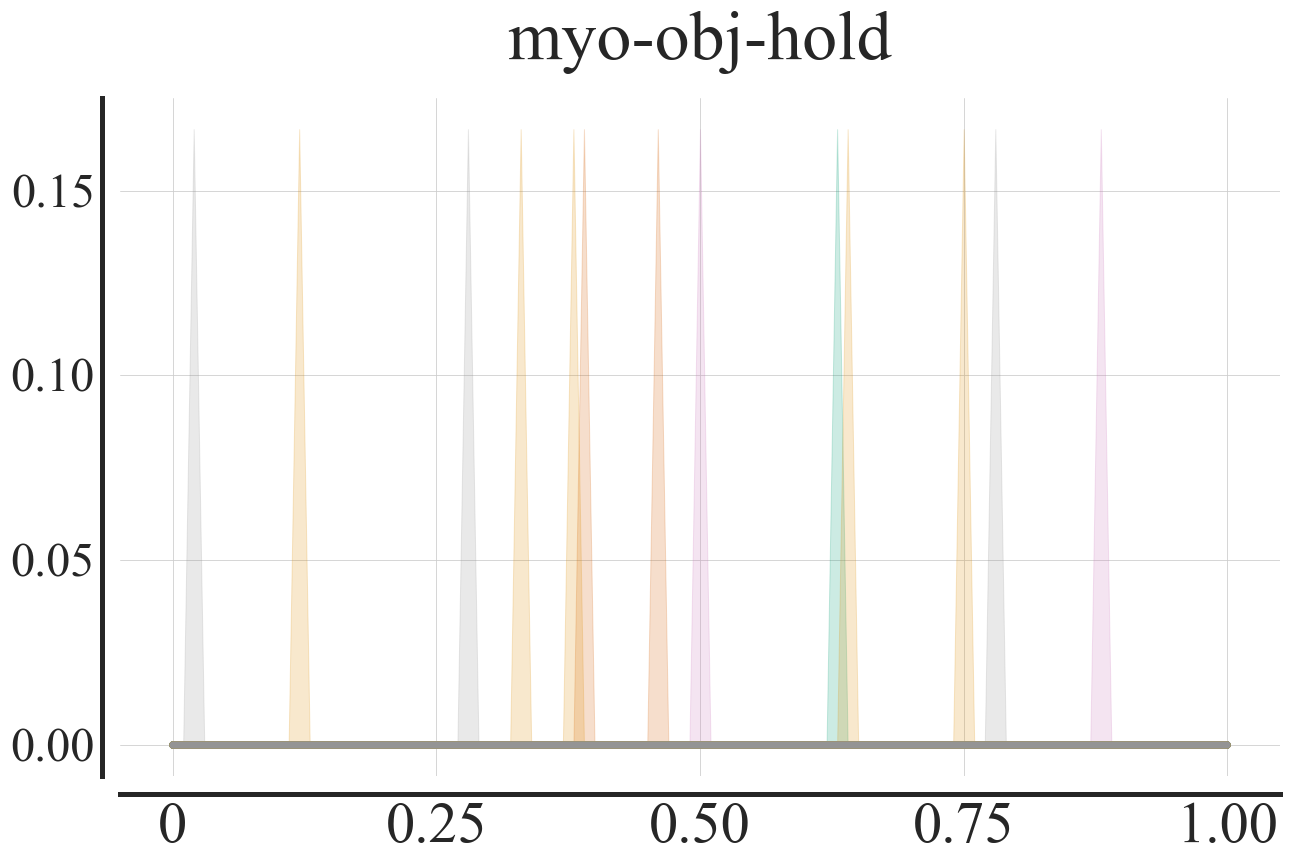}
    \hfill
    \includegraphics[width=0.195\linewidth]{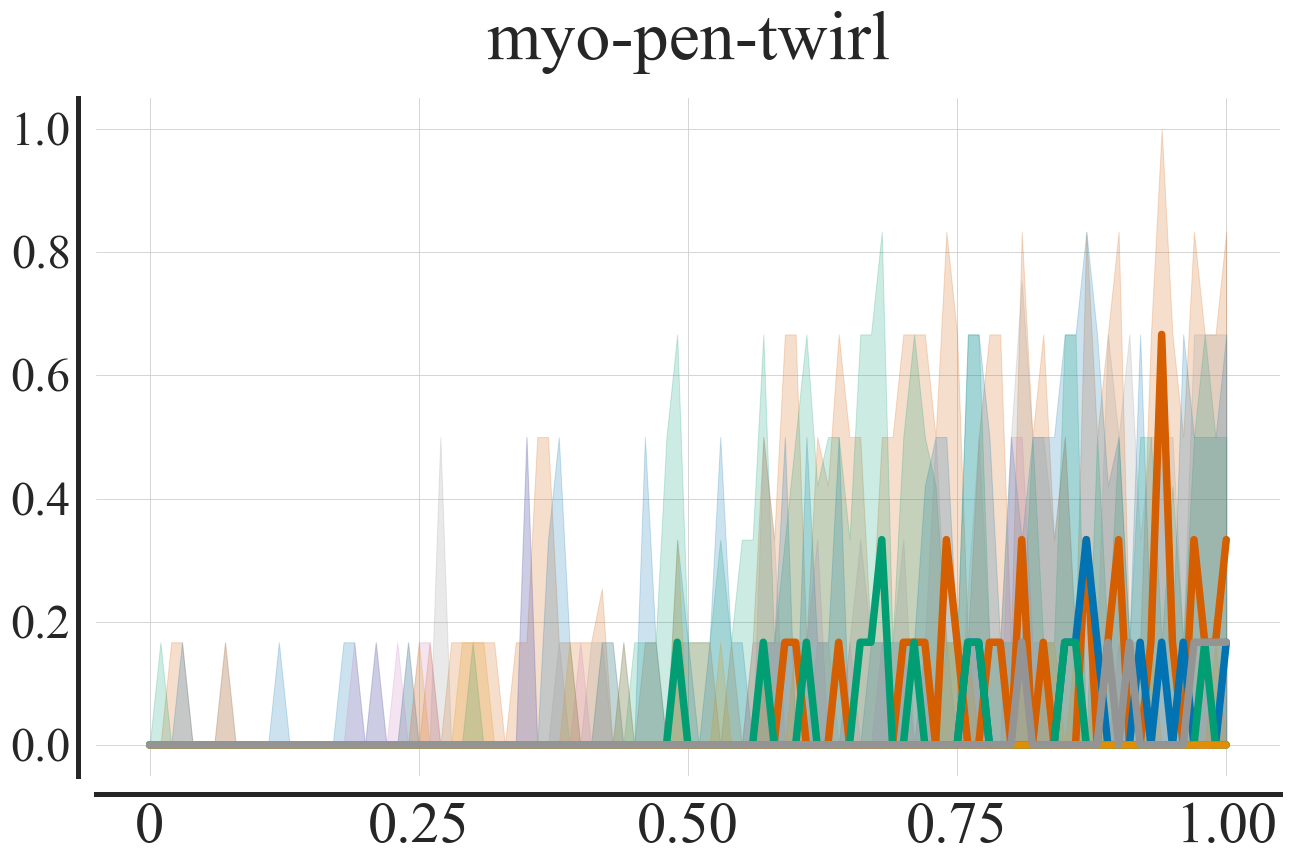}
    \hfill
    \includegraphics[width=0.195\linewidth]{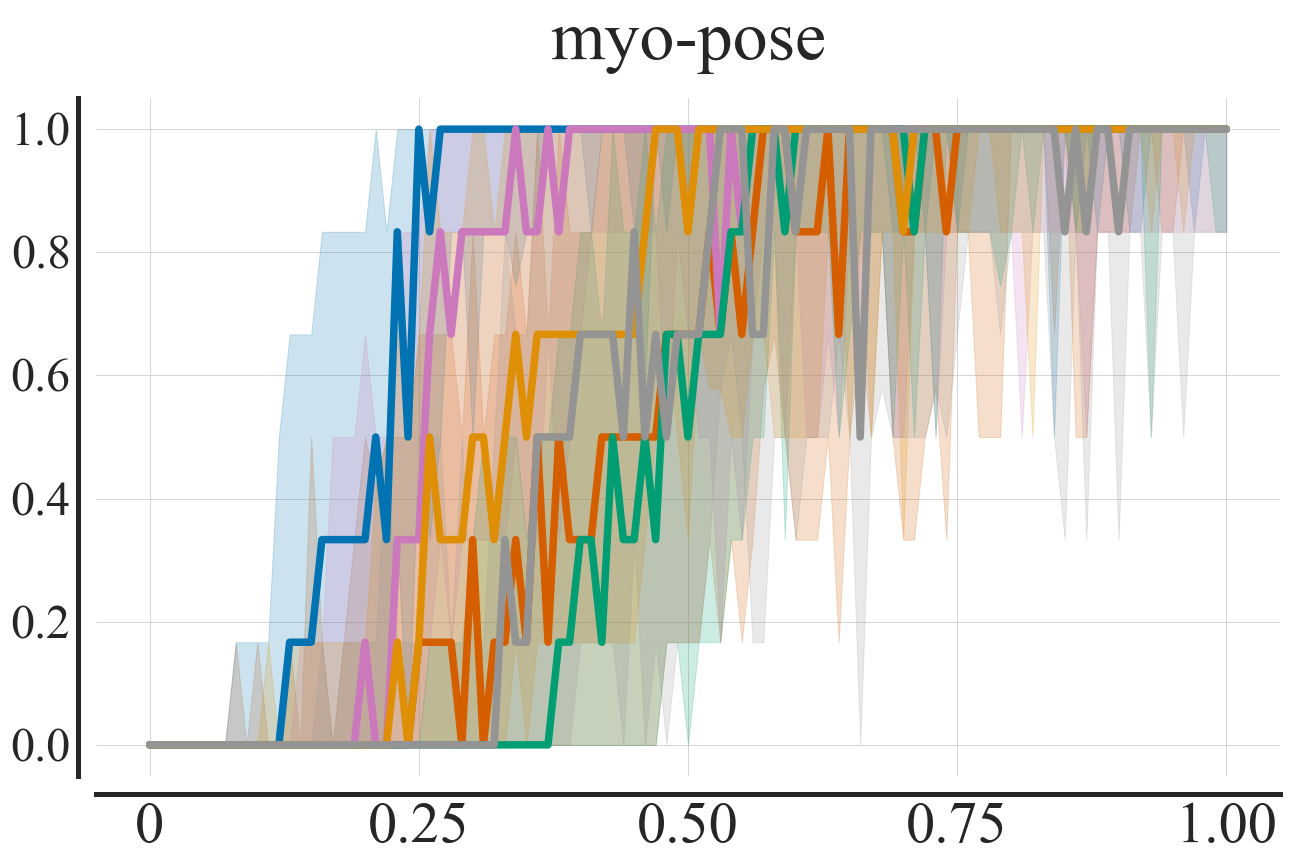}
    \hfill
    \includegraphics[width=0.195\linewidth]{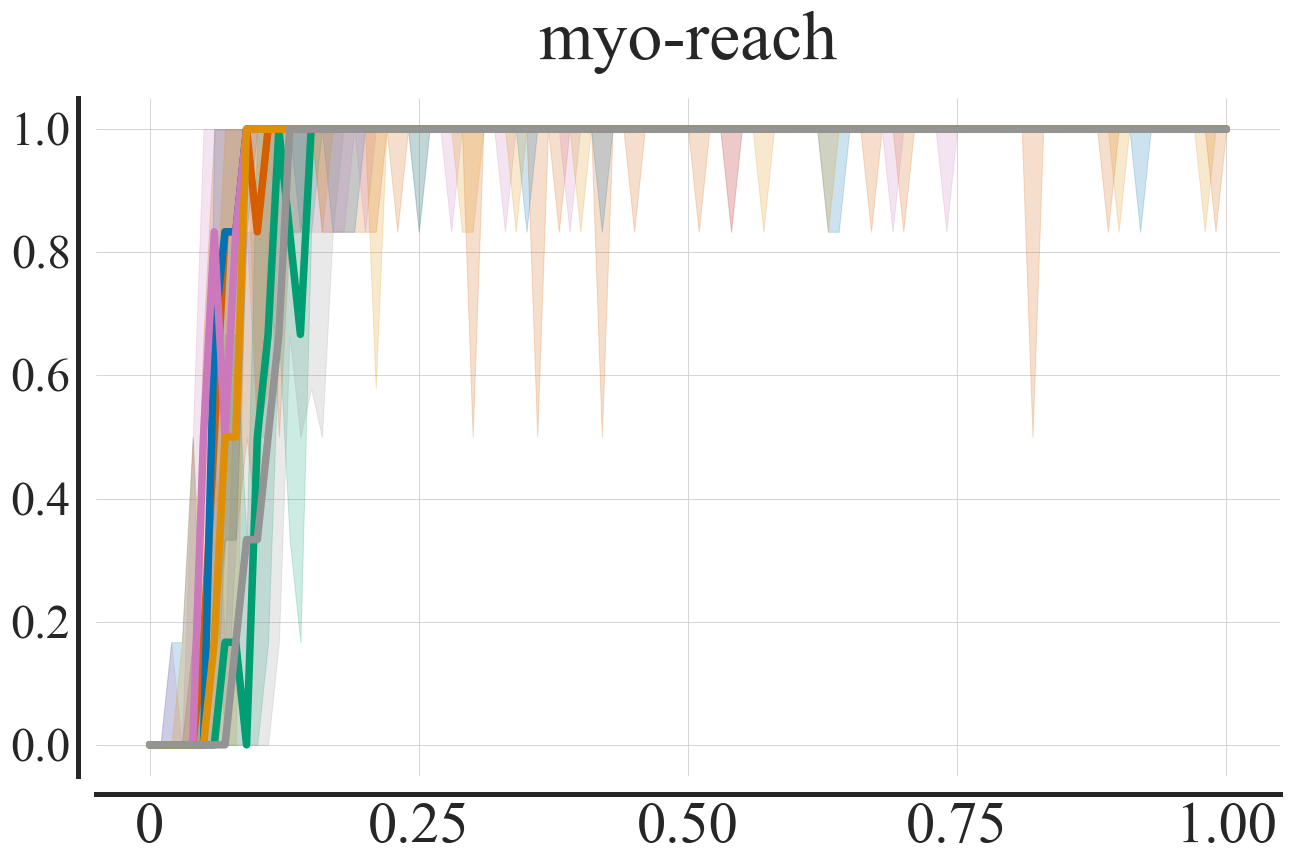}
    \end{subfigure}
\end{minipage}
\begin{minipage}[h]{1.0\linewidth}
    \begin{subfigure}{1.0\linewidth}
    \includegraphics[width=0.195\linewidth]{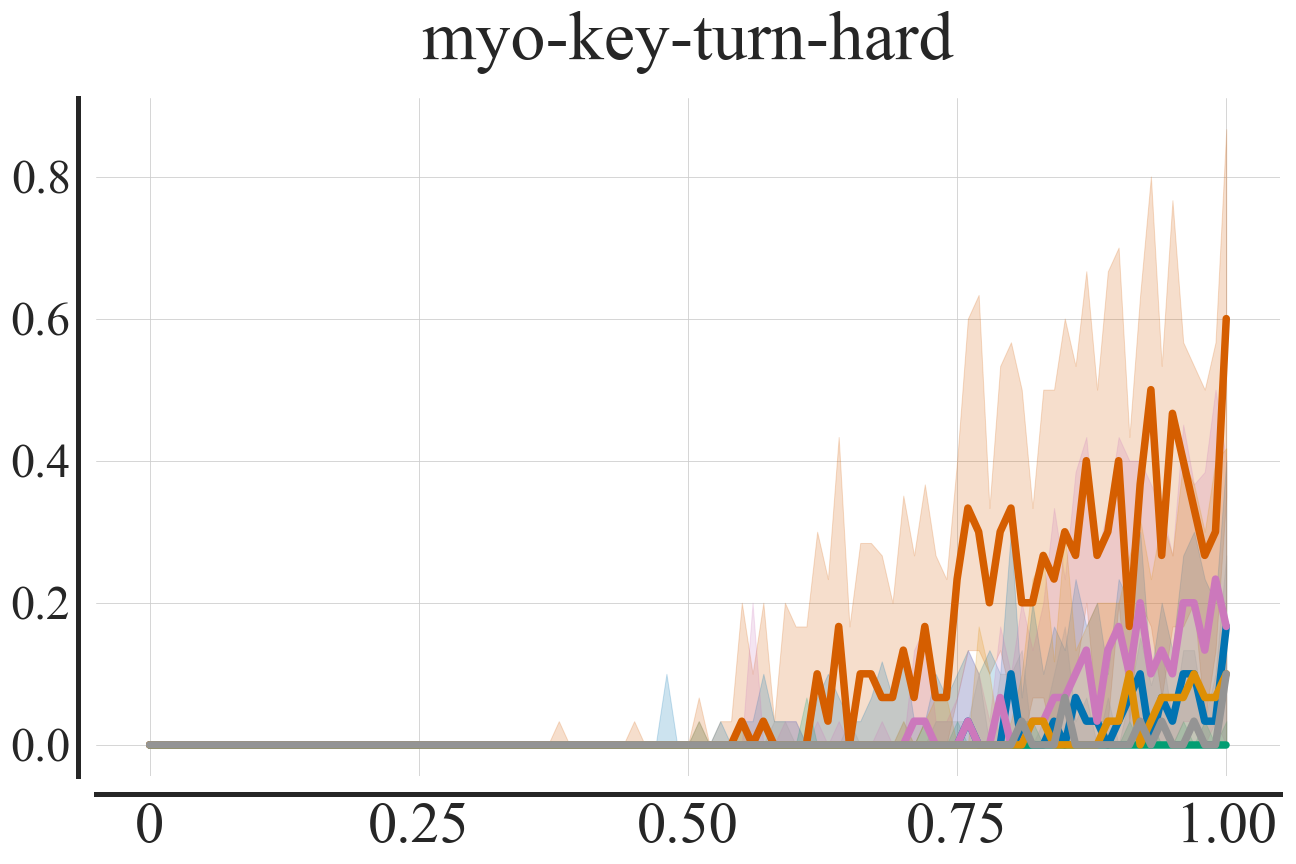}
    \hfill
    \includegraphics[width=0.195\linewidth]{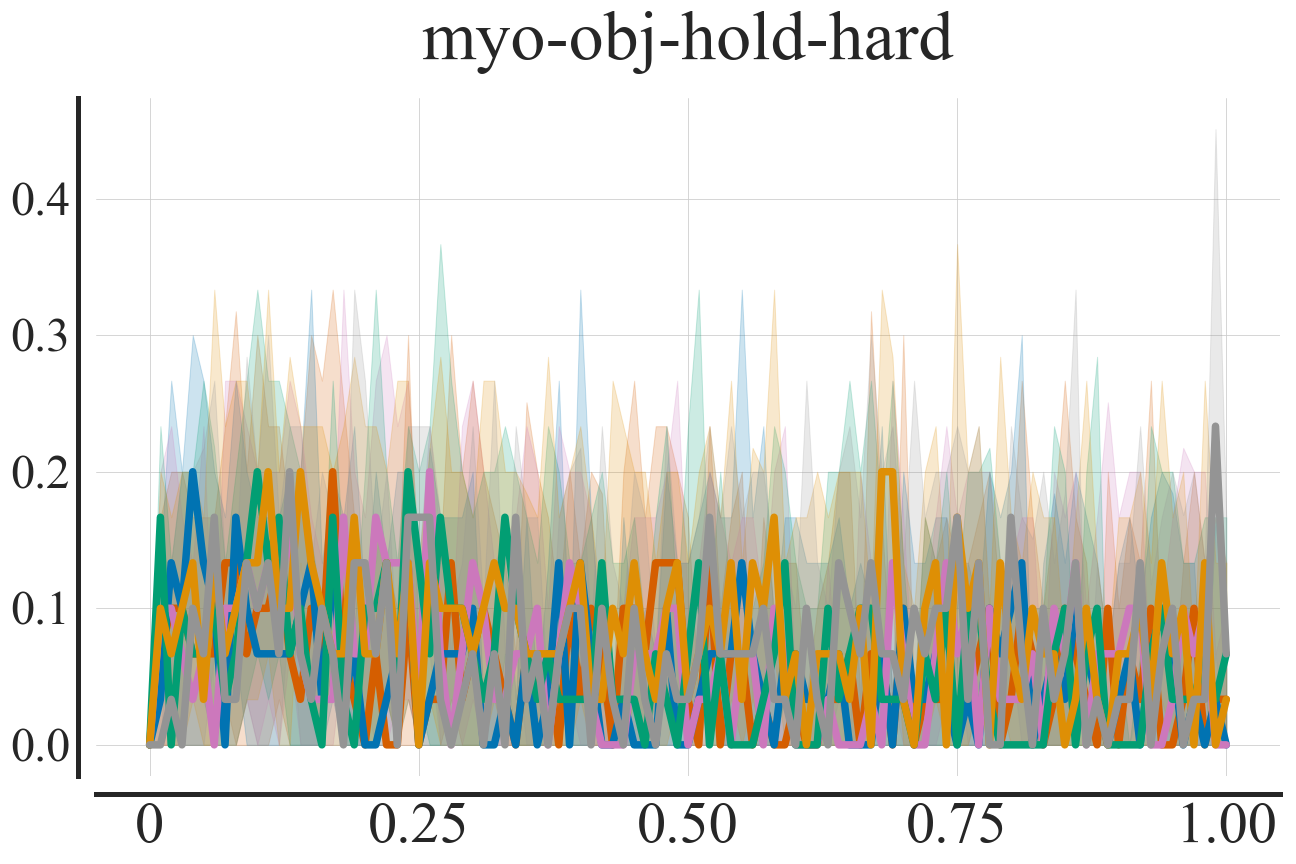}
    \hfill
    \includegraphics[width=0.195\linewidth]{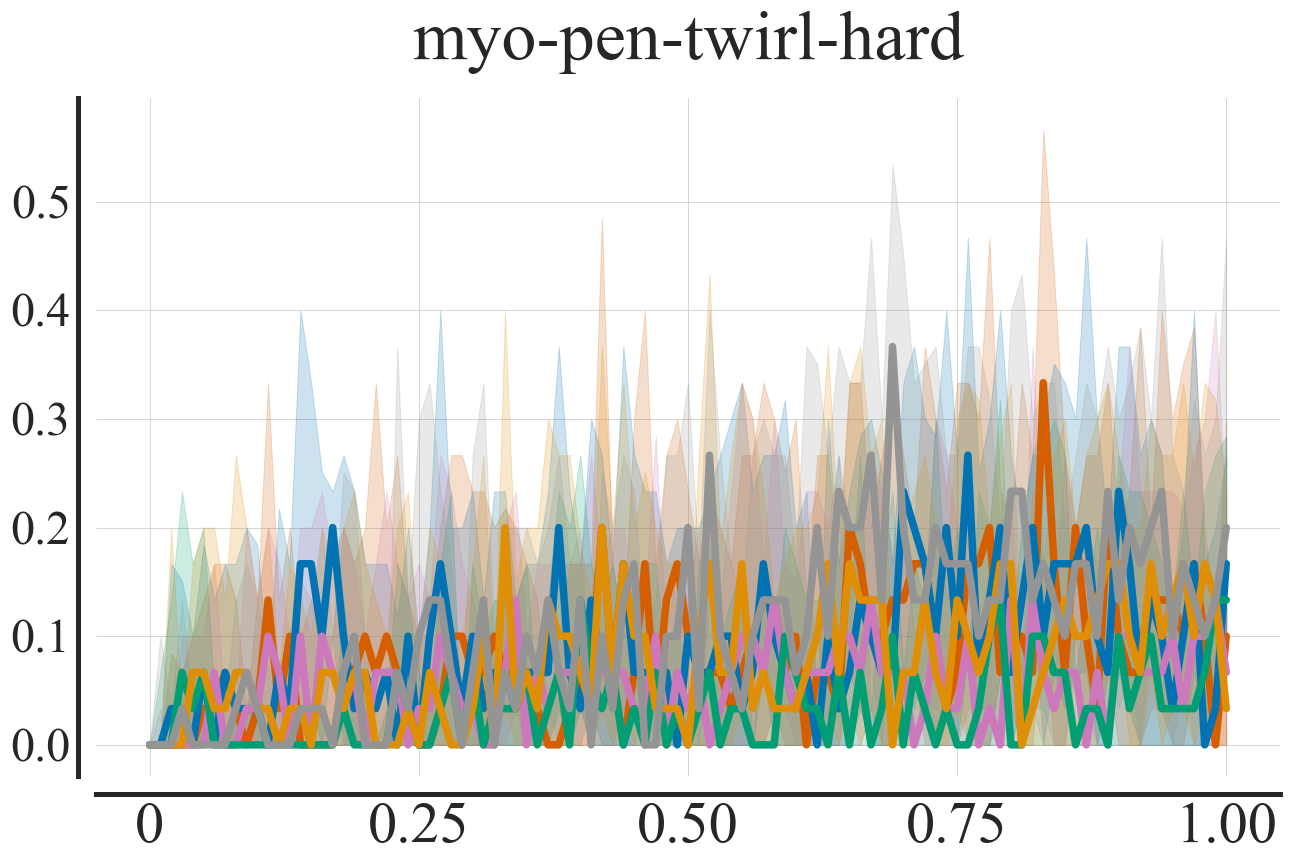}
    \hfill
    \includegraphics[width=0.195\linewidth]{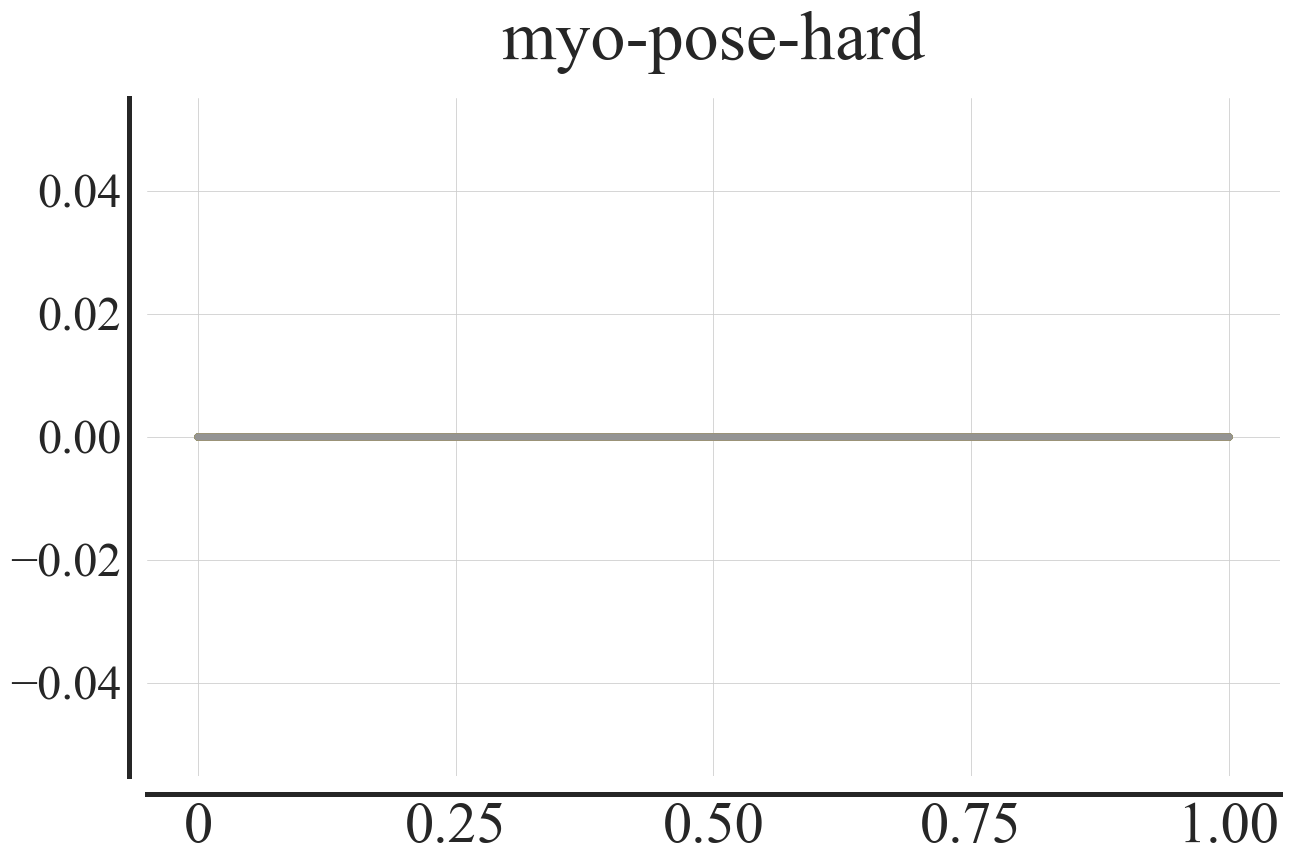}
    \hfill
    \includegraphics[width=0.195\linewidth]{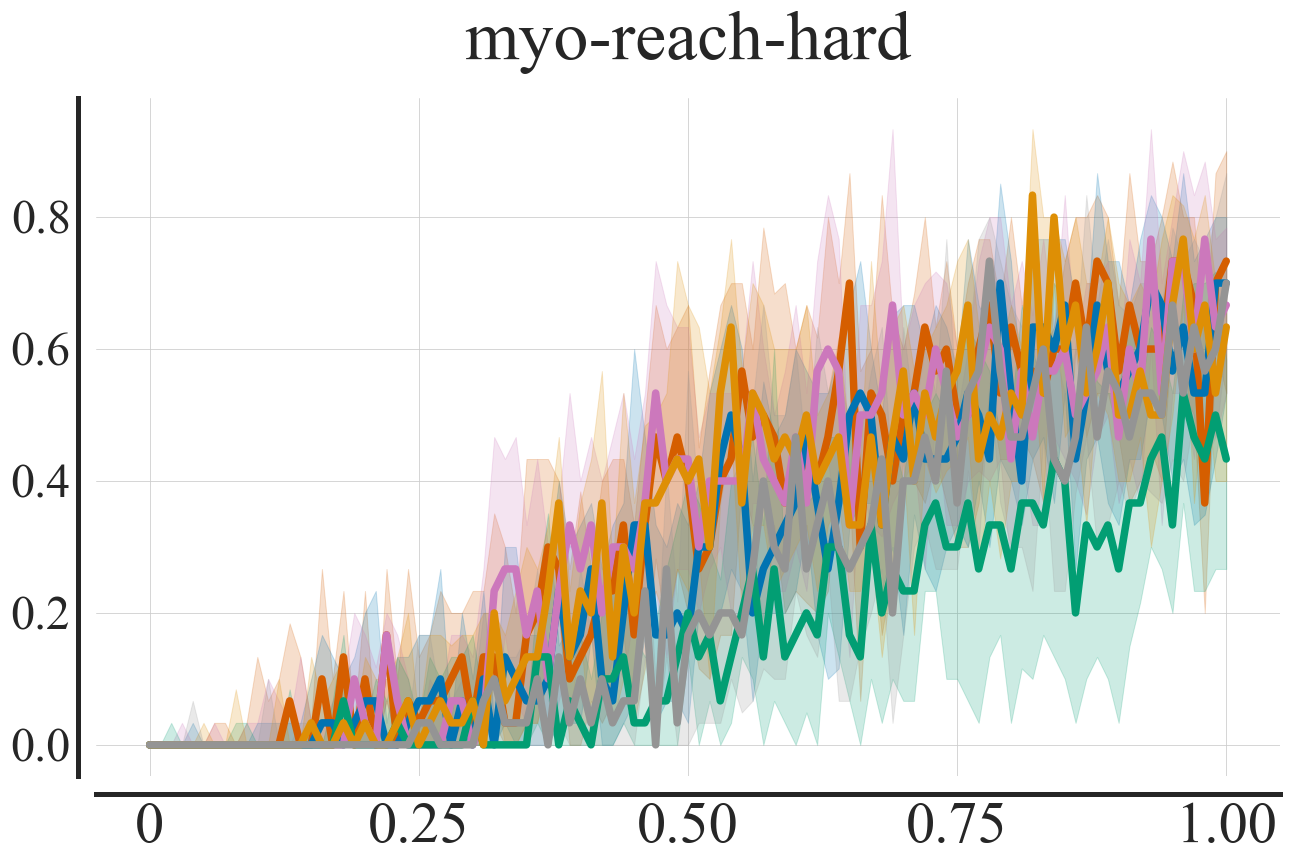}
    \end{subfigure}
\end{minipage}
\begin{minipage}[h]{1.0\linewidth}
    \begin{subfigure}{1.0\linewidth}
    \includegraphics[width=0.195\linewidth]{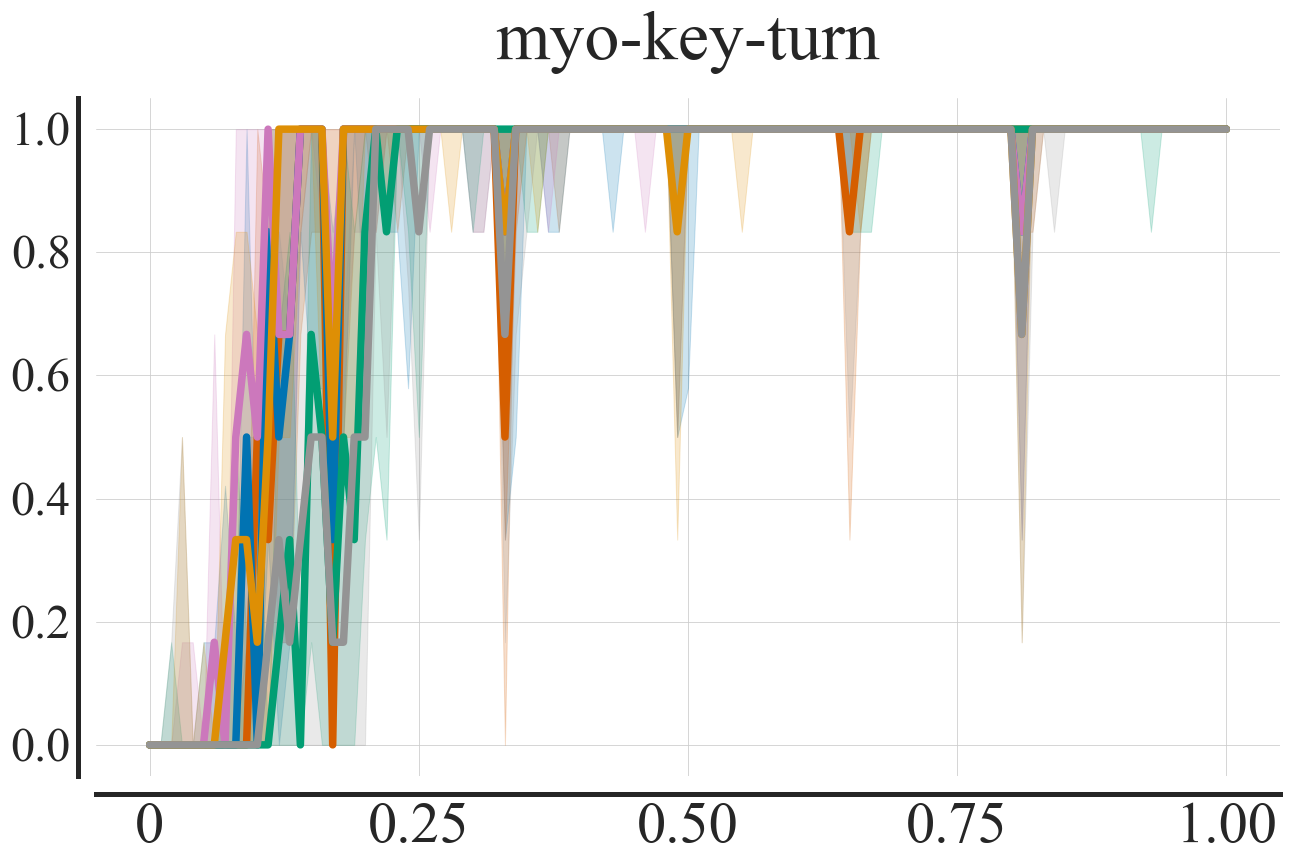}
    \hfill
    \includegraphics[width=0.195\linewidth]{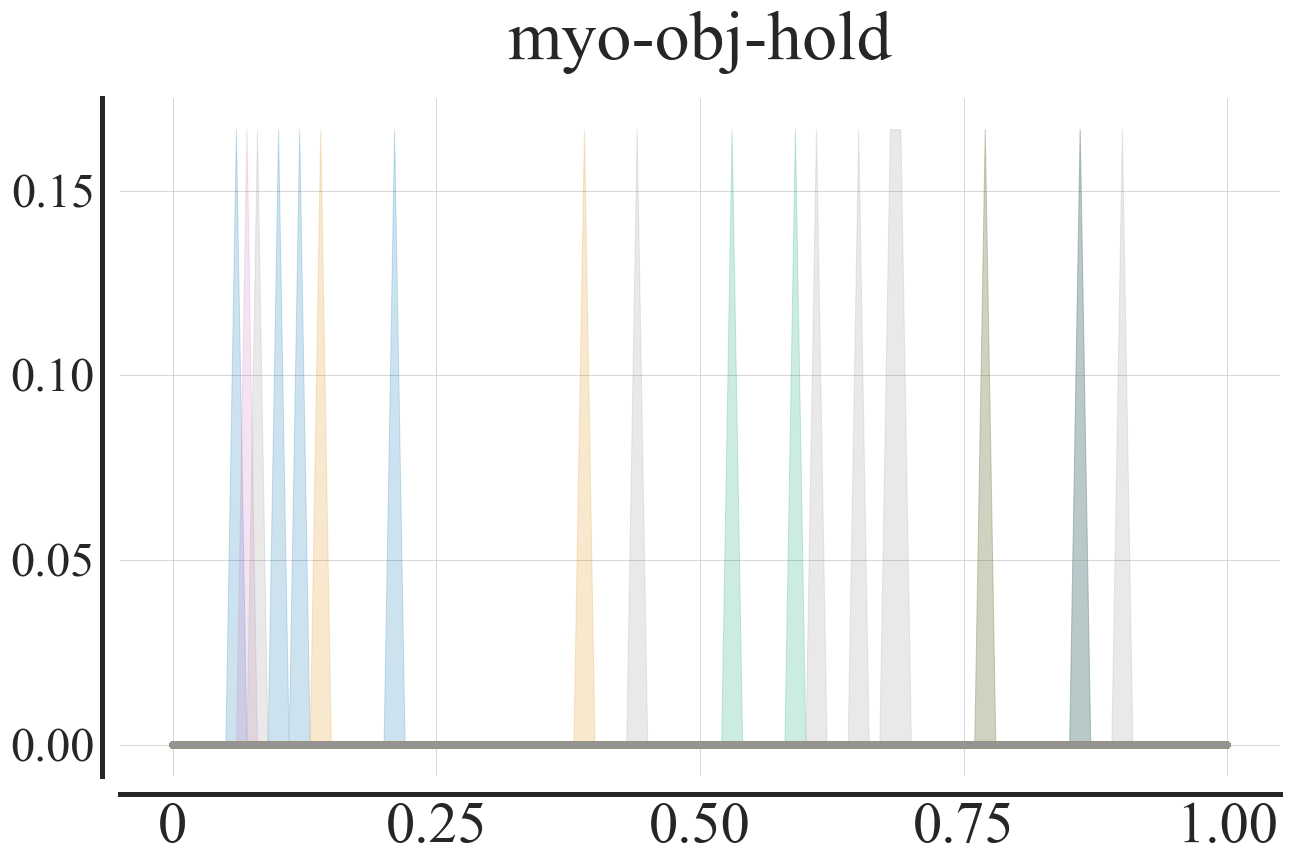}
    \hfill
    \includegraphics[width=0.195\linewidth]{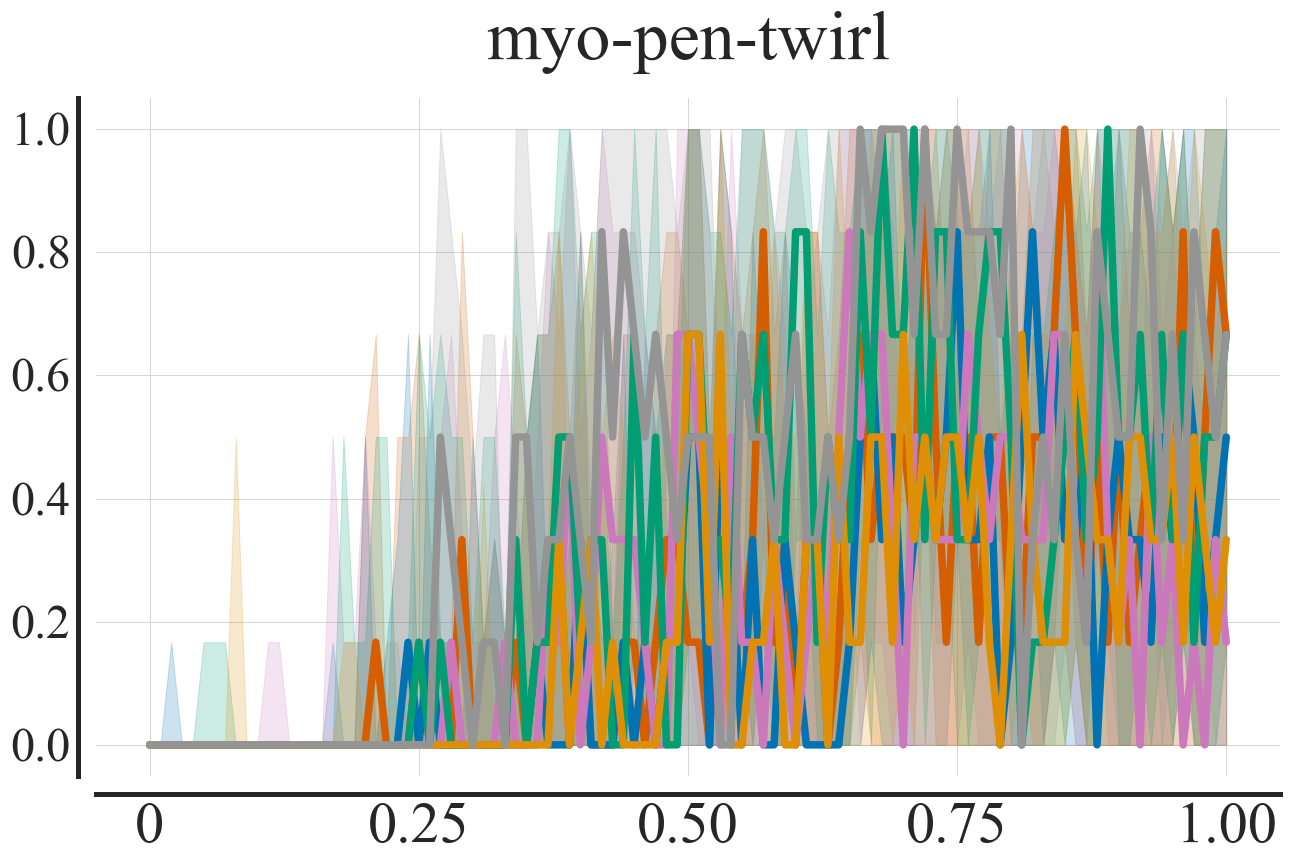}
    \hfill
    \includegraphics[width=0.195\linewidth]{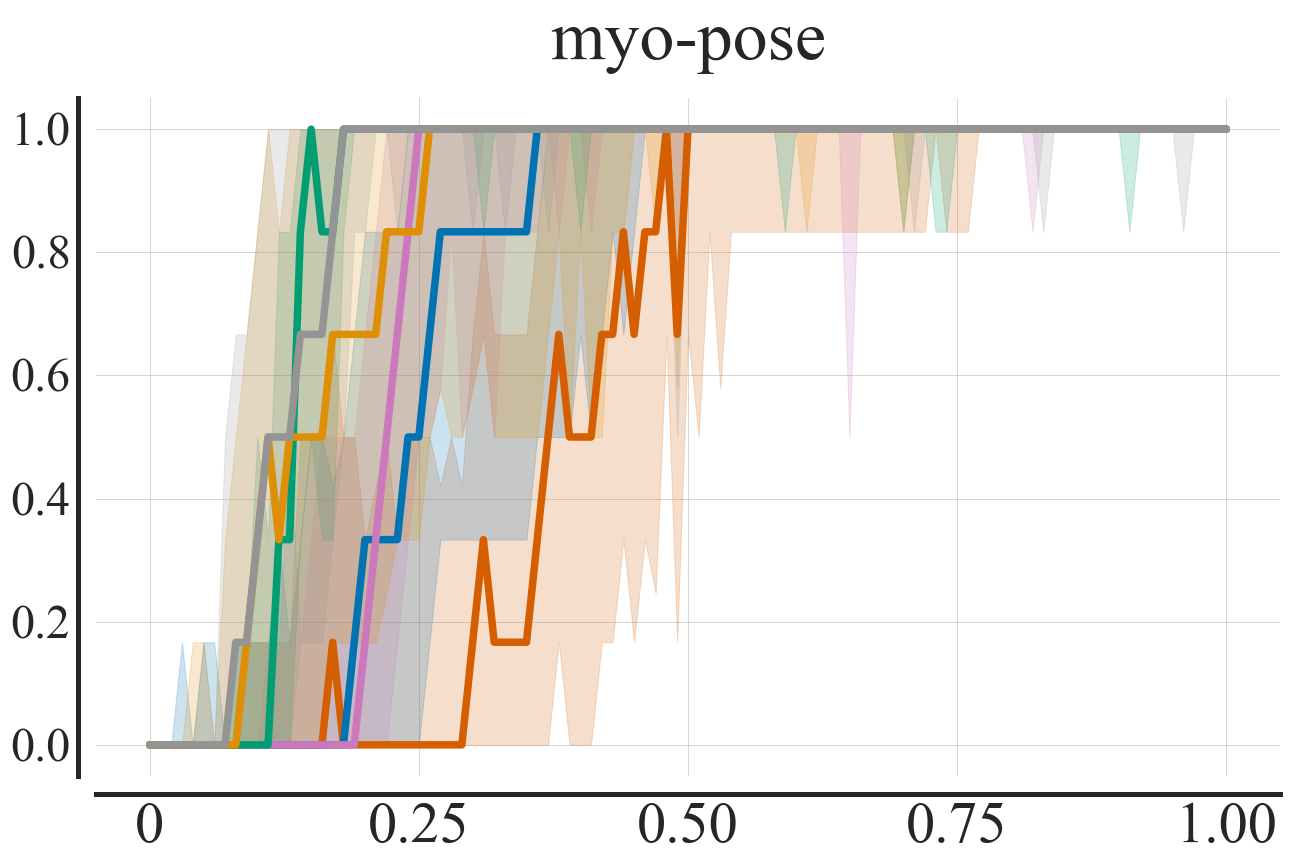}
    \hfill
    \includegraphics[width=0.195\linewidth]{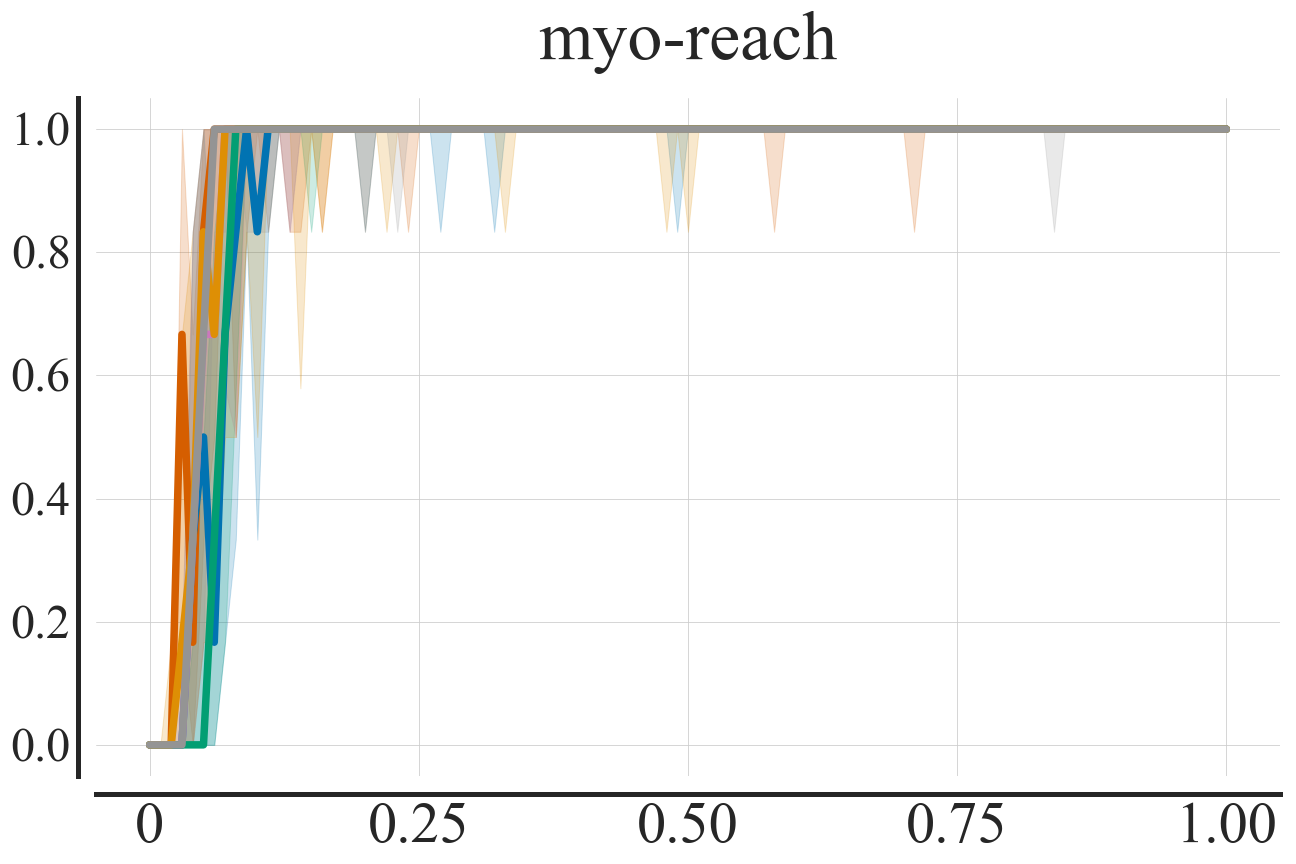}
    \end{subfigure}
\end{minipage}
\begin{minipage}[h]{1.0\linewidth}
    \begin{subfigure}{1.0\linewidth}
    \includegraphics[width=0.195\linewidth]{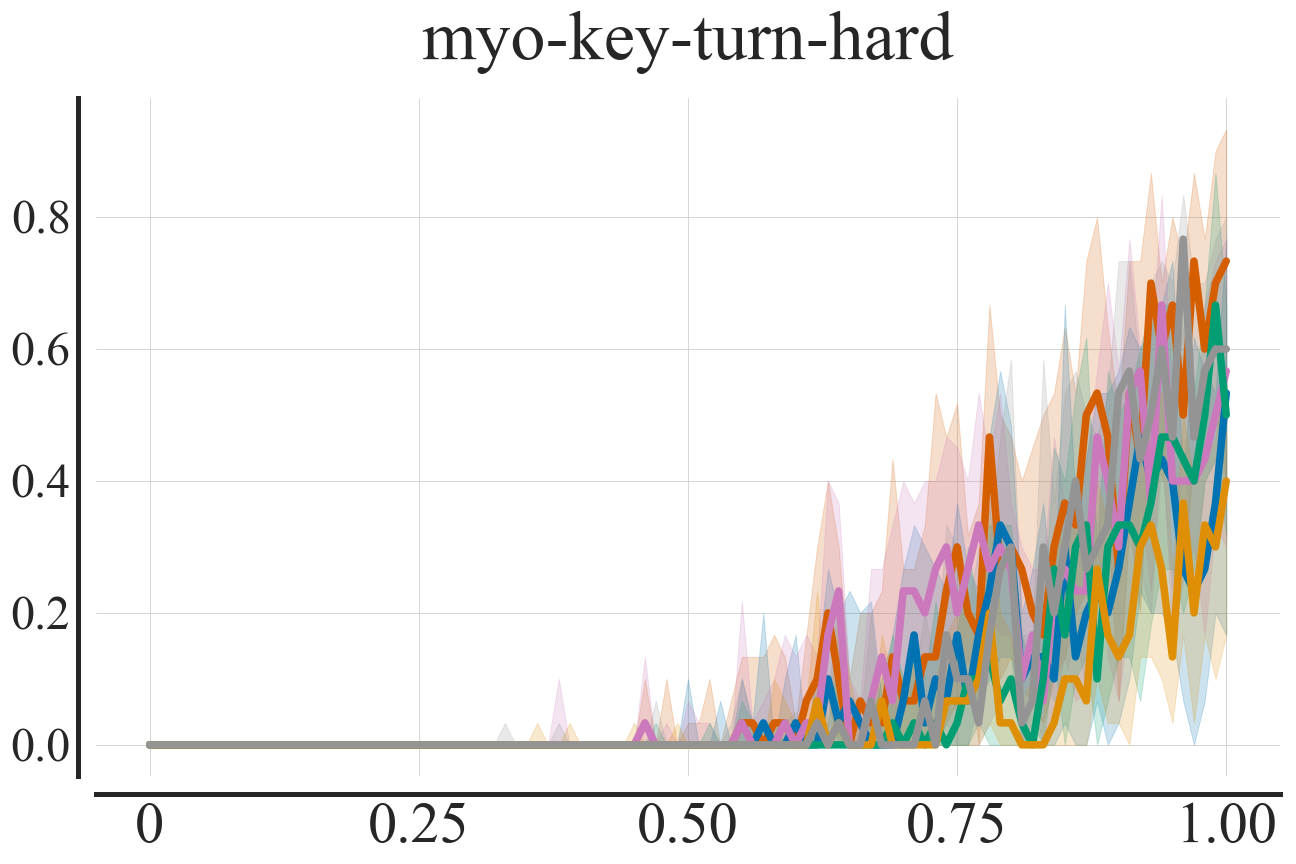}
    \hfill
    \includegraphics[width=0.195\linewidth]{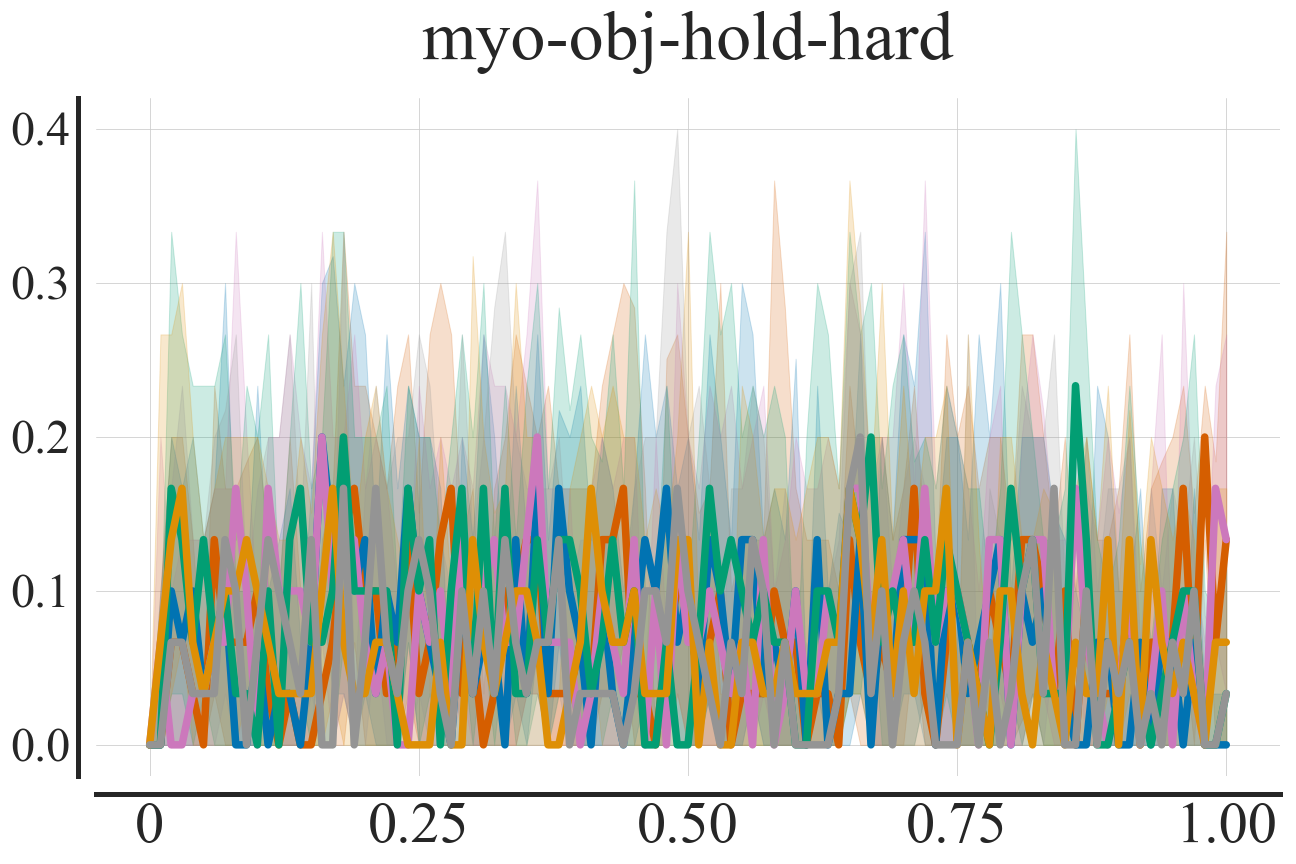}
    \hfill
    \includegraphics[width=0.195\linewidth]{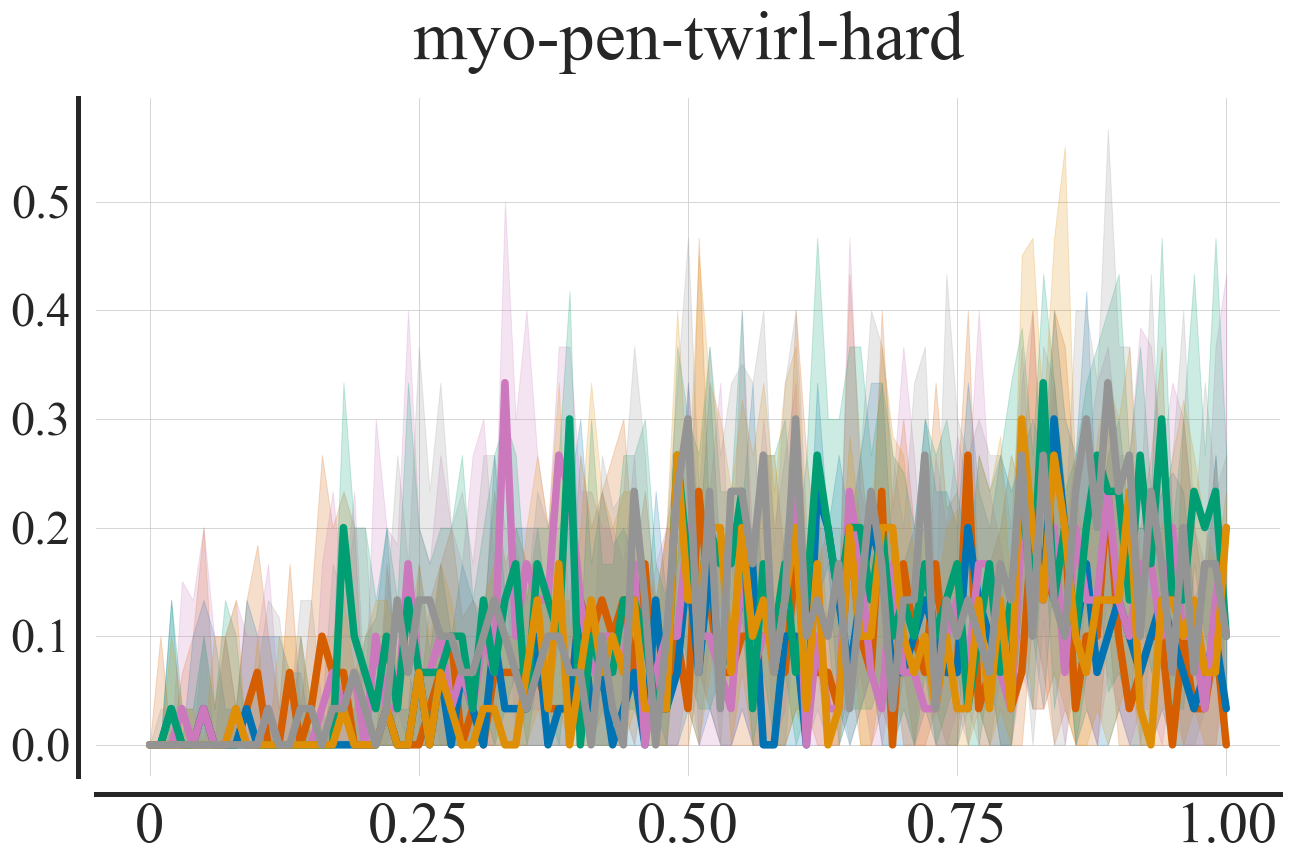}
    \hfill
    \includegraphics[width=0.195\linewidth]{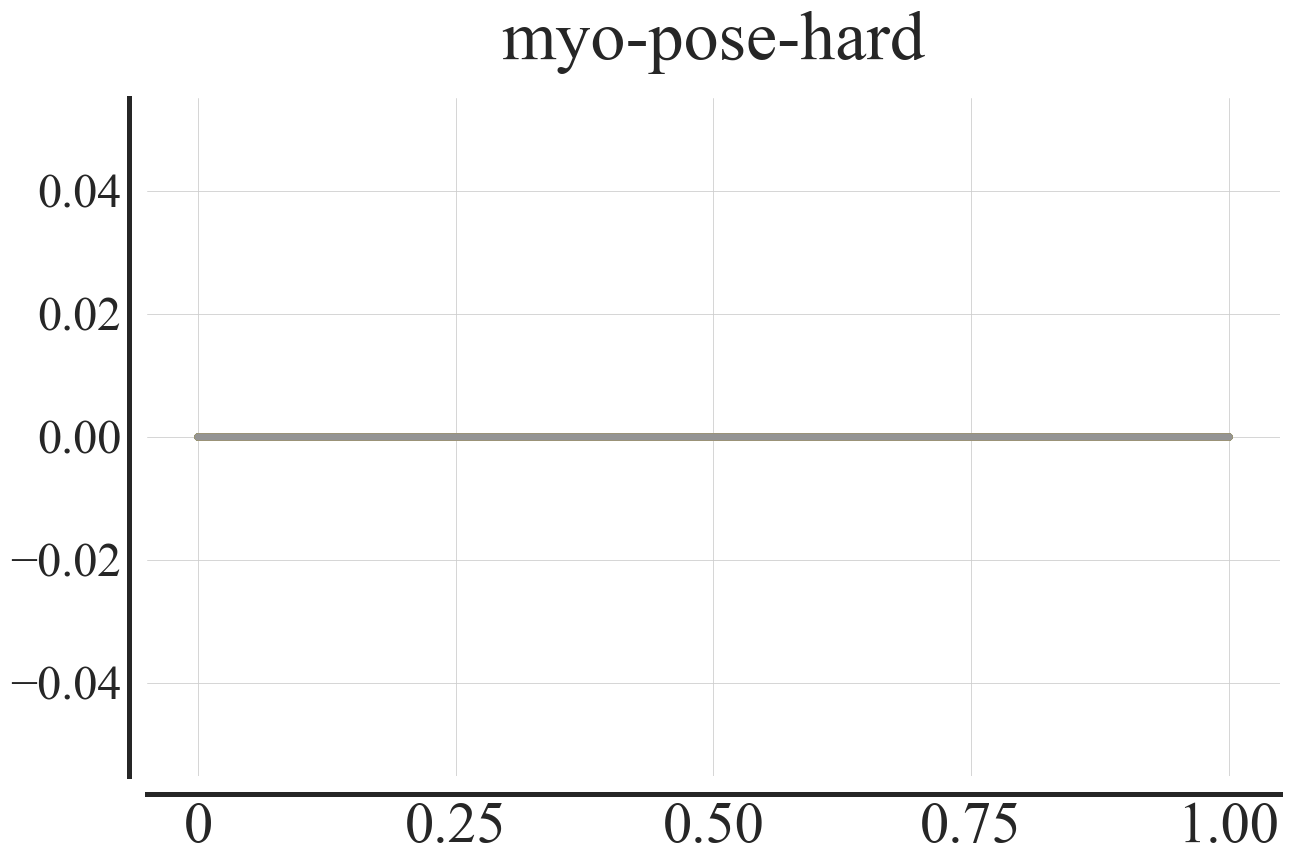}
    \hfill
    \includegraphics[width=0.195\linewidth]{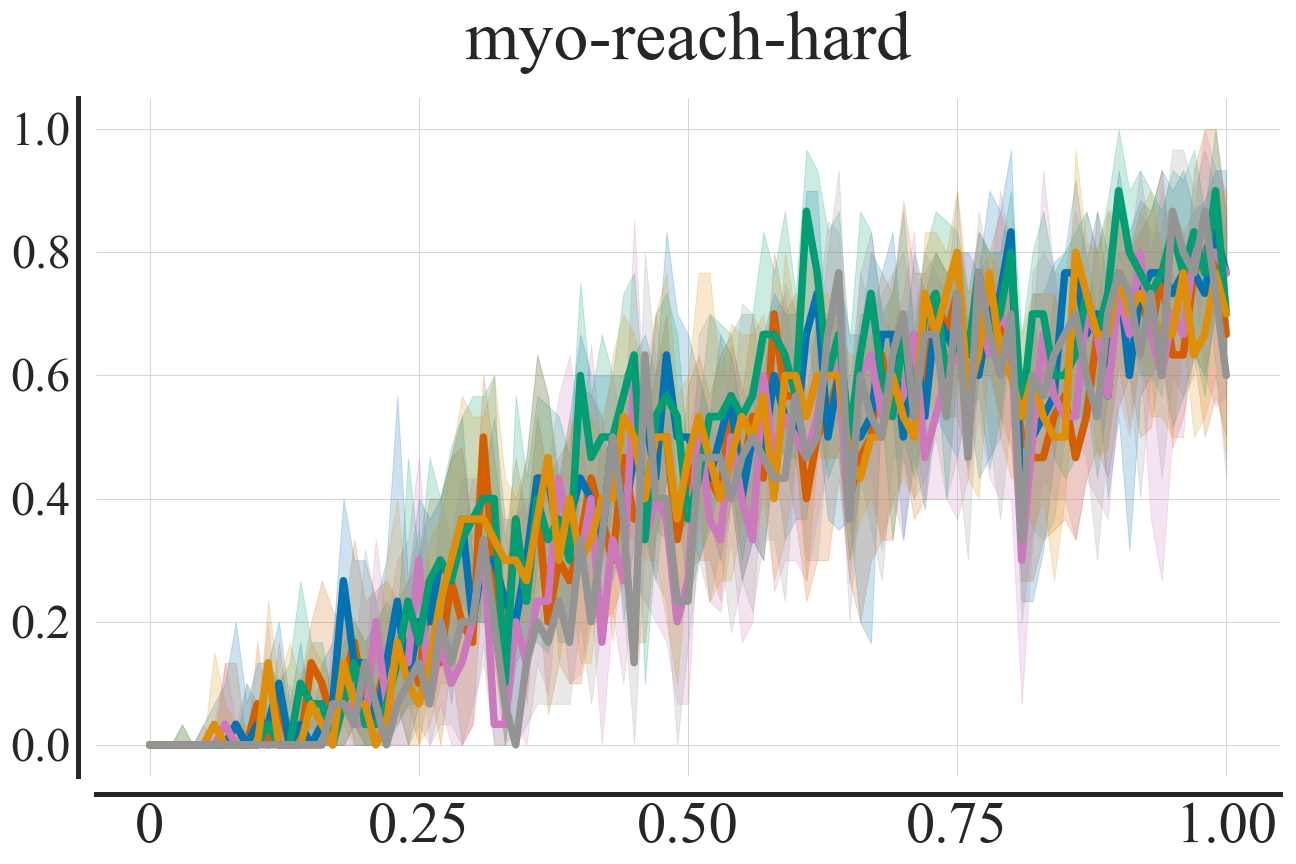}
    \end{subfigure}
\end{minipage}
\caption{Training curves for MYO ($RR=2$ rows $1$ \& $2$, $RR=16$ rows $3$ \& $4$). $Y$-axis denotes IQM success rates and $X$-axis denotes environment steps. 10 seeds per task.}
\label{fig:training_rr2_ms}
\end{center}
\vspace{-0.1in} 
\end{figure}

\end{document}